\documentclass[runningheads]{llncs}
\usepackage{graphicx}

\usepackage{tikz}
\usepackage{comment}
\usepackage{amsmath,amssymb} 
\usepackage{color}
\usepackage{cite}
\usepackage{float}
\usepackage{subcaption}
\usepackage{amsmath}
\usepackage{multirow}  
\usepackage{booktabs}

\DeclareMathOperator*{\argmin}{arg\,min}

\usepackage{hyperref}
\usepackage[capitalize]{cleveref}
\crefname{section}{Sec.}{Secs.}
\Crefname{section}{Section}{Sections}
\Crefname{table}{Table}{Tables}
\crefname{table}{Tab.}{Tabs.}

\usepackage[accsupp]{axessibility}  

\usepackage[export]{adjustbox}

\begin{document}
\pagestyle{headings}
\mainmatter
\def\ECCVSubNumber{6216}  

\title{Compositional Human-Scene Interaction Synthesis with Semantic Control}

\titlerunning{Compositional Human-Scene Interaction Synthesis with Semantic Control}
%
\author{Kaifeng Zhao\inst{1} \and
Shaofei Wang\inst{1} \and
Yan Zhang\inst{1} \and
Thabo Beeler \inst{2} \and
Siyu Tang \inst{1}
}
\authorrunning{K. Zhao et al.}
%
\institute{ETH Zürich \\
\email{\{kaifeng.zhao, shaofei.wang, yan.zhang, siyu.tang\}@inf.ethz.ch} 
\and Google \\
\email{thabo.beeler@gmail.com}
}
\maketitle

\begin{abstract}
Synthesizing natural interactions between virtual humans and their 3D environments is critical for numerous applications, such as computer games and AR/VR experiences.
Recent methods mainly focus on modeling geometric relations between 3D environments and humans, where the high-level semantics of the human-scene interaction has frequently been ignored. 
%
%
Our goal is to synthesize humans interacting with a given 3D scene controlled by high-level semantic specifications as pairs of action categories and object instances, e.g.,~``sit on the chair''.
The key challenge of incorporating interaction semantics into the generation framework is to learn a joint representation that effectively captures heterogeneous information, including human body articulation, 3D object geometry, and the intent of the interaction.
To address this challenge, we design a novel transformer-based generative model, in which the articulated 3D human body surface points and 3D objects are jointly encoded in a unified latent space, and the semantics of the interaction between the human and objects are embedded via positional encoding. 
Furthermore, inspired by the compositional nature of interactions that humans can simultaneously interact with multiple objects, we define interaction semantics as the composition of varying numbers of atomic action-object pairs. 
Our proposed generative model can naturally incorporate varying numbers of atomic interactions, which enables synthesizing compositional human-scene interactions without requiring composite interaction data. 
%
We extend the PROX dataset with interaction semantic labels and scene instance segmentation to evaluate our method and demonstrate that our method can generate realistic human-scene interactions with semantic control. Our perceptual study shows that our synthesized virtual humans can naturally interact with 3D scenes, considerably outperforming existing methods. 
We name our method \textbf{COINS}, for \textbf{CO}mpositional \textbf{IN}teraction
Synthesis with \textbf{S}emantic Control. Code and data are available at \href{https://github.com/zkf1997/COINS}{\color{blue}{https://github.com/zkf1997/COINS}}.

\end{abstract}

 \begin{figure}[t]
    \centering
    \includegraphics[width=\textwidth]{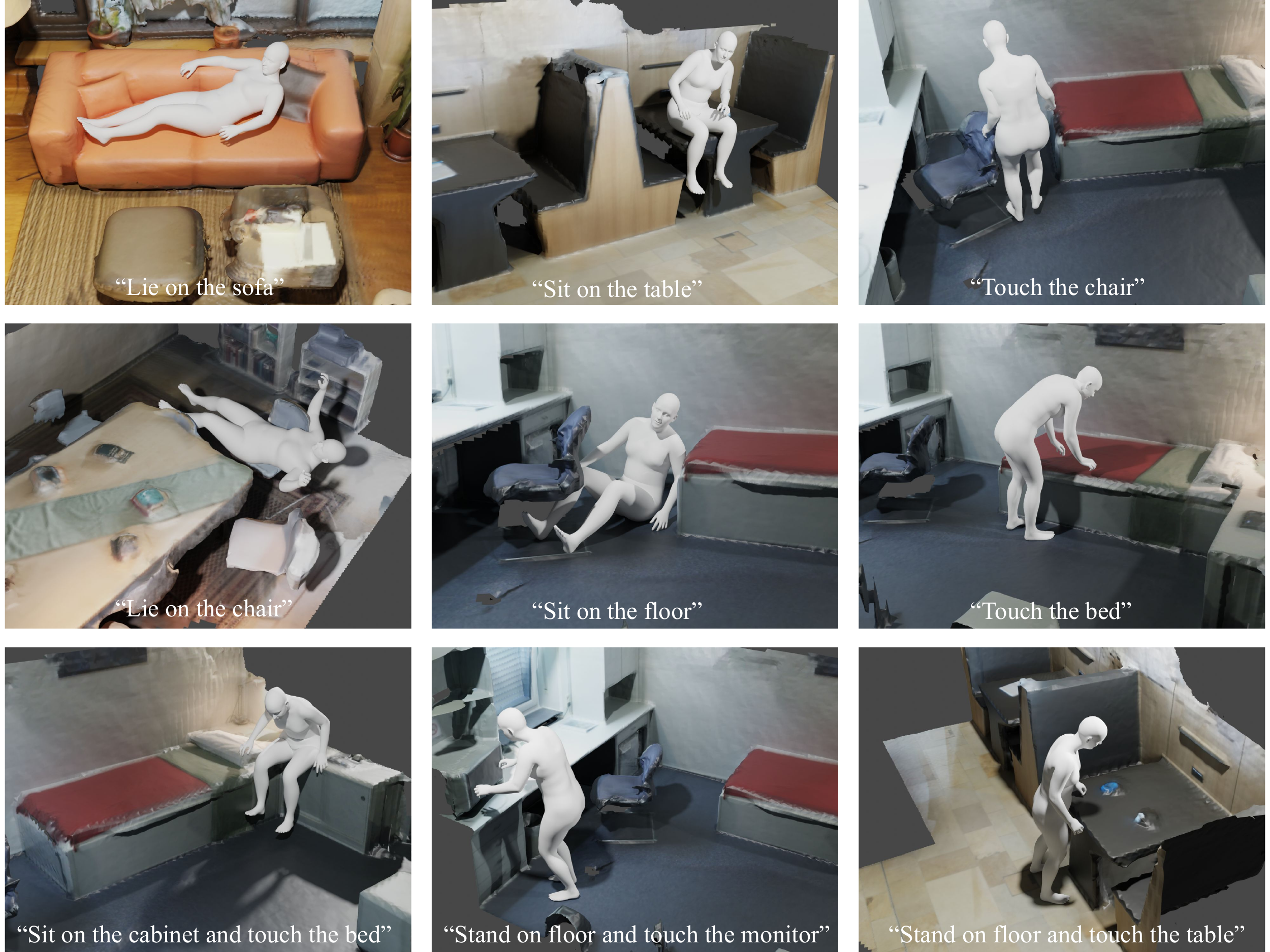}
    
    \caption{Given a pair of action and object instance as the semantic specification, our method generates virtual humans naturally interacting with the object ({\it first row}). Furthermore, our method retargets interactions on unseen action-object combinations ({\it second row}) and synthesizes composite interactions without requiring any corresponding composite training data ({\it third row}).}
    \label{fig:teaser}
\end{figure}

\section{Introduction}
People constantly interact with their surroundings, and such interactions have semantics,  
specifically as combinations of actions and object instances, e.g.,~``sit on the chair" or ``lie on the bed". 
Incorporating and controlling such interaction semantics is critical for creating virtual humans with realistic behavior. 
An effective solution for semantic-aware virtual human generation could advance existing technologies in AR/VR, computer games, and synthetic data generation to train machine learning perception algorithms.
However, existing methods focus on modeling geometric relationships between virtual humans and their 3D environments \cite{PSI:2019, PLACE:3DV:2020, Hassan:CVPR:2021}, and thus are not able to synthesize body-scene interactions with semantic control of human actions and specific interaction objects. This heavily limits their use in practice.

In this paper, we aim to incorporate semantic control into the synthesis of human-scene interactions. 
This is a challenging task due to the following reasons. 
First, semantic-aware human-scene interactions are characterized by body articulation, 3D object geometry, and the intent of the interaction. It is challenging to learn a generalizable representation to capture such heterogeneous information.
Second, given a 3D scene, the space for plausible human-scene interactions is enormous, and obtaining sufficiently diverse training data remains challenging. 


To address these challenges, we introduce a novel interaction synthesis approach that leverages transformer-based generative models and can synthesize realistic 3D human bodies, given a 3D scene and a varying number of action-object pairs. These pairs can specify the intent of the interactions and the interacting objects. The key advantages of our models are two-fold: First, we represent human body articulation and 3D object geometry as tokens in the proposed transformer network, and the interaction semantics are embedded via positional encoding. Combined with a conditional variational auto-encoder (VAE) \cite{kingma2013auto, sohn2015learning}, we learn a unified latent space that captures the distribution of human articulation conditioned on the given 3D objects and the interaction semantics. Second, our models can be used to synthesize composite interactions, as they can naturally incorporate a varying number of 3D objects and interaction semantics.  By training only on irreducible atomic interactions (e.g.,~``sit on the sofa'', ``touch the table''), our model can generate novel composite interactions (e.g.,~``sit on the sofa {\it and} touch the table'') in unseen 3D scenes without requiring  composite training data (\cref{fig:teaser}). 
Furthermore, we decompose the interaction synthesis task into three stages. First, we infer plausible global body locations and orientations given pairs of actions and objects. Second, we generate detailed body articulation and body-scene contact that aligns with the object geometry and interaction semantics.
Lastly, we further refine the local body-scene contact by leveraging the inferred contact map.

To train and evaluate our method, we extend the PROX dataset \cite{PROX:2019} to the PROX-S dataset, which includes 3D instance segmentation, SMPL-X body estimation, and per-frame annotation of interaction semantics.
Since there are no existing methods working on the same task of populating scenes with semantic control, we adapt PiGraph \cite{savva2016pigraphs} and POSA \cite{Hassan:CVPR:2021} to this new task, and train their modified versions on the PROX-S dataset as baselines.
Our perceptual study shows that our proposed method can generate realistic 3D human bodies in a 3D scene with semantic control, significantly outperforming those baselines. 

In summary, our contributions are: (1) a novel compositional representation for interaction with semantics; (2) a generative model for synthesizing diverse and realistic human-scene interactions from semantic specifications; (3) a framework for composing atomic interactions to generate composite interactions; and (4) the extended human-scene interaction dataset PROX-S containing scene instance segmentation and per-frame interaction semantic annotation.

\section{Related Work}
\textbf{Human-Scene Interaction Synthesis.}
Synthesizing  human-scene interactions has been a challenging problem in the computer vision and graphics community. 
Various methods have been proposed to analyze object affordances and generate static human-scene interactions \cite{VladimirGKim2014Shape2PoseHS, savva2014scenegrok, gupta20113d, savva2016pigraphs, grabner2011makes, hu2020predictive, li2019putting}. 
However, the realism of synthesized interactions from these methods is often limited by their oversimplified human body representations. 
The recently published PSI \cite{PSI:2019} uses the  parametric body model SMPL-X \cite{SMPL-X:2019} to put 3D human bodies into scenes, conditioned on scene semantic segmentation and depth map. In addition, geometric constraints are applied to resolve human-scene contact and penetration problems.
PLACE \cite{PLACE:3DV:2020} learns the human-scene proximity and contact in interactions to infer physically plausible humans in 3D scenes. 
POSA \cite{Hassan:CVPR:2021} proposes an egocentric contact feature map that encodes per-vertex contact distance and semantics information. A stochastic model is trained to predict contact feature maps given body meshes, with the predicted contact features guiding the body placement optimization in scenes. 
Wang et al.\cite{wang2021synthesizing} generate human bodies at specified locations and orientations, conditioning on the point cloud of the scene and implicitly modeling the scene affordance. Concurrent work \cite{wang2022towards} first generates scene-agnostic body poses from actions and then places body into scene using POSA, which is similar to the POSA-I baseline we implement.

Apart from being realistic, semantic control is another important goal in synthesizing interactions. 
PiGraph \cite{savva2016pigraphs} learns a probabilistic distribution for each annotated interaction category and synthesizes interactions by selecting the highest scoring samples of fitted categorical distribution. 
%
Conditional action synthesis methods \cite{petrovich2021action,ahuja2019language2pose,ahn2018text2action} generate human motion conditioned on action categories or text descriptions while ignoring human-scene interaction.
NSM \cite{SebastianStarke2019NeuralSM} uses a mixture of experts modulated by the action category to generate human motion of seven predefined categories. SAMP \cite{hassan2021stochastic} extends NSM by incorporating stochastic motion modeling and a GoalNet predicting possible interaction goals given scene objects and action categories. 
NSM and SAMP are limited in interaction semantics modeling since they can only model interactions with one single object or the surrounding environment as a whole. 
Our work, in contrast, models the compositional nature of interaction semantics and can generate composite interactions involving multiple objects simultaneously.

\textbf{Human Behavior Semantics Modelling.}
Understanding human behavior is an important problem that has long received a significant amount of attention. 
Action recognition \cite{simonyan2014two, tran2018closer, kay2017kinetics, ji20123d} classifies human activities into predefined action categories. 
Language driven pose and motion generation \cite{ahn2018text2action, ahuja2019language2pose, tevet2022motionclip} incorporates natural languages that can represent more expressive and fine-grained action semantics.
Human-object interaction detection \cite{AbhinavGupta2009ObservingHI, BangpengYao2010ModelingMC, GeorgiaGkioxari2018DetectingAR}  models interactions as triplets of (human, action, object), detecting humans and objects from images along with the relations between them.

One key observation is that humans can perform multiple actions simultaneously and such behaviors have composite semantics.
Composite semantics are often modeled as the combination of basic semantic units, like how phrases are composed by words in natural languages. Semantic composition is termed as compound in linguistics \cite{lieber2011oxford, plag2018word} and studied in natural language processing \cite{mitchell2008vector, yin2020sentibert, mineshima2015higher}.
PiGraph \cite{savva2016pigraphs} describes composite human-scene interaction semantics with a set of (verb, noun) pairs.
BABEL \cite{BABEL:CVPR:2021} uses natural language labels to annotate human motion capture data \cite{AMASS:ICCV:2019, de2009guide, IjazAkhter2015PoseconditionedJA, ChristianMandery2015TheKW, NikolausFTroje2002DecomposingBM} and allow multiple descriptions existing in a single frame to model composite semantics. 
These works model semantics on the object category level and ignore the differences among instances of the same category. Our work goes beyond this limitation to model interaction semantics with specific object instances, which is critical for solving ambiguity in scenes with multiple objects of the same category.

\section{Method}

\subsection{Preliminaries}
\paragraph{Body Representation.}
We represent the 3D human body using a simplified SMPL-X model, consisting of vertex locations, triangle faces, and auxiliary per-vertex binary contact features, denoted as $\mathbf{B}=(\mathbf{V}, \mathbf{F}, \mathbf{C})$. 
For computational efficiency, we downsample the original SMPl-X mesh to have 655 vertices $\mathbf{V} \in \mathbb{R}^{655 \times 3}$ and 1296 faces $\mathbf{F} \in \{1 \cdots 655\}^{1296 \times 3}$ using the sampling method of \cite{COMA:ECCV18}. 
The binary contact features $\mathbf{C} \in \{0, 1\}^{655}$ indicate whether each body vertex is in contact with scene objects \cite{Hassan:CVPR:2021}. Specifically, we consider a body vertex is in contact if its distance to scene objects is below a threshold value. 
We introduce the auxiliary contact features because human-object contact is a strong cue for interaction semantics, e.g., ``touch'' implies contact on hands. We use the contact features to further refine human-scene interaction in post-optimization.

\paragraph{3D Scene Representation.}
We represent a 3D scene as a set of object instances, denoted as $\mathbf{S}=\{o^{i}\}_{i=1}^O$ where $O$ is the number of object instances.
Each object instance $o^i \in \mathbb{R}^{8192 \times 9}$ is an oriented point cloud of 8192 points with attributes of location, color, and normal. 
We represent objects as point clouds to incorporate the geometry information and enable generalization to open-set objects that do not belong to predefined object categories, unlike previous works \cite{savva2016pigraphs, Hassan:CVPR:2021, PSI:2019} that only use object category information. 
Moreover, the effect of intra-class object variance on interactions can be learned from the object geometry to improve fine-grained human-scene contacts. 

\paragraph{Compositional Interaction Representation.}
Human-scene interactions involve three types of components: human bodies, actions, and objects.
Inspired by the observation that humans can perform multiple interactions at the same time, we propose a compositional interaction representation $\mathbf{I}=(\mathbf{B}, \{(a^i, o^i)\}_{i=1}^{M})$, where $\mathbf{B}$ denotes the human body, and $\{(a^i, o^i)\}_{i=1}^{M}$ is a set of atomic interactions.
We use the term atomic interaction to refer to the basic, irreducible building block for interaction semantics, and each atomic interaction  $(a^i, o^i)$ is a pair of a one-hot encoded action and an object instance, e.g., ``sit on a chair''.
In addition, interactions can comprise more than one such atomic interaction unit, since one can simultaneously interact with multiple objects, e.g., ``sit on a chair and type on a keyboard''. 
We refer to such interactions consisting of more than one atomic interaction as composite interactions. 

\subsection{Interaction Synthesis with Semantic Control}
Synthesizing interactions with semantic control requires generating human bodies interacting with given scene objects in a natural and semantically correct way.  
Specifically, given a scene $S$ and interaction semantics specified as $\{(a^i, o^i)\}_{i=1}^{M}$, the task is to generate a human body $\mathbf{B}$ that perceptually satisfies the constraints of performing action $a^i$ with object $o^i$. 

\begin{figure}[t]
    \centering
    \includegraphics[width=\textwidth]{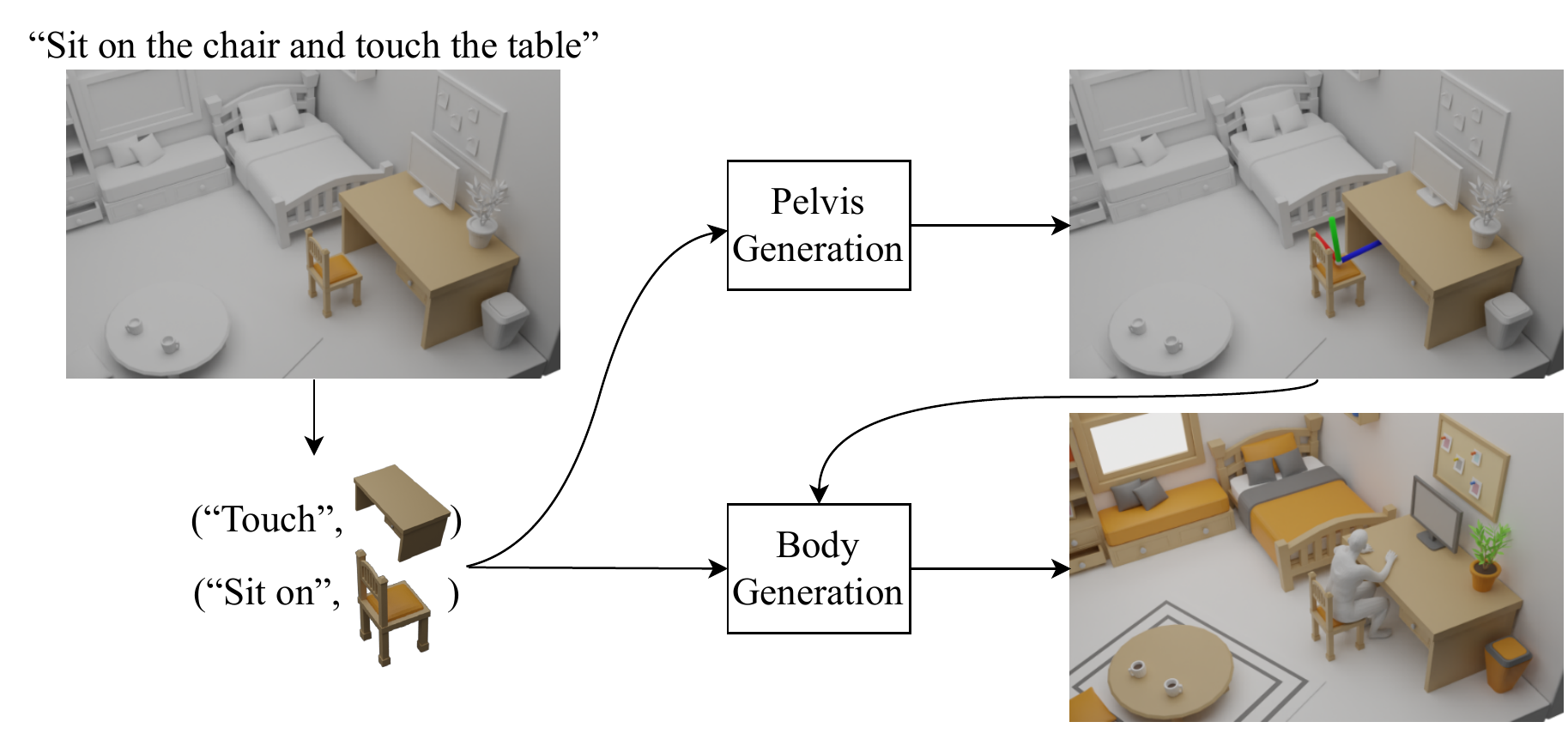}

    \caption{Illustration of our COINS method for human-scene interaction synthesis with semantic control. Given a 3D scene and semantic specifications of actions (e.g.,~``sit on'', ``touch'') paired with object instances (highlighted with color), COINS first generates a plausible human pelvis location and orientation and then generates the detailed body.}
    \label{fig:pipeline}
\end{figure}

We propose a multi-stage method for interaction synthesis with semantic control as illustrated in \cref{fig:pipeline}. Given interaction semantics specified as action-object pairs, we first infer possible pelvis locations and orientations.
Then we transform objects to the generated pelvis coordinate frame and sample body and contact map in the local space.
%
We separately train two transformer-based conditional VAE (cVAE) \cite{kingma2013auto, sohn2015learning} models for pelvis and body generation, and name them as PelvisVAE and BodyVAE respectively. 
Lastly, we apply an interaction-based optimization that is guided by contact features to refine the generated interactions.


\paragraph{Architecture.}
We propose a transformer-based conditional variational auto-encoder (cVAE) architecture that can handle the heterogeneous inputs of our compositional interaction representation containing body mesh vertices, object points without predefined topology, and interaction semantics.
Specifically, object points and body vertices are represented as token inputs to the transformer, and learnable action embeddings are added to paired object tokens as positional encoding. 

\begin{figure}[t]
    \centering
    \includegraphics[width=\textwidth]{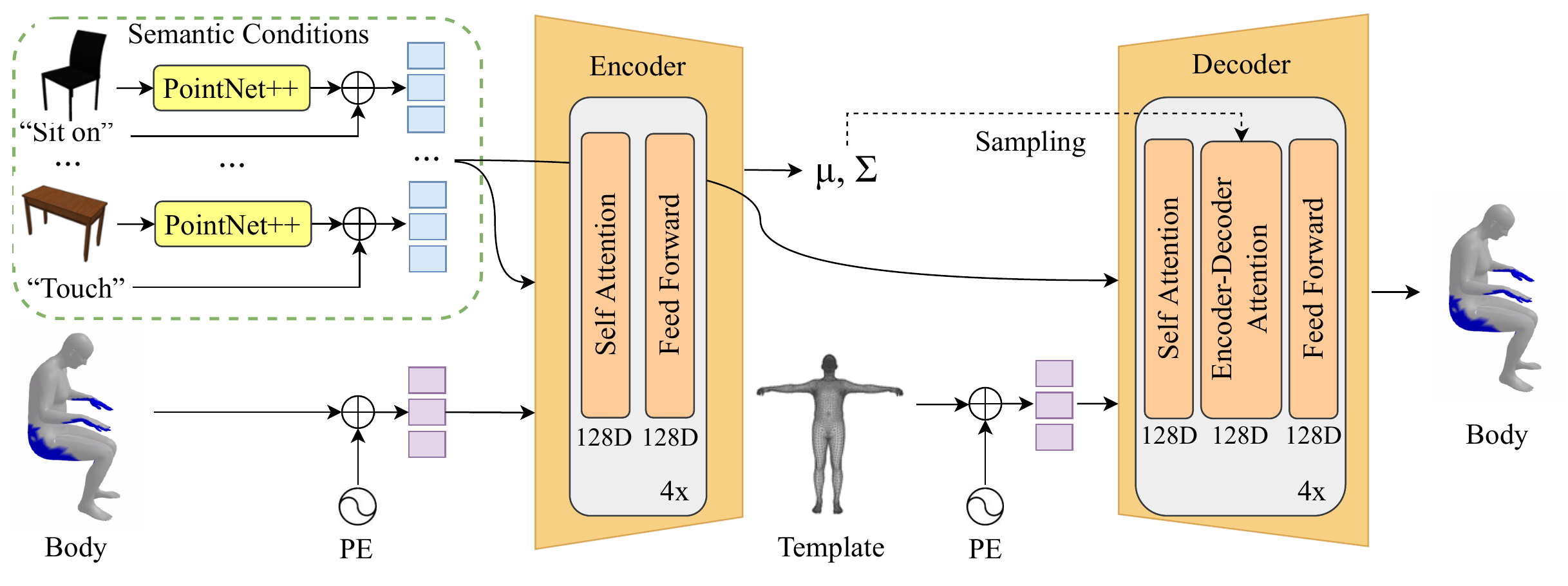}

    \caption{Illustration of BodyVAE architecture. Human body and objects are represented as tokens of the transformers while each action category is mapped to a learnable embedding and added to the corresponding object tokens. Modules within the dashed box are the semantic conditions of a varying number of action-object pairs. 
    We denote addition with $\bigoplus$ and positional encoding with ``PE''.
    We visualize vertices with positive contact features as blue on the body mesh.}
    \label{fig:architecture}
\end{figure}

The BodyVAE architecture is illustrated in \cref{fig:architecture}. The PelvisVAE shares a similar architecture and we refer readers to the Supp.~Mat. for illustration.

Our cVAE networks take varying numbers of action-object pairs as conditional inputs. Each action category is mapped to a learnable embedding, and each object point cloud is first converted to 256 local keypoints using PointNet++~\cite{qi2017pointnetplusplus}. The action embedding is then added to the tokens of the paired object as positional encoding. 
Compared to treating actions as separate tokens~\cite{petrovich2021action}, encoding actions into the positional encoding for object tokens can regulate each action to only affect the paired object, e.g., the action ``sit on'' is only related to the sofa, but not the table, in the example of \cref{fig:architecture}.
%
The human body input comprises mesh vertices with location and auxiliary contact features. We apply a linear layer to map vertices to tokens and vertex index based positional encoding to encode body topology. 

The transformer encoder maps input interactions into a latent space and the decoder predicts human bodies with contact features from conditional action-object pairs and the sampled latent code.
On top of the encoder transformer network, we apply average pooling on the output human tokens followed by a linear layer to predict the latent normal distribution. 
At the decoder, we use a template body with person-dependent shape parameters for body tokens. Using the personalized template body makes our model generalize to different body shapes and enables explicit control of the body shape.
The sampled latent code is used for keys and values of the encoder-decoder attention layers~\cite{vaswani2017attention}.

BodyVAE shares the same transformer-based architecture and the same conditional input of action-object pairs as PelvisVAE. There are mainly two differences between BodyVAE and PelvisVAE. First, BodyVAE represents a human as a body mesh with contact features, while PelvisVAE represents a human as the pelvis joint location and orientation. Second, all input objects are in the scene coordinate frame in PelvisVAE while input objects are transformed to the local pelvis coordinate frame in BodyVAE.

To make our sampled bodies compatible with the SMPL-X body model and ensure a valid body shape, we jointly train a Multi-Layer Perceptron (MLP) \cite{rumelhart1985learning} with the BodyVAE to regress the corresponding SMPL-X body parameters from the sampled body meshes. 
We refer to the Supp.~Mat.~for more details on architecture and the training losses.

\paragraph{Interaction-Based Optimization.}
Our model is trained with the pseudo ground truth body estimation \cite{PROX:2019,Zhang:ICCV:2021}, which does not guarantee natural human-scene contact. Therefore, the generated interactions can also be unnatural. We follow \cite{PSI:2019, PLACE:3DV:2020, Hassan:CVPR:2021} to apply interaction-based optimization to refine the interactions.
We use generated SMPL-X parameters as initialization and optimize body translation $t$, global orientation $R$ and pose $\theta$. The optimization objective is given by:

\begin{equation}
E(t, R, \theta) = \mathcal{L}_{interaction} + \mathcal{L}_{coll} + \mathcal{L}_{reg},
\end{equation}
where $\mathcal{L}_{interaction}$ encourages body vertices with positive contact features to have zero distance to objects, $\mathcal{L}_{coll}$ is the scene SDF based collision term defined in \cite{PSI:2019}, and the $\mathcal{L}_{reg}$ term penalizes $t, R$ and $\theta$ deviating from the initialization. We refer to the Supp.~Mat.~for detailed definitions.

\subsection{Compositional Interaction Generation}
Compositional interaction generation aims to generate composite interactions using models trained only on atomic interactions. 
Compositional interaction generation reflects the compositional nature of interaction semantics and can save the need to collect composite interaction data. 
Although our transformer-based generative model can synthesize bodies given composite interaction semantics, e.g.~``sit on the sofa and touch the wall'', it requires collecting data and training for all composite interactions we want to synthesize. 
However, collecting composite interaction data is exponentially more expensive than atomic interactions considering the vast possible combinations of atomic interactions.
Therefore, it is desirable to have an automatic way to generate composite interactions requiring only atomic interaction data. 

To that end, we propose a simple yet effective two-step approach to generating composite interactions using models trained only on atomic interactions: (1) Compositional pelvis generation, where we formulate the task of generating pelvis frames given composite interaction semantics as finding the intersection of pelvis distributions of atomic interactions. 
%
(2) Compositional body generation, where we leverage attention masks derived from contact statistics in transformer inference to compose atomic interactions at the body parts level. 
\paragraph{Compositional Pelvis Generation.}
We formulate compositional pelvis generation as finding the intersection of pelvis distributions of component atomic interactions. 
Using our PelvisVAE trained for atomic interactions, we can decode the prior standard Gaussian distribution to distributions of plausible pelvis frames given atomic interaction semantics. 
Each component atomic interaction of a composite interaction determines a pelvis frame distribution in scenes, and intuitively the plausible pelvis of the composite interaction should lie in the intersection of all pelvis distributions.
Therefore, we propose a compositional pelvis generation method where we sample pelvis frames for each atomic interaction, optimize all pelvis frames to be close in space, and average the optimized pelvis frames as the generated composite pelvis frame.

Specifically, we denote the decoder of the PelvisVAE as a function $\mathcal{G}(z^i|a^i, o^i)$ mapping a latent code $z^i$ to a pelvis frame conditioned on the action-object pair of $(a^i, o^i)$, we obtain the final pelvis frame from the composition of $M$ atomic interactions from the following optimization problem:

\begin{equation}
    \Tilde{Z} = \argmin_{Z=\{z^i\}_{i=1}^M} \sum_{1 \leq i < M} \sum_{i < j \leq M} \left|\mathcal{G}(z^i|a^i, o^i) - \mathcal{G}(z^j|a^j, o^j)\right| - \lambda \sum_{1 \leq i \leq M} log(p(z^i)),
\end{equation}


 $Z$ denotes the set of $M$ randomly sampled latent codes for $M$ component atomic interactions, and $\lambda$ is a tunable weight. The initial values of the latent codes are randomly sampled from the prior distribution. 
 The motivation is to optimize the latent codes of $M$ component atomic interactions to make their decoded pelvis frames as close as possible in the scene while maintaining high probabilities in each component pelvis distribution. 
 We take the average of the $M$ optimized component pelvis frames as the sampled composite pelvis frame.

\paragraph{Compositional Body Generation.}
\label{sec:compo_body}
For compositional body generation, we use the attention mechanism of transformer networks to compose atomic interactions at body part level.
Although we use BodyVAE trained only on atomic interaction data, we can input more than one atomic interaction and achieve compositional body generation with additional attention masks derived from contact statistics. 

\begin{figure}[t]
    \centering
    \includegraphics[width=\textwidth]{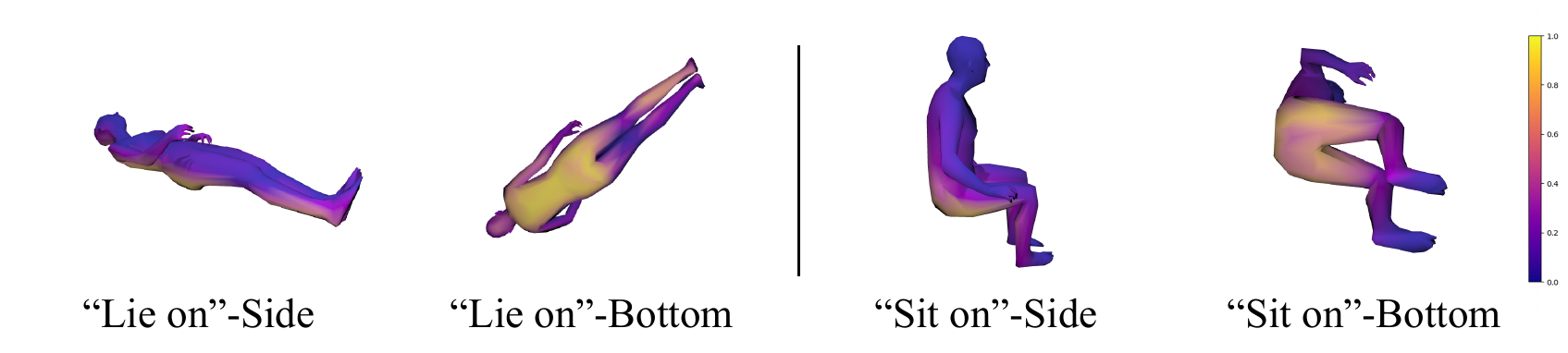}
    
    \caption{Visualization of per-vertex contact probability for two actions from the side and bottom views. Brighter colors indicate higher contact probability with objects. We randomly sample a pose within the action category for visualization. }
    \label{fig:heatmap}
\end{figure}

Our method is inspired by the observation from contact probability statistics that interaction semantics can often be determined by contact on certain body parts instead of the whole body, e.g.,~sitting is characterized by contact around the hip area. 
Specifically, we calculate the per-vertex contact probability for each action using training data. The vertex contact probability for one action is defined as the ratio of the number of positive contact features of the vertex and the number of all contact features of the vertex in the interaction data of the specified action. 
Heatmap visualization of such categorical contact statistics is shown in \cref{fig:heatmap}. High contact probability indicates a high correlation between the body vertex area and action semantics. 

Inspired by the observed action-part correlation, we propose leveraging the attention mechanism to compose atomic interactions according to three body parts: upper limbs, lower limbs, and torso. Given composite interaction semantics for body generation, we calculate the sum of contact probability at each part for each action. If the sum of contact probability of one action at one body part exceeds those of other actions by a threshold, we mask the attention between the part and all other atomic interactions to be zero. 
Taking the composite interaction ``sit on chair and touch table'' as an example, touching has a significantly greater sum of contact probability on upper limbs so the attention between upper limb tokens and chair (objects not paired with touching) tokens will be masked as zero.

We input such attention masks and component atomic interactions to the transformer for compositional body generation. The derived attention masks can naturally compose the component atomic interactions via BodyVAE.

\section{Experiments}

\subsection{Dataset}
\begin{figure}[t]
    \centering
    \begin{subfigure}[t]{.32\textwidth}
        \includegraphics[width=\textwidth]{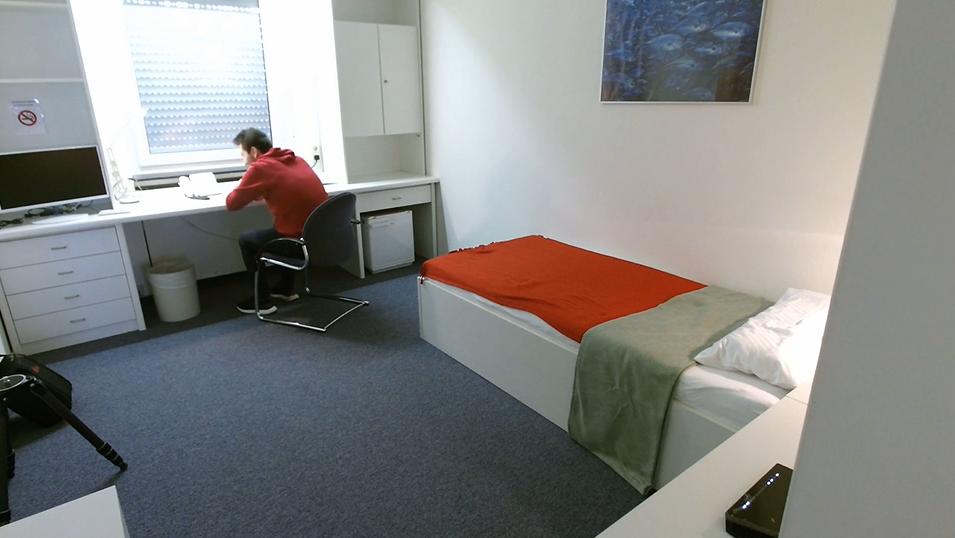}
    \end{subfigure}
    \begin{subfigure}[t]{.32\textwidth}
        \includegraphics[width=\textwidth]{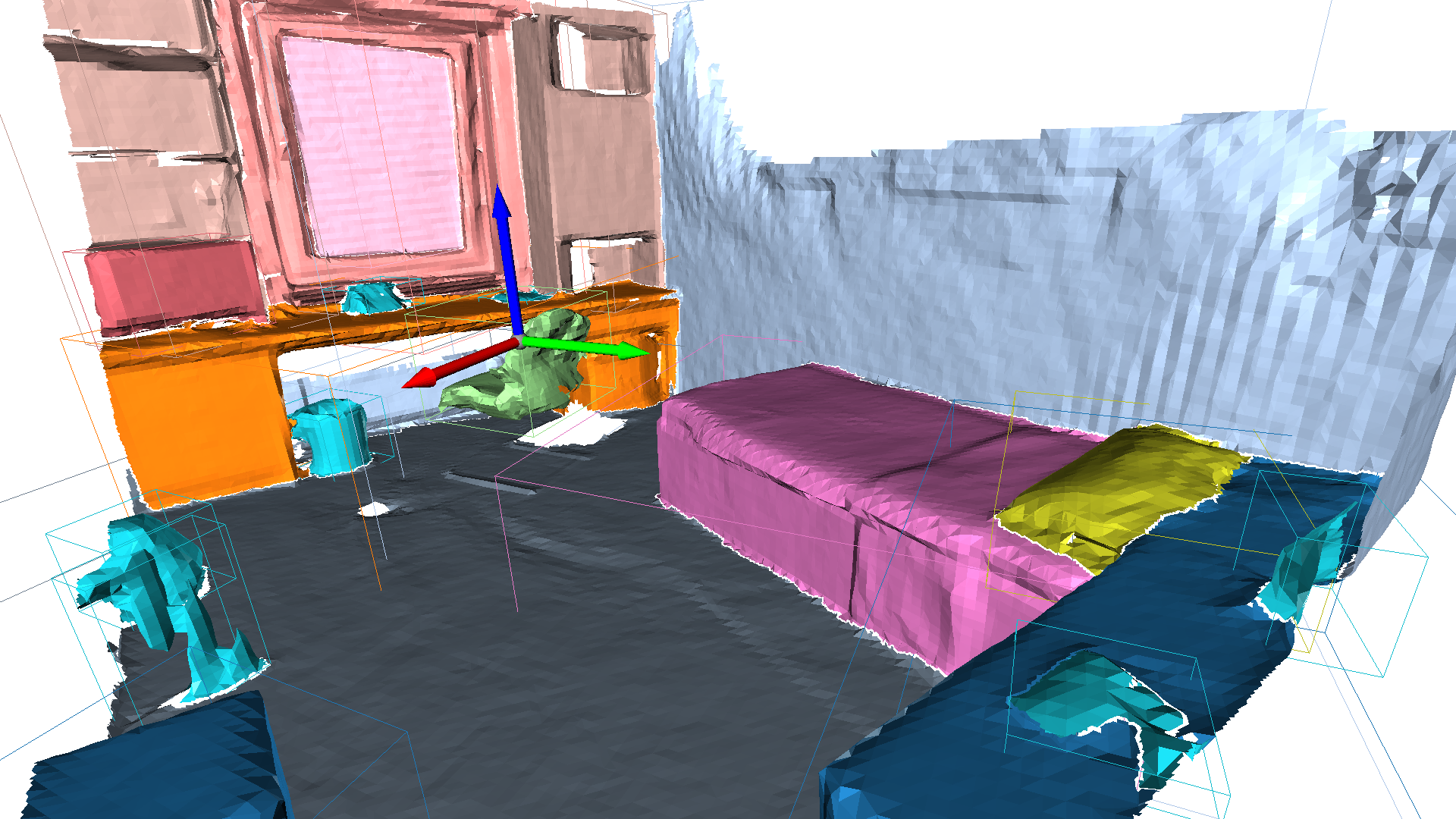}
    \end{subfigure}
    \begin{subfigure}[t]{.32\textwidth}
        \includegraphics[width=\textwidth]{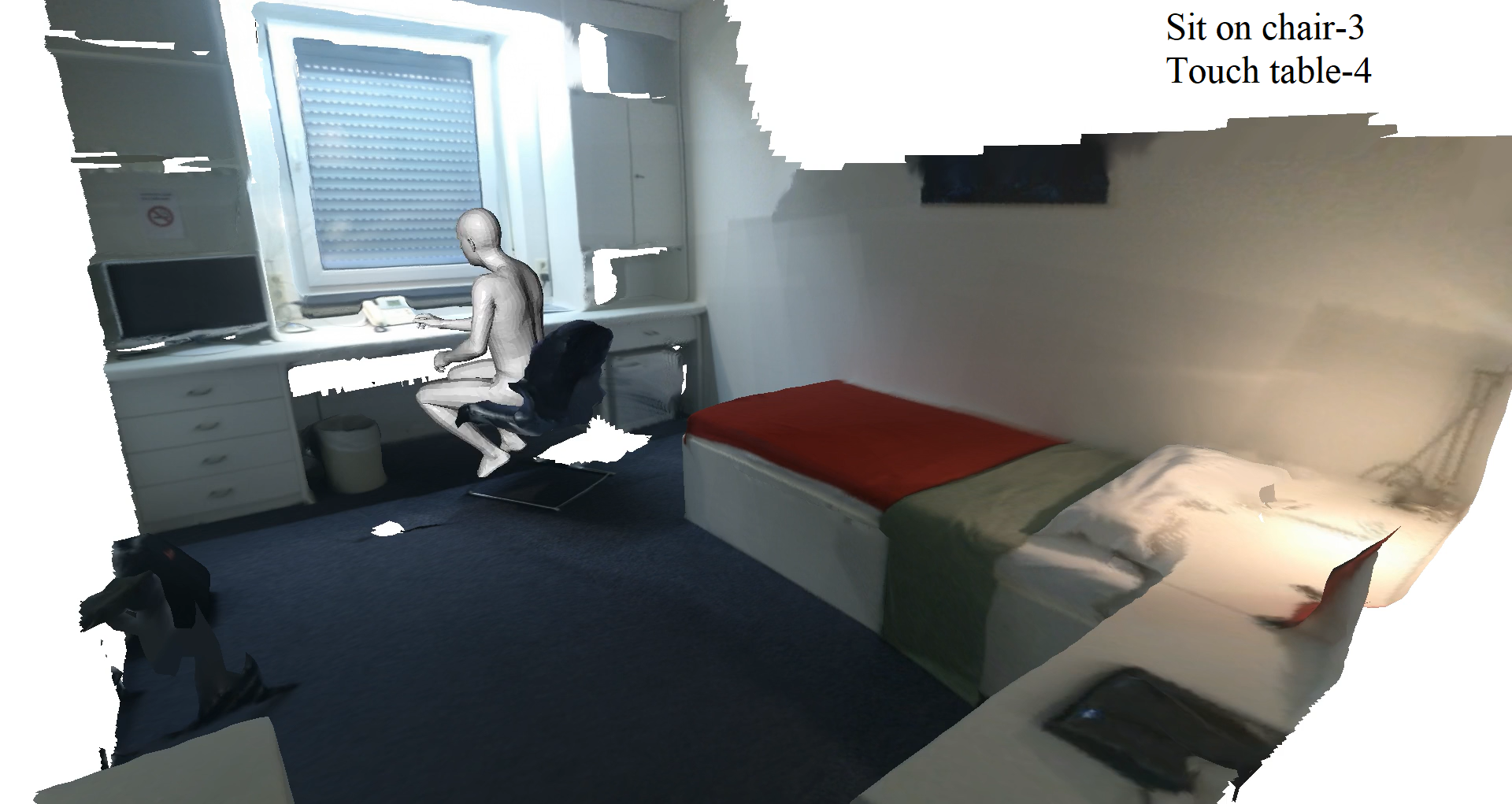}
    \end{subfigure}
    \caption{Illustration of the PROX-S dataset. From left to right, it shows the RGB image, 3D scene reconstruction with instance segmentation, and the fitted SMPL-X human body with per-frame interaction semantic labels. }
    \label{fig:proxs}
\end{figure}

We annotate a new dataset called PROX-S on top of the PROX \cite{PROX:2019} dataset to evaluate interaction synthesis with semantic control since there is no existing dataset suitable for this task. 
The PROX-S dataset contains (1) 3D instance segmentation of all 12 PROX scenes; (2) 3D human body reconstructions within the scenes; and (3) per-frame interaction semantic labels in the form of action-object pairs, as shown in \cref{fig:proxs}. 
We refer to the Supp.~Mat. for the detailed process description and statistics. 

\subsection{Evaluation Metrics}

\paragraph{Semantic Contact.} 
Compared to the semantics-agnostic body-scene contact score used in \cite{PSI:2019}, we propose a semantic contact score to evaluate whether the generated interactions adhere to the semantic specifications characterized by proper body-scene contact. 
Using the action-specific body vertex contact probability introduced in \cref{sec:compo_body} as weights, we calculate a weighted sum of binary contact features for each body in the training data and name the sum as the semantic contact score.
This action-dependent semantic contact score reflects how body contact features align with the semantics of one action. We use the 20 percentile semantic contact score of training data bodies as the threshold score for each action to filter low-quality training interaction data with inaccurate contact.
When evaluating a generated interaction sample, we calculate the weighted sum of contact features and compare it to the action-specific threshold score. The semantic contact score is one if the sum is higher than the threshold and zero otherwise. 

\paragraph{Physical Plausibility.} 
We evaluate the physical plausibility of generated interactions using the non-collision metric from \cite{PSI:2019, PLACE:3DV:2020, Engelmann21CVPR,Hassan:CVPR:2021}. 
%
%
It is calculated as the ratio of the number of body vertices with non-negative scene SDF values and the number of all body vertices.

\paragraph{Diversity.} Following \cite{PSI:2019}, we perform K-Means clustering to cluster the SMPL-X parameters of generated bodies into 50 clusters and evaluate the diversity using the entropy of the cluster ID histogram and the mean distance to cluster centers. 

\paragraph{Perceptual Study.} 
We evaluate the perceptual realism and semantic accuracy of generated human-scene interactions by conducting perceptual studies on Amazon Mechanical Turk (AMT).
We perform a binary-choice perceptual study where samples of the same interaction semantics from different sources are shown and the turkers are asked to choose the more natural one. 
For semantic accuracy, we instruct the turkers to rate the consistency between shown interaction samples and semantic labels from 1 (strongly disagree) to 5 (strongly agree).
For more details regarding the perceptual studies, please refer to the Supp.~Mat.

\subsection{Baselines}
There is no existing method that directly applies to the problem of generating 3D human bodies given a 3D scene and interaction semantics. Therefore, we modify two state-of-the-art interaction synthesis methods for our tasks and train their models using the same PROX-S dataset. 

\paragraph{PiGraph-X.}
PiGraph~\cite{savva2016pigraphs} is the most related work to our knowledges, which simultaneously generates scene placement and human skeletons from interaction category specifications. 
We remove the scene placement step of the original paper and replace the skeleton body representation with the more realistic SMPL-X body model.
%
%
We refer to this modified PiGraph variant as PiGraph-X.
For compositional interaction generation, we extend PiGraph by sampling basic bodies from all comprised atomic interactions and using the average similarity score to those atomic interactions. 
We denote this PiGraph extension as PiGraph-X-C.

\paragraph{POSA-I.}
POSA\cite{Hassan:CVPR:2021} generates human-scene interactions using optimization guided by per-vertex contact features.
However, POSA lacks global control of the generated contact features and cannot generate human bodies conditioned on interaction semantics. 
To incorporate global semantics of human-scene interaction, we modify POSA to generate human bodies with vertex-level contact features from interaction semantics in the format of action-noun pairs. 
We denote the semantic-controllable POSA variant as POSA-I. 
\subsection{Interaction Synthesis with Semantic Control}
\label{sec:exp_semantic}
We evaluate the synthesized interactions in terms of perceptual realism, perceptual semantic accuracy, semantic contact, physical plausibility, and diversity. 
Table \ref{tab:binary} and \ref{tab:metrics} show the results of the perceptual studies on binary realism and other quantitative metrics, respectively. Our method significantly outperforms the baselines in the binary realism study and even comes close to the body poses reconstructed from the real human-scene interactions, which we consider as the pseudo ground truth. Our method also achieves higher perceptual semantic ratings than all the baselines. Quantitatively, our method achieves the highest semantic contact score (0.93) which significantly outperforms the baselines.
Our method performs comparably on physical plausibility as POSA-I and significantly better than PiGraph-X. 
For diversity metrics, our method has a larger cluster entropy and smaller cluster size compared to the baselines. 
We observe more frequent unnatural interaction samples from PiGraph-X and POSA-I, which accounts for their larger cluster sizes.
Note that our method is a generative pipeline where we evaluate every randomly sampled result, while both baselines need to search thousands of samples of global placement in scenes. 

\begin{table}[t]
\scriptsize
\caption{Binary Realism Perceptual Study.}
    \begin{subtable}{.45\textwidth}
        \centering
        \caption{Interaction Synthesis with Semantic Control.}
        \begin{tabular}{cc}
            \toprule
            \multicolumn{2}{c}{ \% users rated as ``more realistic''}   \\ \hline
             Ours  & POSA-I \\
             \textbf{63.8} \%   &  36.2 \%          \\ \hline
             Ours & PiGraph-X  \\
             \textbf{79.7} \%  &  20.3  \%       \\ \hline
             Ours & PseudoGT \\
            45.3 \%                      &  \textbf{54.7} \%              \\ 
            \bottomrule
        \end{tabular}
    \end{subtable}%
    \hfill
    \begin{subtable}{.45\textwidth}
        \centering
        \caption{Interaction Synthesis via Semantic Composition.}
        \begin{tabular}{cc}
            \toprule
            \multicolumn{2}{c}{ \% users rated as ``more realistic''}   \\ \hline
             Ours-C & PiGraph-X-C  \\
             \textbf{81.2} \%  &  18.2  \%       \\ \hline
             Ours-C & PseudoGT \\
            42.4 \%                      &  \textbf{57.6} \%              \\ 
            \bottomrule
        \end{tabular}
    \end{subtable}
    
    \label{tab:binary}
\end{table}

\begin{table}[t]
    \caption{Evaluation of semantics, physical plausibility and diversity.}
    \footnotesize
    \centering
    \resizebox{\textwidth}{!}
    {
        \begin{tabular}{@{\extracolsep{4pt}}lccccccc}
        \hline
       & Semantic Accuracy  & Semantic Contact  & Non-Collision & Entropy & Cluster Size & Samples \\
        \hline
        PiGraph-X  & 3.90 $\pm$ 0.93 & 0.84  & 0.85 & 3.67 & 1.07 & 25K\\
        POSA-I & 4.03 $\pm$ 0.85& 0.67 & \textbf{0.98} & 3.75 & \textbf{1.08} & $>1K$ \\
        Ours & \textbf{4.17} $\pm$ \textbf{0.71}& \textbf{0.93} & 0.97 & \textbf{3.77} & 0.83 & \textbf{1} \\
        
        \hline
        \end{tabular}
    }
    
    \label{tab:metrics}
\end{table}

\begin{figure}
  \captionsetup[subfigure]{labelformat=empty}
  \centering
  \includegraphics[width=\textwidth]{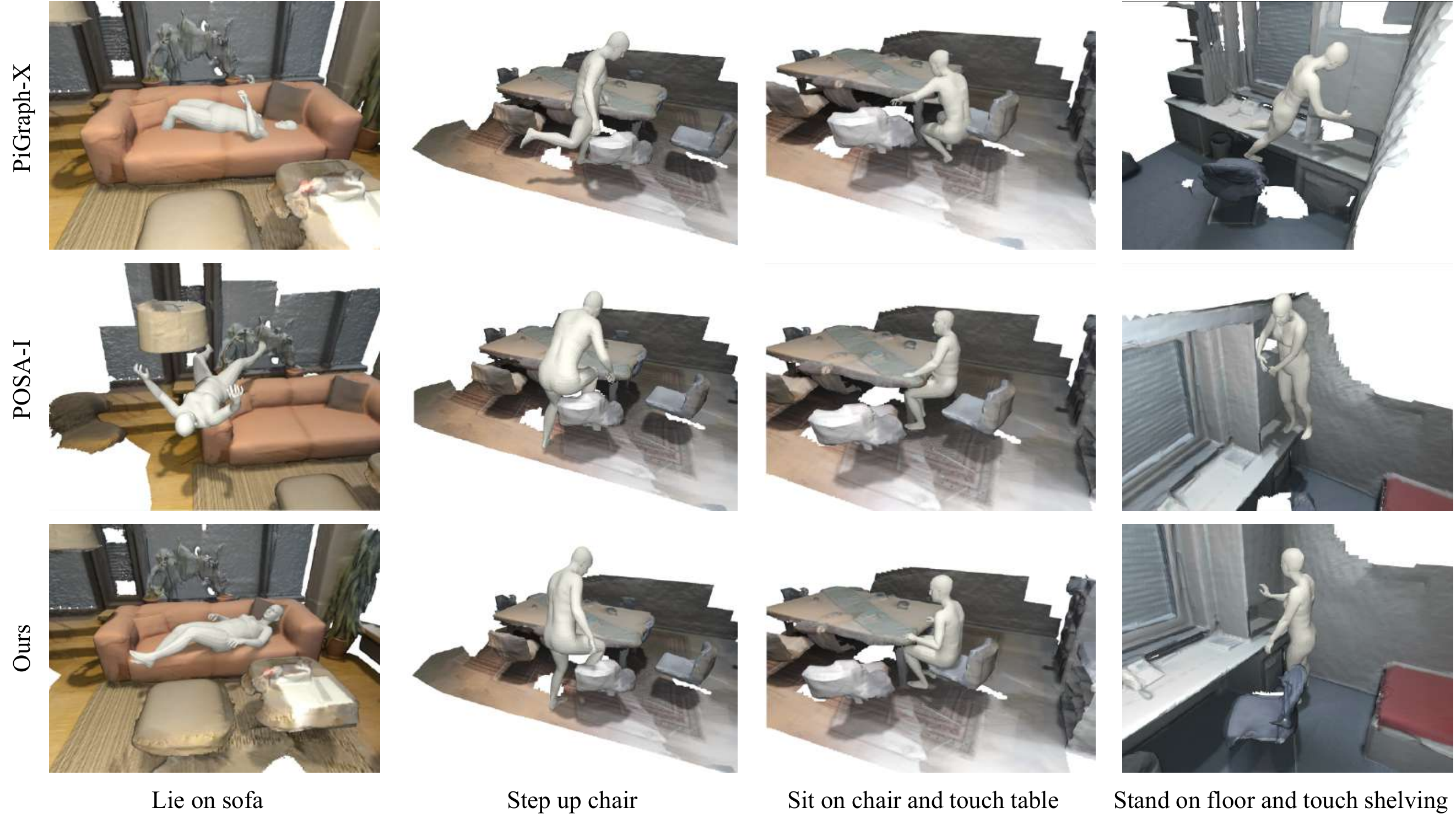}
    \caption{Qualitative comparison of interactions generated with our method and two baselines. Each row shows interactions generated from one method. From left to right, the columns correspond to different interaction semantics of ``lying on the sofa'', ``stepping up on the chair'', ``sitting on the chair and touching the table'', and ``standing on the floor and touching the shelving''.  }
        \label{fig:comparison}
\end{figure}

We show the qualitative comparison of interactions generated from ours and two baselines in  \cref{fig:comparison}.
PiGraph-X naively models the human body as independent joints and therefore suffers from unnatural body poses, poor human-scene contact, and penetration. 
POSA-I generates bodies with physically plausible interactions, but since it needs to find plausible body placement in the scene, its optimization can easily get stuck at local minimums that are not semantically accurate and have inferior human-scene contact. 
In contrast, our method learns how humans interact with objects given different interaction semantics and generates realistic interactions that are both physically plausible and human-like.
\subsection{Compositional Interaction Generation}

\begin{figure}[t]
    \centering
    \includegraphics[width=\textwidth]{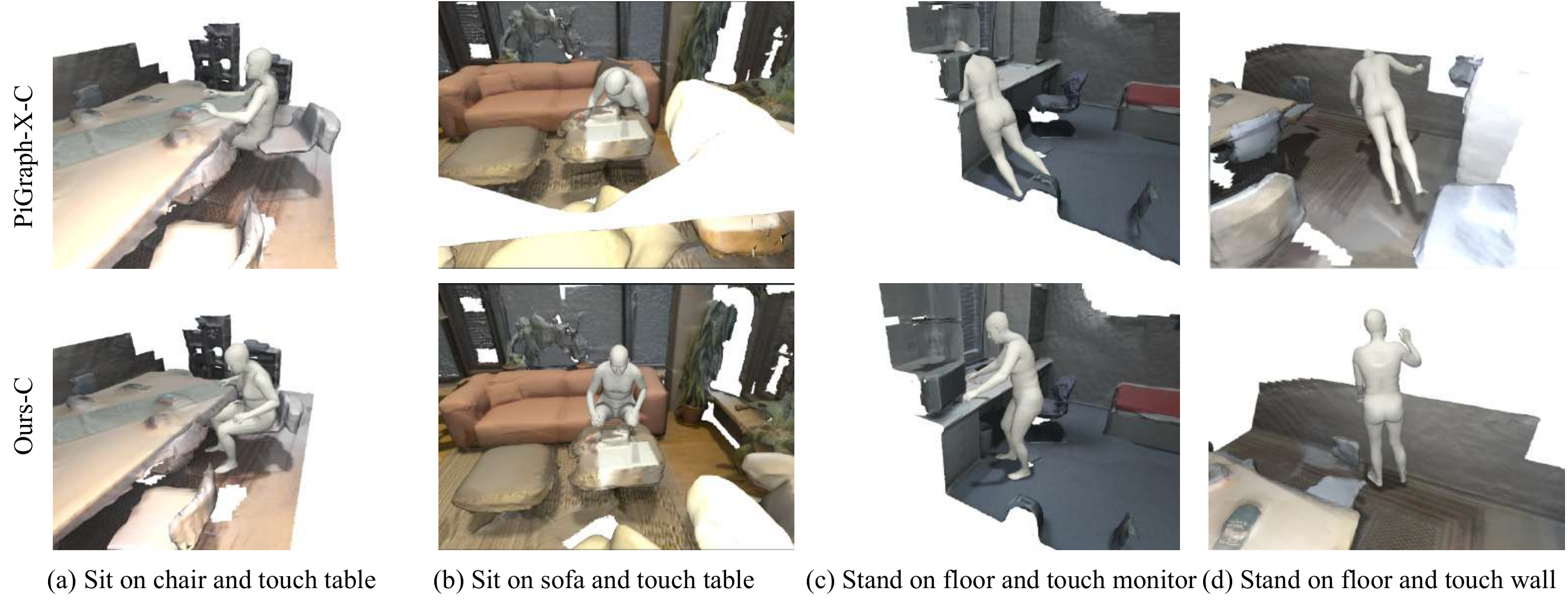}

    \caption{Qualitative comparison of composite interactions generated using ours-C and PiGraph-X-C. Each row shows interactions generated from one method. 
    }
    \label{fig:comparison-novel-composition}
\end{figure}

We conduct the same evaluation for novel composite interaction generation as introduced in \cref{sec:exp_semantic}.
We compare our method to the PiGraph-X-C extension for compositional interaction generation.
POSA-I is not included in this part since it can not be easily extended for semantic composition.
In the forced-alternative-choice (binary-choice) perceptual study (\cref{tab:binary} (b)), our method significantly outperforms the baseline. 
The semantic accuracy, semantic contact, non-collision and diversity comparisons are shown in \cref{tab:composition_metrics}. Our method achieves significantly better semantic accuracy, physical plausibility, and sample efficiency. 
The slight inferior semantic contact score is because the combinations of actions and objects in the test scenes may not always afford natural contact, like touching an incomplete wall in \cref{fig:comparison-novel-composition} (d). Touching the wall and floor simultaneously generates higher contact scores but perceptually unnatural poses to humans.

We refer to the Supp.~Mat.~for \textbf{more qualitative results, failures, ablation study, and limitations}.

\begin{table}[t]
    \caption{Evaluation of semantics, physical plausibility and diversity  for novel compositional interaction generation.}
    \centering
    \resizebox{\textwidth}{!}
    {
        \begin{tabular}{@{\extracolsep{4pt}}lccccccc}
        \hline
       & Semantic Accuracy  & Semantic Contact  & Non-Collision & Entropy & Cluster Size & Samples \\
        \hline
        PiGraph-X-C  & 3.88 $\pm$ \textbf{1.15} & \textbf{0.80}  & 0.84 & 3.72 & \textbf{1.09} & 25K\\
        Ours-C & \textbf{3.95} $\pm$ 1.18 & 0.76 & \textbf{0.98} & \textbf{3.80} & 0.68 & \textbf{1} \\
        
        \hline
        \end{tabular}
    }

    \label{tab:composition_metrics}
\end{table}

\section{Conclusion}
In this paper, we introduce a method to generate humans interacting with 3D scenes given interaction semantics as action-object pairs. We propose a compositional interaction representation and transformer-based VAE models to stochastically generate human-scene interactions. We further explored generating composite interactions by composing atomic interactions, which does not demand training on composite interaction data. Our method is a step towards creating virtual avatars interacting with 3D scenes. \newline
\textbf{Acknowledgements:} We sincerely acknowledge the anonymous reviewers for their insightful suggestions. We thank Francis Engelmann for help with scene segmentation and proofreading, and Siwei Zhang for providing body fitting results.
This work was supported by the SNF grant 200021 204840.

\clearpage
%
%
\bibliographystyle{splncs04}
\bibliography{egbib}
\clearpage

\title{Compositional Human-Scene Interaction Synthesis with Semantic Control}
\titlerunning{Compositional Human-Scene Interaction Synthesis with Semantic Control}
%
\author{Kaifeng Zhao\inst{1} \and
Shaofei Wang\inst{1} \and
Yan Zhang\inst{1} \and
Thabo Beeler \inst{2} \and
Siyu Tang \inst{1}
}
\authorrunning{K. Zhao et al.}
%
\institute{ETH Zürich \\
\email{\{kaifeng.zhao, shaofei.wang, yan.zhang, siyu.tang\}@inf.ethz.ch} 
\and Google \\
\email{thabo.beeler@gmail.com}
}
\appendix
\renewcommand{\thefigure}{S\arabic{figure}} 
\maketitle
In the supplementary, we first provide method details, including architecture illustrations, training losses, and compositional body generation in Sec.~\ref{sec:method_detail}.
We then elaborate on experiment details in Sec.~\ref{sec:expr_detail}.
In Sec.~\ref{sec:ablation}, we present ablation studies on different body representations and the two-stage generation framework.
In Sec.~\ref{sec:more_results}, we show more qualitative results and discuss typical failure cases and limitations.

\section{Method Details}
\label{sec:method_detail}
\subsection{Architecture Details}
\subsubsection{PelvisVAE.}
\begin{figure}[t]
    \centering
    \includegraphics[width=\textwidth]{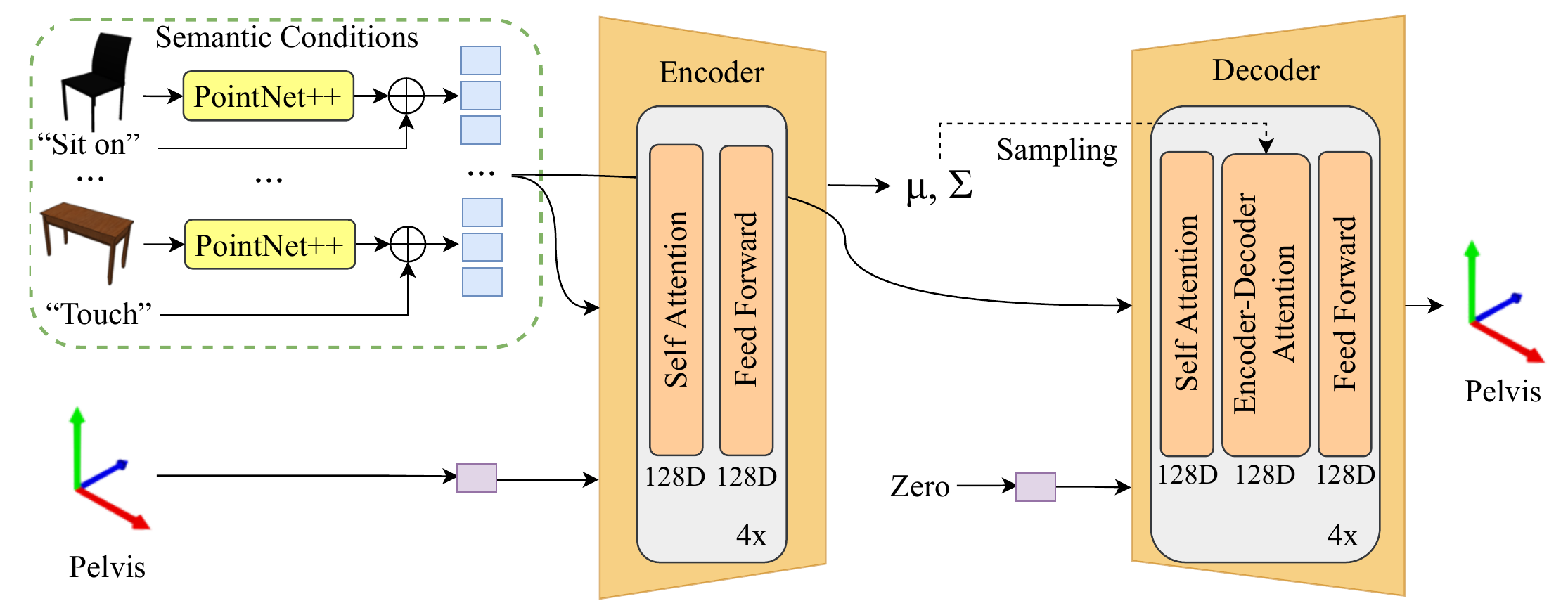}
    
    \caption{Detailed architecture of PelvisVAE.  }
    \label{fig:detail_vaes}
\end{figure}

We illustrate the detailed architecture of PelvisVAE in \cref{fig:detail_vaes}.
The PelvisVAE encoder and decoder use a stack of 2 transformer layers with an embedding dimension of 64.
PelvisVAE represents a human as a pelvis frame of location and orientation. At the decoder, PelvisVAE takes a zero vector as the body token. 

\subsubsection{PointNet++.}
\label{sec:pointnet2}
\begin{figure}[t]
    \centering
    \includegraphics[width=\textwidth]{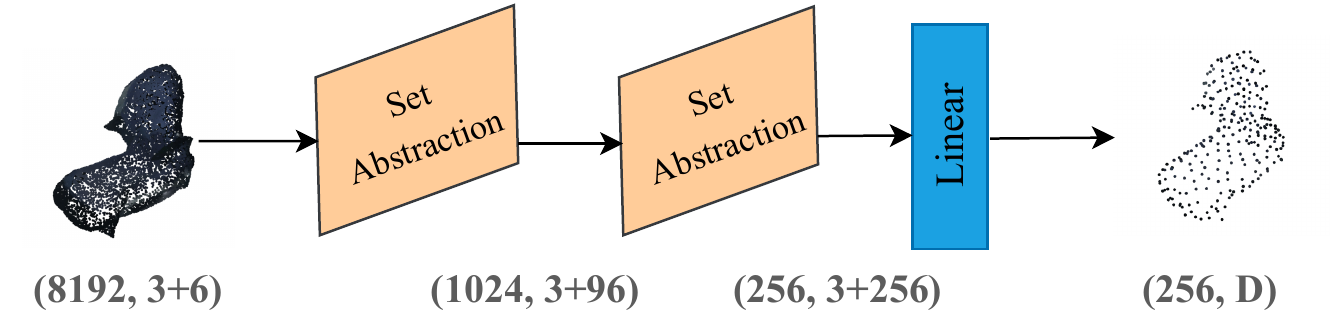}
    \caption{The architecture of the PointNet++ object encoder, where D denotes the embedding dimension used by transformers. }
    \label{fig:detail_pointnet2}
\end{figure}

We train a PointNet++ \cite{qi2017pointnetplusplus} network to extract sparse key points from point clouds, which reduces the number of object nodes to a suitable level for transformers.
We show the architecture of the PointNet++ object encoder in \cref{fig:detail_pointnet2}.
The PointNet++ module comprises two set abstraction layers to extract 256 key points for each object. 
The output points features are projected to vectors of dimension 128 for BodyVAE and dimension 64 for PelvisVAE using a linear layer. 

\subsubsection{SMPL-X Regressor}

\begin{figure}[t]
    \centering
    \includegraphics[width=\textwidth]{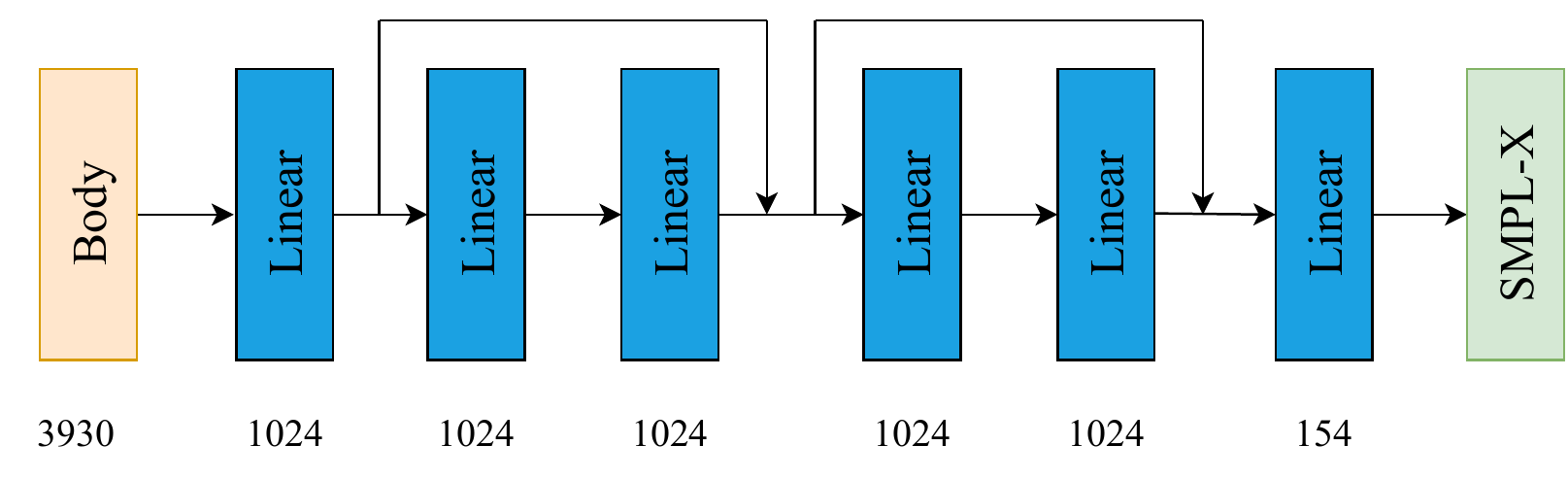}
    \caption{The architecture of the SMPL-X regressor. }
    \label{fig:detail_regressor}
\end{figure}

We train a multi-layer perceptron (MLP) to regress SMPL-X parameters from body meshes as shown in \cref{fig:detail_regressor}. The MLP has six linear layers, and we employ residual connections. The MLP outputs the SMPL-X body poses using the 6D continuous representation \cite{zhou2019continuity}, body shape parameters, and the first six hand pose PCA components for each hand. 
Following \cite{kolotouros2019convolutional}, We detach the gradients of the reconstructed body from the computational graph and use the concatenation of the reconstructed body and a template T-pose body as inputs to the MLP.

\subsection{Training Loss}
Our BodyVAE is trained to minimize the $\mathcal{L}_{body}$ loss:

\begin{equation}
\mathcal{L}_{body} =  \mathcal{L}_{interaction} +  \mathcal{L}_{mesh} + \mathcal{L}_{feature} + \mathcal{L}_{KL} + \mathcal{L}_{regress},
\end{equation}
where the terms represent the interaction loss, body mesh reconstruction loss, contact feature reconstruction loss, the Kullback-Leibler divergence, and the SMPL-X regression loss. Weights for each term are omitted for simplicity.
The term $\mathcal{L}_{interaction}$ encourages the vertices with predicted positive contact feature to have zero distance to the input objects:
\begin{equation}
\mathcal{L}_{interaction} = \sum_{i=1}^{655} \hat{c}^i \cdot \min_{v_o \in V_o}||\hat{v}^i - v_o ||_2,
\label{equ:interaction}
\end{equation}
where $\hat{c}^i$ and $\hat{v}^i$ denote the predicted contact feature and location of body vertex $i$ respectively, and $V_o=\bigcup\limits_{i=1}^{M} o^i$ denotes the set of all points of input interaction objects.

The body mesh reconstruction loss consists of the vertex coordinate loss $\mathcal{L}_{vertex}$, the surface normal loss $\mathcal{L}_{normal}$, the edge length loss $\mathcal{L}_{edge}$ following \cite{wang2018pixel2mesh, Choi_2020_ECCV_Pose2Mesh}, and the normal consistency loss $\mathcal{L}_{consistency}$ that regularizes the normal of adjacent faces to change smoothly, which helps in particular with details of the hands. The loss terms are defined by:
\begin{equation}
\mathcal{L}_{mesh} = \mathcal{L}_{vertex} +  \mathcal{L}_{normal} + \mathcal{L}_{edge}  + \mathcal{L}_{consistency}.
\end{equation}

\begin{equation}
\mathcal{L}_{vertex} = ||\hat{V} - V||_1,
\end{equation}

\begin{equation}
\mathcal{L}_{normal} = \sum_{f \in F} \sum_{(i,j)\in f}\left| \langle n_f,  \frac{\hat{v^i} - \hat{v^j}}{||\hat{v^i}-\hat{v^j}||_2}\rangle\right|,
\end{equation}

\begin{equation}
\mathcal{L}_{edge} = \sum_{f \in F} \sum_{(i,j)\in f} \left| 1 - \frac{||\hat{v^i} - \hat{v^j}||_2} {||v^i - v^j||_2}  \right|,
\end{equation}

\begin{equation}
\mathcal{L}_{consistency} = \sum_{f^i, f^j \in F, f^i \cap f^j \neq \emptyset} 1-\langle\hat{n}_{f^i}, \hat{n}_{f^j}\rangle,
\end{equation}
where $V$, $F$ denote the vertices and faces of input body mesh, $n_{f}$ denotes the normal of triangle $f \in F$ and $\hat{(\cdot)}$ denotes the corresponding reconstructions.

The contact feature reconstruction loss is calculated as the binary cross-entropy loss (BCE) between reconstructed $\hat{C}$ and ground truth $C$ contact features:

\begin{equation}
\mathcal{L}_{feature} = BCE(\hat{C}, C).
\end{equation}
We use the robust Kullback-Leibler divergence (KL) \cite{zhang2020perpetual,zhang2021we} to avoid posterior collapse:
\begin{equation}
\label{equ:kl}
\mathcal{L}_{KL} = \Psi(KL(q(z|\mathcal{I})||\mathcal{N}(0, I))),
\end{equation}
where $\Psi$ is the Charbonnier function \cite{charbonnier1994two} $\Psi(s)=\sqrt{1+s^2}-1$.
The SMPL-X parameter regression loss $\mathcal{L}_{regress}$ comprises parameter error and vertex error:
\begin{equation}
	\mathcal{L}_{regress} = ||\theta - \hat{\theta}||_2 + |\mathcal{M}(\theta) -  \mathcal{M}(\hat{\theta})|,
\end{equation}
where $\theta$ and $\hat{\theta}$ are GT and predicted SMPL-X body parameters respectively, and $\mathcal{M}$ denotes the SMPL-X model mapping parameters to mesh vertices.

The PelvisVAE is trained with the following losses:
\begin{equation}
\mathcal{L}_{pelvis} = \mathcal{L}_{transl} + \mathcal{L}_{orient} + \mathcal{L}_{KL},
\end{equation}
where $\mathcal{L}_{transl}$ and $\mathcal{L}_{orient}$ denote the reconstruction loss of pelvis joint location and orientation, and $\mathcal{L}_{KL}$ is the robust Kullback-Leibler divergence loss from \cref{equ:kl}.

\subsection{Interaction-Based Optimization.}
We use the sampled SMPL-X parameters as initialization and optimize body translation $t$, global orientation $R$ and pose $\theta$. The optimization objective is given by:

\begin{equation}
E(t, R, \theta) = \mathcal{L}_{interaction} + \mathcal{L}_{coll} + \mathcal{L}_{reg},
\end{equation}
The interaction term $\mathcal{L}_{interaction}$ is defined in \cref{equ:interaction}. 
The scene collision term is defined as $\mathcal{L}_{coll}=\sum_{i=1}^{655}\Psi(v^i)$, with $\Psi(v^i)$ denoting the signed distance of vertex i to the scene. For computational efficiency, we use a precomputed SDF grid for each scene and use interpolation to get SDF value for body vertices.
The regularization term $\mathcal{L}_{reg}=|t-t_{init}| + |R - R_{init}| + ||\theta - \theta_{init}||_2$ penalizes $t, R$ and $\theta$ deviating from their initialization.

\subsection{Implementation Details}
Our implementation is based on PyTorch \cite{paszke2019pytorch}. We use the Adam optimizer \cite{diederik2014adam} with the learning rate 3e-4 and batch size of 8 for training all models.
For PelvisVAE, we use weights of 3, 1, 1 for $\{\mathcal{L}_{transl}, \mathcal{L}_{orient}, \mathcal{L}_{KL}\}$. For BodyVAE, we use weights of {1, 1, 0.1, 0.2, 0.05} for $\{\mathcal{L}_{interaction}$, $\mathcal{L}_{vertex}$, $\mathcal{L}_{normal}$, $\mathcal{L}_{edge}$, $\mathcal{L}_{consistency}\}$. We use weight 1 for $\mathcal{L}_{KL}$ and apply the weight annealing scheme \cite{bowman2015generating}. For SMPL-X regressor, we use weights of 1 for vertex and body pose reconstruction and 0.1 for shape parameter and hand PCA components reconstruction. 
our contact features used a threshold object distance of 5cm. 
We use latent dimensions of 6 and 128 for PelvisVAE and BodyVAE, respectively.

For the interaction-based optimization, we respectively use weights of 1 for interaction term $\mathcal{L}_{interaction}$, 32 for collision term $\mathcal{L}_{coll}$, 0.1 for translation and orientation regularization and 32 for pose regularization.

For composite pelvis sampling,  we use the Adam optimizer with a learning rate of 0.1 and 100 optimization steps. We scale the sum of log probability of latent codes with the weight of 0.05 to balance the influence of pelvis frame difference and probability.

\section{Experiment Details}
\label{sec:expr_detail}
\subsection{Dataset Collection }

We extend the PROX-S dataset based on PROX \cite{PROX:2019}, which contains 3D reconstructions of 12 static scenes and RGB-D recordings of natural human-scene interactions captured using a Kinect sensor. 
The pseudo-ground truth body fittings of PROX recordings are estimated using \cite{PROX:2019, Zhang:ICCV:2021}.
To obtain object instance segmentation and interaction semantics,  we further process the PROX dataset to get: (1) 3D instance segmentation in all the PROX scenes and (2) per-frame interaction semantic labels in the form of action-object pairs.

\begin{figure}[t]
    \captionsetup[subfigure]{labelformat=empty}
	\centering
    \begin{subfigure}[t]{0.45\textwidth}
        \includegraphics[width=\textwidth]{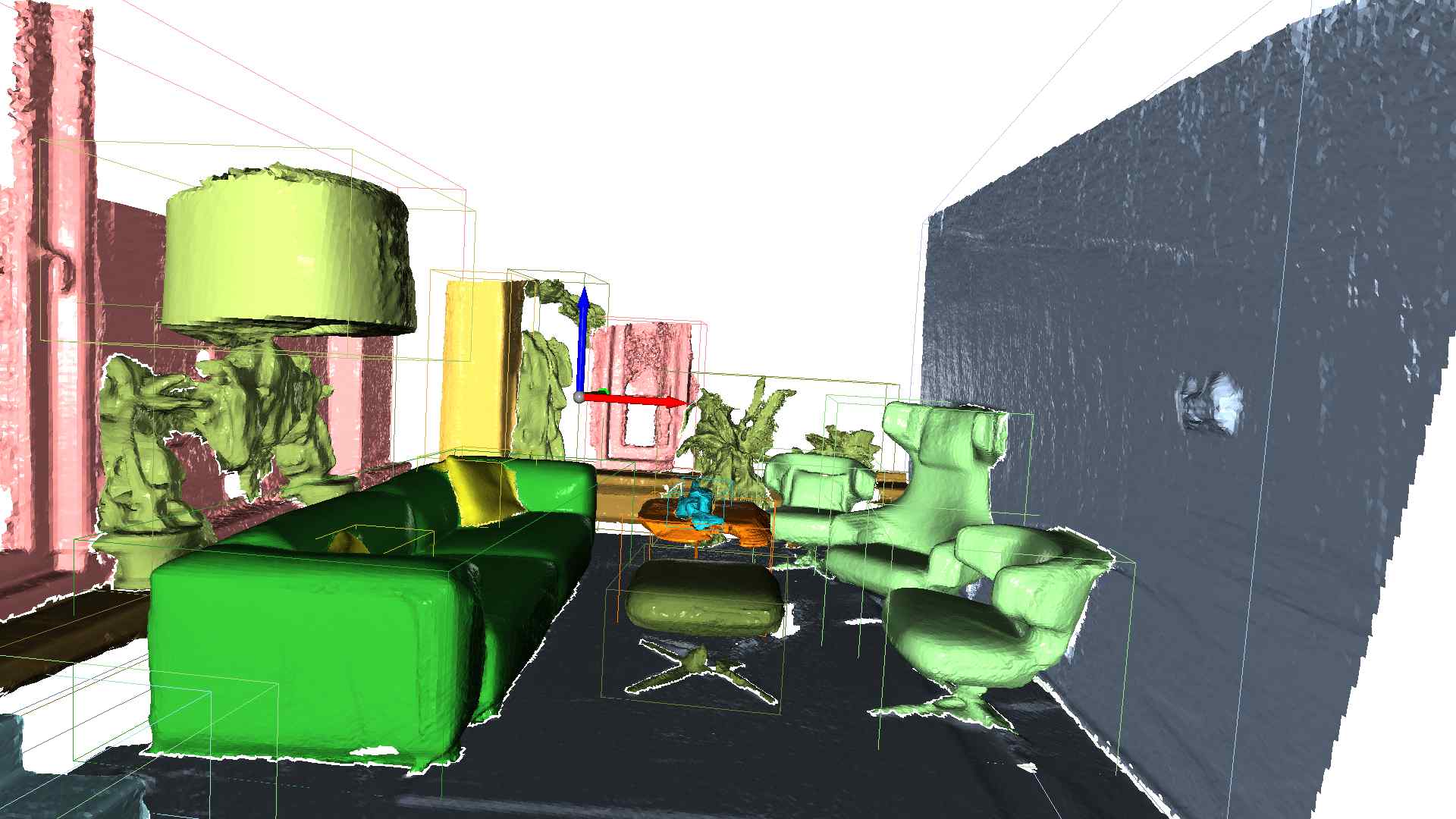}    
        \caption{N3OpenArea}
    \end{subfigure}
    \begin{subfigure}[t]{0.45\textwidth}
        \includegraphics[width=\textwidth]{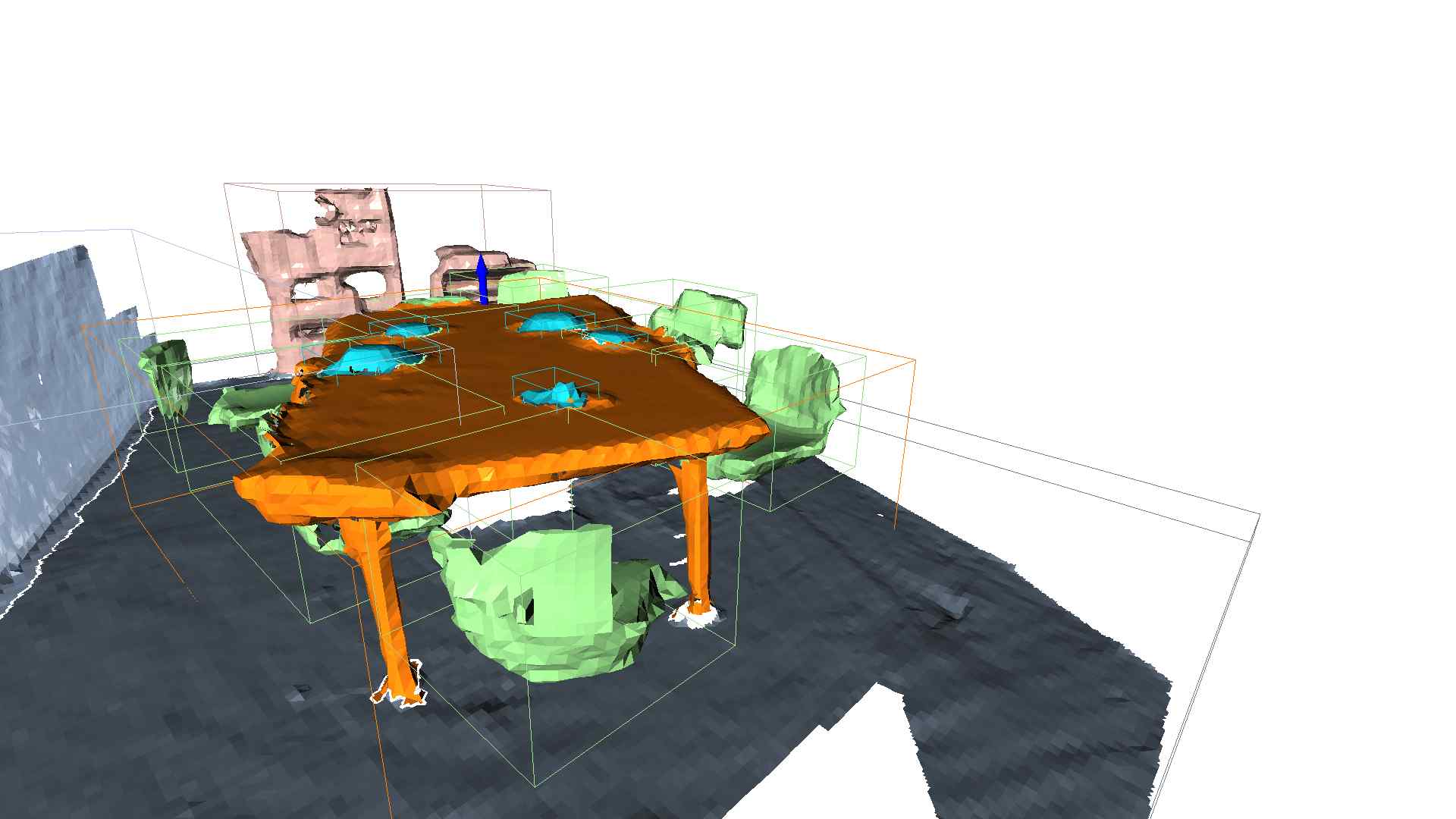}    
        \caption{MPH1Library}
    \end{subfigure}
    \begin{subfigure}[t]{0.45\textwidth}
        \includegraphics[width=\textwidth]{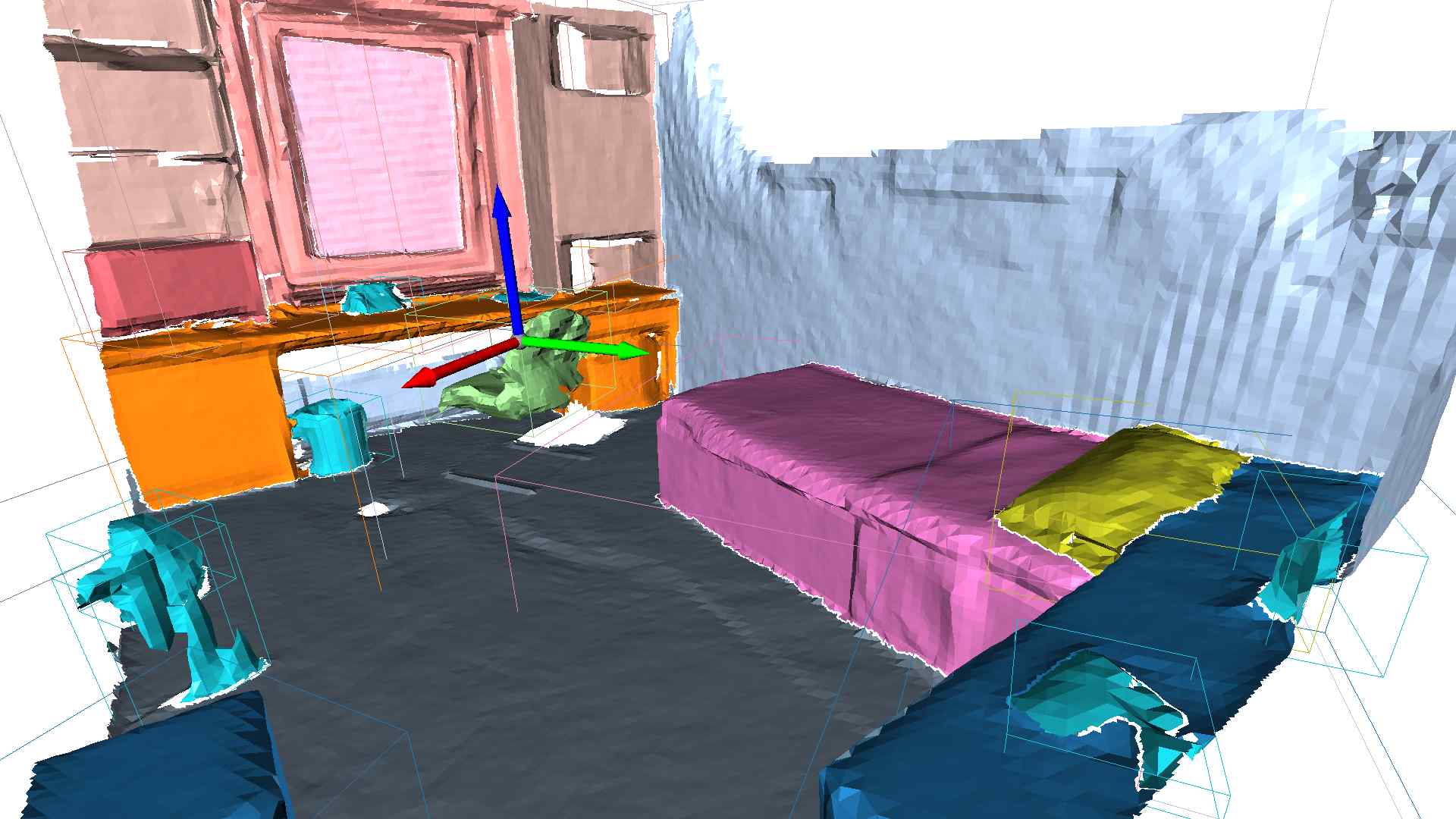}    
        \caption{MPH16}
    \end{subfigure}
    \begin{subfigure}[t]{0.45\textwidth}
        \includegraphics[width=\textwidth]{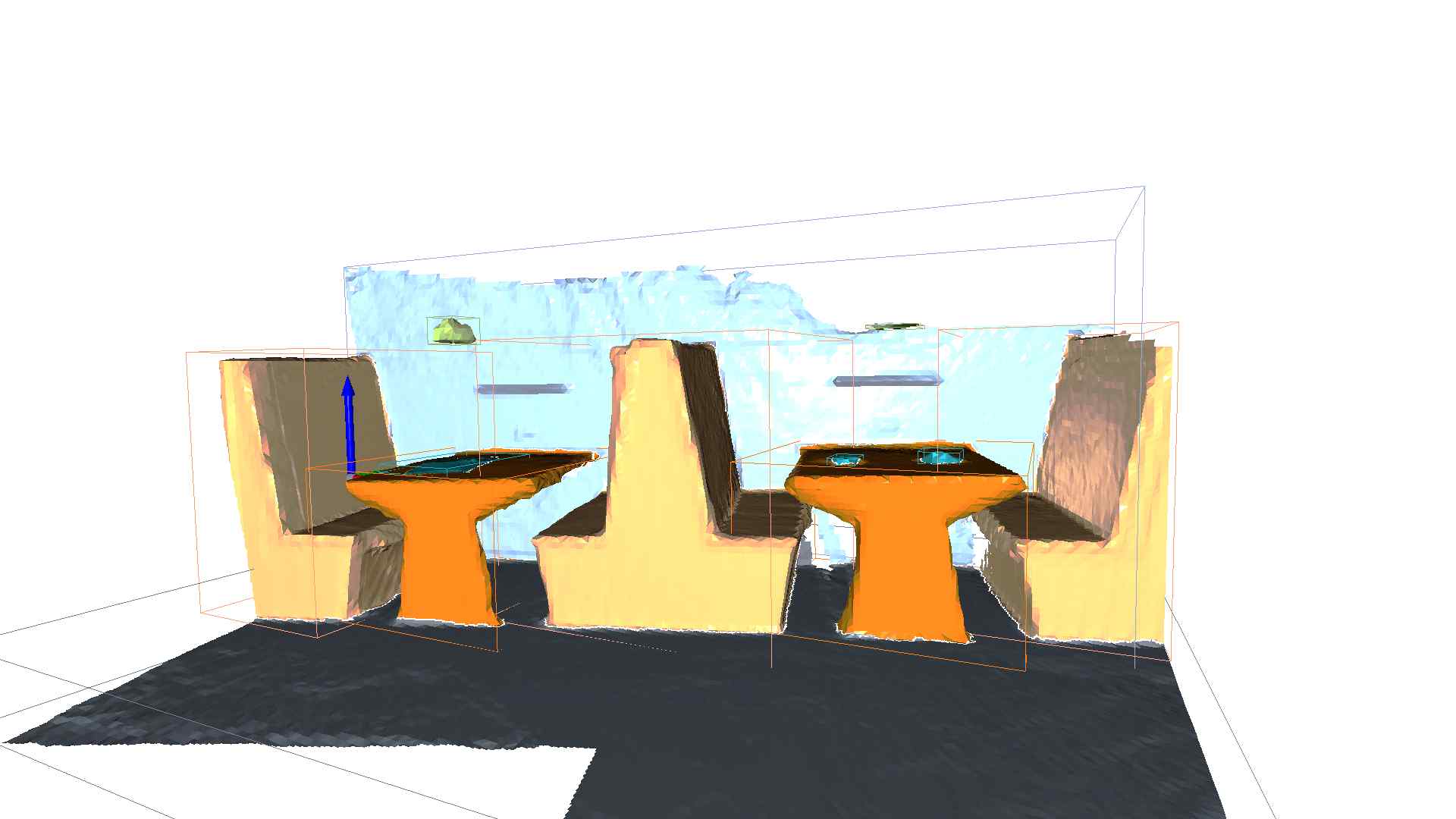}    
        \caption{N0SittingBooth}
    \end{subfigure}
    
    \caption{Visualization of instance segmentation results in the 4 test scenes. The objects are colored according to their semantic categories.}
    \label{fig:instance}
\end{figure}

We first conduct instance segmentation for the 12 scenes based on the scene semantic annotation provided in the PROX-E \cite{PSI:2019} dataset. 
Specifically, we split the scene into multiple possible over-segmented instances using connected components analysis.
Then we manually annotate the instances in the scenes. 
The instance segmentation results of the 4 test scenes are visualized in ~\cref{fig:instance}.

\begin{figure}[t]
	\centering
    \includegraphics[width=\textwidth]{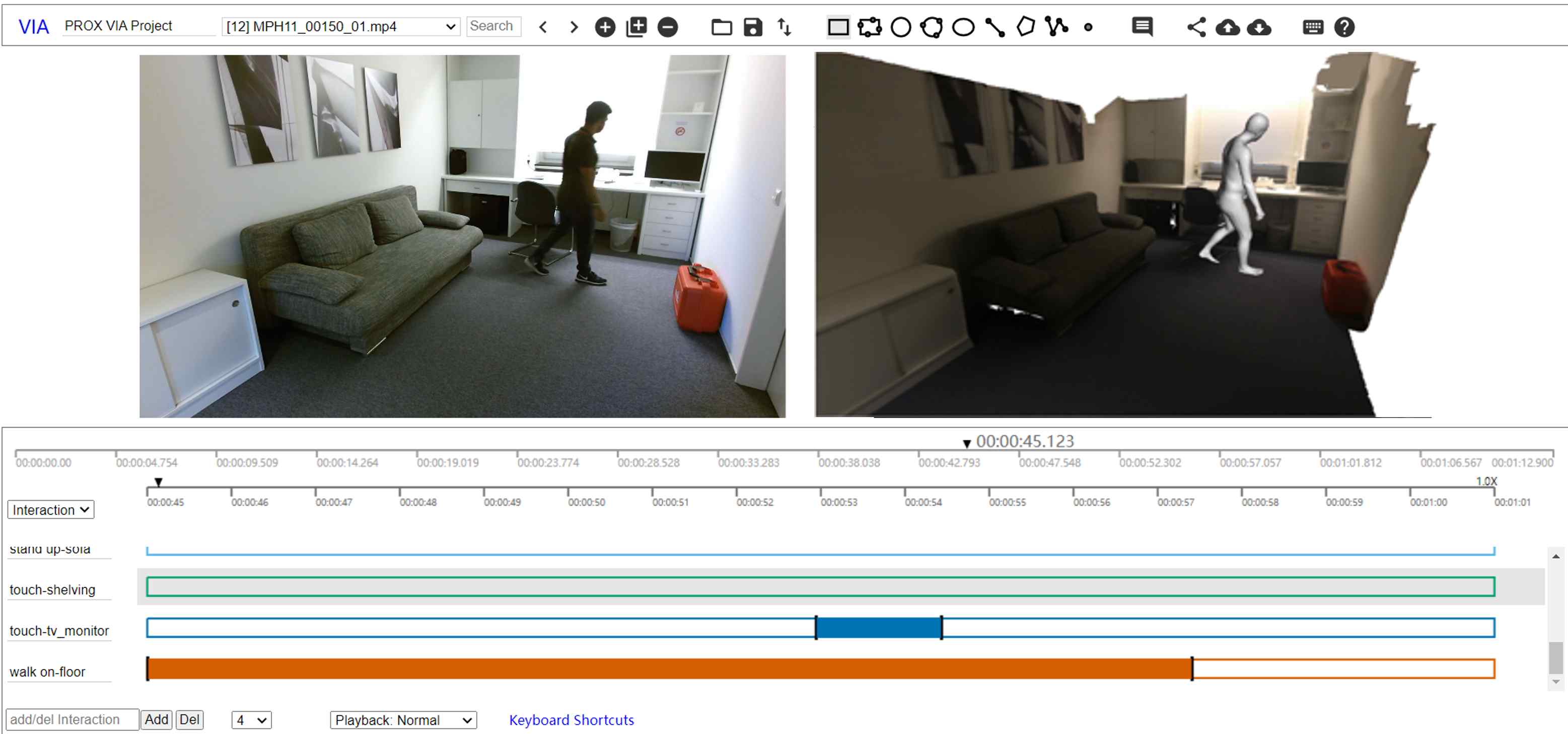}
    \caption{Our annotation tool for annotating interaction semantics. RGB recordings and the visualization of SMPL-X fitting are displayed side-by-side. Annotators are instructed to label the intervals containing interactions using action-noun pairs.}
    \label{fig:interface}
\end{figure}

To obtain interaction semantics, we densely annotated the PROX dataset using the VIA video annotation tool \cite{dutta2019vgg}. The annotators are asked to label all intervals containing interactions specified by the action-noun pairs.  
The annotation tool is illustrated in \cref{fig:interface}.
Note that multiple interaction labels can exist in a single frame if the human interacts with multiple objects. 
%
We further retrieve interaction object instance ID using the scene segmentation and object category annotation. When there is more than one object instance of the annotated category in one scene, we assign the object label to the instance with the closest distance to action-related body parts.

The PROX-S dataset contains around 32K frames of human-scene interactions from 43 sequences recorded in 12 indoor scenes. 
We follow the dataset split in \cite{PSI:2019} to use ‘MPH16’, ‘MPH1Library’, ‘N0SittingBooth’, and ‘N3OpenArea’ as test scenes and the remaining eight scenes for training. 
The training data comprises a total of 17 different actions and 42 categories of interactions defined as action-object pairs. 
We evaluate interaction synthesis with semantic control on about 150 different combinations of action and object instances in the four test scenes. 

\subsection{Perceptual Study}
\begin{figure}[t]
    \centering
    \begin{subfigure}[t]{0.8\textwidth}
        \includegraphics[width=\textwidth]{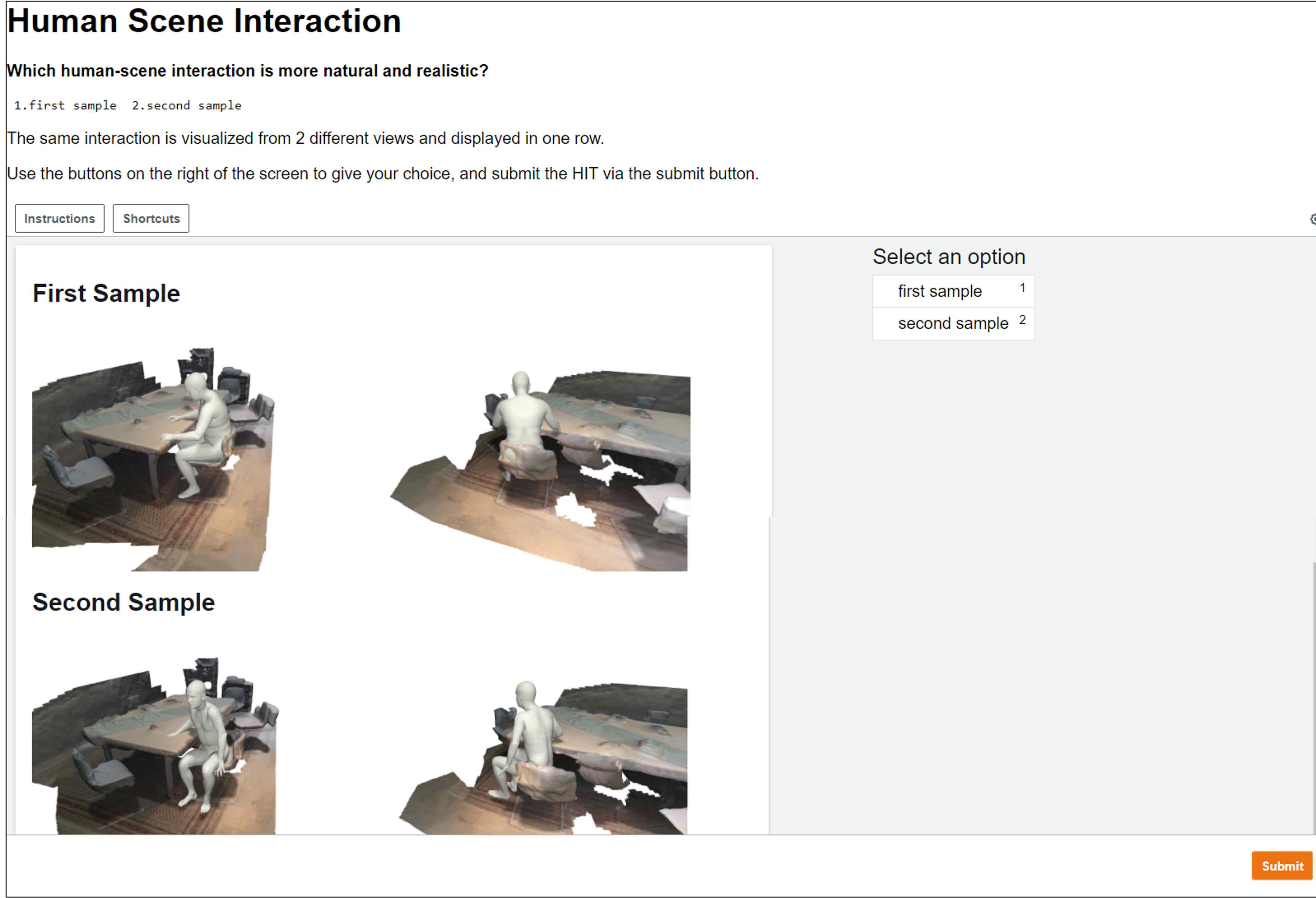}
        \caption{Binary perceptual study}
    \end{subfigure}
    \begin{subfigure}[t]{0.8\textwidth}
        \includegraphics[width=\textwidth]{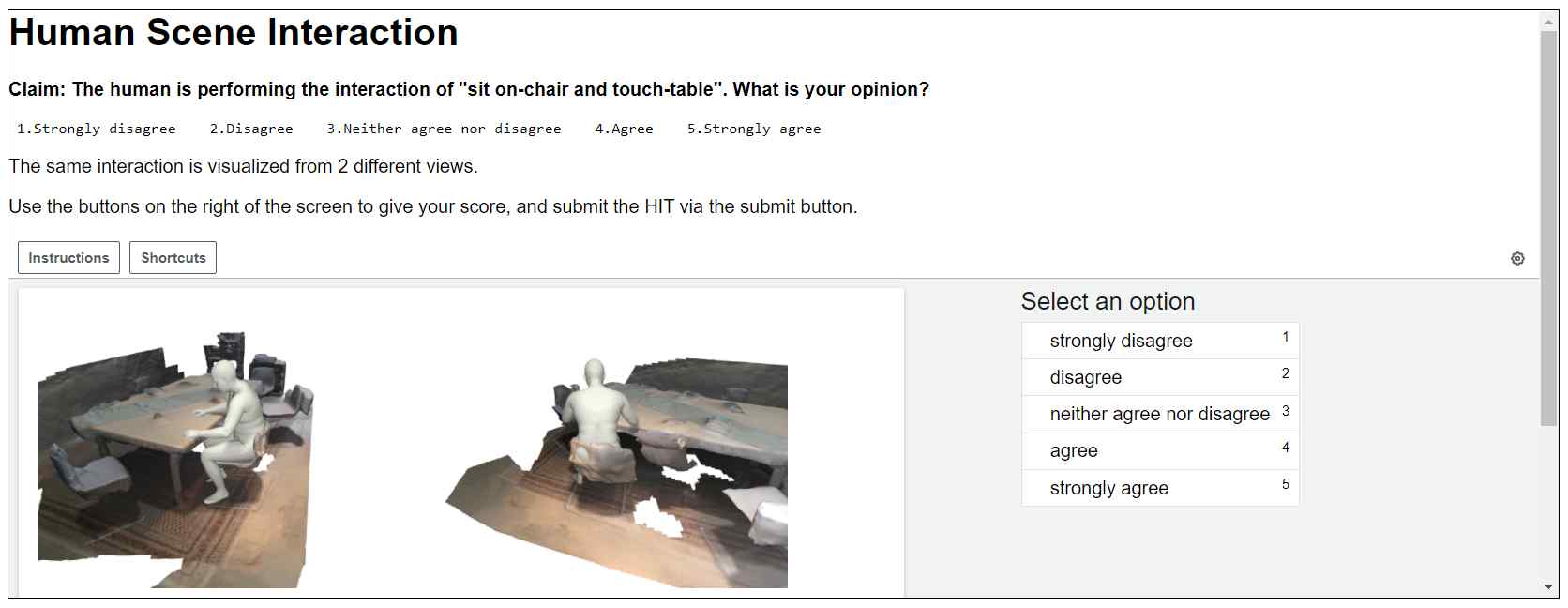}
        \caption{Unary perceptual study}
    \end{subfigure}

    \caption{AMT User interfaces for the perceptual studies.}
    \label{fig:amt}
\end{figure}

We conduct binary perceptual studies to evaluate the interaction naturalness and unary perceptual studies to evaluate the semantic accuracy of the generated interactions on Amazon Mechanical Turk (AMT). The AMT interfaces of the two perceptual studies are illustrated in \cref{fig:amt}.
For the binary perceptual studies, we uniformly select 160 samples from PROX pseudo ground truth with varying semantic labels and respectively generate 160 random samples with the same semantic labels using our method and two baselines. We render each interaction with two different views and compare out method against the two baselines and pseudo ground truth. 
During the study, participants are instructed to select one sample that they think is more realistic from two samples generated with different methods. 
Each comparison task is distributed to three participants for evaluation.

For the unary perceptual studies, we sample and render one interaction for the 155 combinations of actions and objects in our test scenes. 
These interaction samples are shown to the participants with the interaction semantic labels and the participants are instructed to rate the semantic accuracy from 1 (strongly disagree) to 5 (strongly agree).

\section{Ablation Study}
\label{sec:ablation}
We investigate the influence of body representations and the two-stage generation design.

We compare three body representations of joint locations (JL), joints locations and orientations (JLO), and mesh vertices locations (VL) in BodyVAE by evaluating the semantic contact score of sampled interactions without optimization, and the consistency error between generated body and corresponding SMPL-X body defined as:
\begin{equation}
	\mathcal{L}_{body} = |\mathbf{B} -  \mathcal{M}(\theta, \beta)|,
\end{equation}
where $\mathbf{B}$ denotes the generated body joints (JL and JLO) or vertices locations(VL), and $\mathcal{M}(\cdot)$ denotes the SMPL-X body model that yields body joints and vertices given body pose $\theta$ and shape parameter $\beta$. We use the pose and shape parameters predicted by the SMPL-X regressor for JL and VL. We do not train a regressor for JLO and directly use the input shape parameter of the template body and the generated joint orientations to pose the SMPL-X body.
\begin{table}[t]
    \caption{Evaluation of body representation choices.}
    \centering
    \begin{tabular}{c c c}
        \hline
           & Semantic Contact $\uparrow$ & Body Consistency (m) $\downarrow$ \\
        \hline
        Vertex Location     & \textbf{0.72} & \textbf{0.01}    \\
        Joint Location      & 0.71 & 0.02    \\
        Joint Location and Rotation      & 0.69 & 0.04    \\
        \hline
    \end{tabular}
    \label{tab:unary}
\end{table}

Directly generating joint locations with orientations in JLO leads to a lower semantic contact score of 0.69 and a significantly worse body consistency error of 0.04m. This is because generating joint locations and orientations together by networks does not directly yield
valid SMPL-X bodies, as the bone lengths defined by the
generated joint locations may not correspond to valid human skeletons. 
Regressing joint rotations from joint or vertex locations leads to better performance and vertex location representation achieves the best semantic contact score of 0.72 and body consistency error of 0.01m.
Our result indicates that regressing SMPL-X body parameters from body mesh vertices is easier than from a skeleton of joints and generates slightly better human-object contact. 

To quantify the importance of the two-stage design in interaction generation, we train two models for one-stage generation which directly predicts the human body given objects in the original scene coordinate system and re-centered scene coordinate system, respectively. The re-centered scene coordinate system translates the origin to the average of interaction object points. 
%
The semantic contact score of interactions generated from the one-stage models using original and re-centered scene coordinates rapidly drops to 0.20 and 0.31 respectively, compared to 0.72 of the two-stage method.
Our result shows that learning to jointly predict global interaction location, orientation, and body pose is much more difficult and the two-stage design is necessary for generating interactions with natural human-scene contact.

\section{More Qualitative Results}
\label{sec:more_results}
\begin{figure}[t]
    \captionsetup[subfigure]{labelformat=empty}
    \centering
    \begin{subfigure}[t]{0.23\textwidth}
         \includegraphics[width=\textwidth]{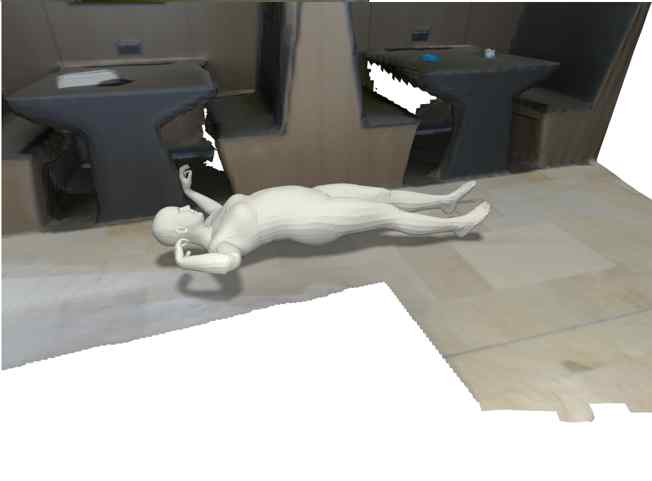}
         \caption{``lie on the floor''}
    \end{subfigure}
    \begin{subfigure}[t]{0.23\textwidth}
         \includegraphics[width=\textwidth]{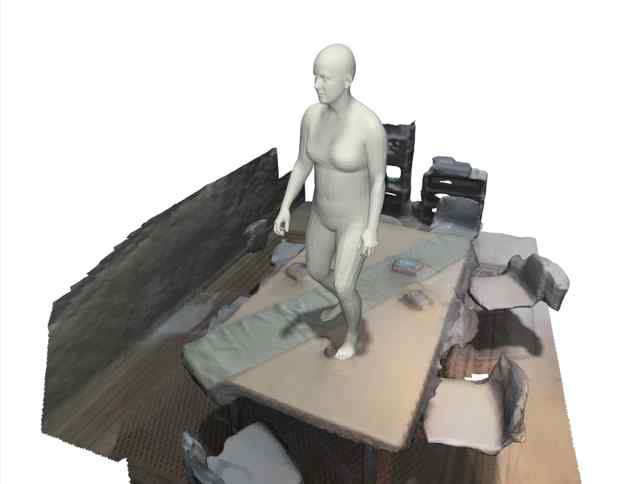}
         \caption{``walk on the table''}
    \end{subfigure}
     \begin{subfigure}[t]{0.23\textwidth}
         \includegraphics[width=\textwidth]{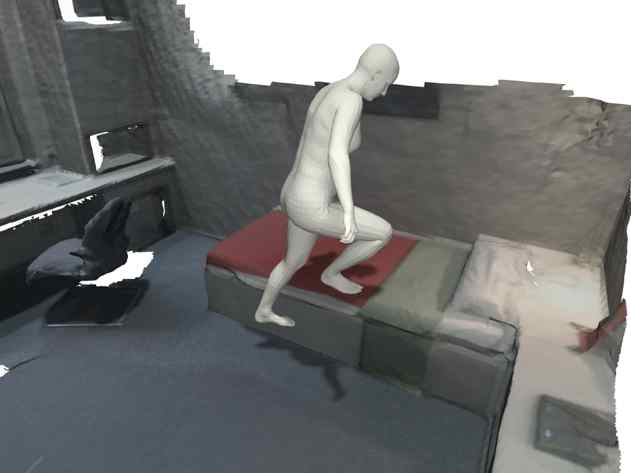}
         \caption{``step up the bed''}
    \end{subfigure}
    \begin{subfigure}[t]{0.23\textwidth}
         \includegraphics[width=\textwidth]{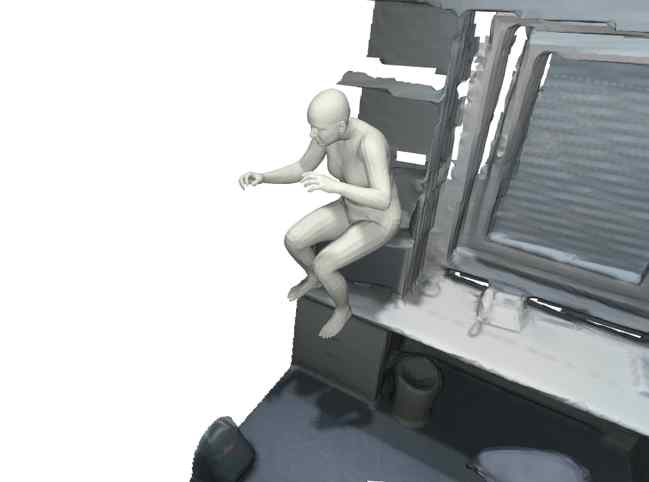}
         \caption{``sit on the monitor''}
    \end{subfigure}
    
    \caption{Novel combinations of actions and objects generated by our method that were not part of the training data.}
    \label{fig:novel}
\end{figure}

\subsection{Retarget Novel Objects.} 
Our method can naturally retarget learned interactions to unseen objects with similar geometry and affordance because we use the point cloud object representation which encodes object shapes, instead of the object category in previous works.
Figure \ref{fig:novel} shows some created novel interactions that are not seen in training.
It demonstrates the generalization capability of our method and the potential for synthesizing interactions with open-set objects beyond predefined object categories.
Moreover, our method also creates some interesting interactions that are less possible in the real world, e.g., sitting on a monitor. 

\subsection{Explicit Body Shape Control.}
\begin{figure}[t]
    \captionsetup[subfigure]{labelformat=empty}
    \centering
    \adjincludegraphics[width=0.22\textwidth, trim={0 {.5\height} 0 0}, clip]{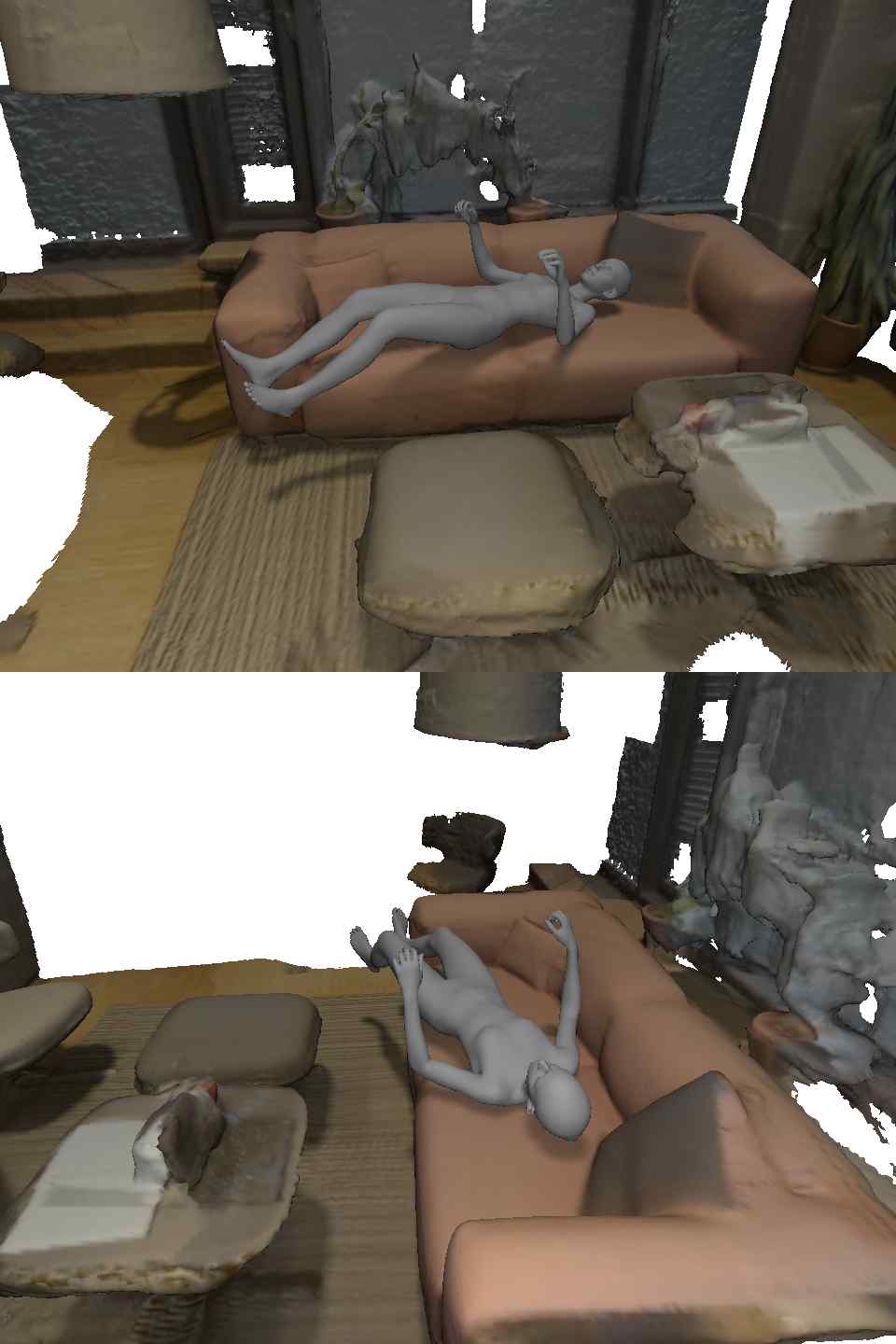}
    \adjincludegraphics[width=0.22\textwidth, trim={0 {.5\height} 0 0}, clip]{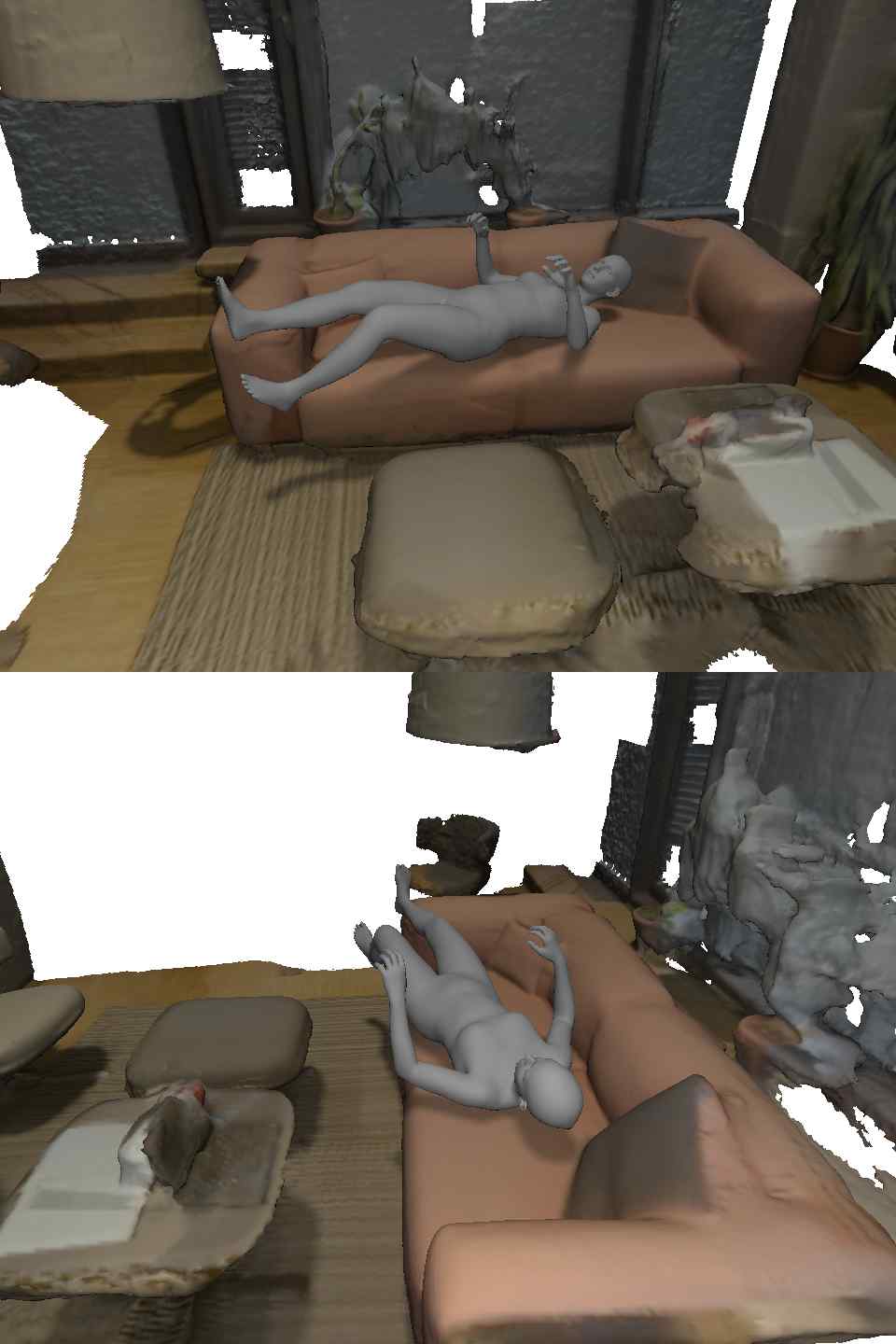}
    \adjincludegraphics[width=0.22\textwidth, trim={0 {.5\height} 0 0}, clip]{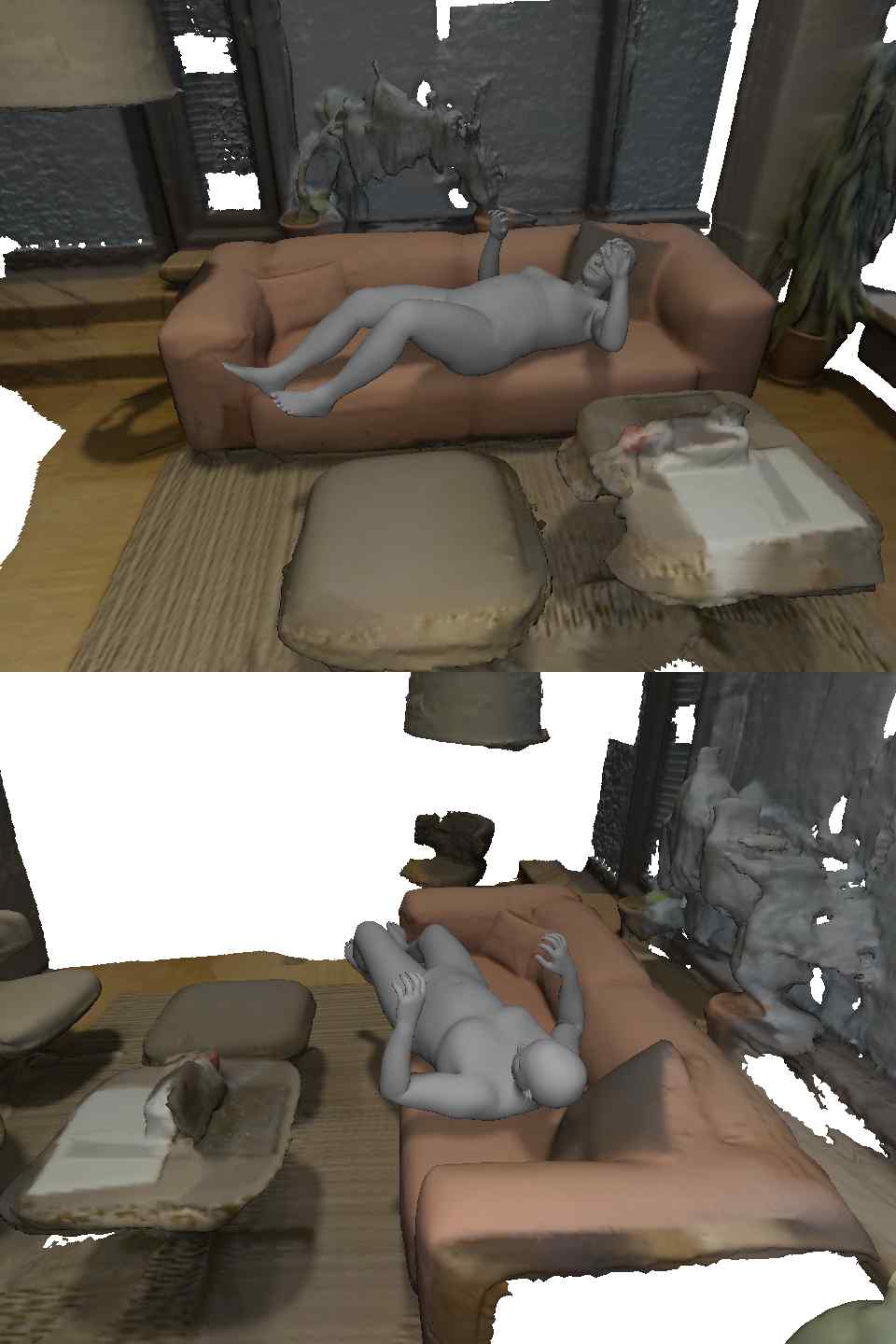}
    \adjincludegraphics[width=0.22\textwidth, trim={0 {.5\height} 0 0}, clip]{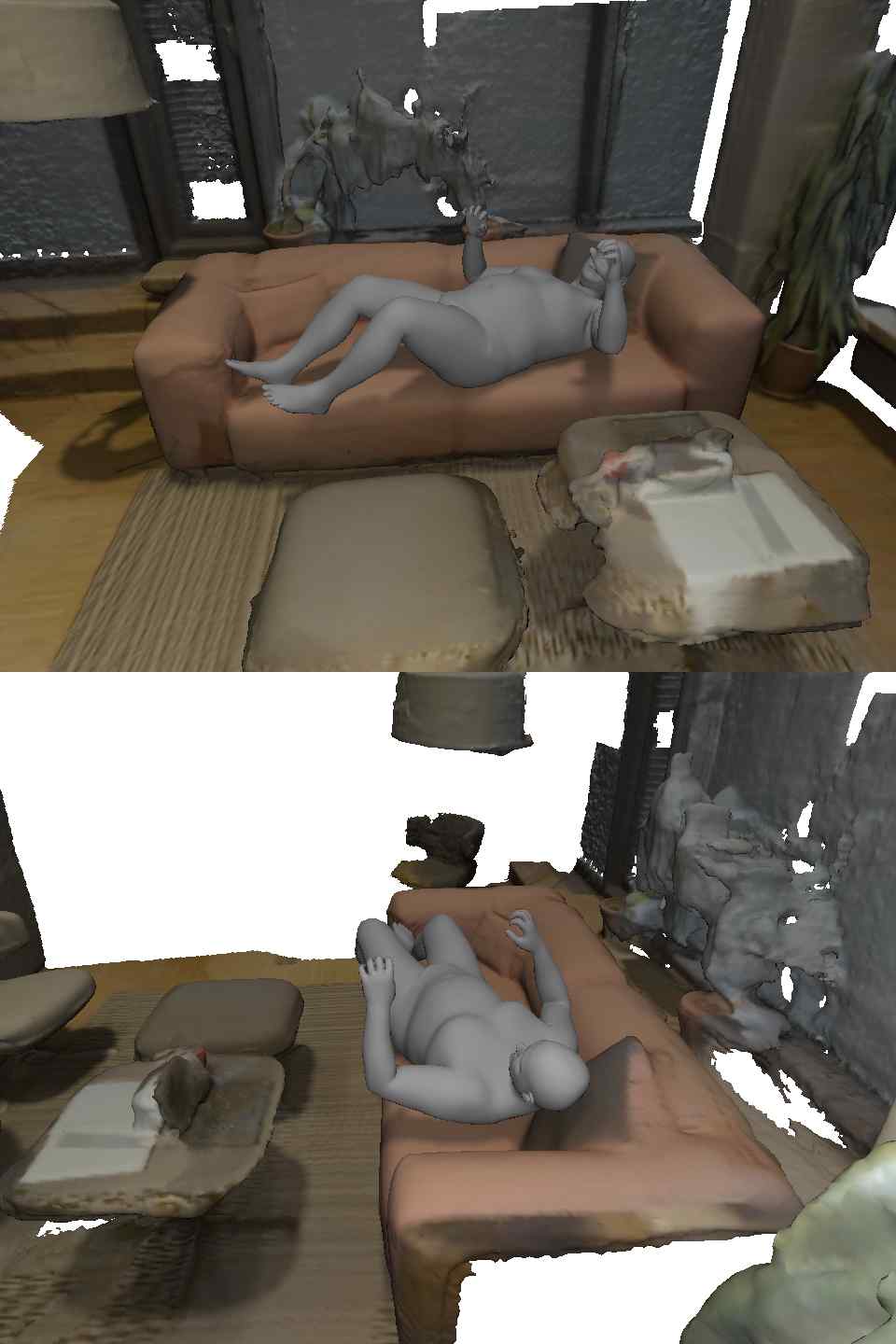}
    
    \caption{Interaction synthesis results with explicitly controlled varying human body shapes, where the extremely thin and heavy bodies are not seen during training.}
    \label{fig:shape}
\end{figure}

Our method features explicit body shape control in interaction generation, which is achieved by using personalized body templates.
Figure \ref{fig:shape} shows interaction synthesis results where we control the SMPL-X body shape parameters changing from -3 to 3. 
Note that the extremely thin and heavy bodies are not seen during training and our method generalizes to such extreme shapes.


\subsection{Random Interaction Samples}
\begin{figure}[t]
    \centering
    \begin{subfigure}[t]{\textwidth}
        \rotatebox{90}{lie on sofa}
        \includegraphics[width=0.15\textwidth]{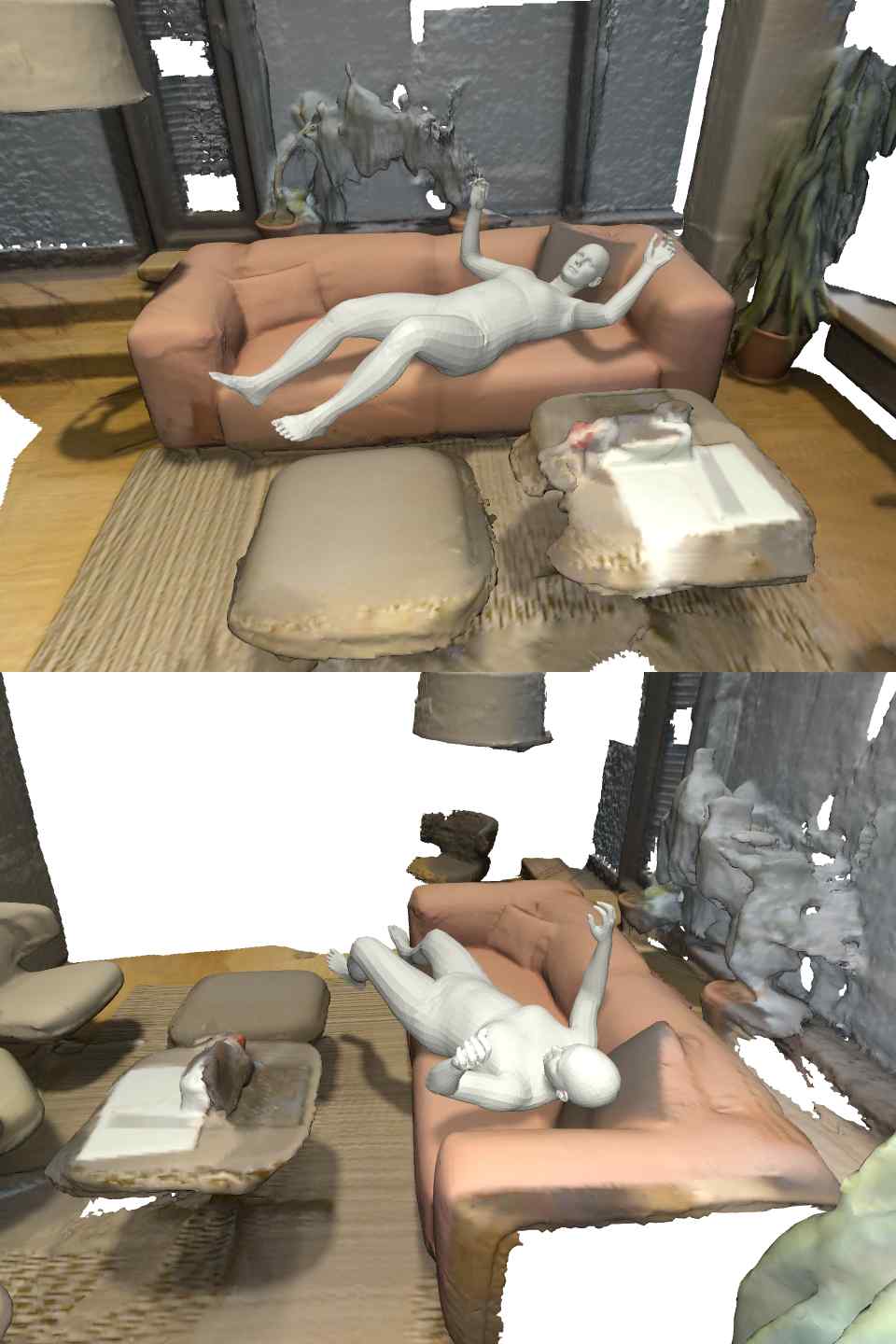}
        \includegraphics[width=0.15\textwidth]{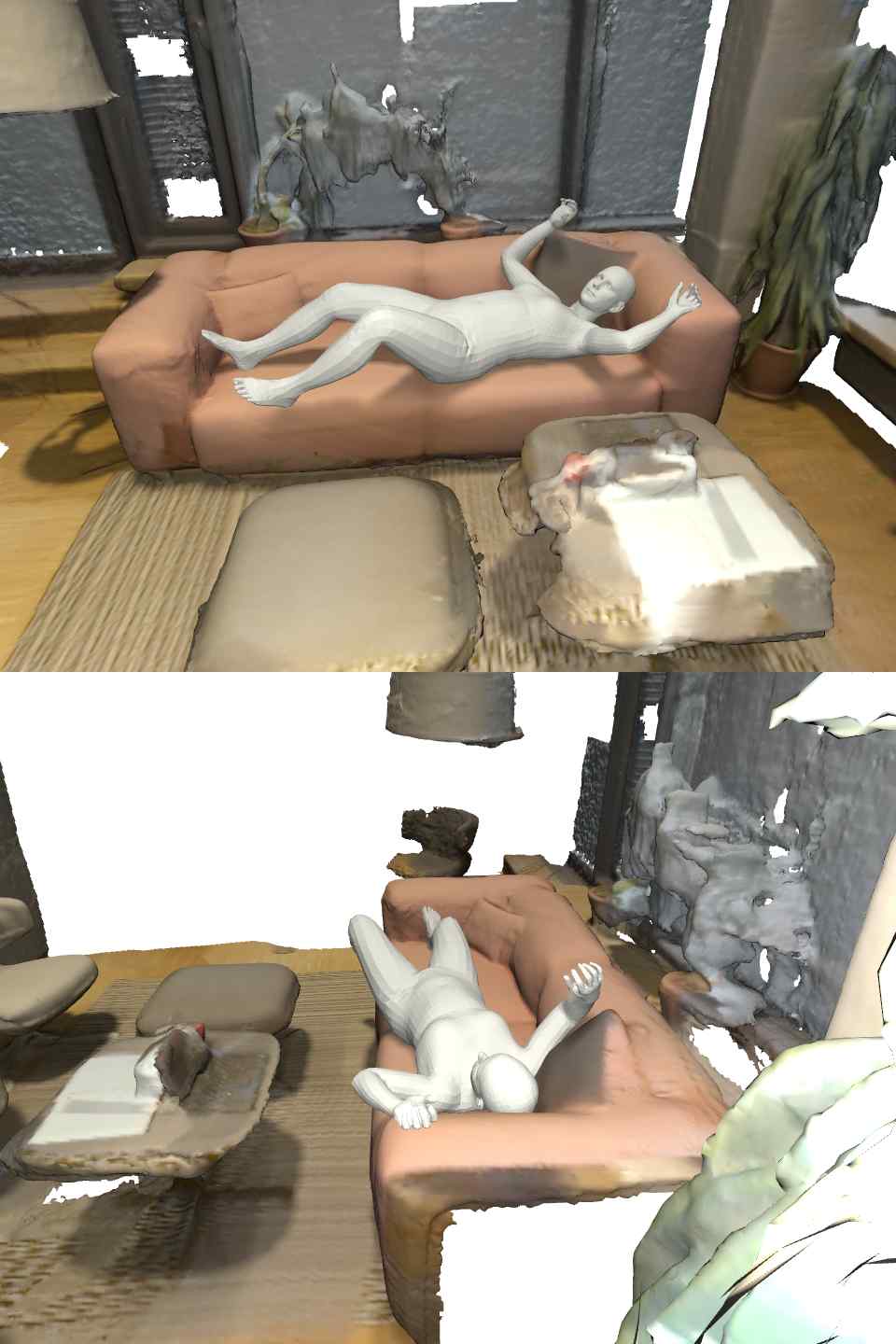}
        \includegraphics[width=0.15\textwidth]{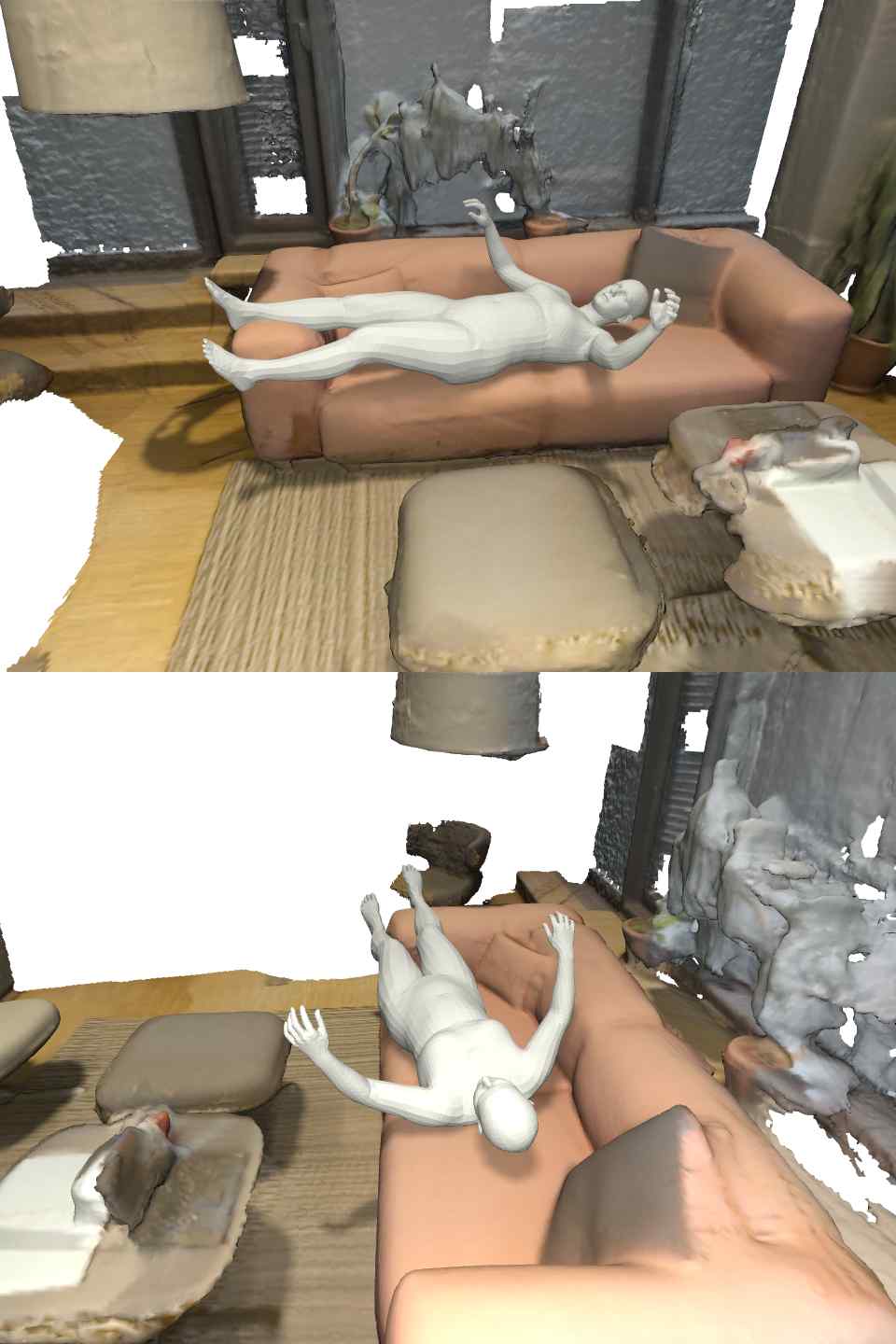}
        \includegraphics[width=0.15\textwidth]{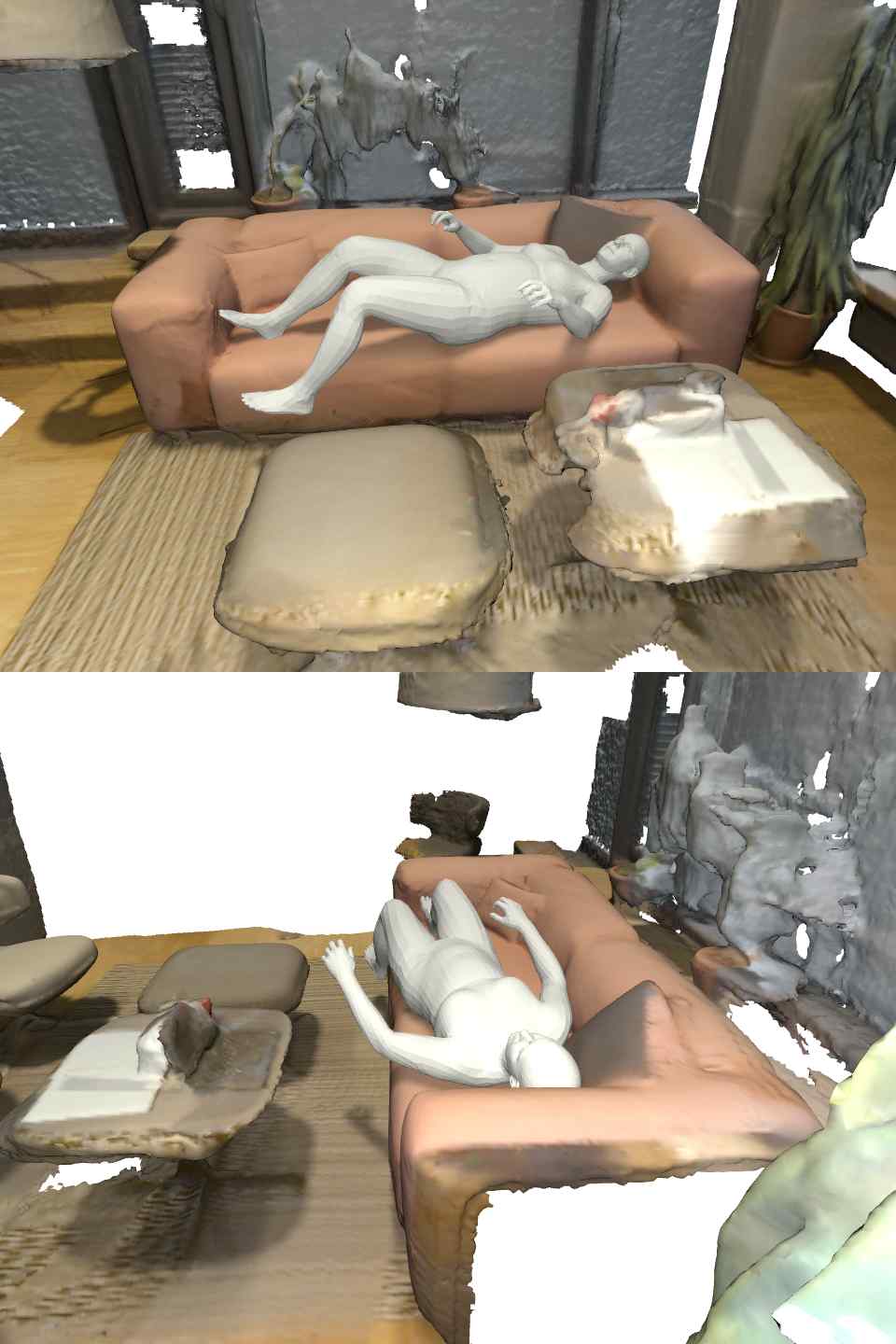}
        \includegraphics[width=0.15\textwidth]{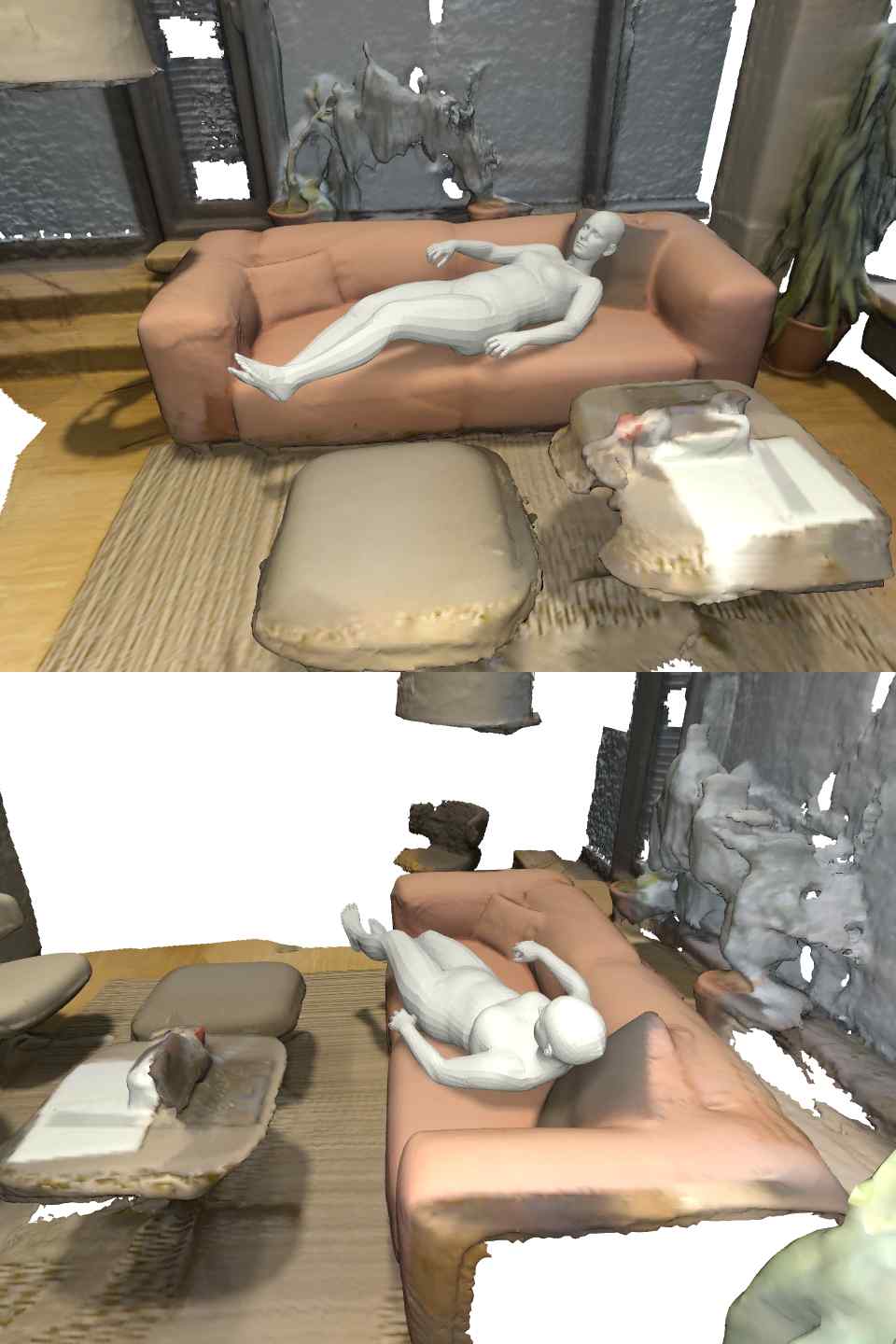}
        \includegraphics[width=0.15\textwidth]{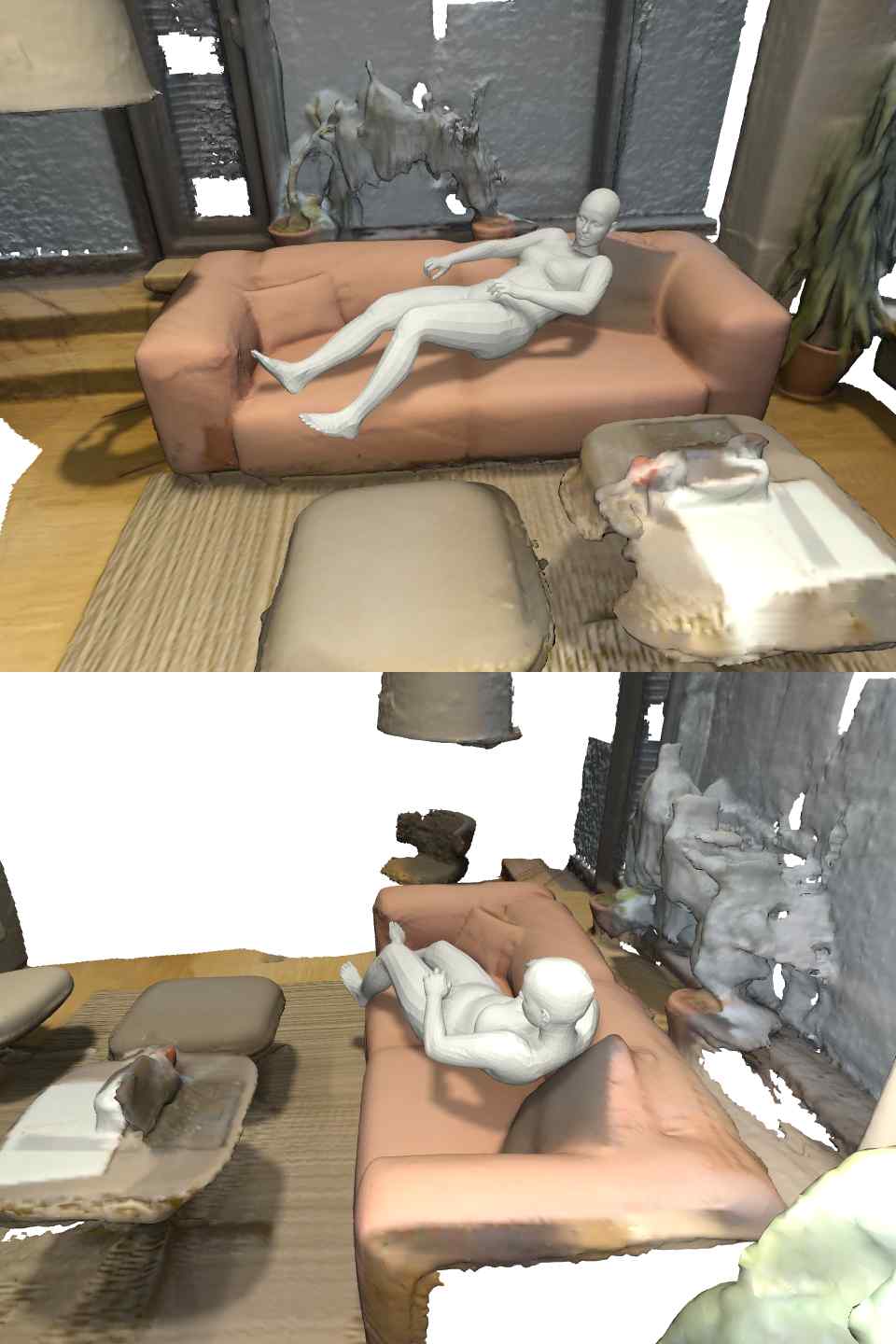}
    \end{subfigure}
    \begin{subfigure}[t]{\textwidth}
        \rotatebox{90}{lie down sofa}
        \includegraphics[width=0.15\textwidth]{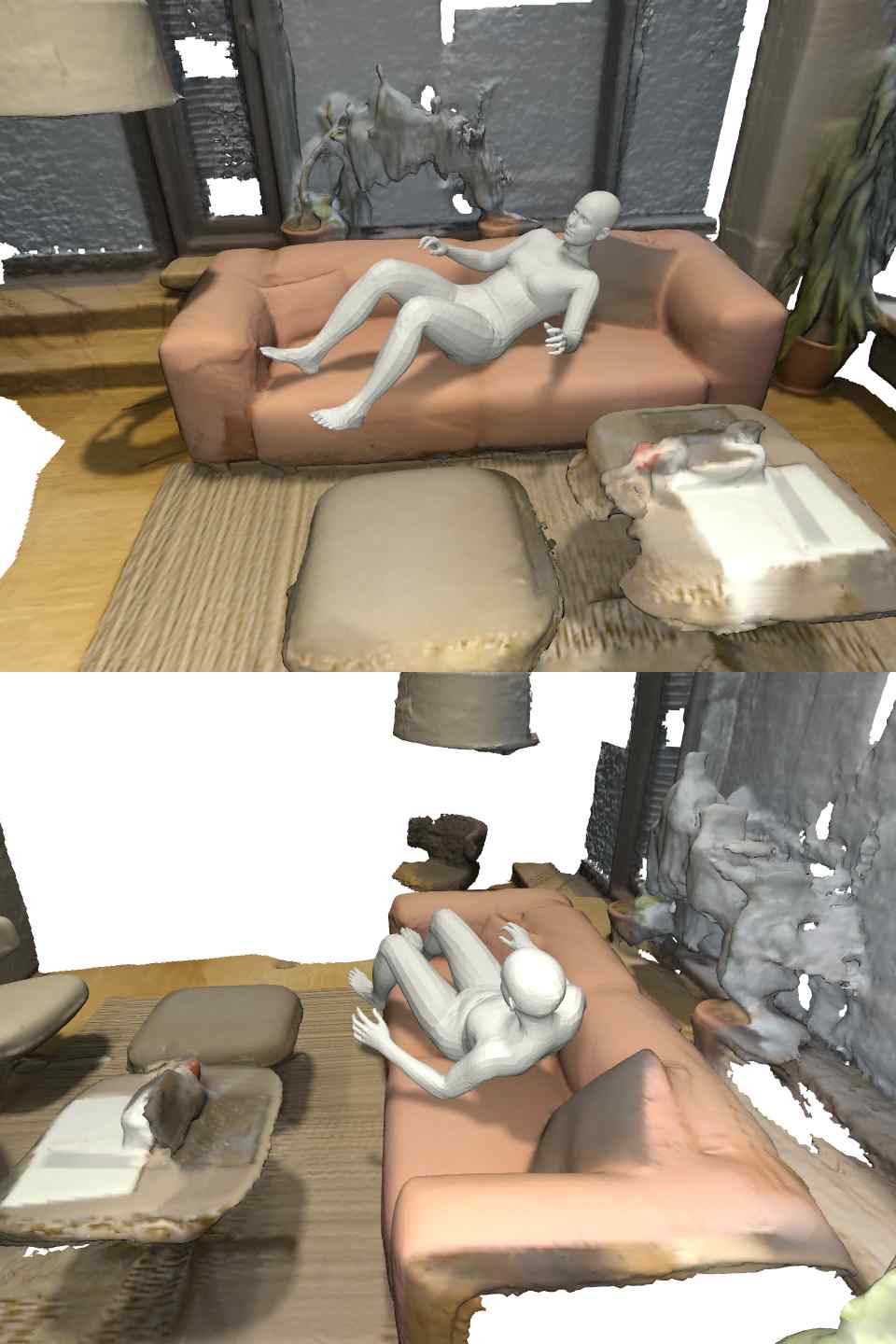}
        \includegraphics[width=0.15\textwidth]{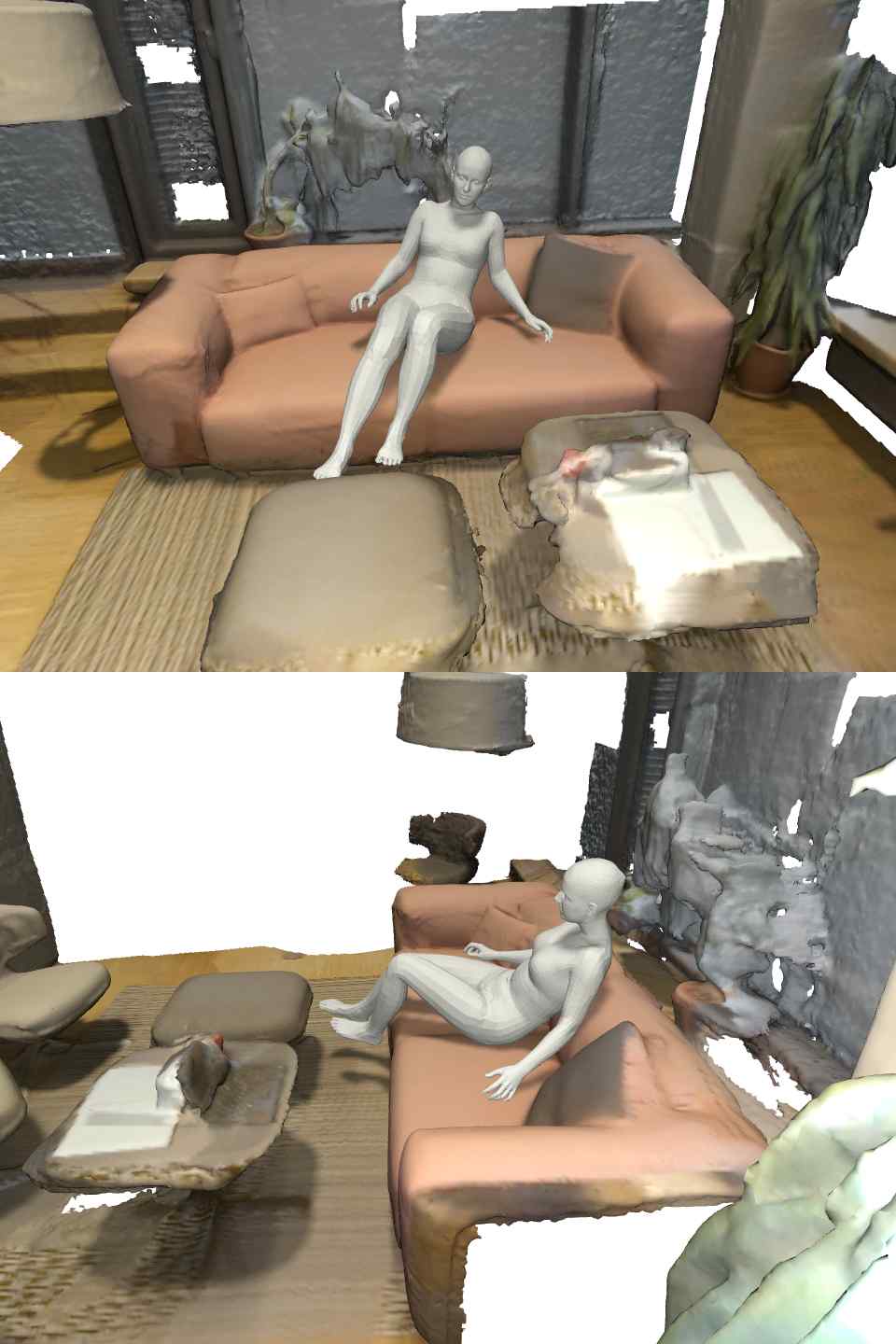}
        \includegraphics[width=0.15\textwidth]{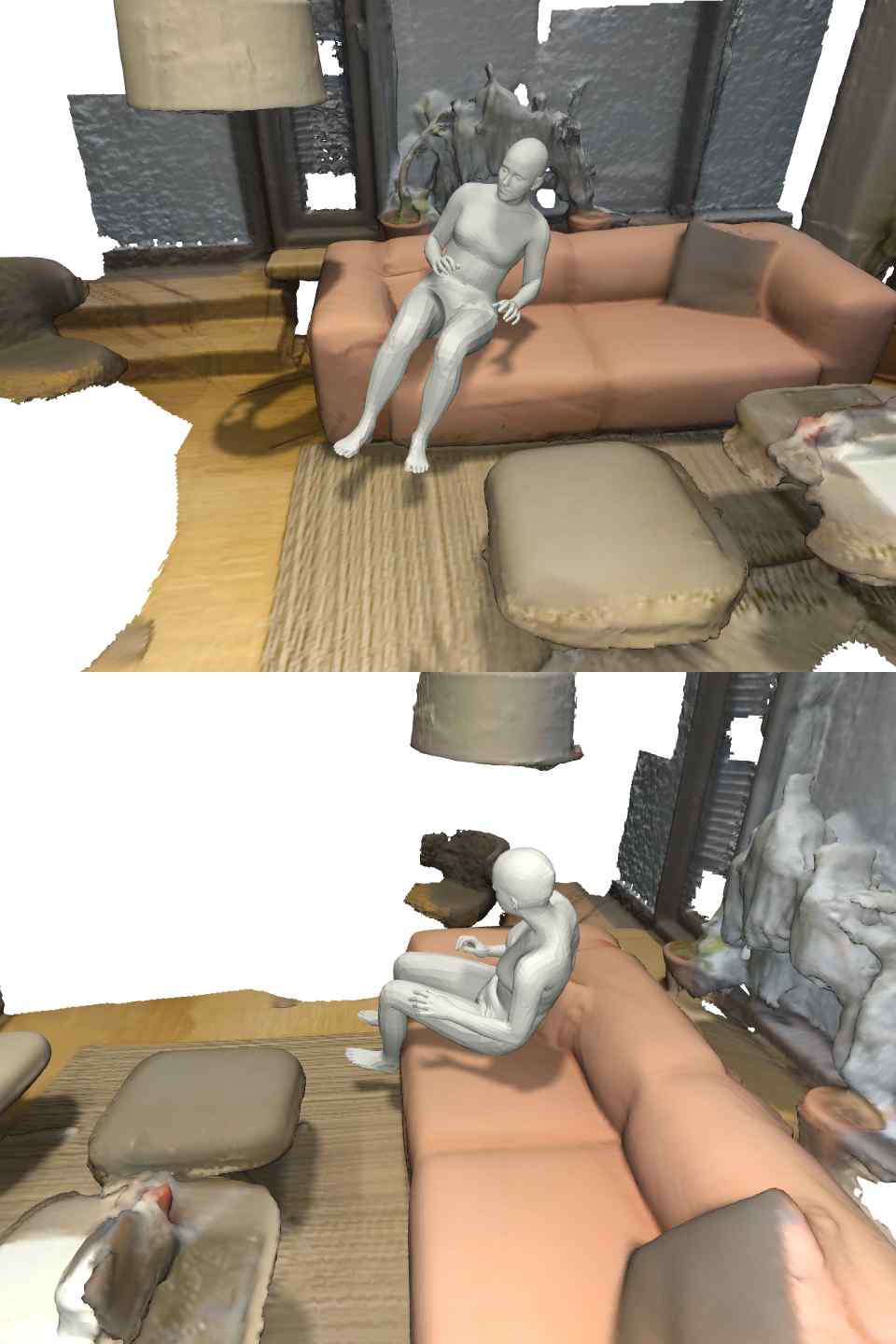}
        \includegraphics[width=0.15\textwidth]{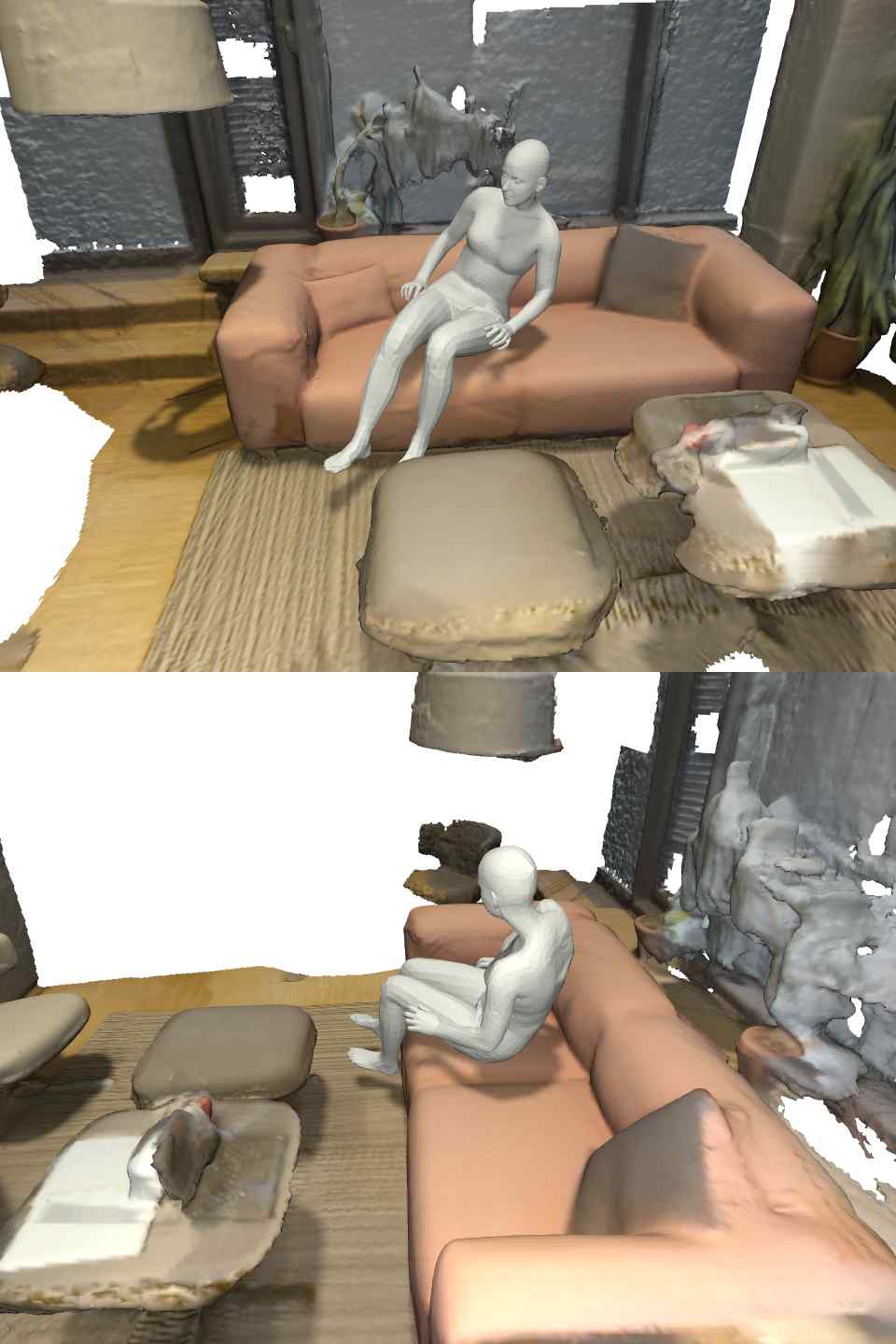}
        \includegraphics[width=0.15\textwidth]{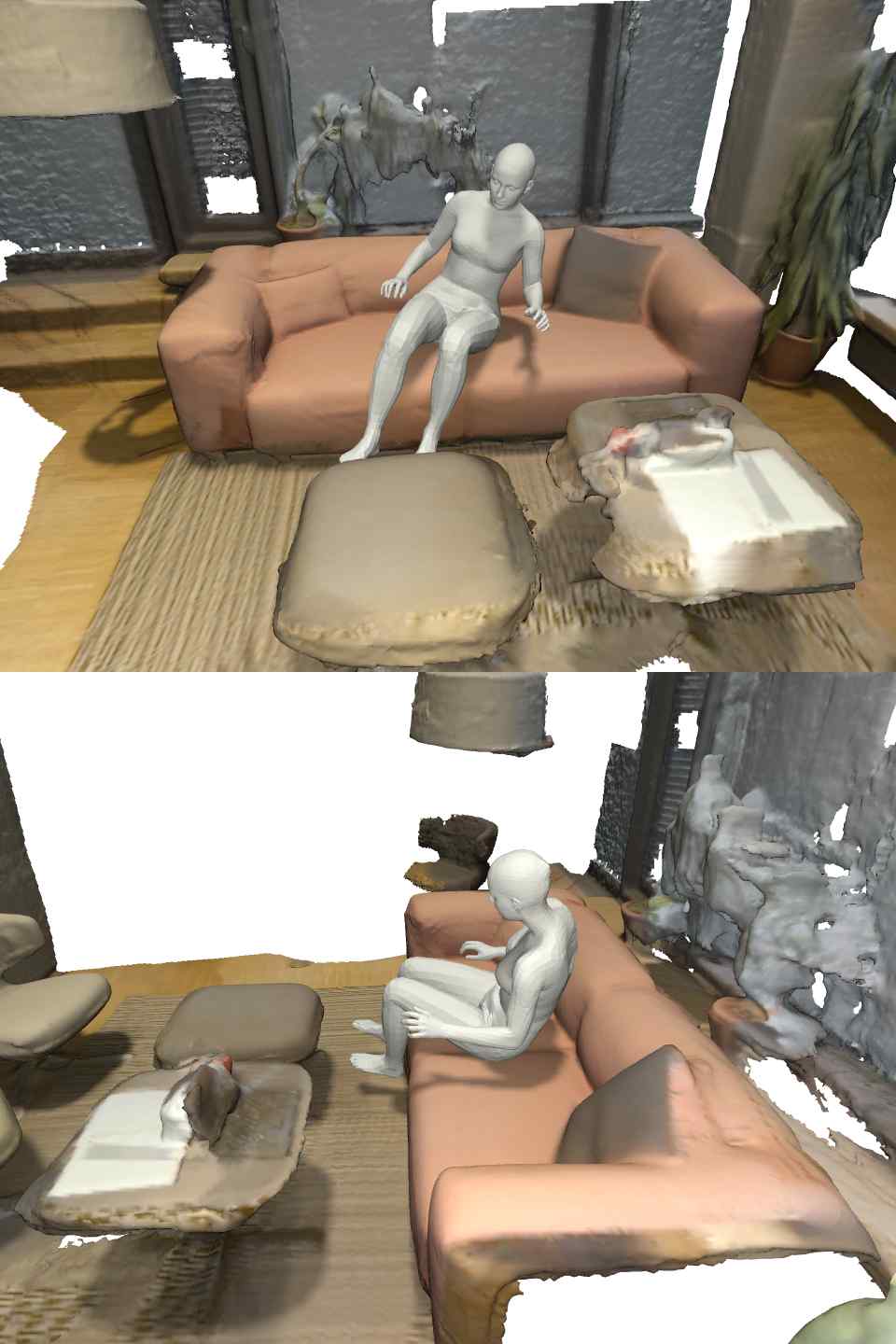}
        \includegraphics[width=0.15\textwidth]{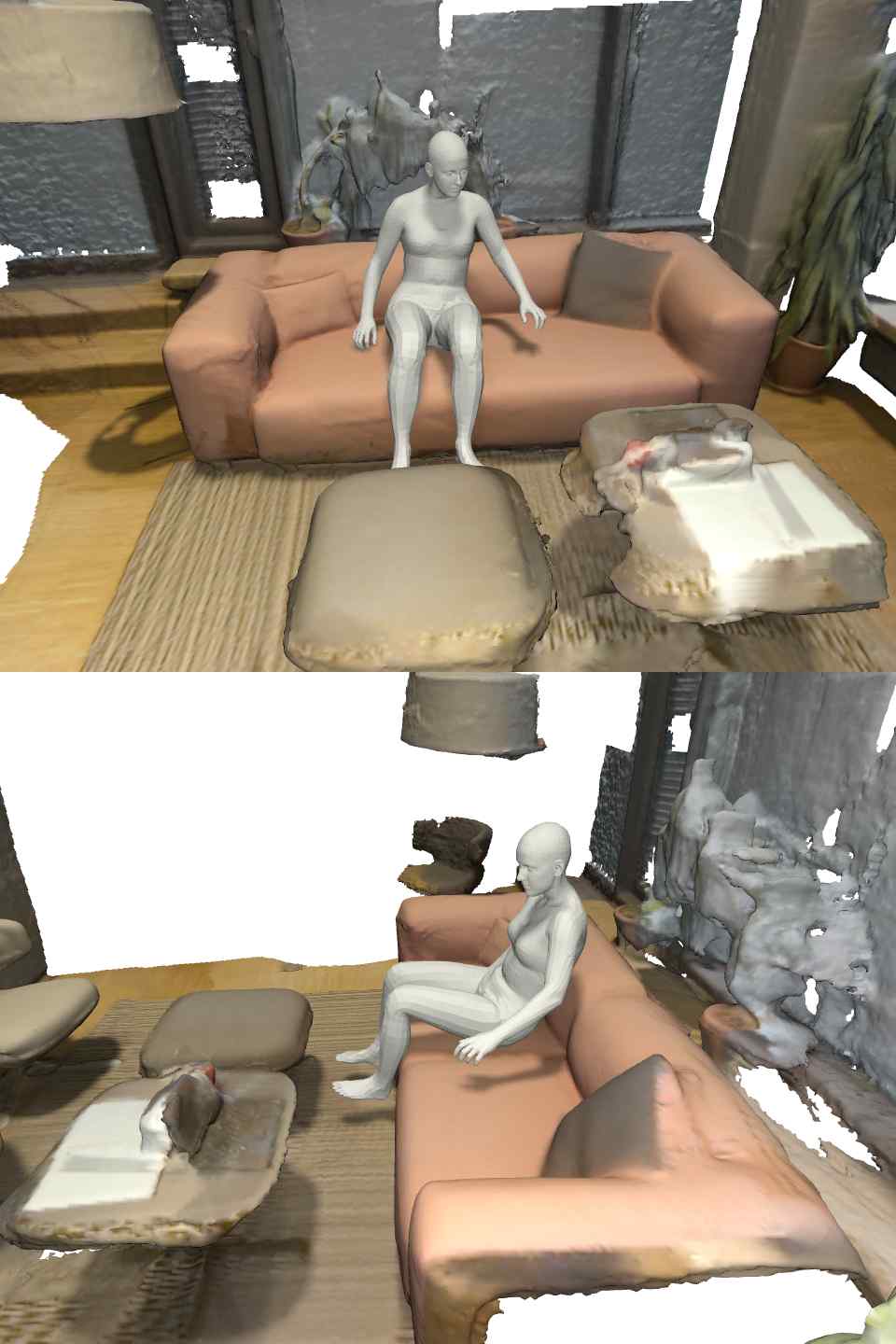}
        
    \end{subfigure}
    \begin{subfigure}[t]{\textwidth}
        \rotatebox{90}{touch chair}
        \includegraphics[width=0.15\textwidth]{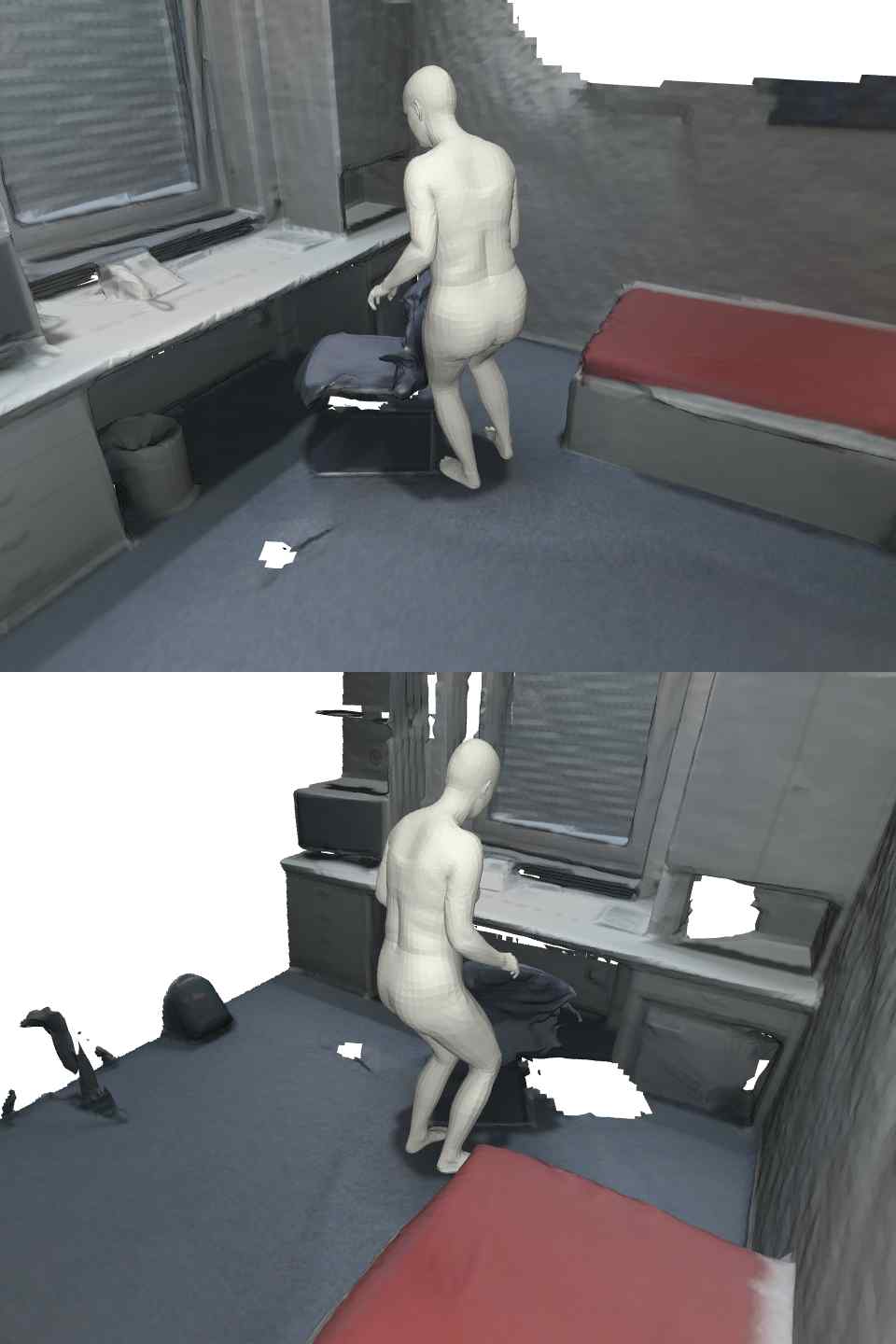}
        \includegraphics[width=0.15\textwidth]{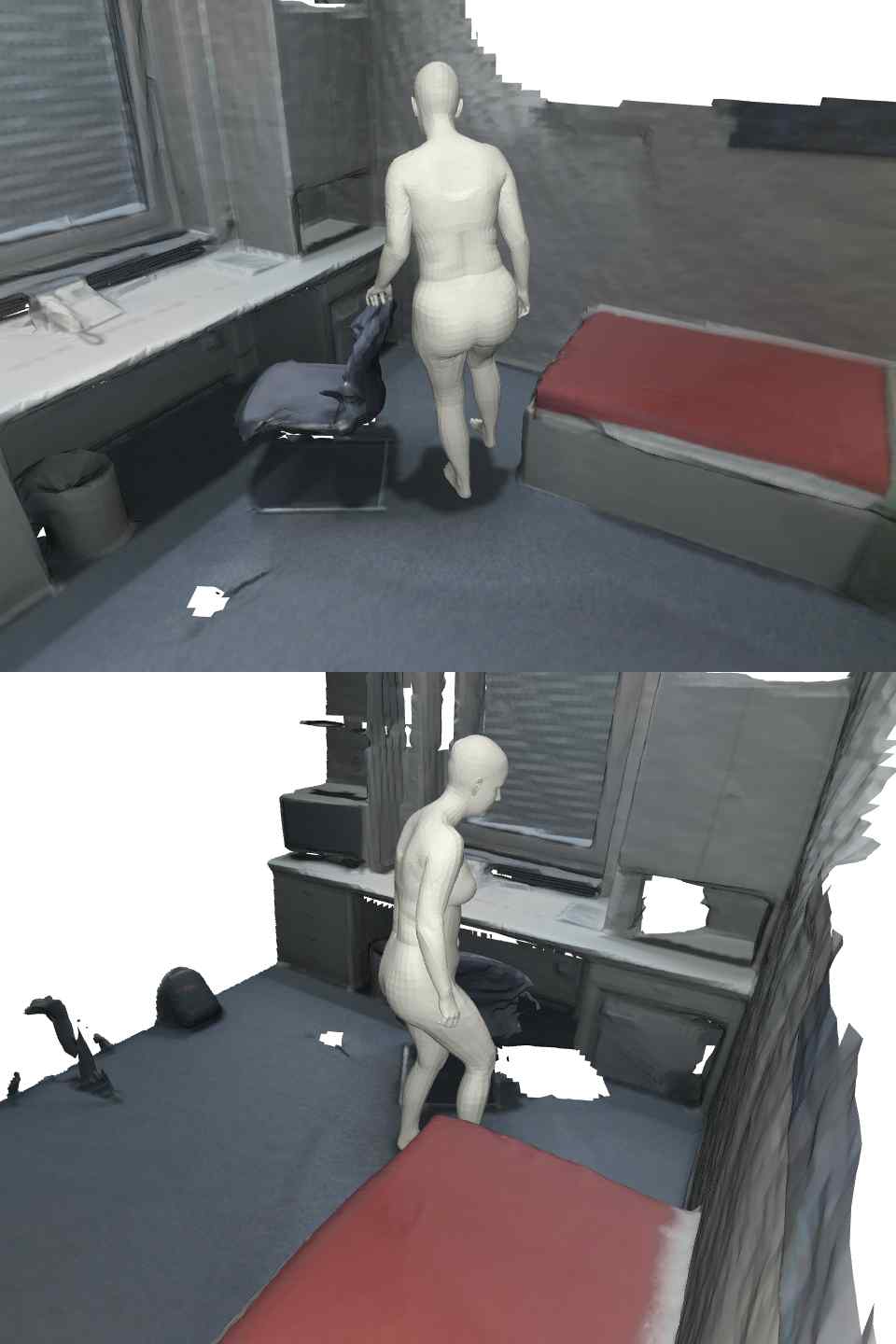}
        \includegraphics[width=0.15\textwidth]{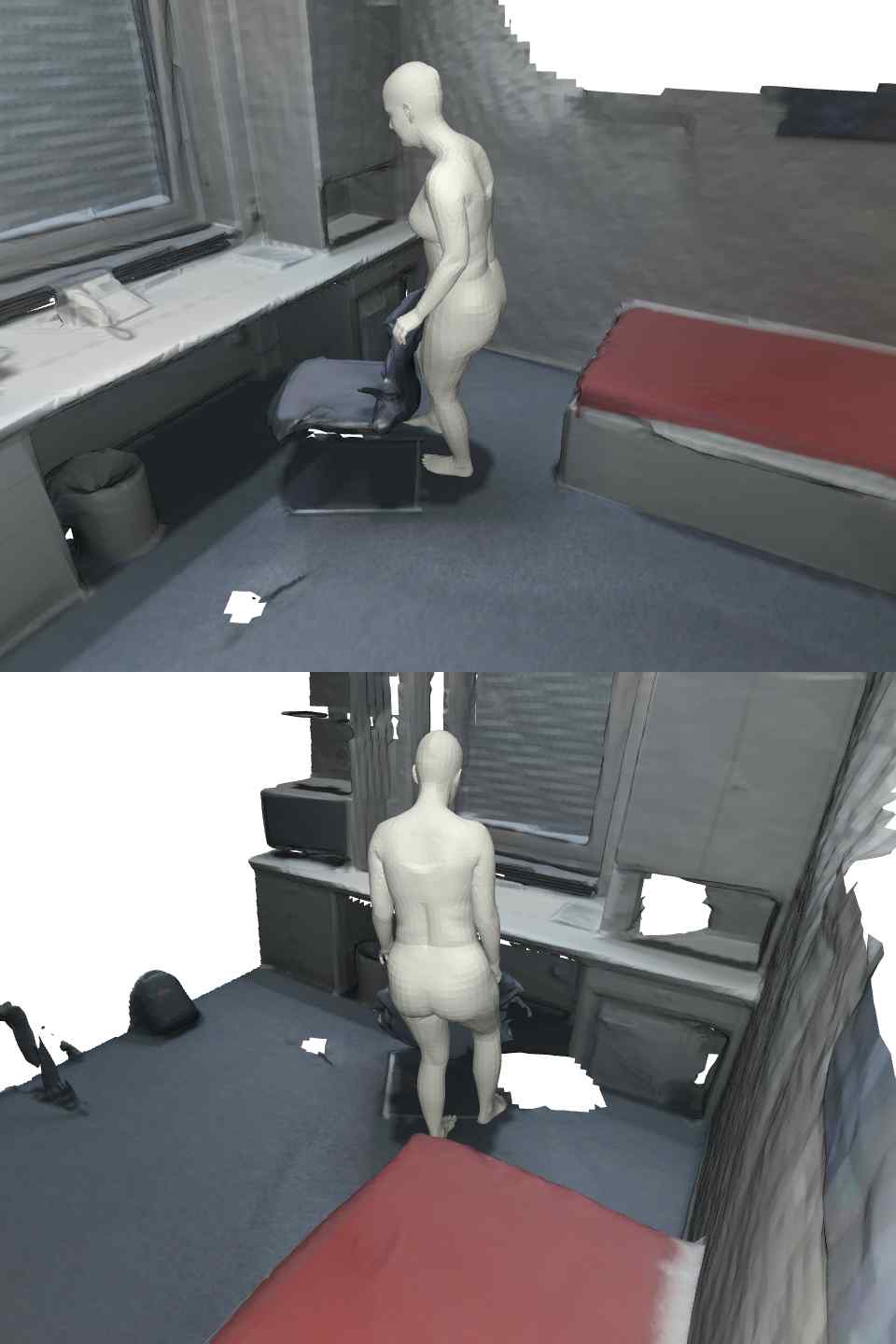}
        \includegraphics[width=0.15\textwidth]{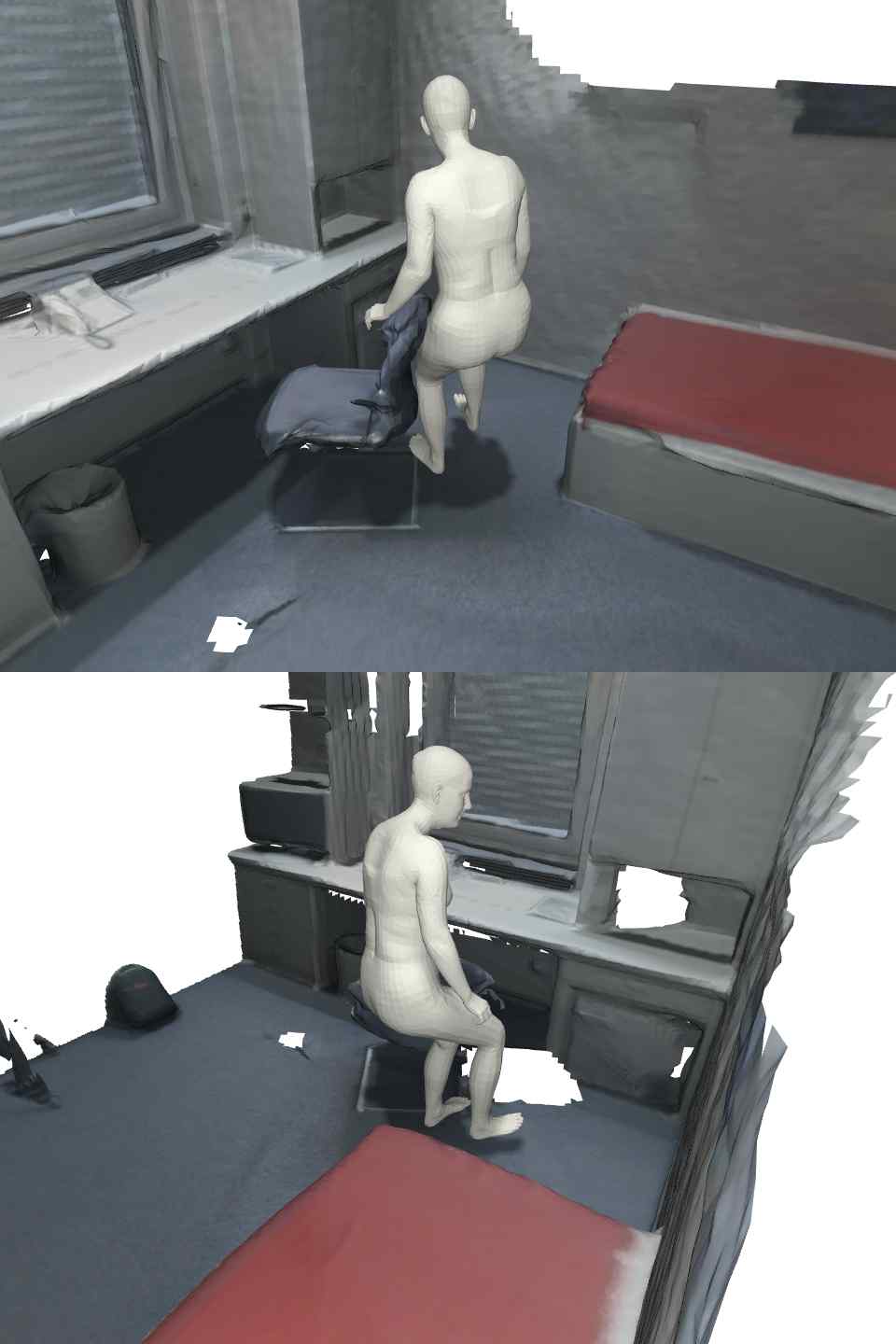}
        \includegraphics[width=0.15\textwidth]{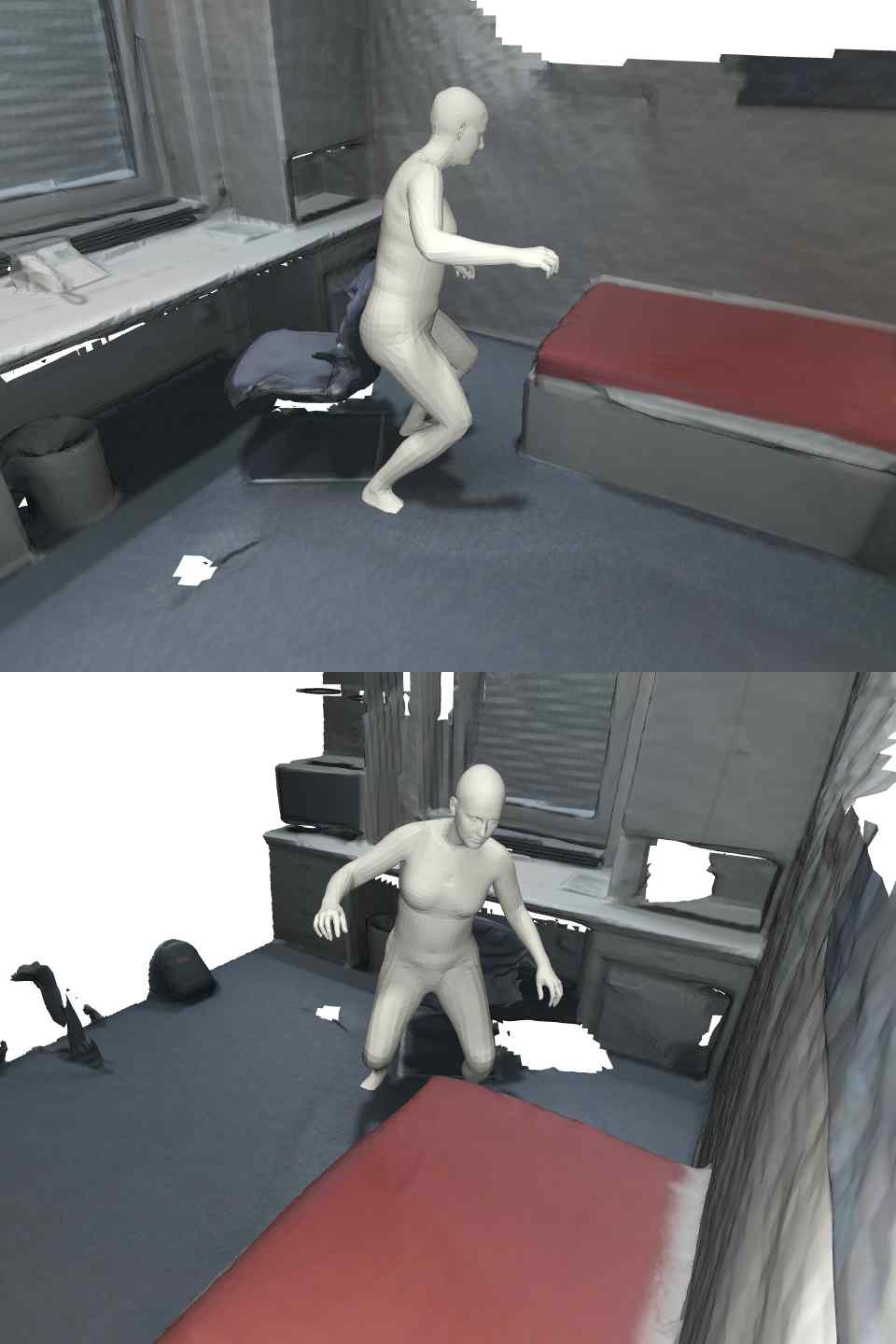}
        \includegraphics[width=0.15\textwidth]{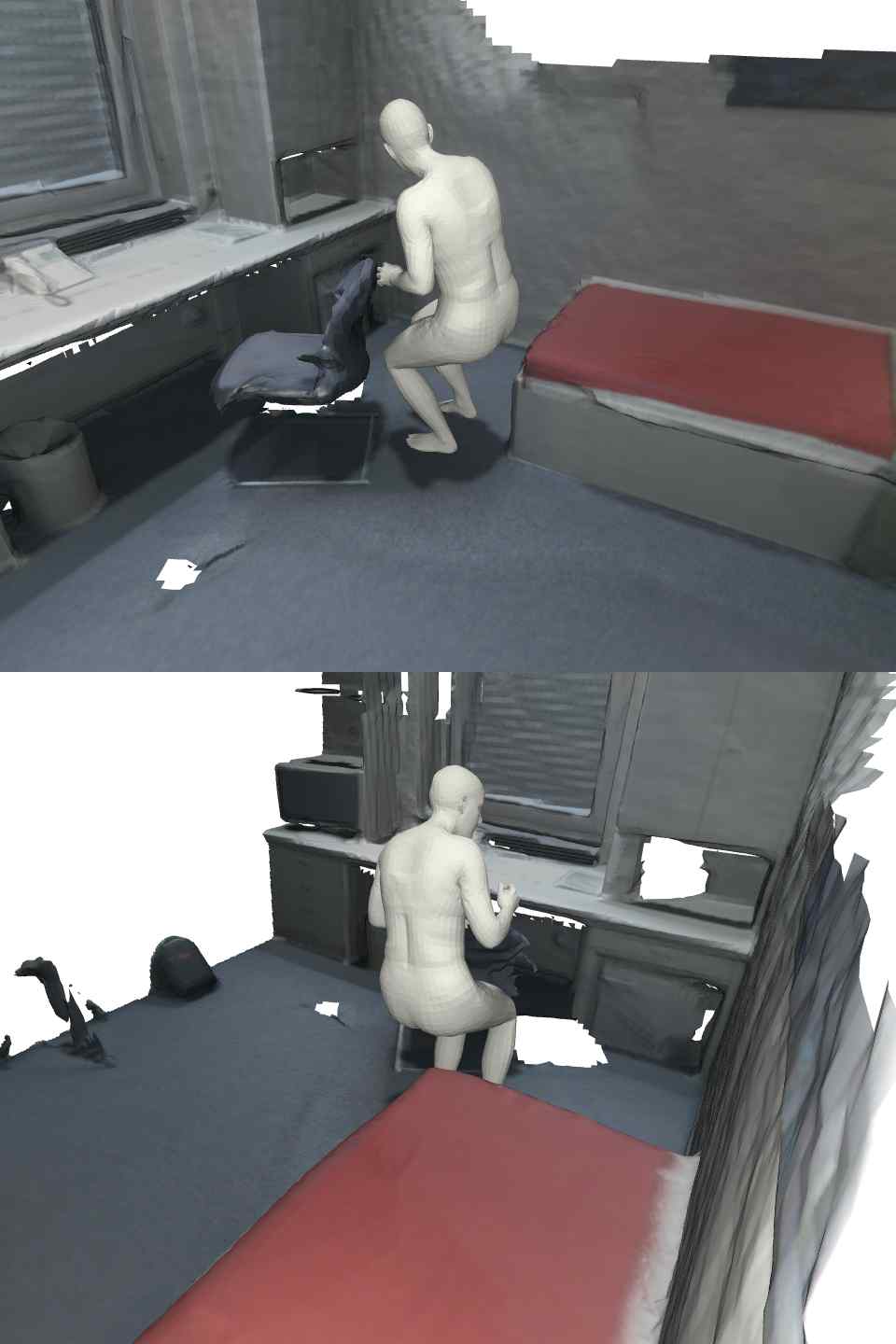}
    \end{subfigure}
    \begin{subfigure}[t]{\textwidth}
        \rotatebox{90}{sit down chair}
        \includegraphics[width=0.15\textwidth]{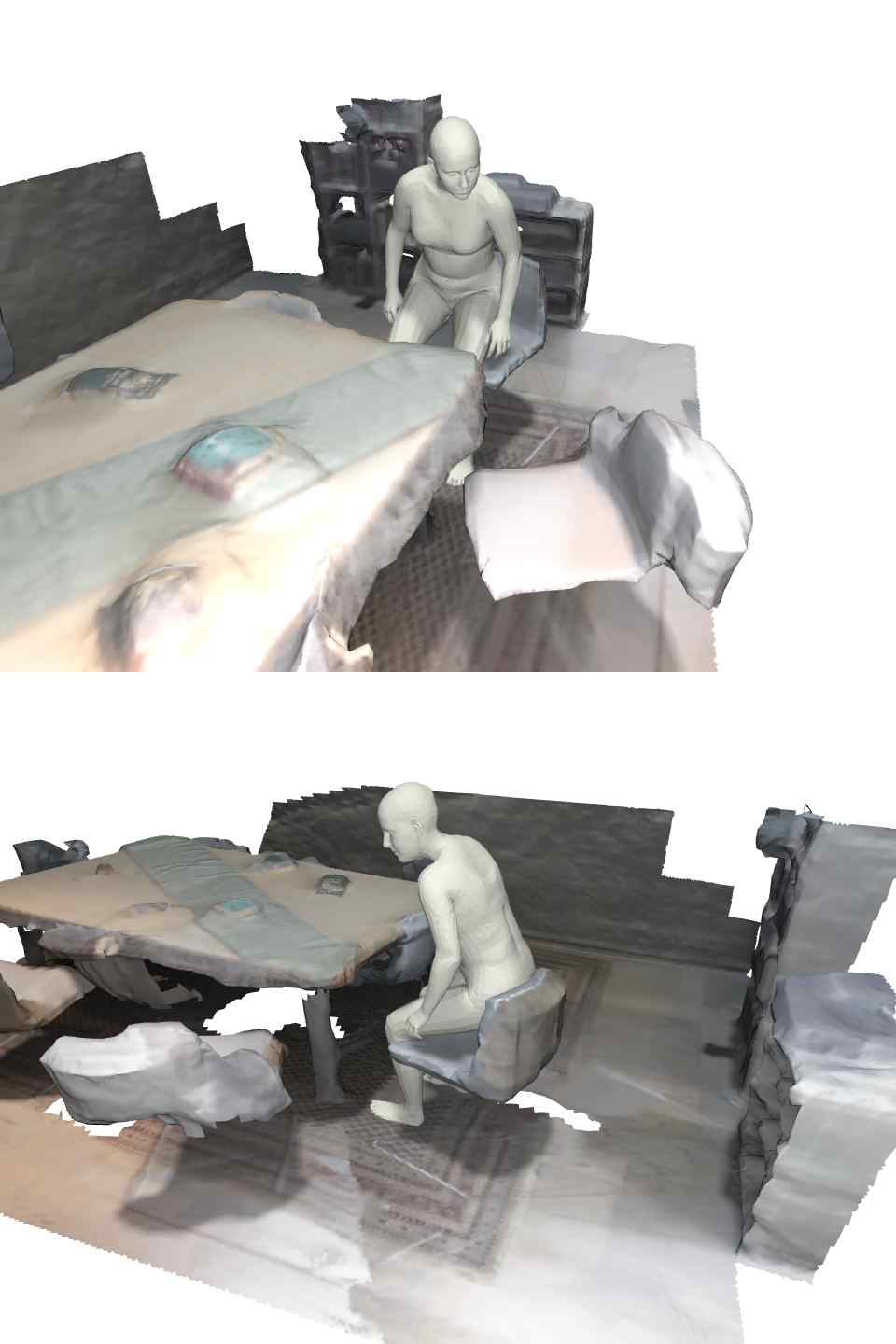}
        \includegraphics[width=0.15\textwidth]{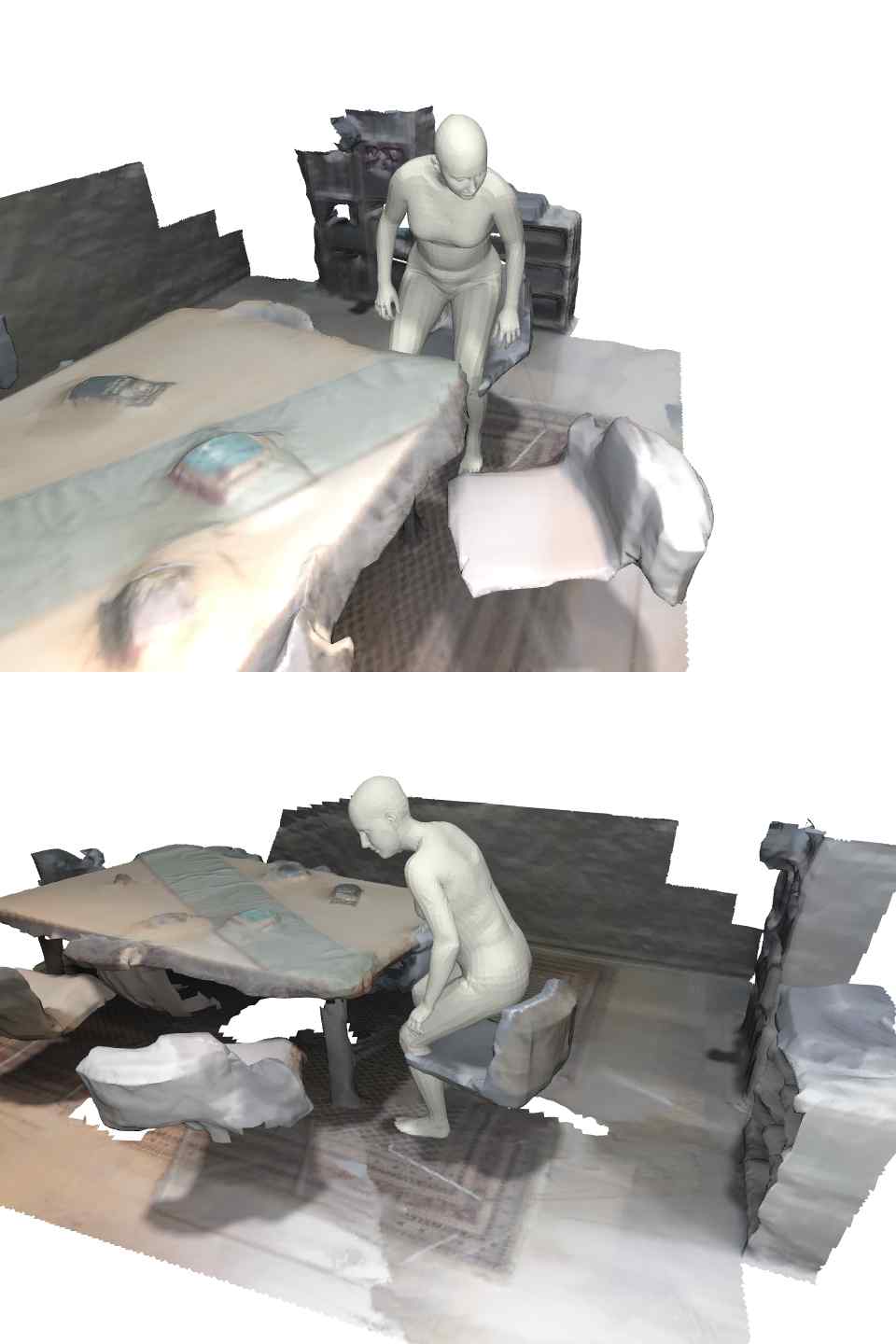}
        \includegraphics[width=0.15\textwidth]{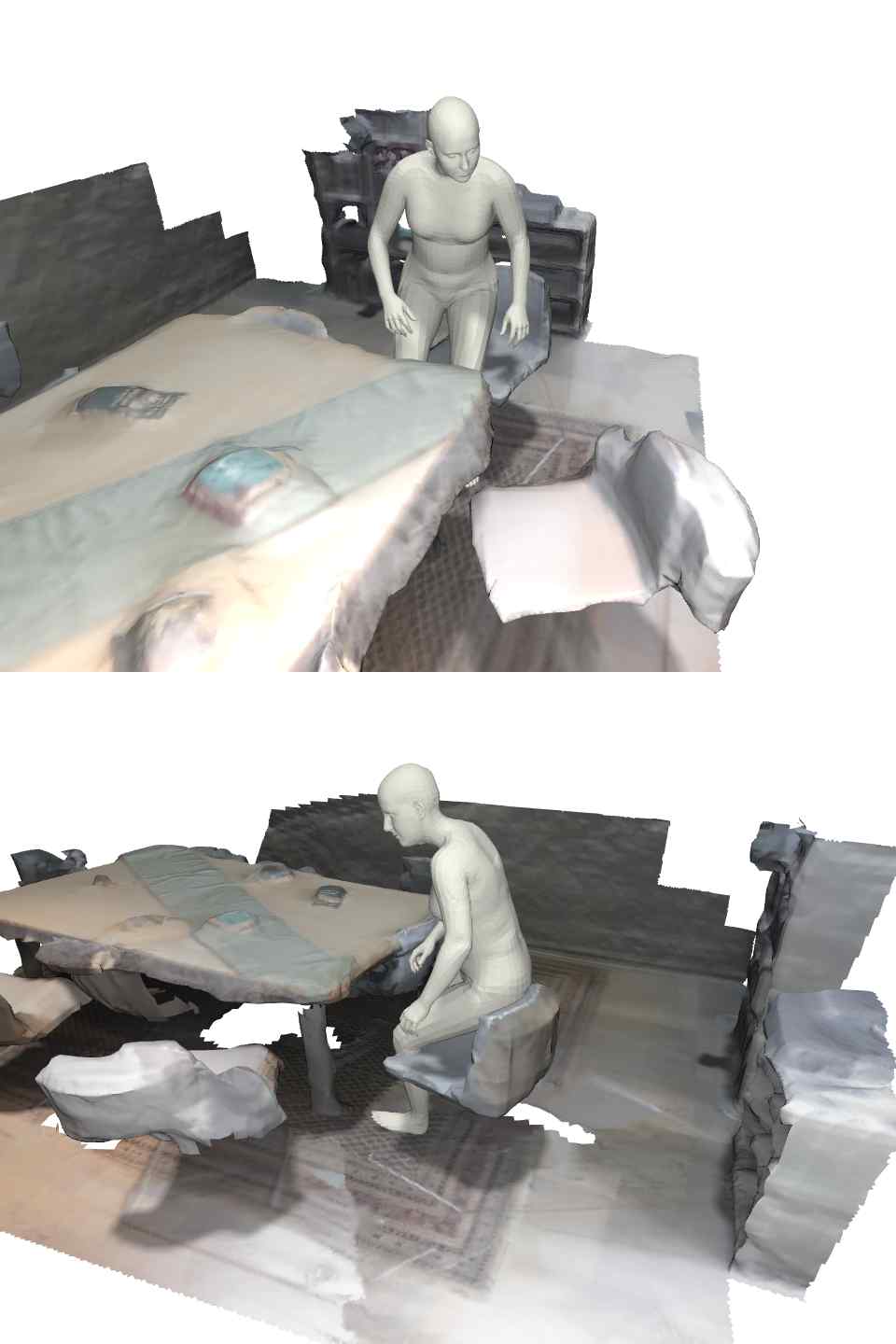}
        \includegraphics[width=0.15\textwidth]{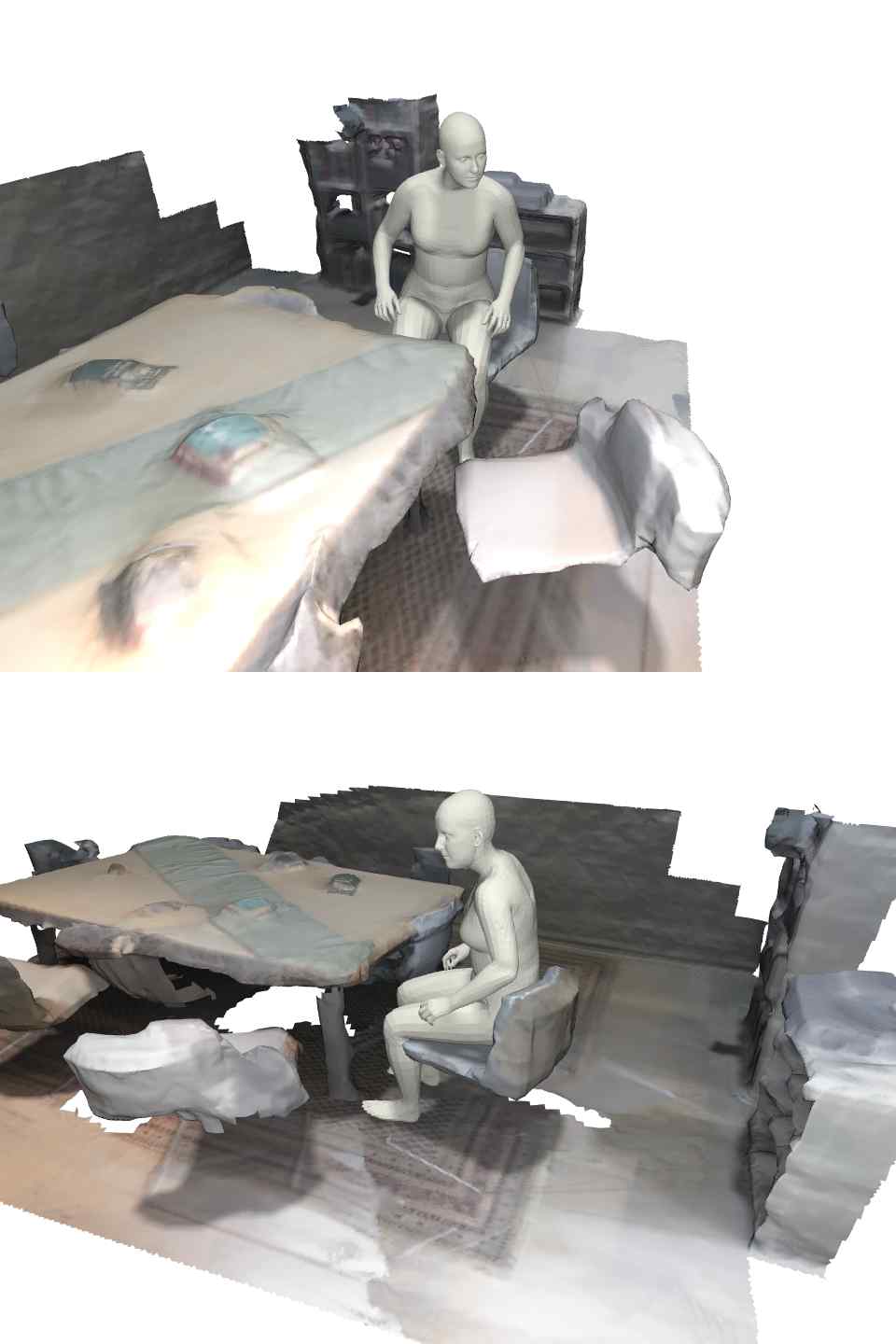}
        \includegraphics[width=0.15\textwidth]{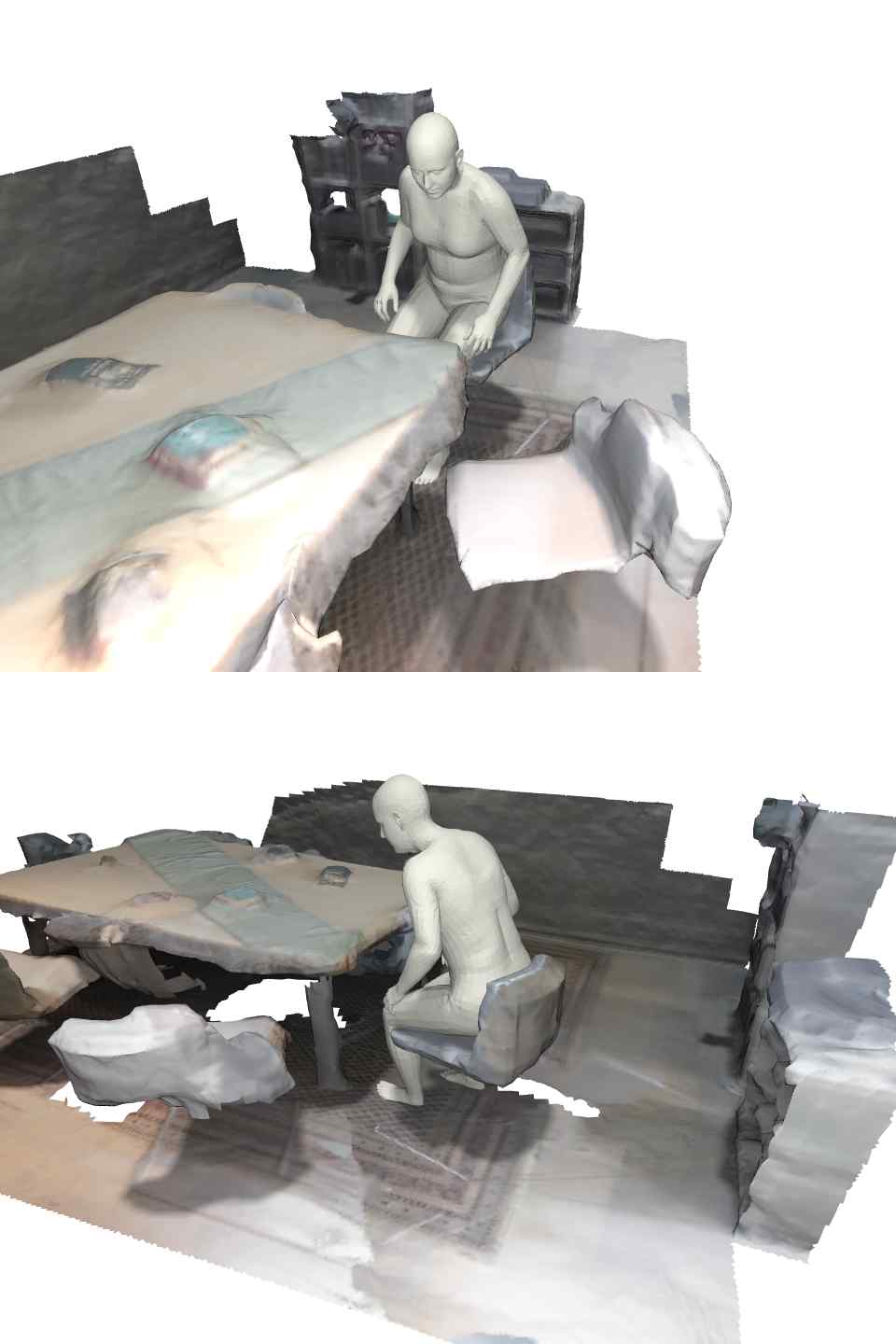}
        \includegraphics[width=0.15\textwidth]{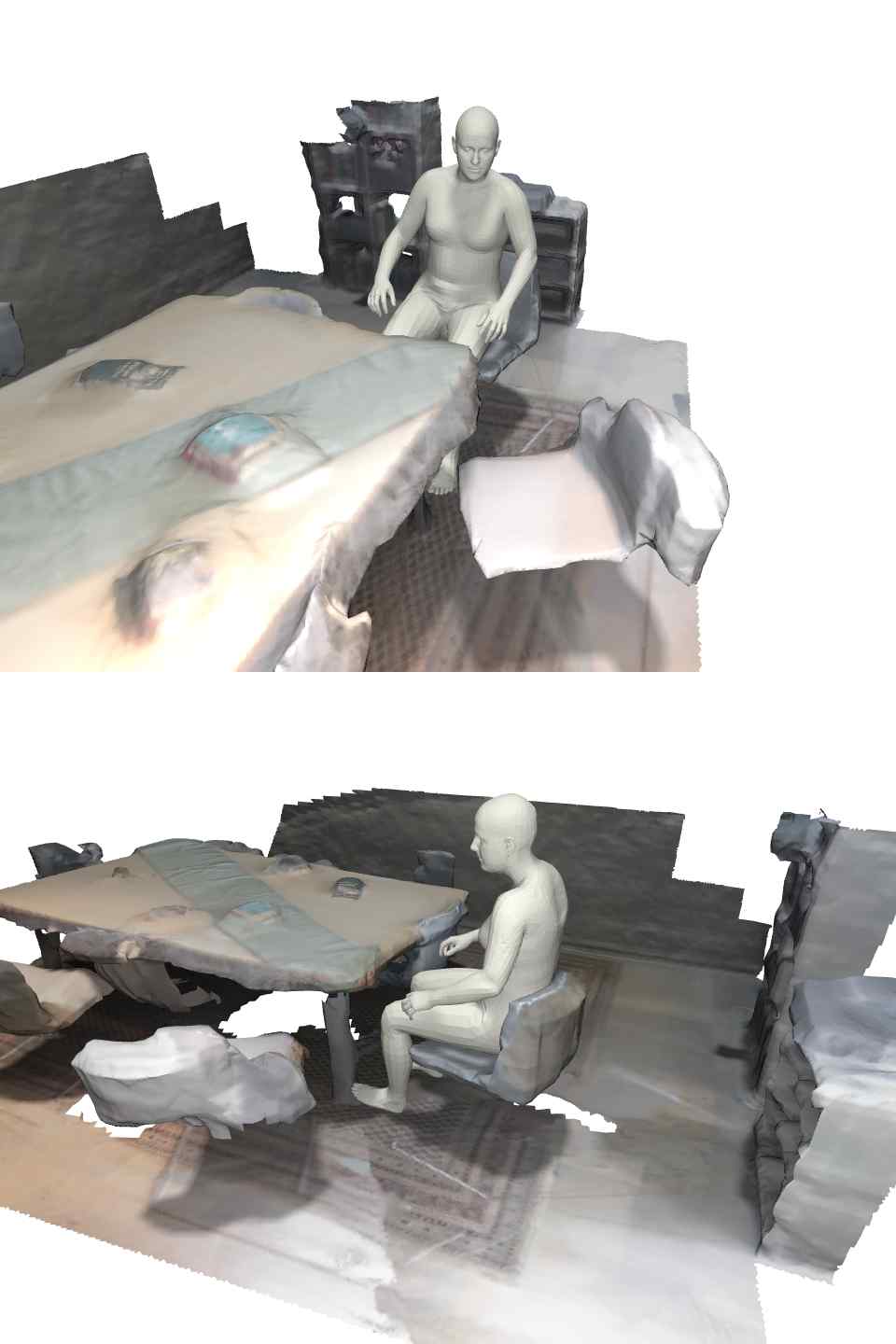}
    \end{subfigure}
    \begin{subfigure}[t]{\textwidth}
        \rotatebox{90}{touch monitor}
        \includegraphics[width=0.15\textwidth]{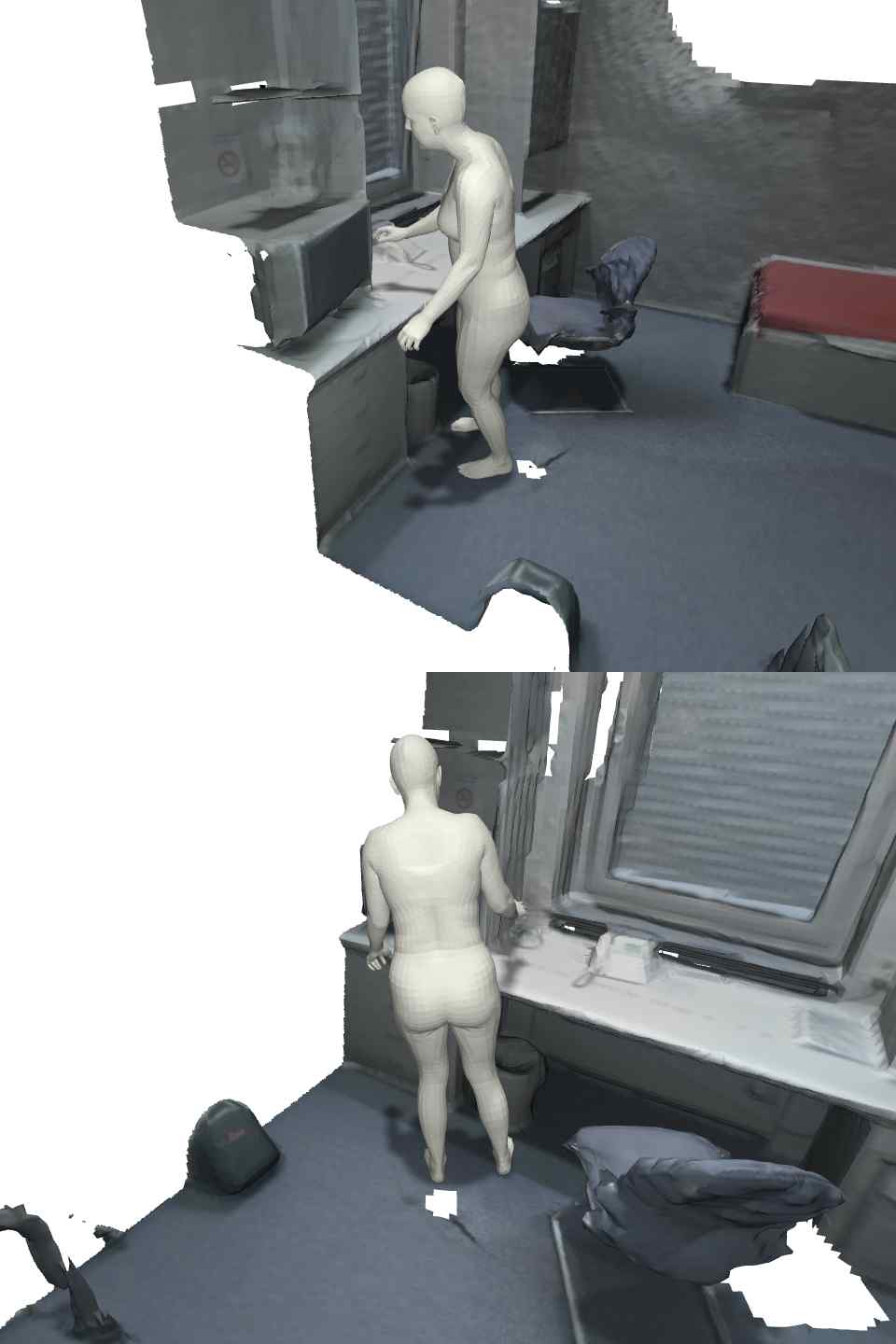}
        \includegraphics[width=0.15\textwidth]{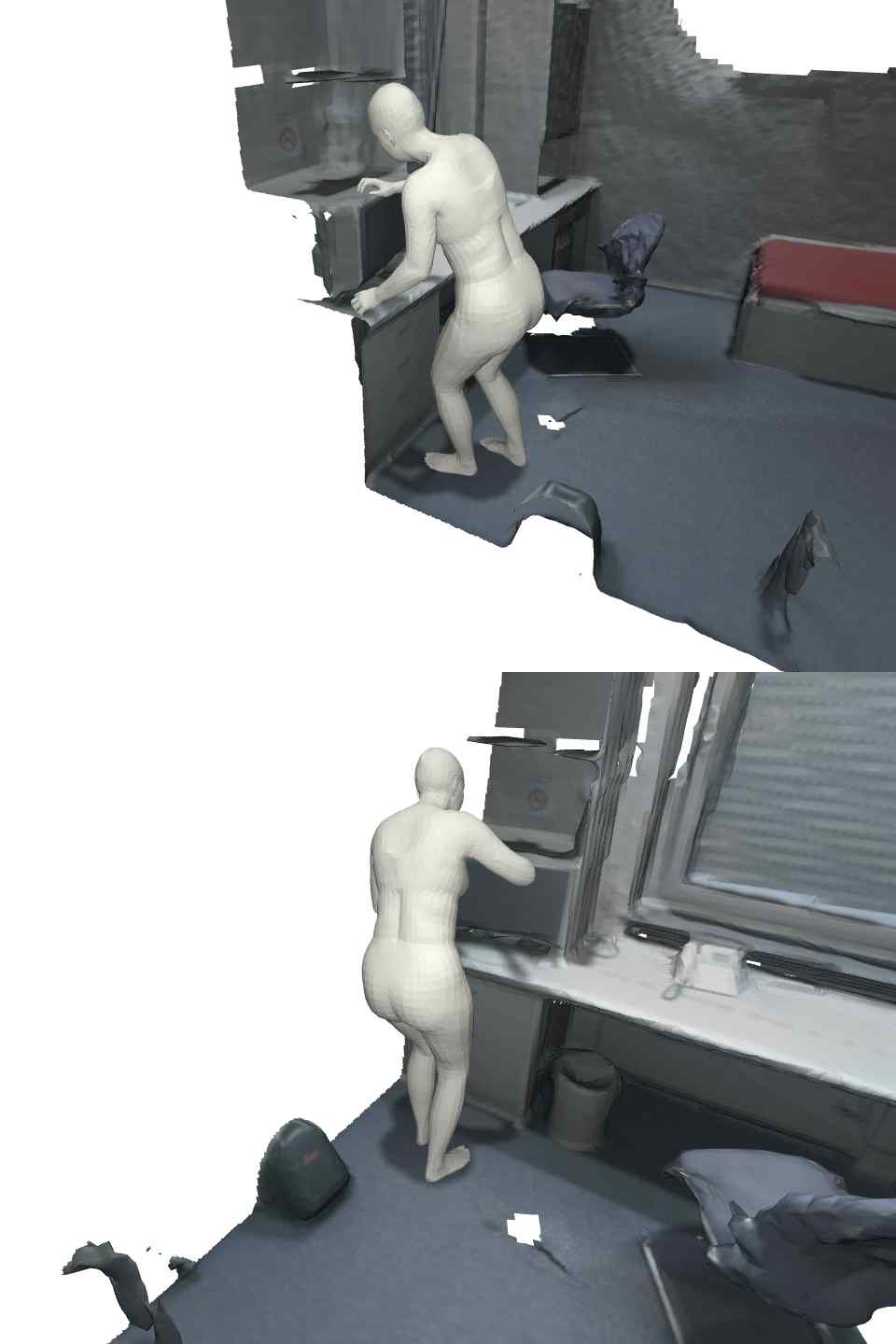}
        \includegraphics[width=0.15\textwidth]{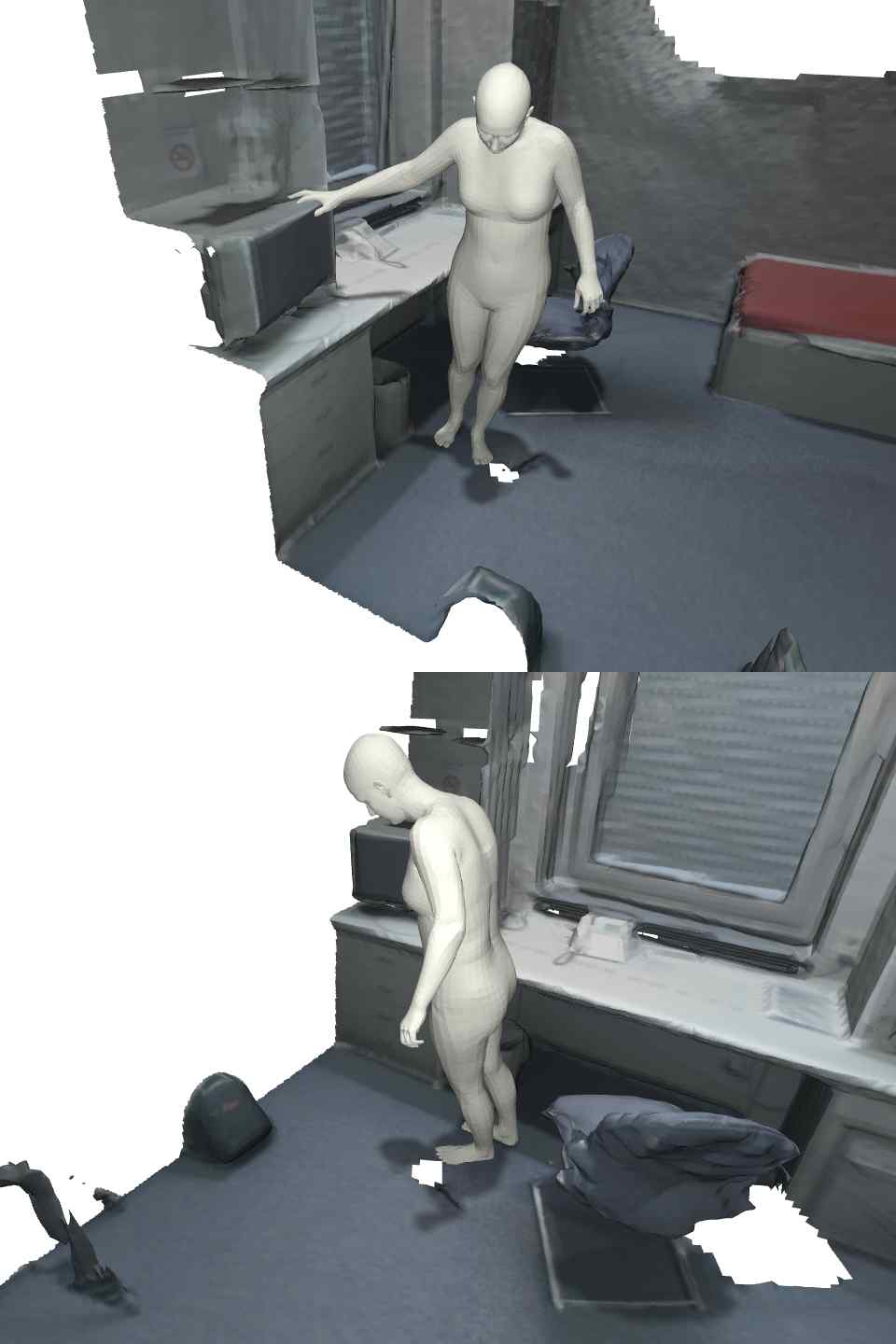}
        \includegraphics[width=0.15\textwidth]{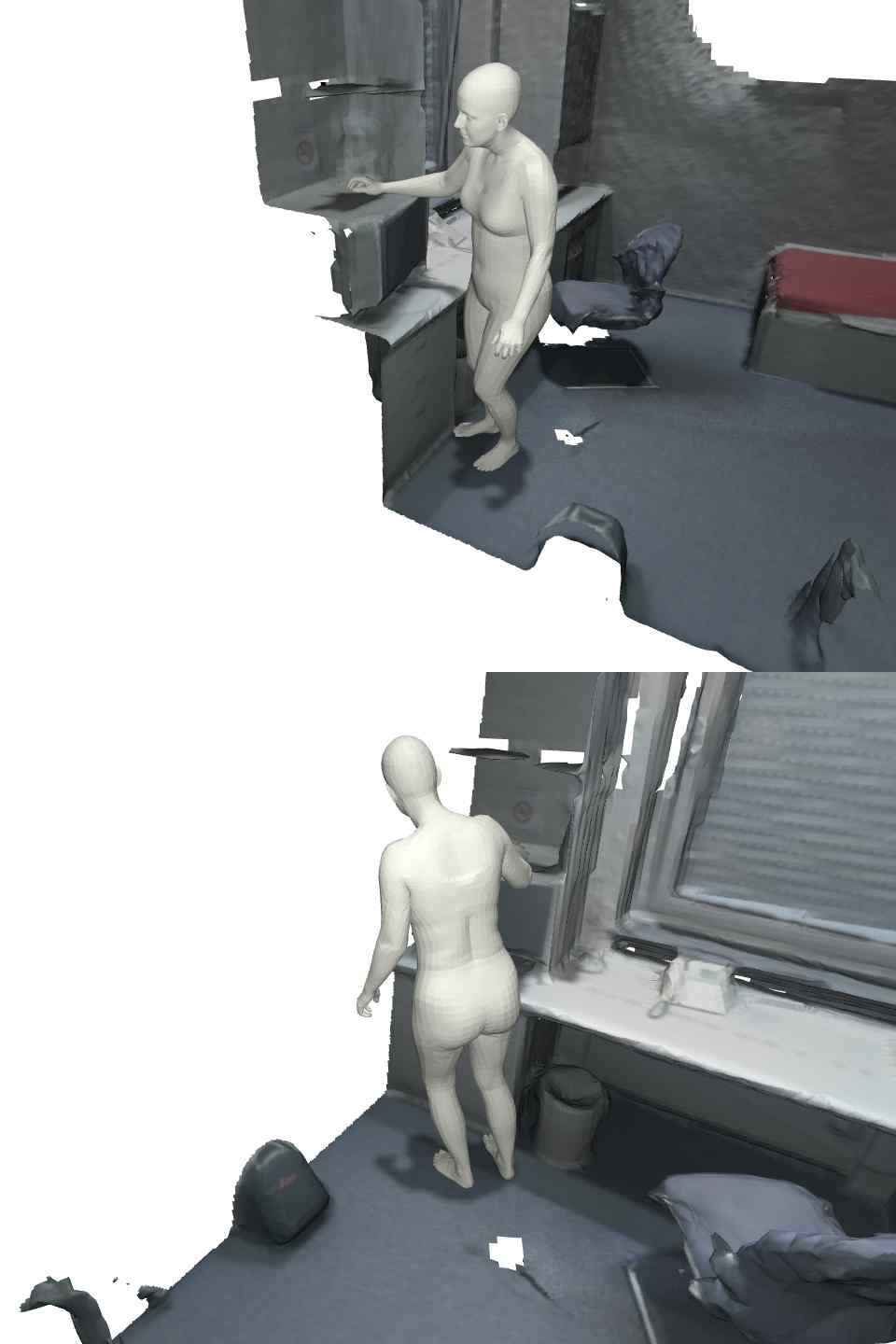}
        \includegraphics[width=0.15\textwidth]{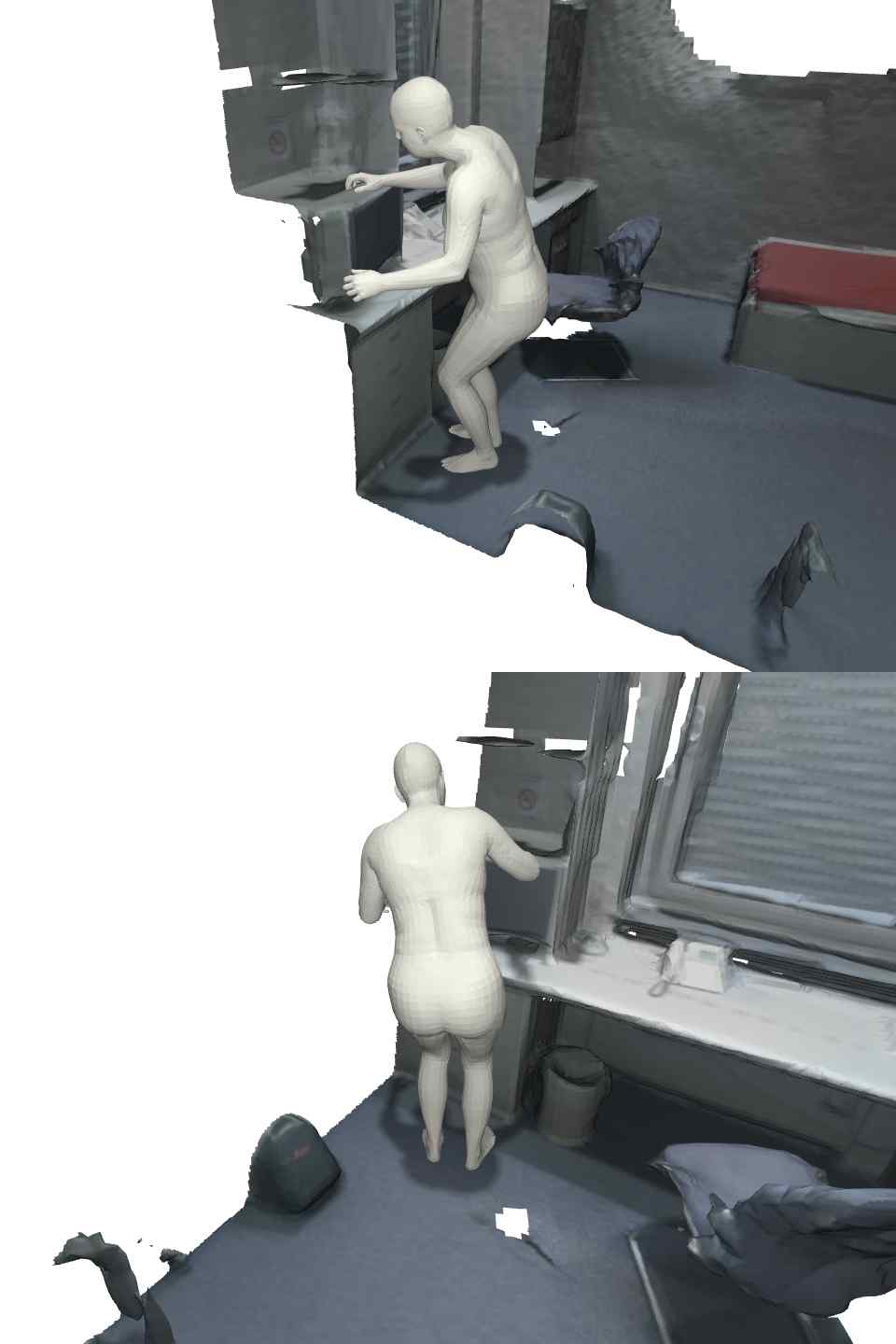}
        \includegraphics[width=0.15\textwidth]{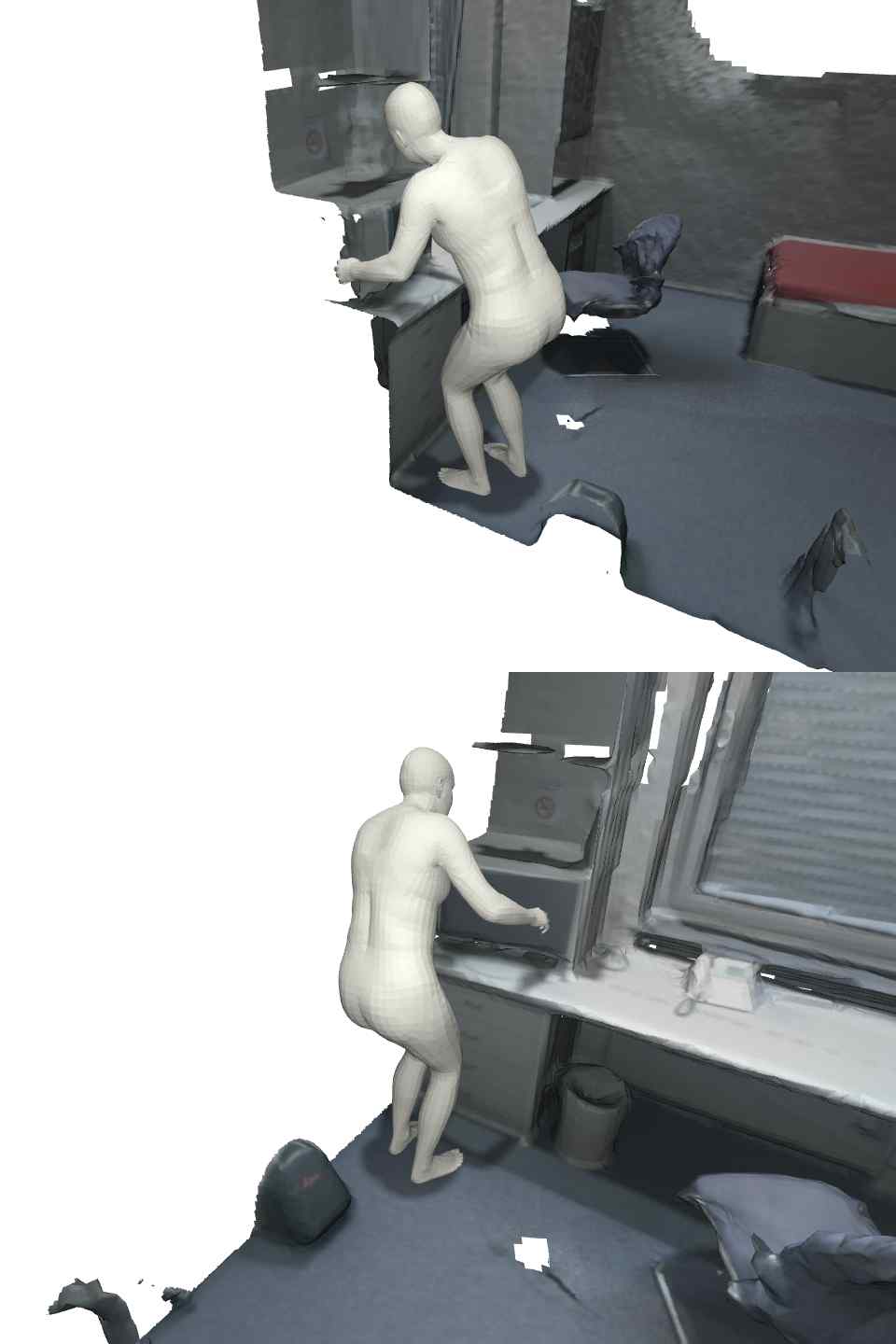}
    \end{subfigure}
    \begin{subfigure}[t]{\textwidth}
        \rotatebox{90}{sit on cabinet}
        \includegraphics[width=0.15\textwidth]{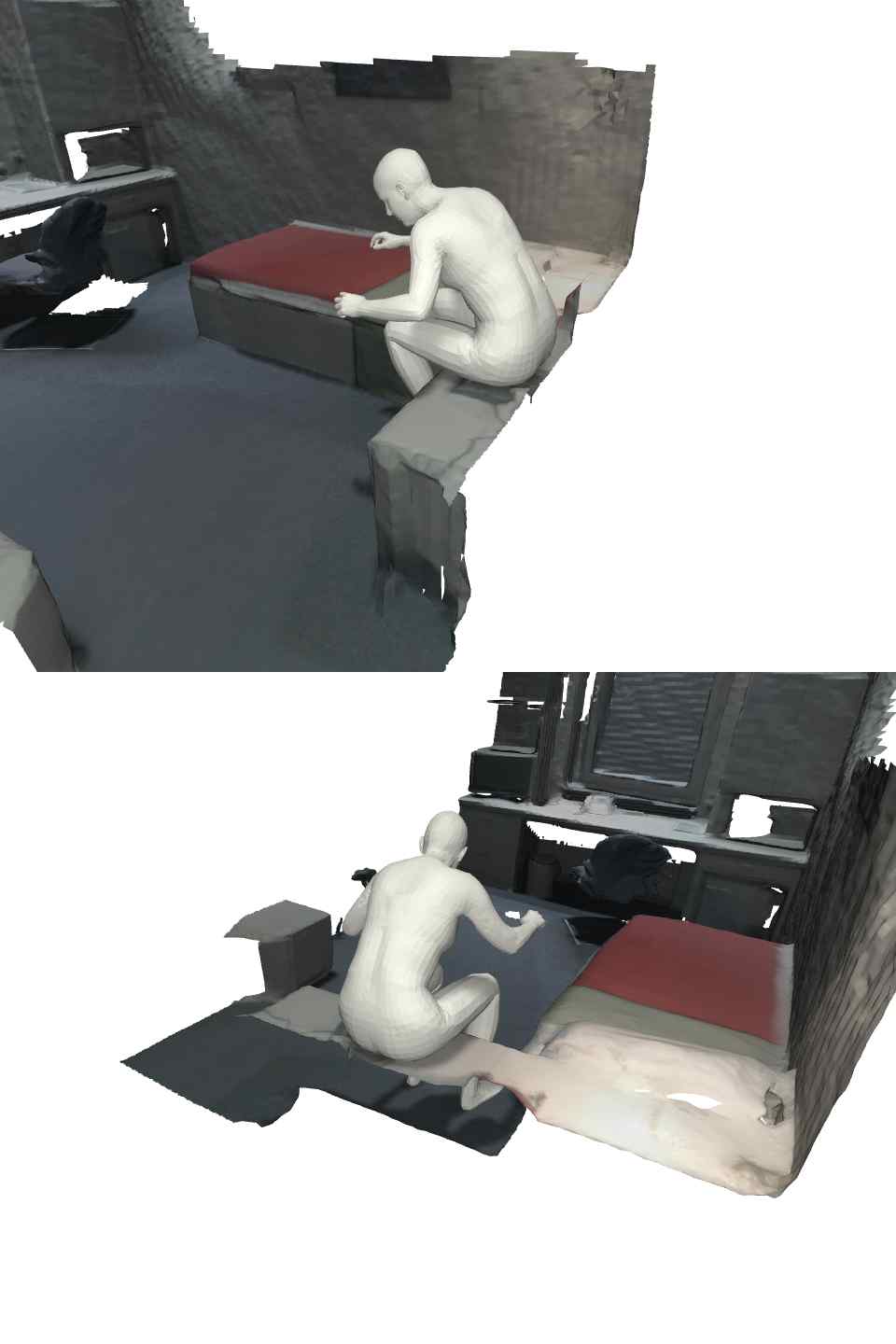}
        \includegraphics[width=0.15\textwidth]{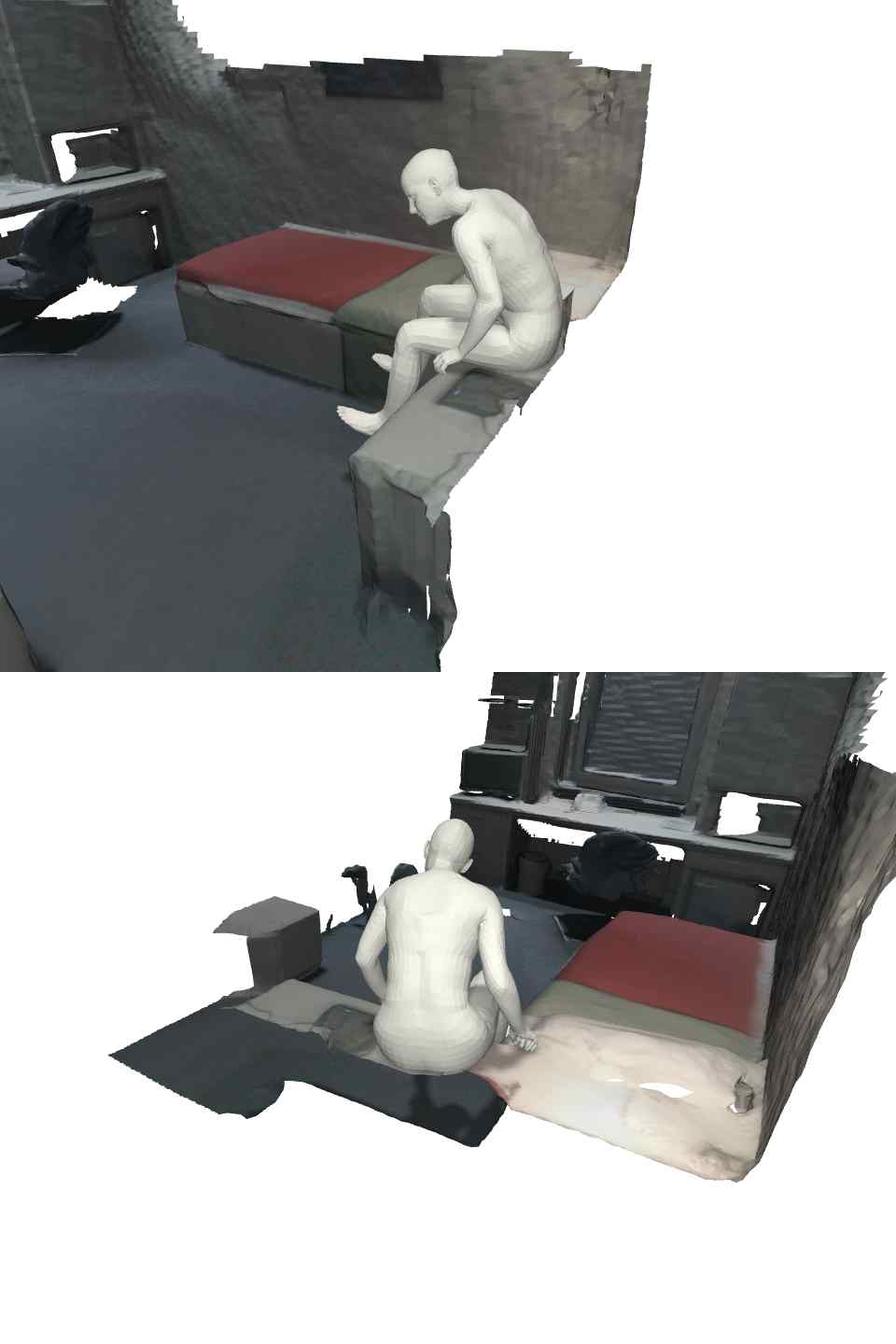}
        \includegraphics[width=0.15\textwidth]{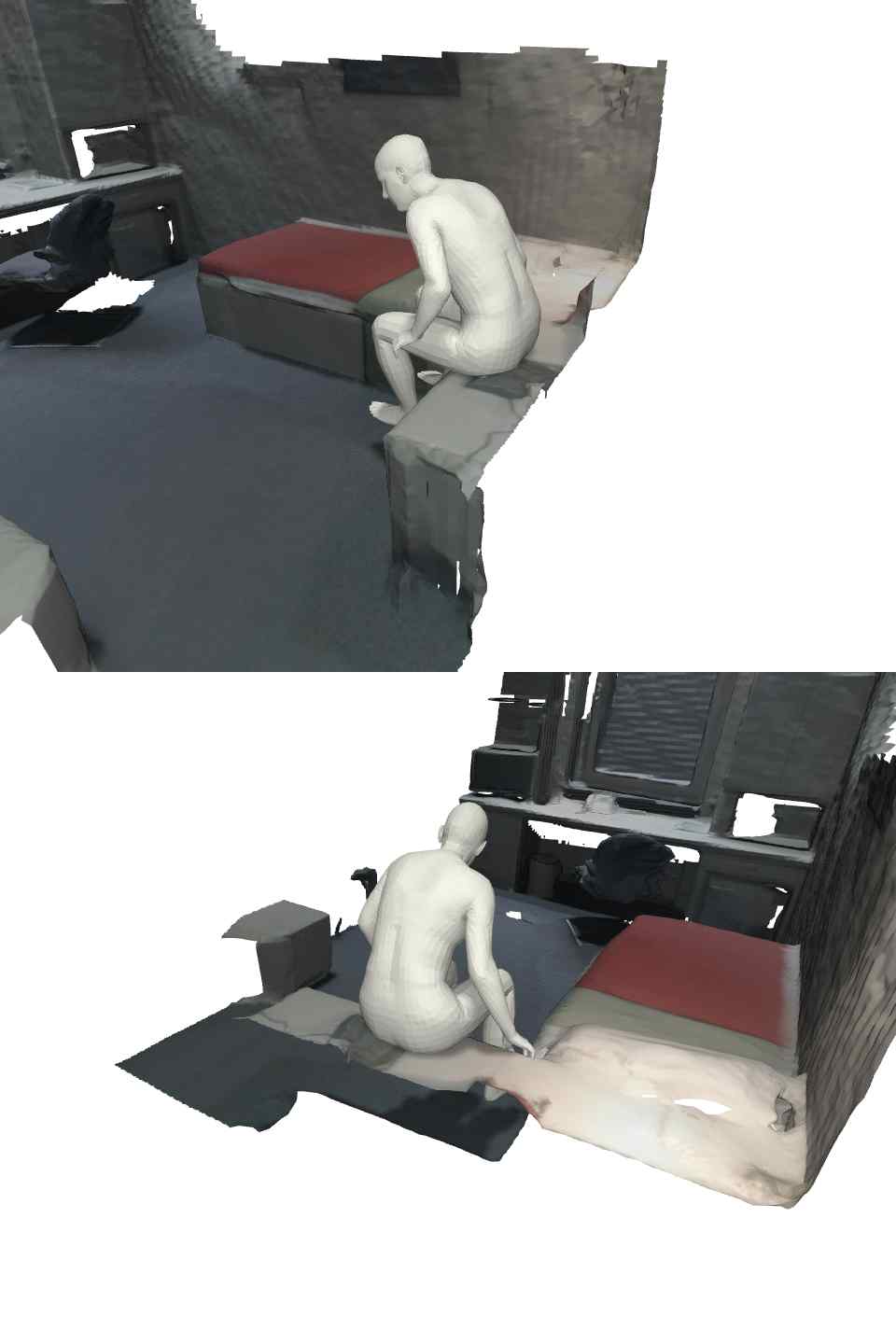}
        \includegraphics[width=0.15\textwidth]{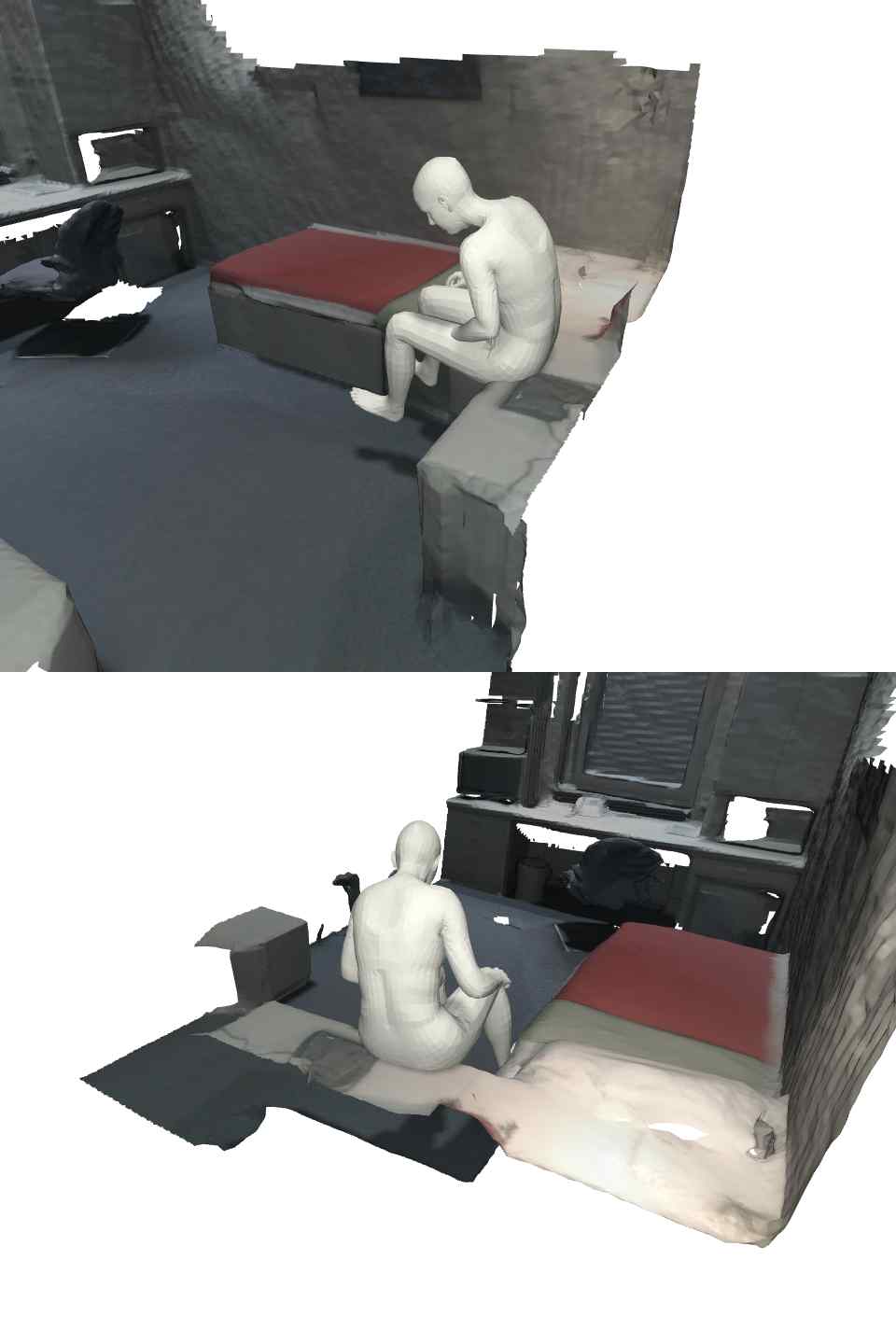}
        \includegraphics[width=0.15\textwidth]{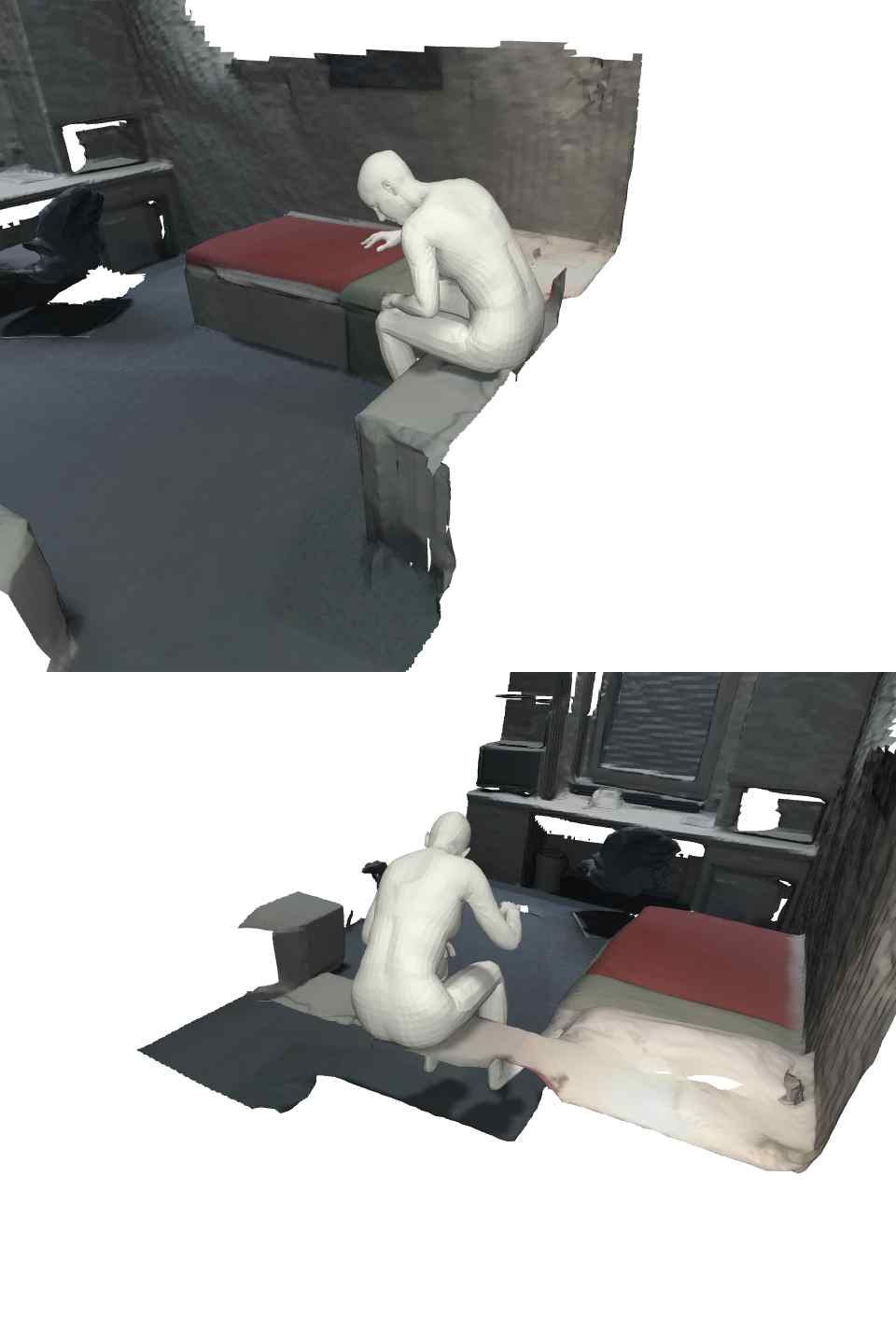}
        \includegraphics[width=0.15\textwidth]{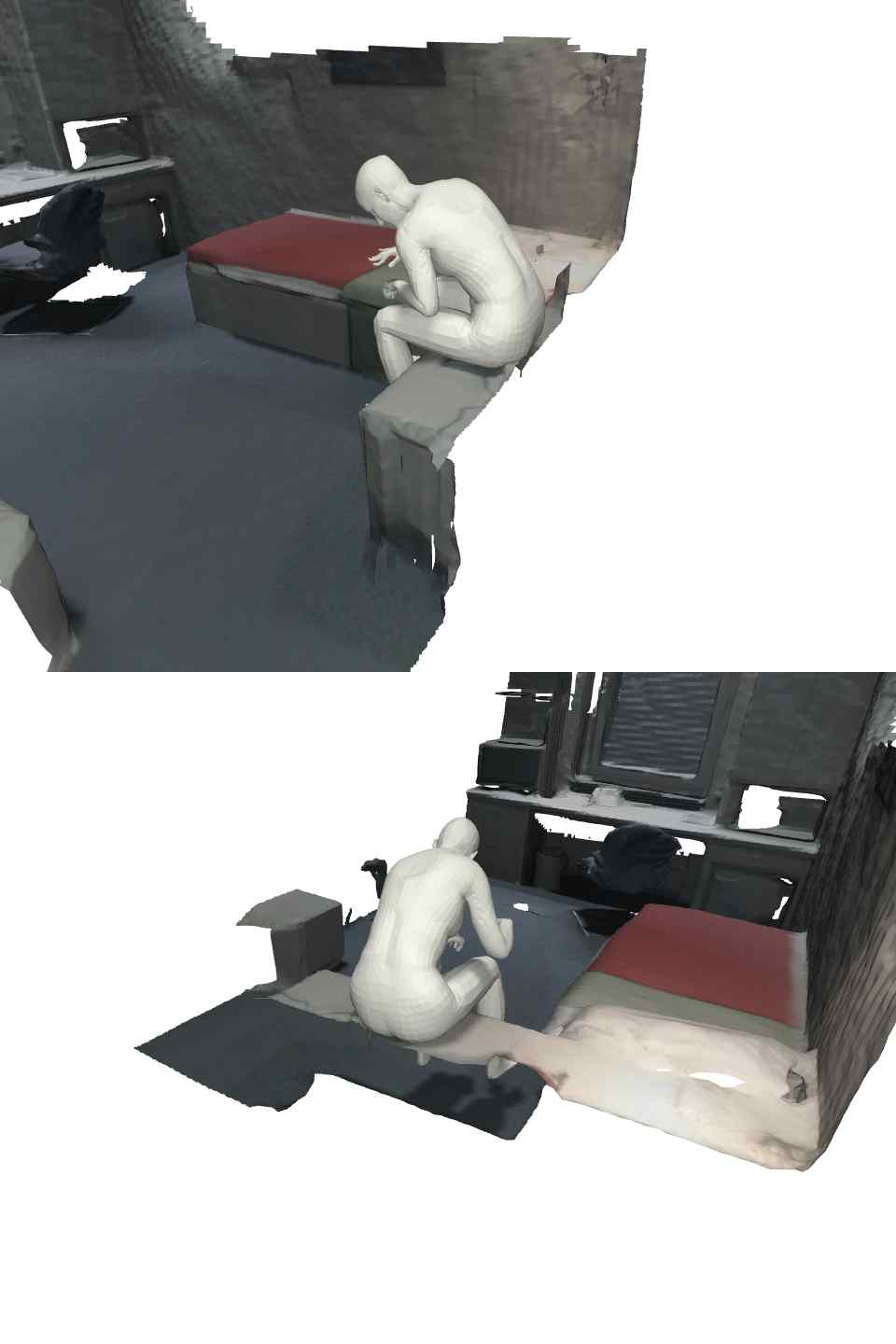}
    \end{subfigure}
    \caption{\textcolor{blue}{Randomly} sampled interactions from our method. Each interaction is rendered with two views.}
    \label{fig:random}
\end{figure}

We show more random interaction samples from our method in \cref{fig:random}. Our method generates natural and diverse human-scene interactions.

\subsection{Interaction Refinement}
\begin{figure}[t]
    \centering
    \begin{subfigure}[t]{0.22\textwidth}
        \includegraphics[width=\textwidth]{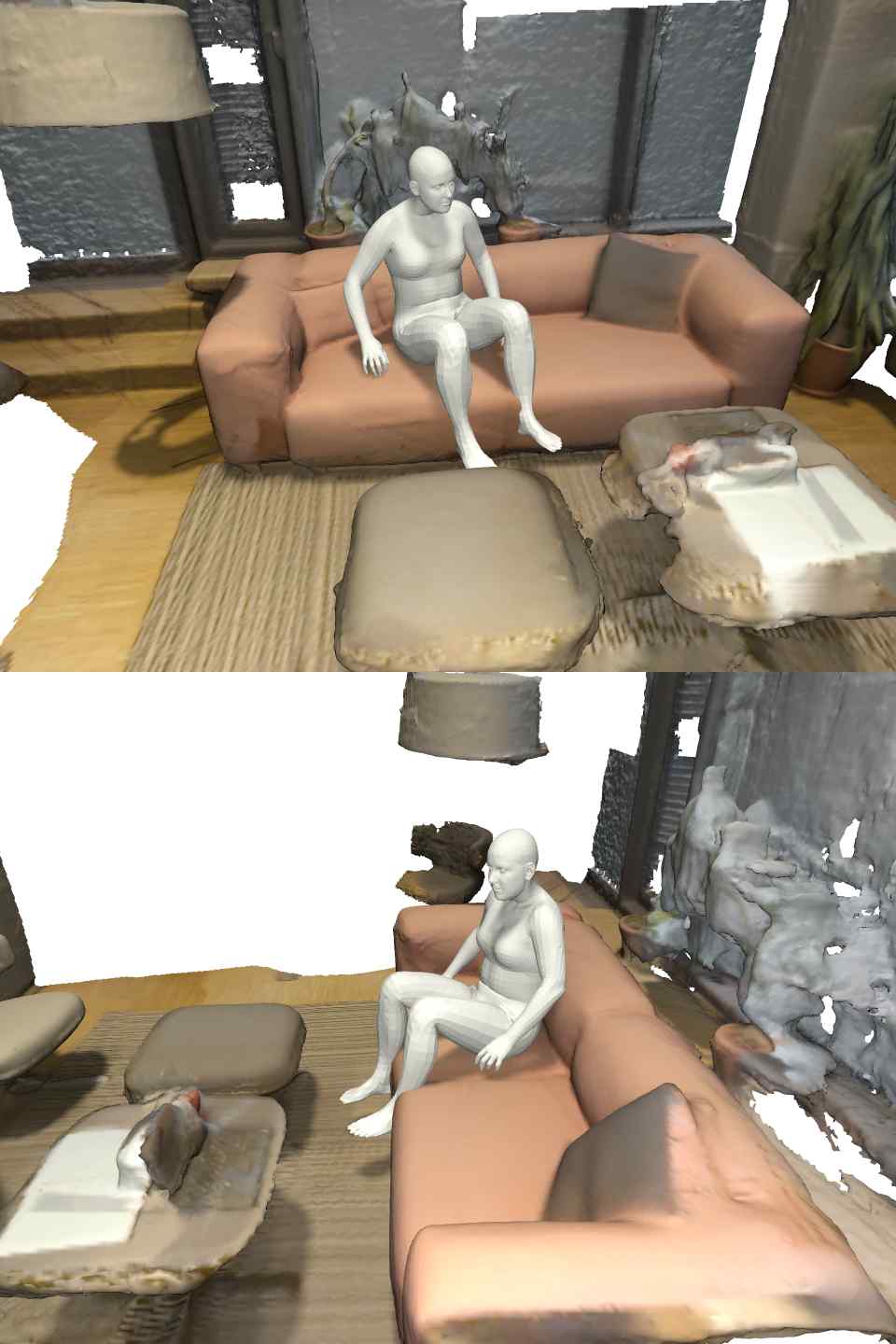}
       \caption{before}
    \end{subfigure}
    \begin{subfigure}[t]{0.22\textwidth}
         \includegraphics[width=\textwidth]{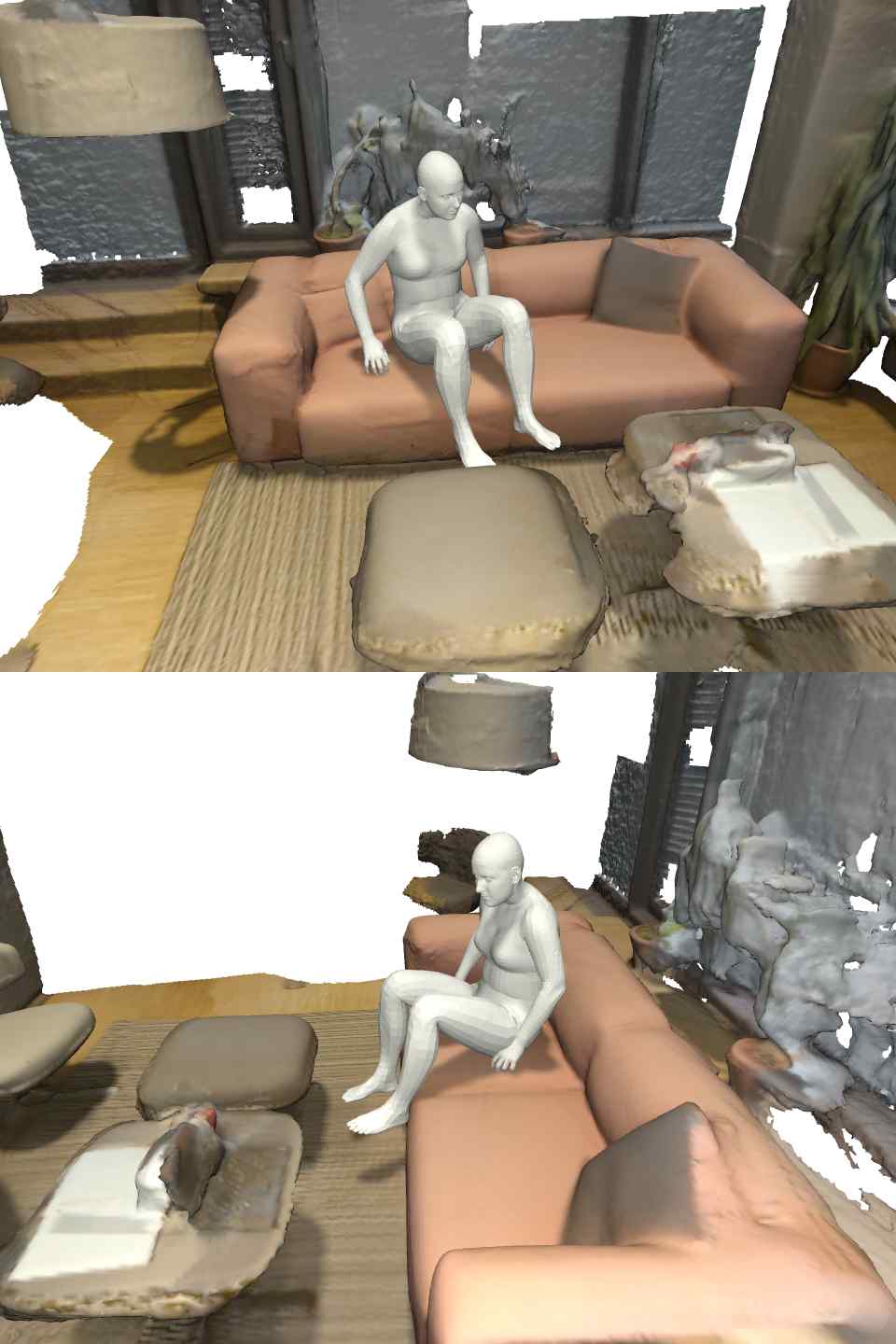}
         \caption{after}
    \end{subfigure}
    \vrule
    \hspace{1pt}
    \begin{subfigure}[t]{0.22\textwidth}
         \includegraphics[width=\textwidth]{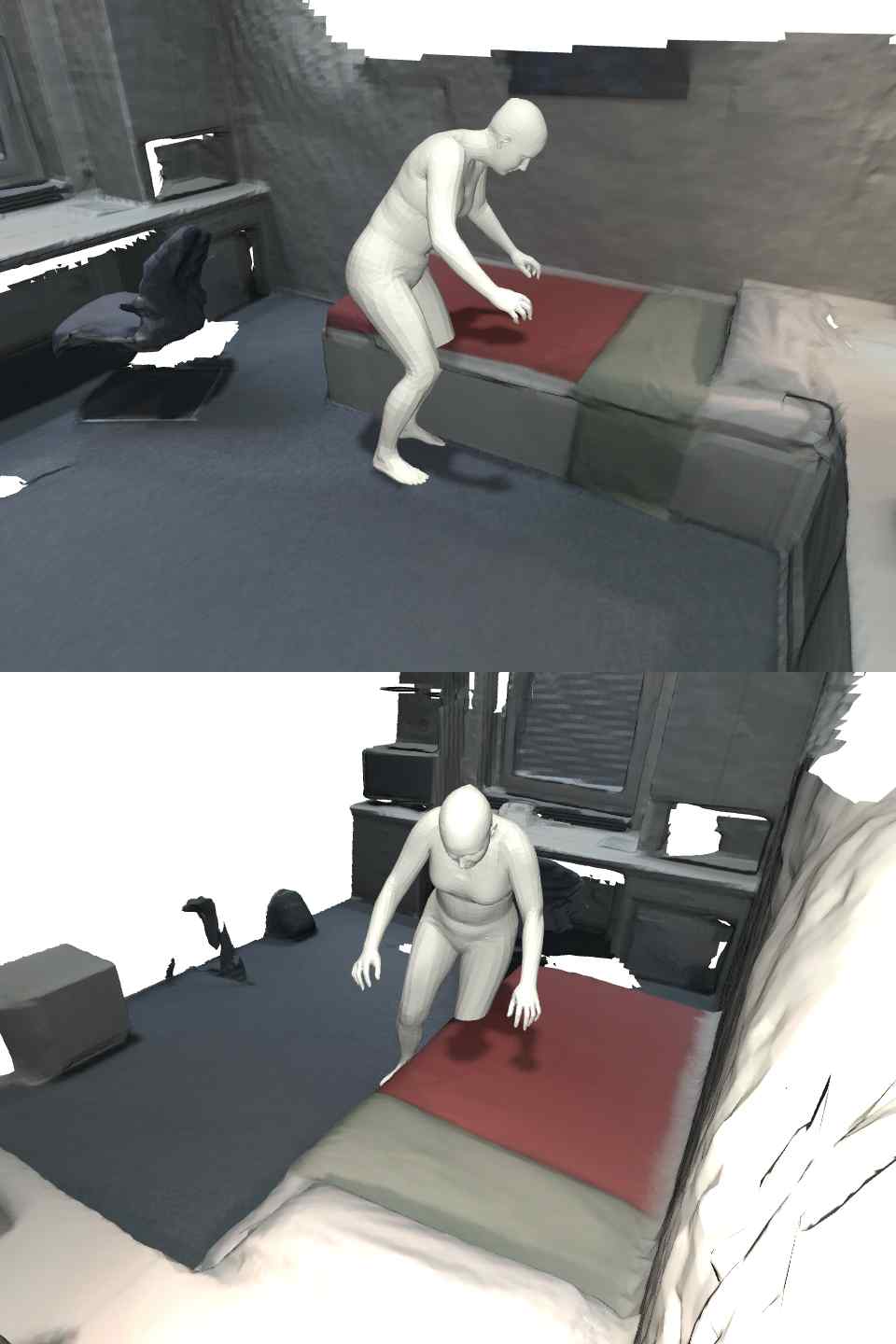}
         \caption{before}
    \end{subfigure}
    \begin{subfigure}[t]{0.22\textwidth}
        \includegraphics[width=\textwidth]{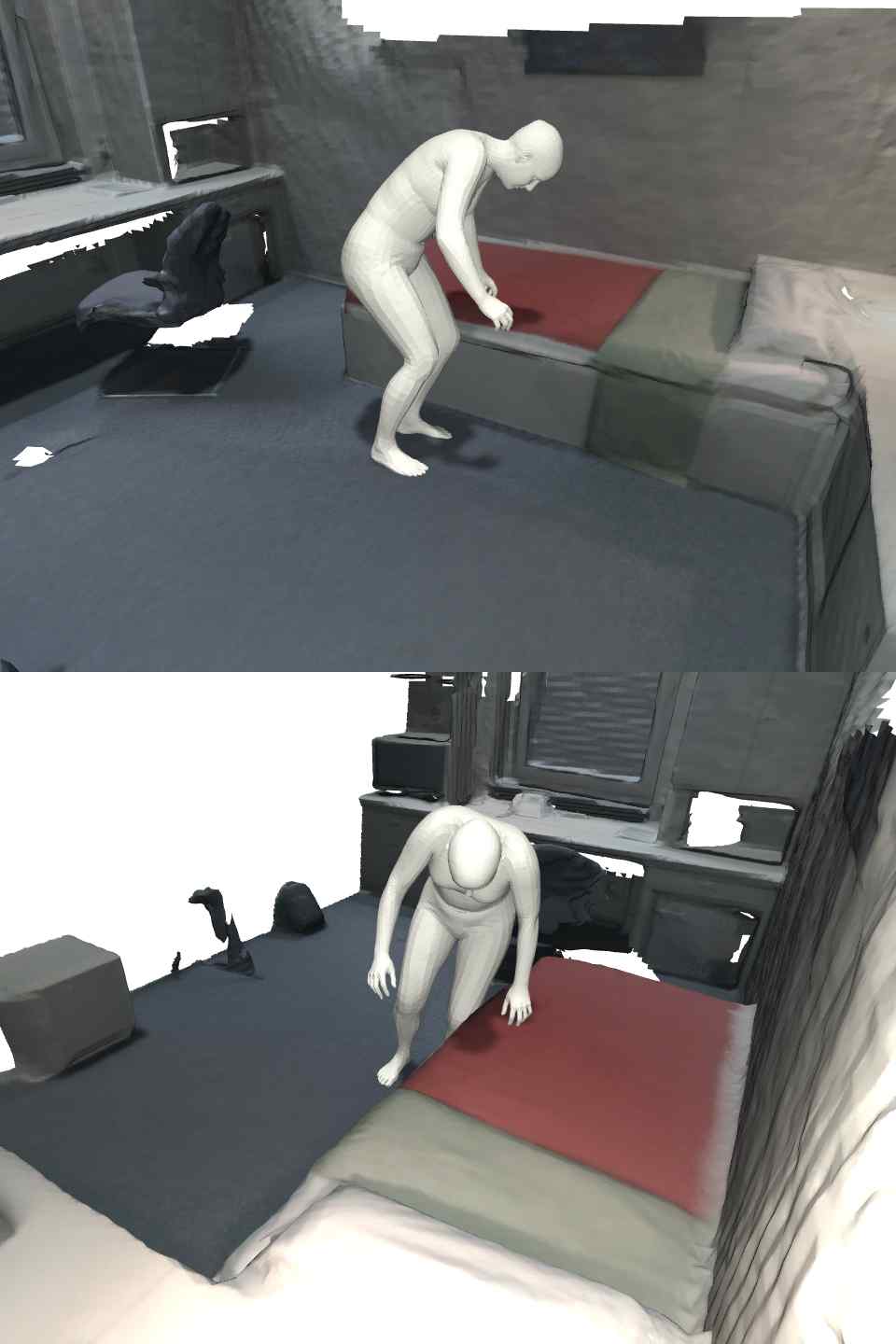}
        \caption{after}
    \end{subfigure}
    
    \caption{Illustration of interaction-based optimization where (b) and (d) are the optimized results of (a) and (c). The human-scene penetration and contact are improved after the optimization.}
    \label{fig:refine}
\end{figure}

We demonstrate how the interaction-based optimization improves human-scene penetration and contact in ~\cref{fig:refine}.

\subsection{Compositional Interaction Generation}
\begin{figure}[t]
    \centering
    \begin{subfigure}[t]{\textwidth}
        \rotatebox{90}{\tiny sit on sofa+touch table}
        \includegraphics[width=0.15\textwidth]{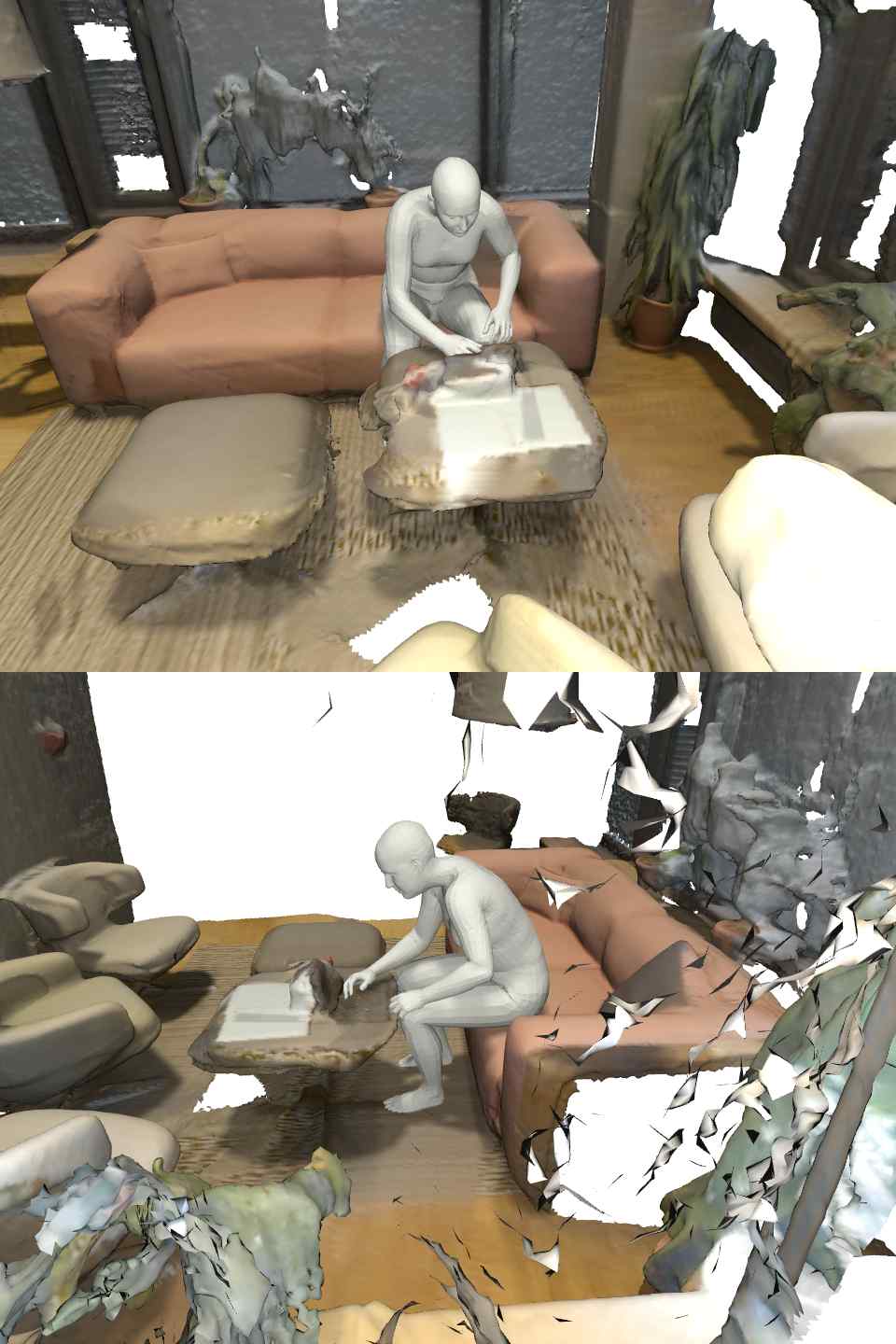}
        \includegraphics[width=0.15\textwidth]{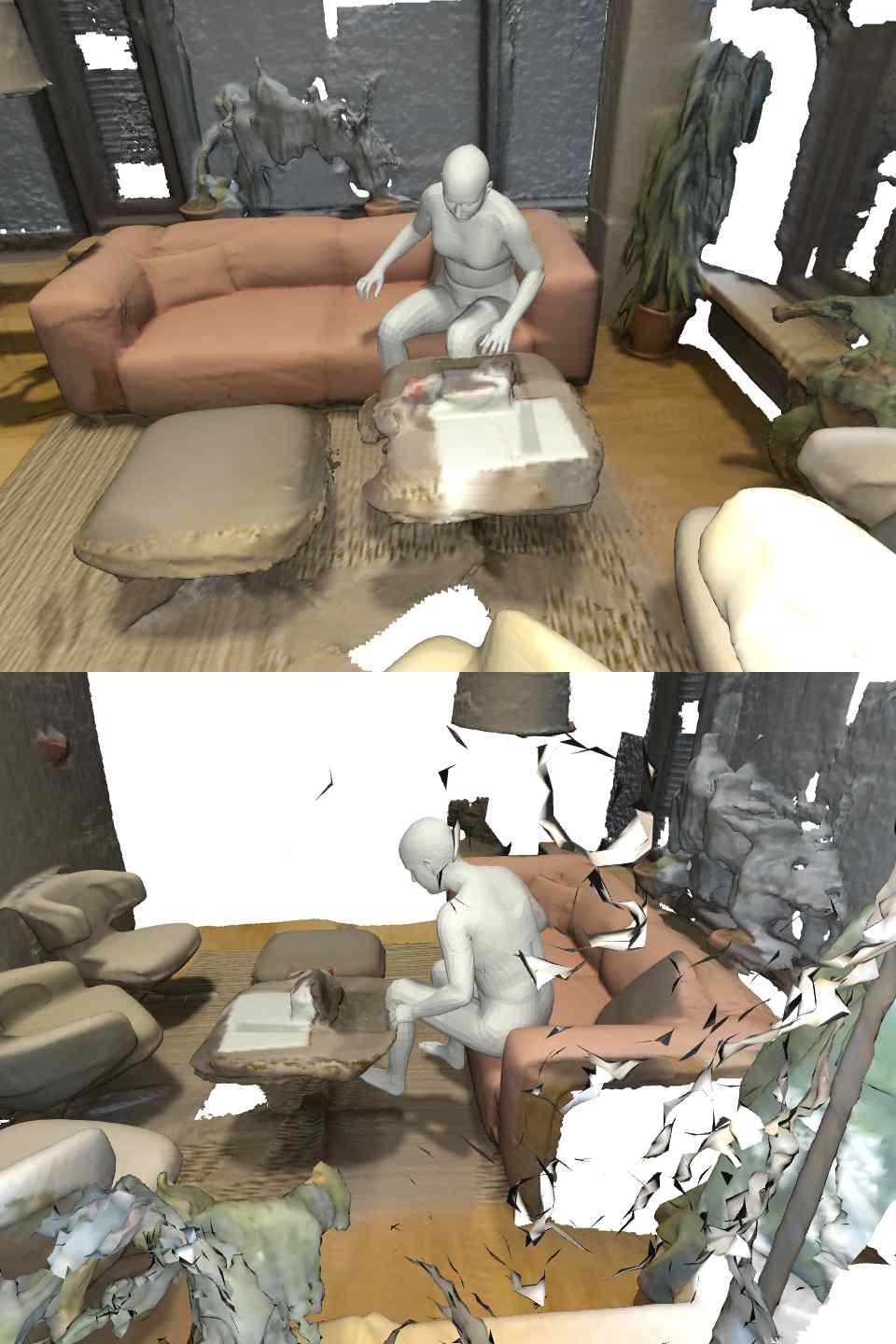}
        \includegraphics[width=0.15\textwidth]{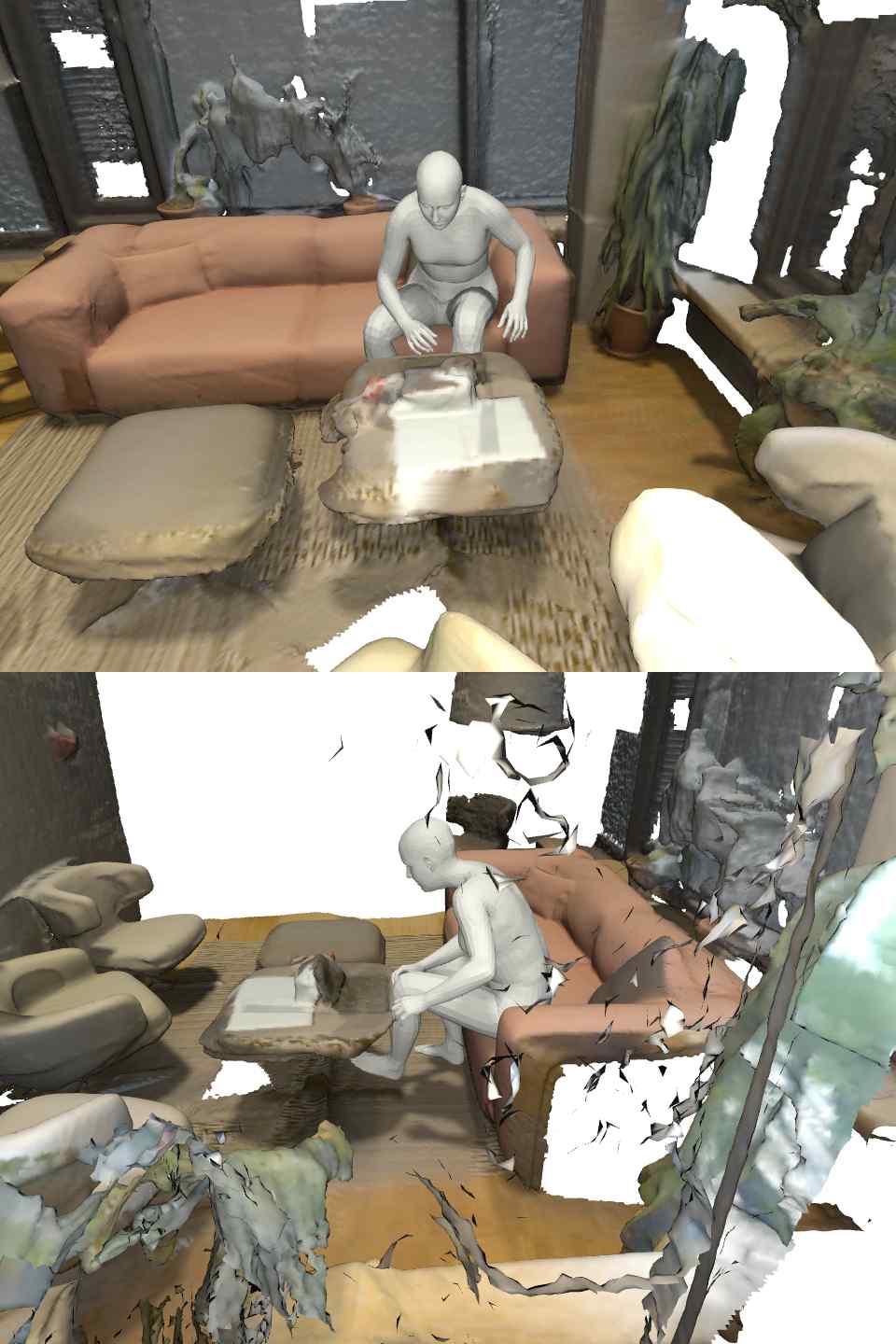}
        \includegraphics[width=0.15\textwidth]{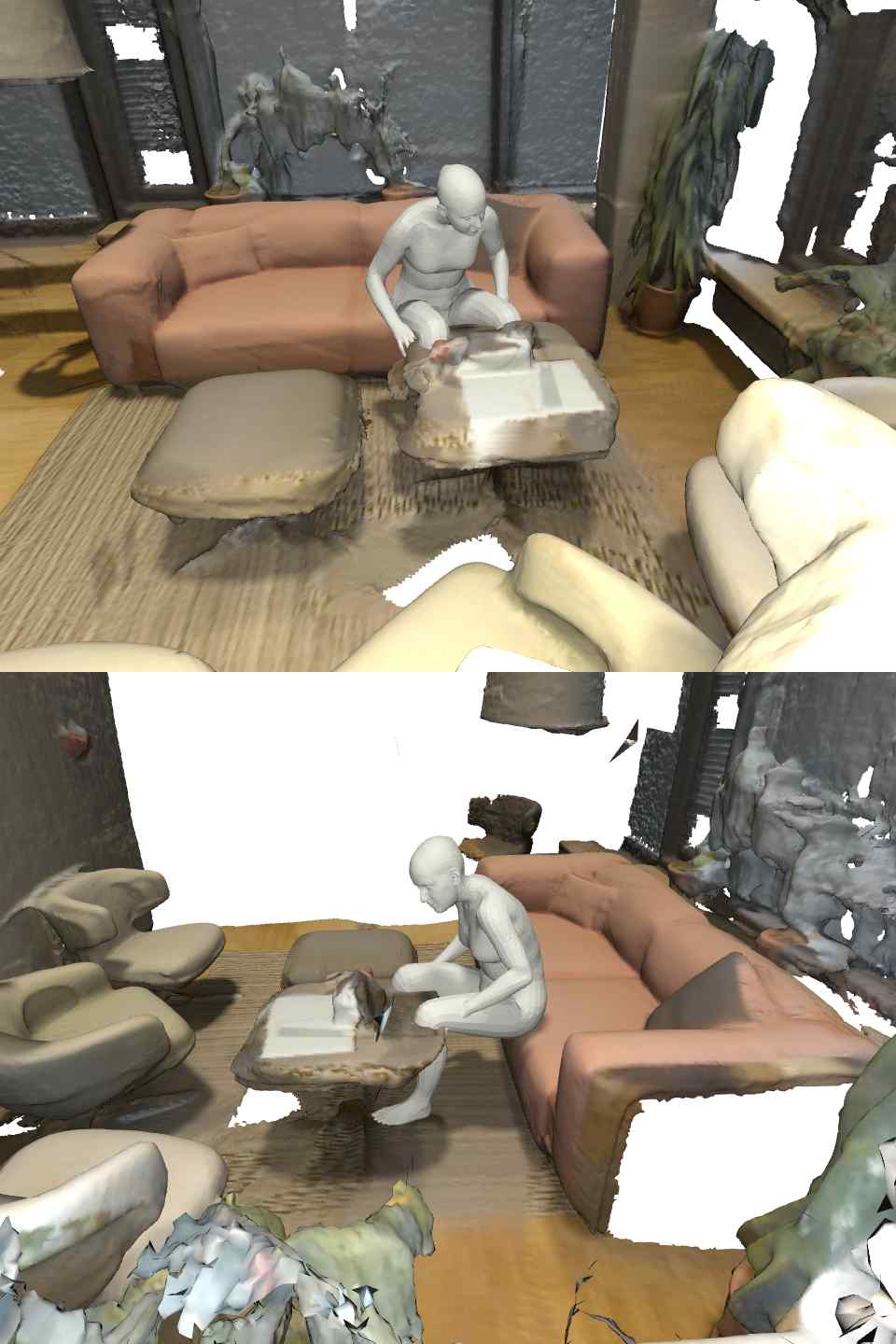}
        \includegraphics[width=0.15\textwidth]{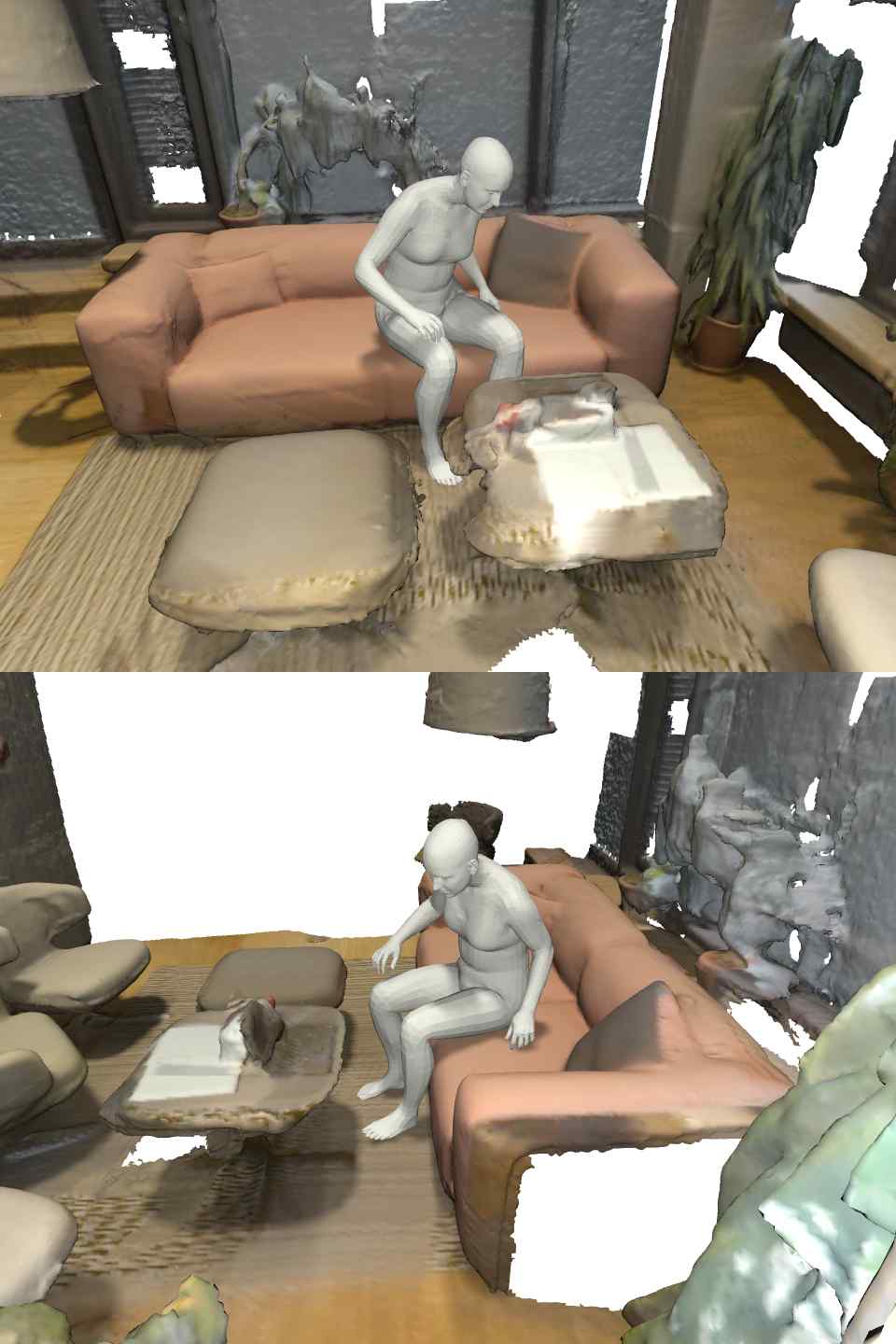}
        \includegraphics[width=0.15\textwidth]{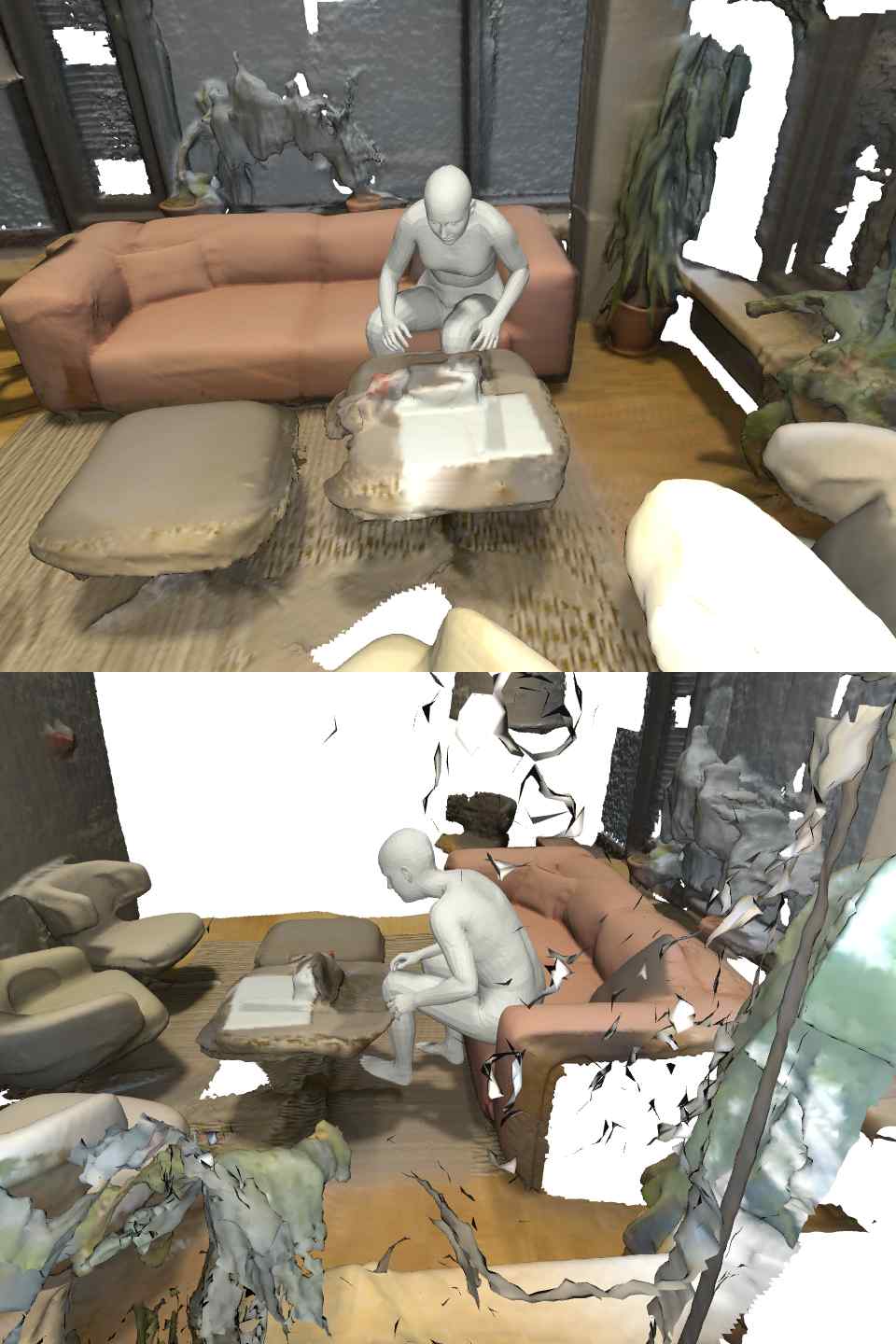}
    \end{subfigure}
    
    \vspace{1em}
    
    \begin{subfigure}[t]{\textwidth}
        \rotatebox{90}{\tiny stand on floor+touch bed}
        \includegraphics[width=0.15\textwidth]{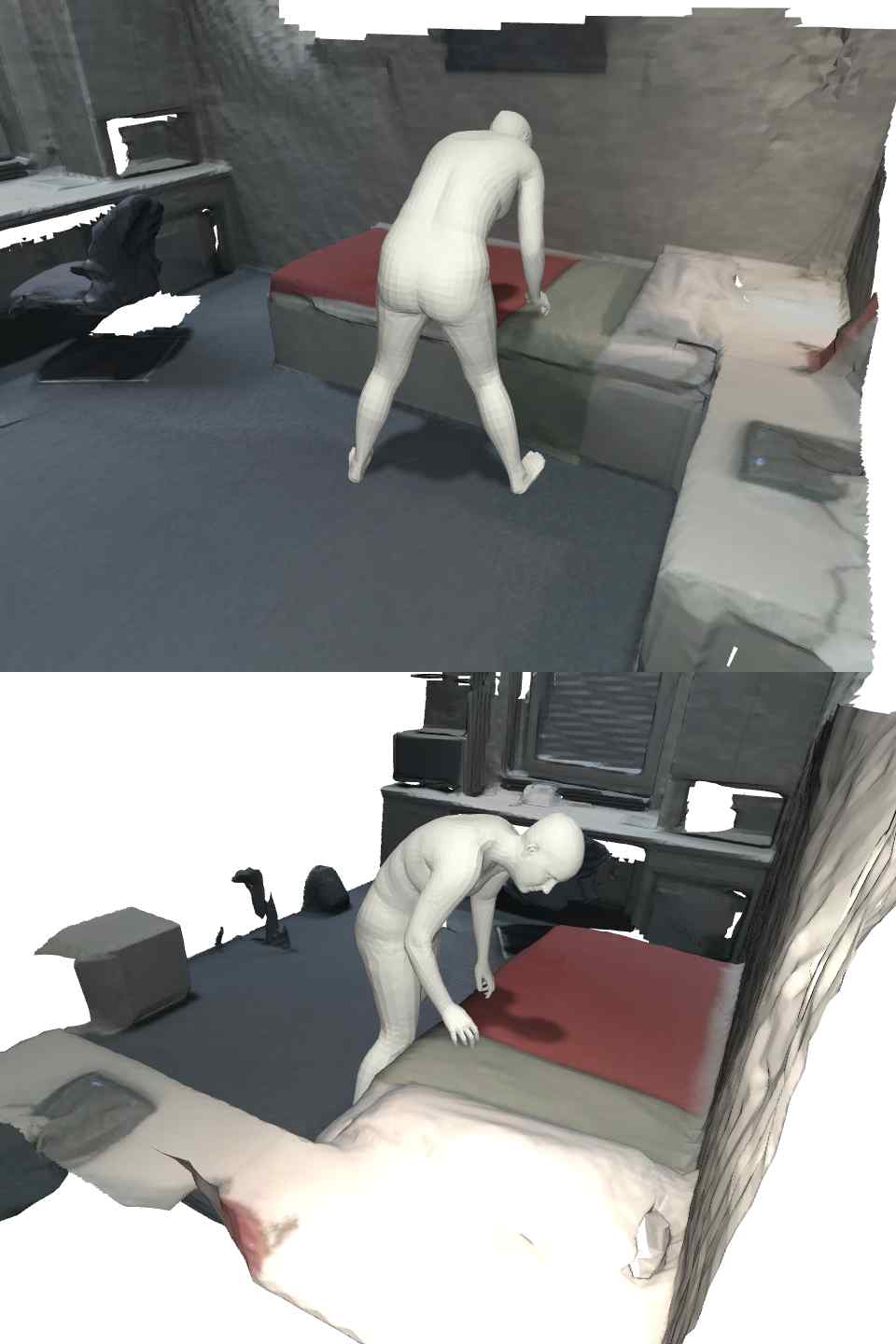}
        \includegraphics[width=0.15\textwidth]{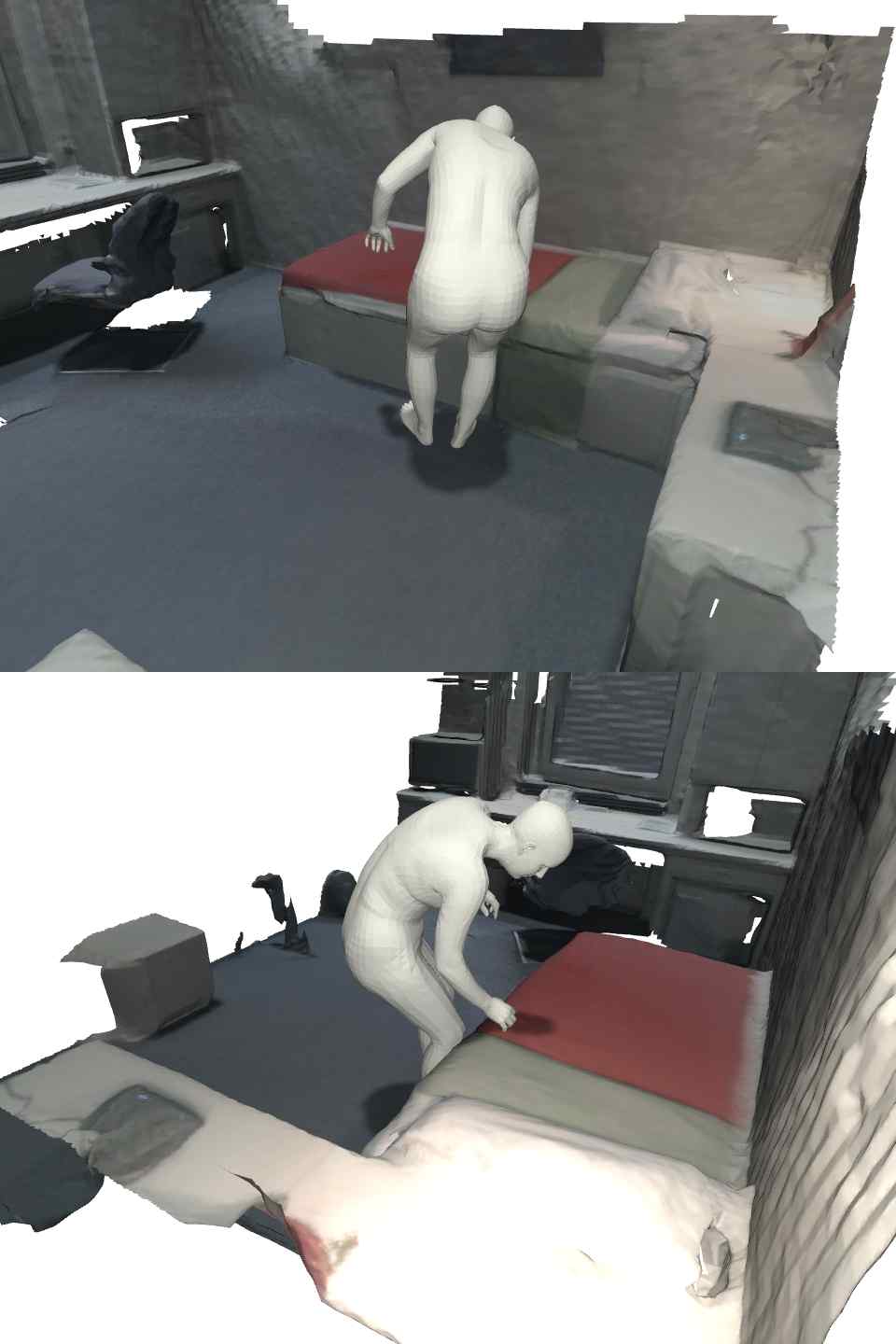}
        \includegraphics[width=0.15\textwidth]{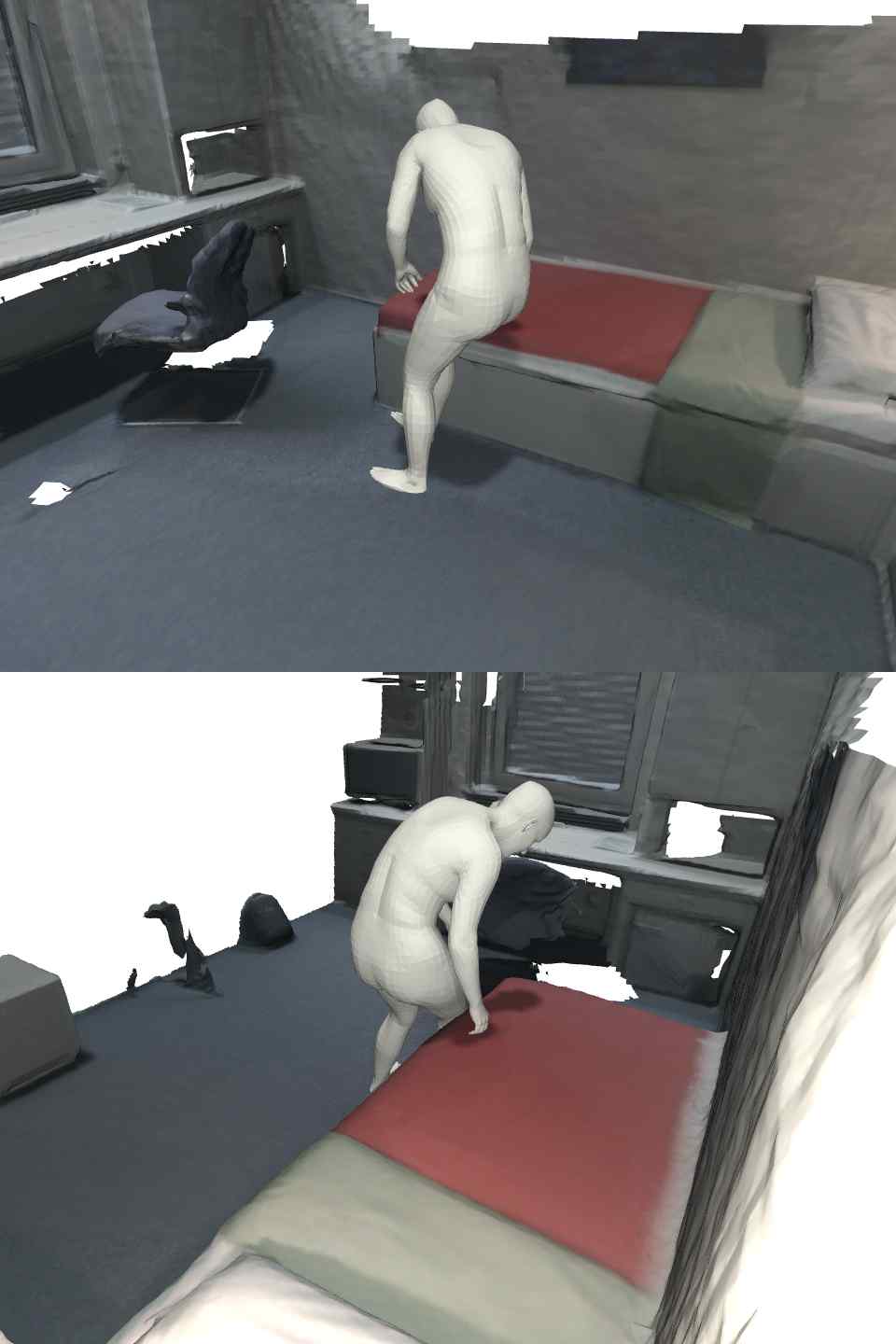}
        \includegraphics[width=0.15\textwidth]{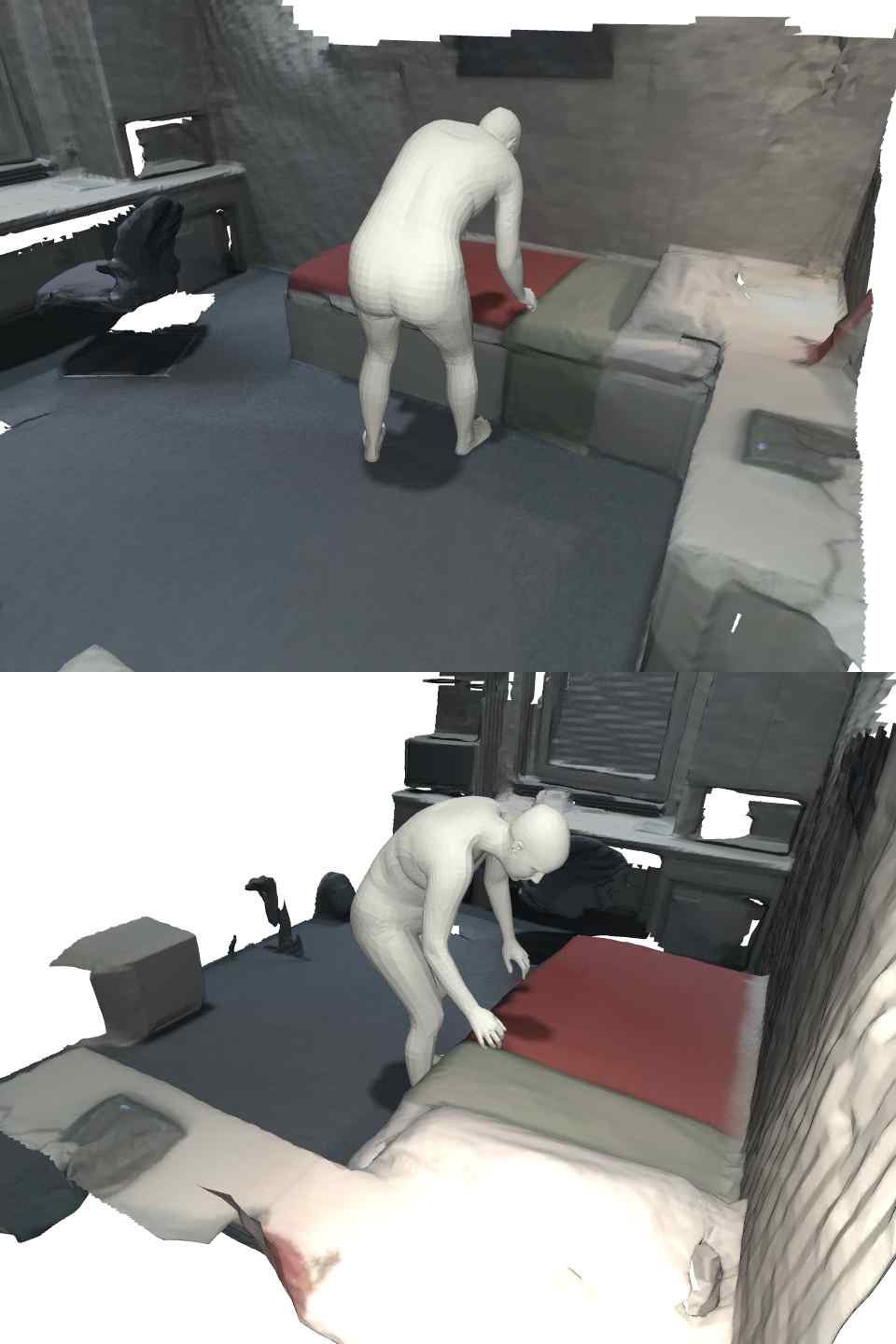}
        \includegraphics[width=0.15\textwidth]{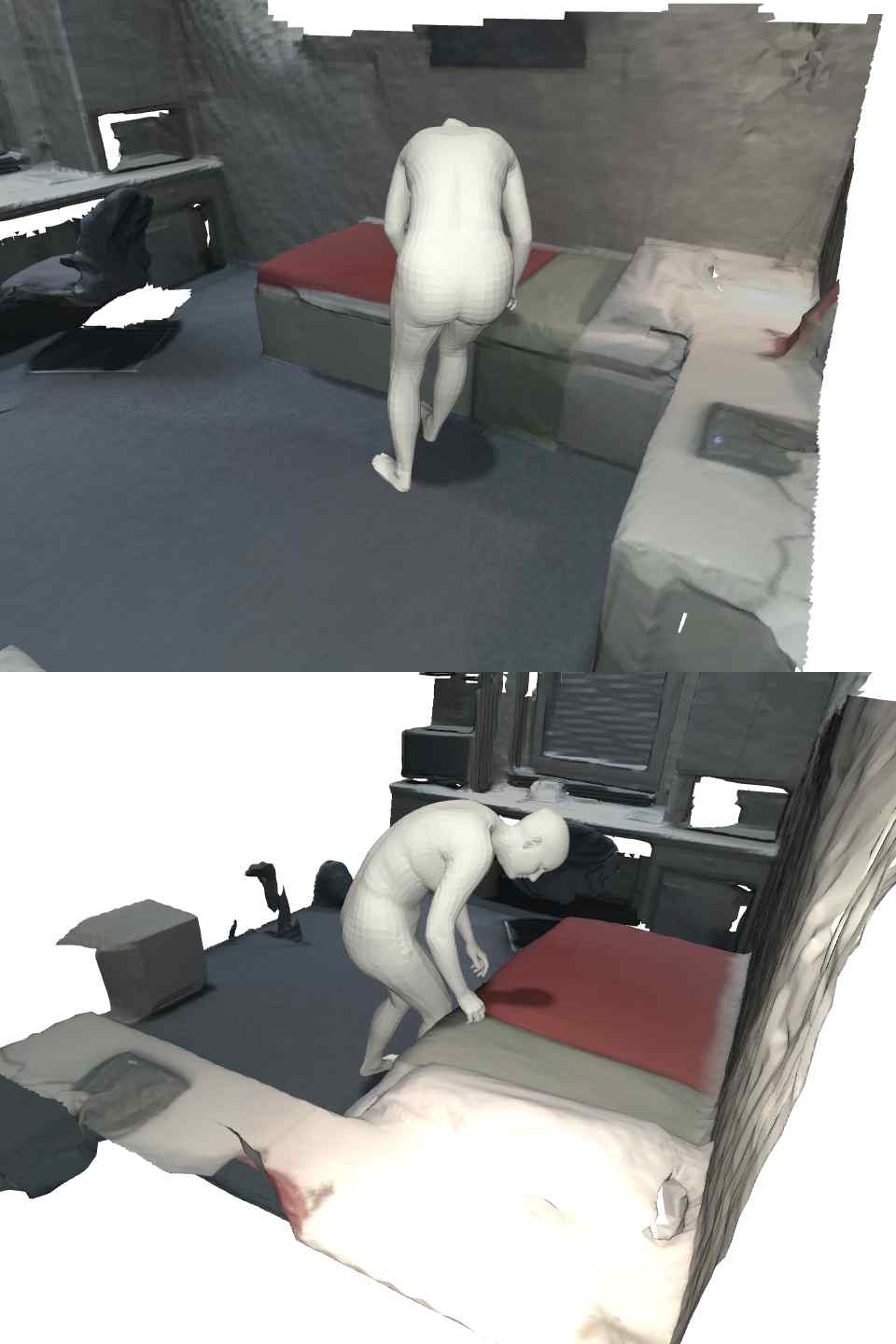}
        \includegraphics[width=0.15\textwidth]{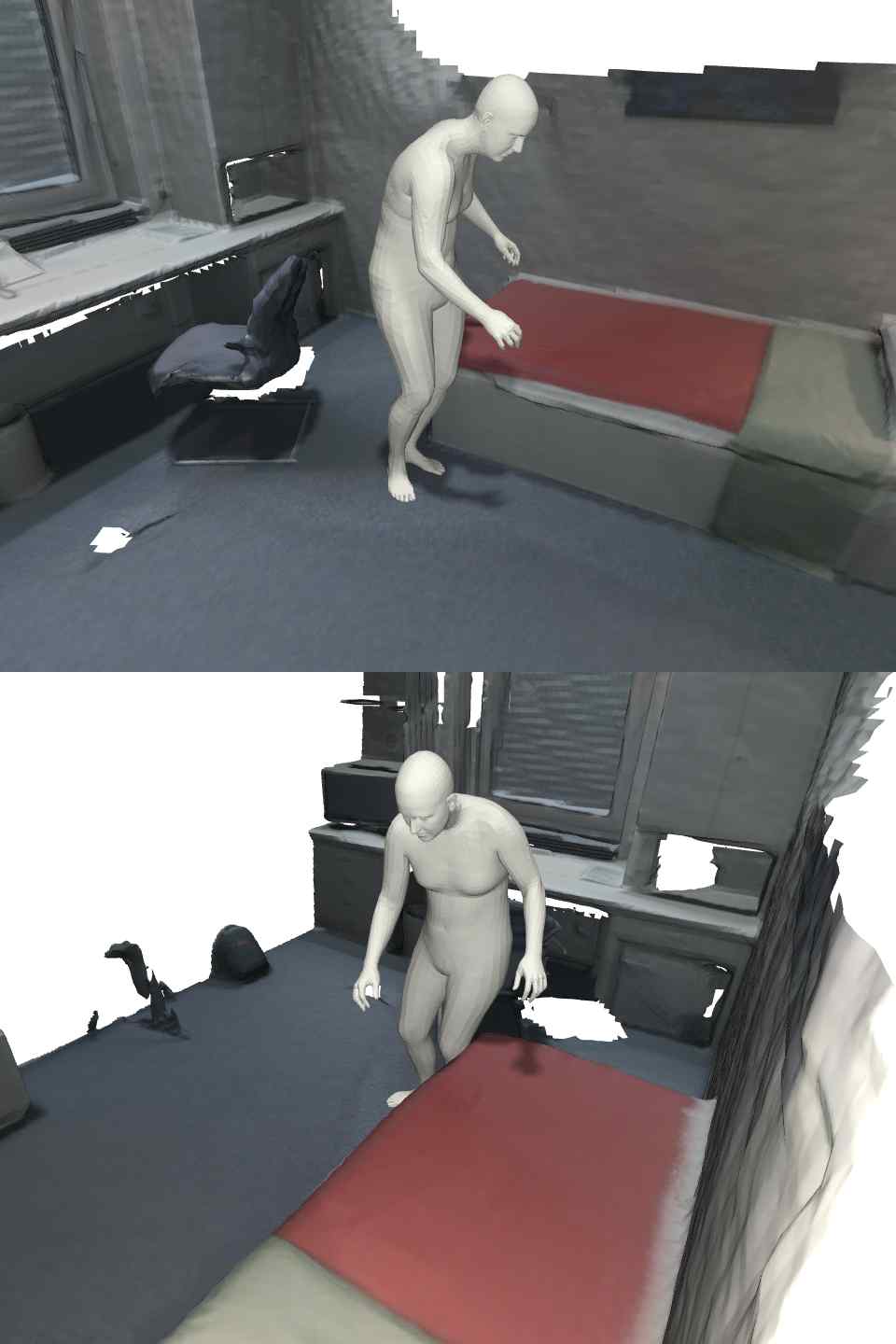}
    \end{subfigure}
    
    \vspace{1em}
    
    \begin{subfigure}[t]{\textwidth}
        \rotatebox{90}{\tiny sit on table+touch seating}
        \includegraphics[width=0.15\textwidth]{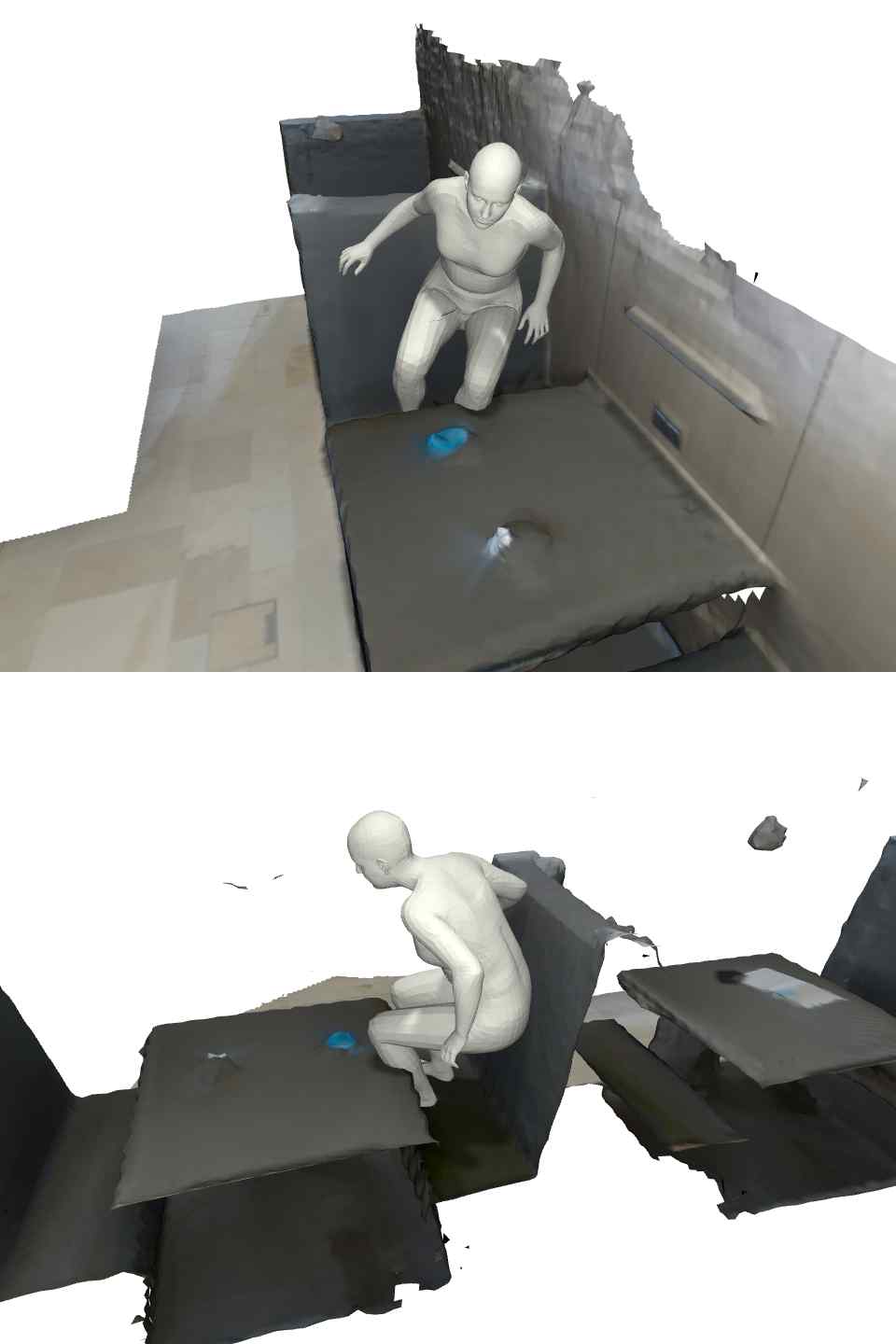}
        \includegraphics[width=0.15\textwidth]{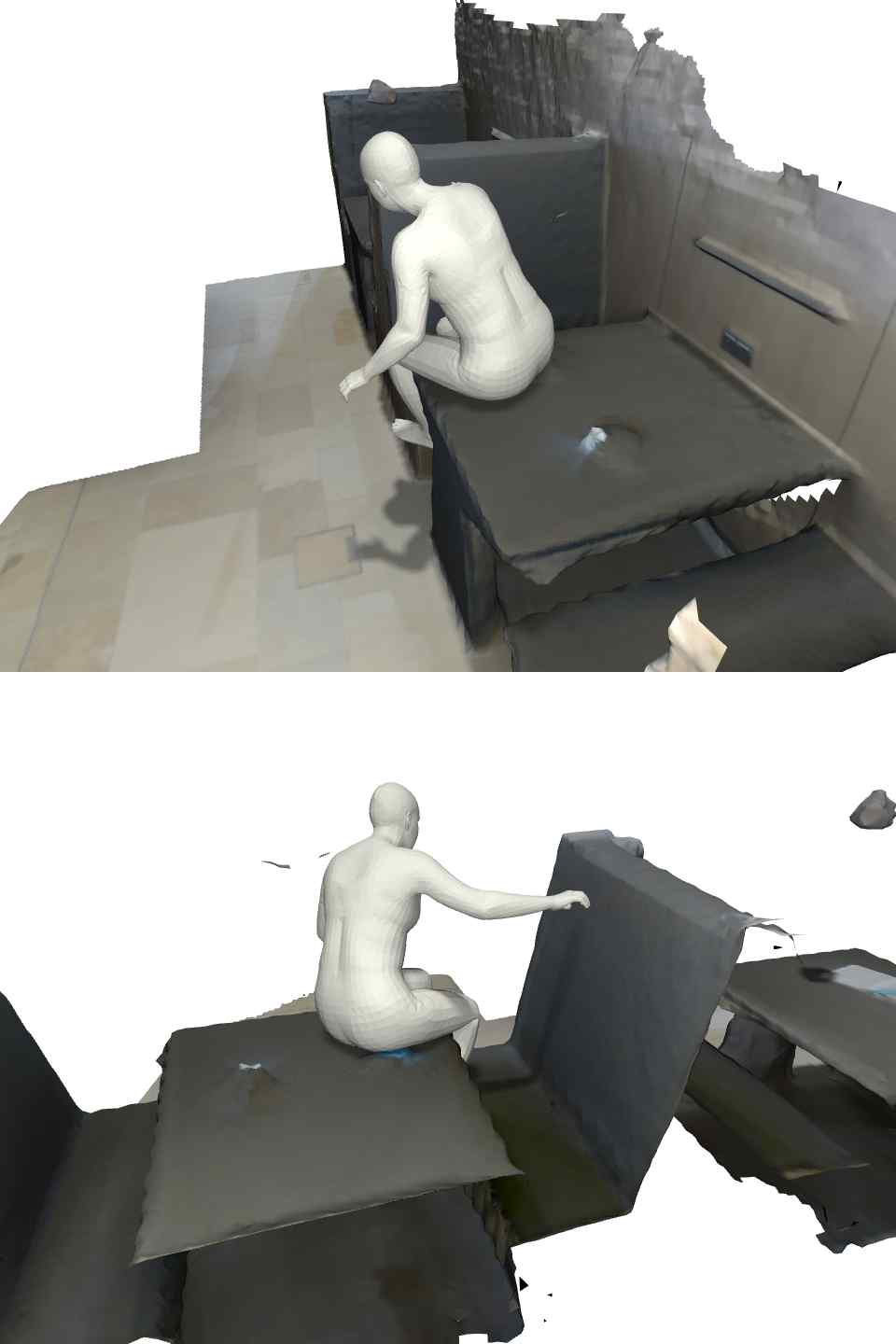}
        \includegraphics[width=0.15\textwidth]{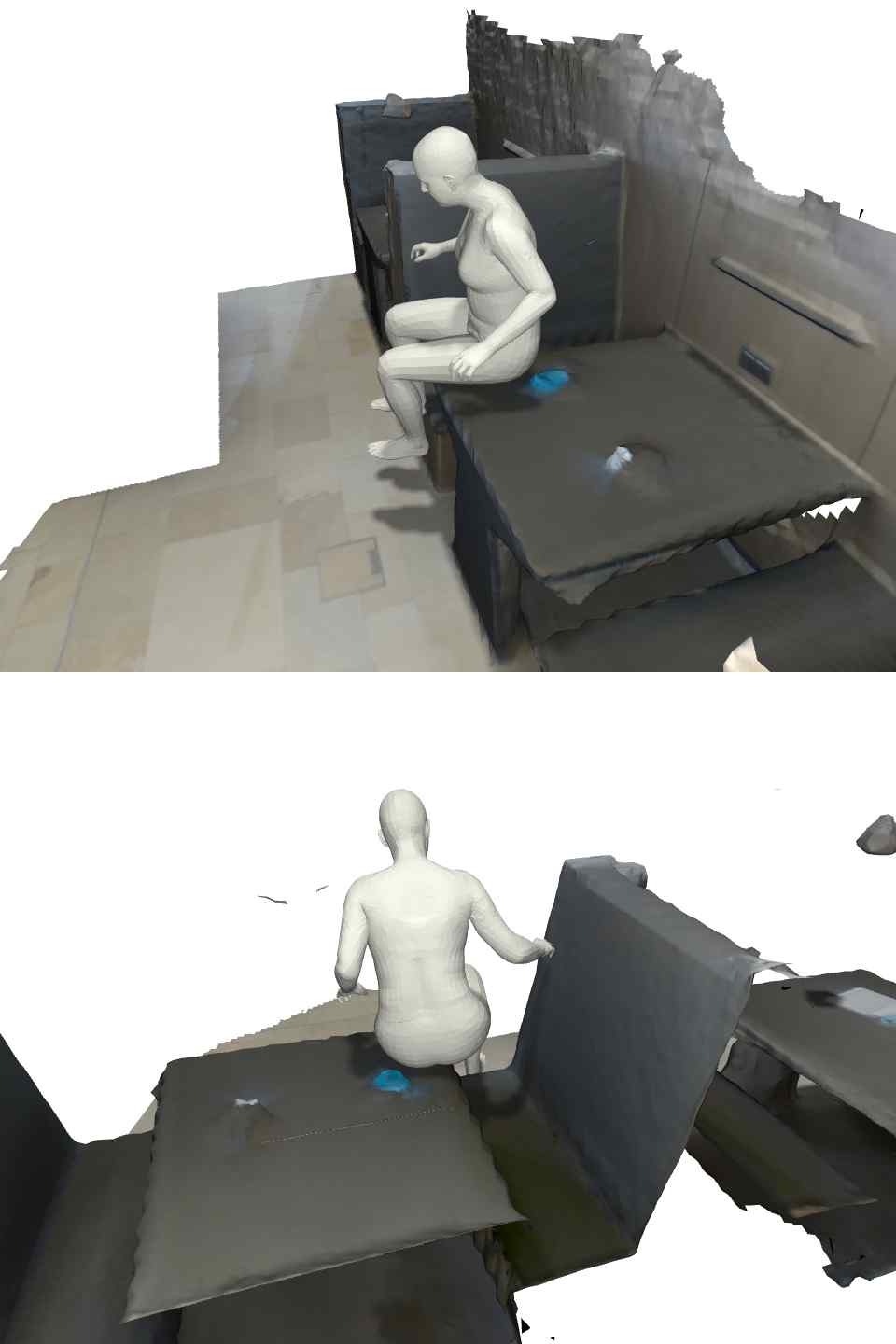}
        \includegraphics[width=0.15\textwidth]{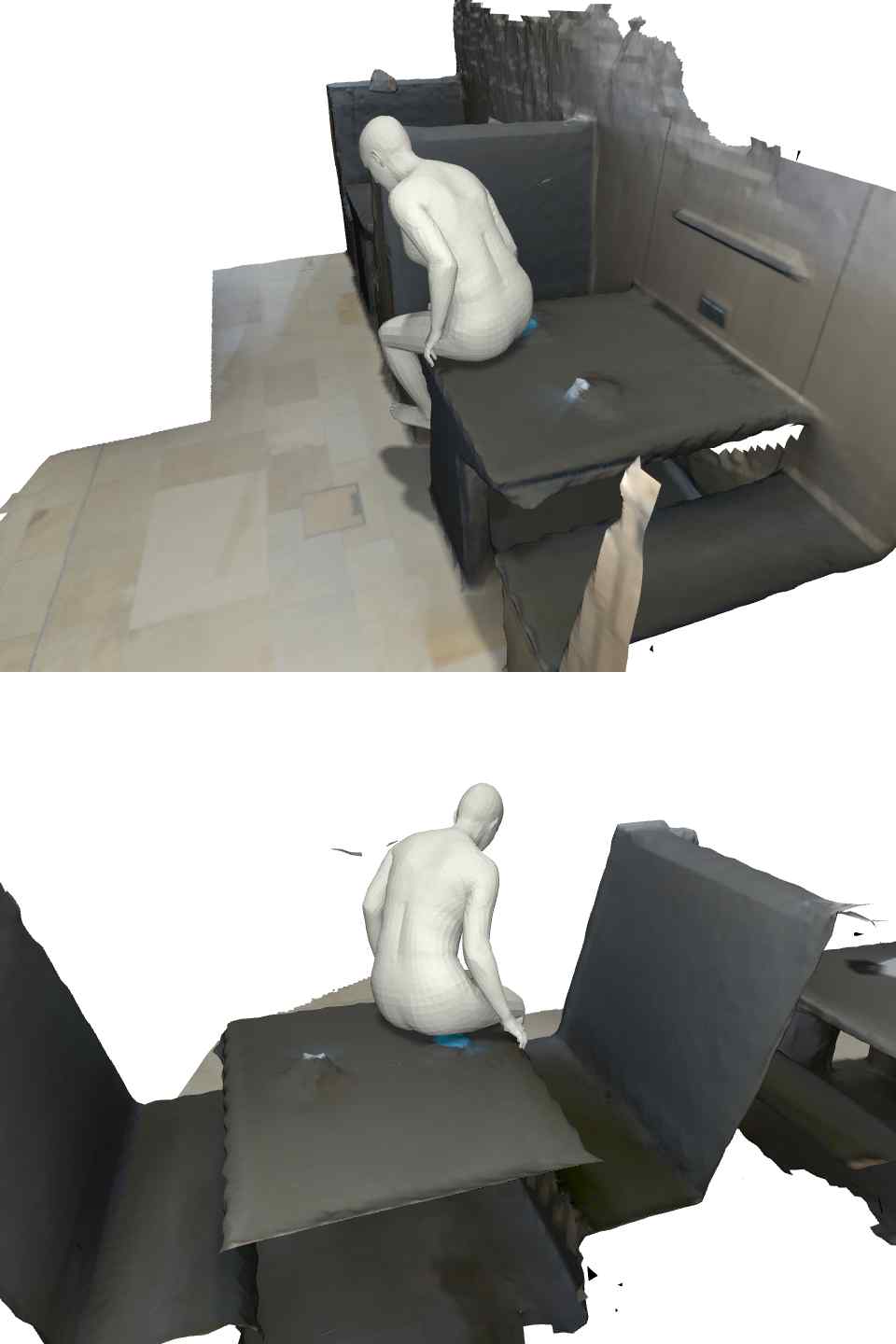}
        \includegraphics[width=0.15\textwidth]{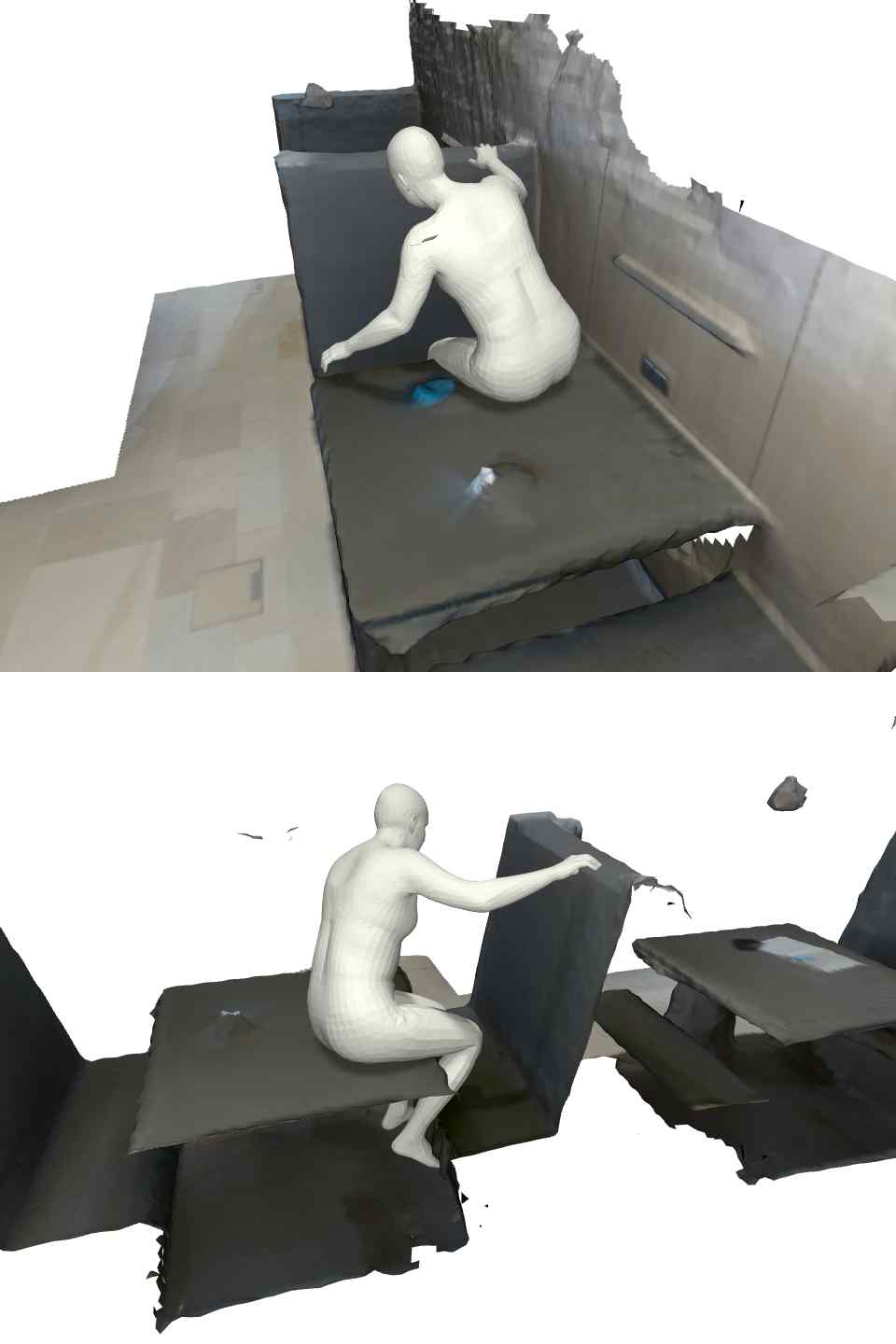}
        \includegraphics[width=0.15\textwidth]{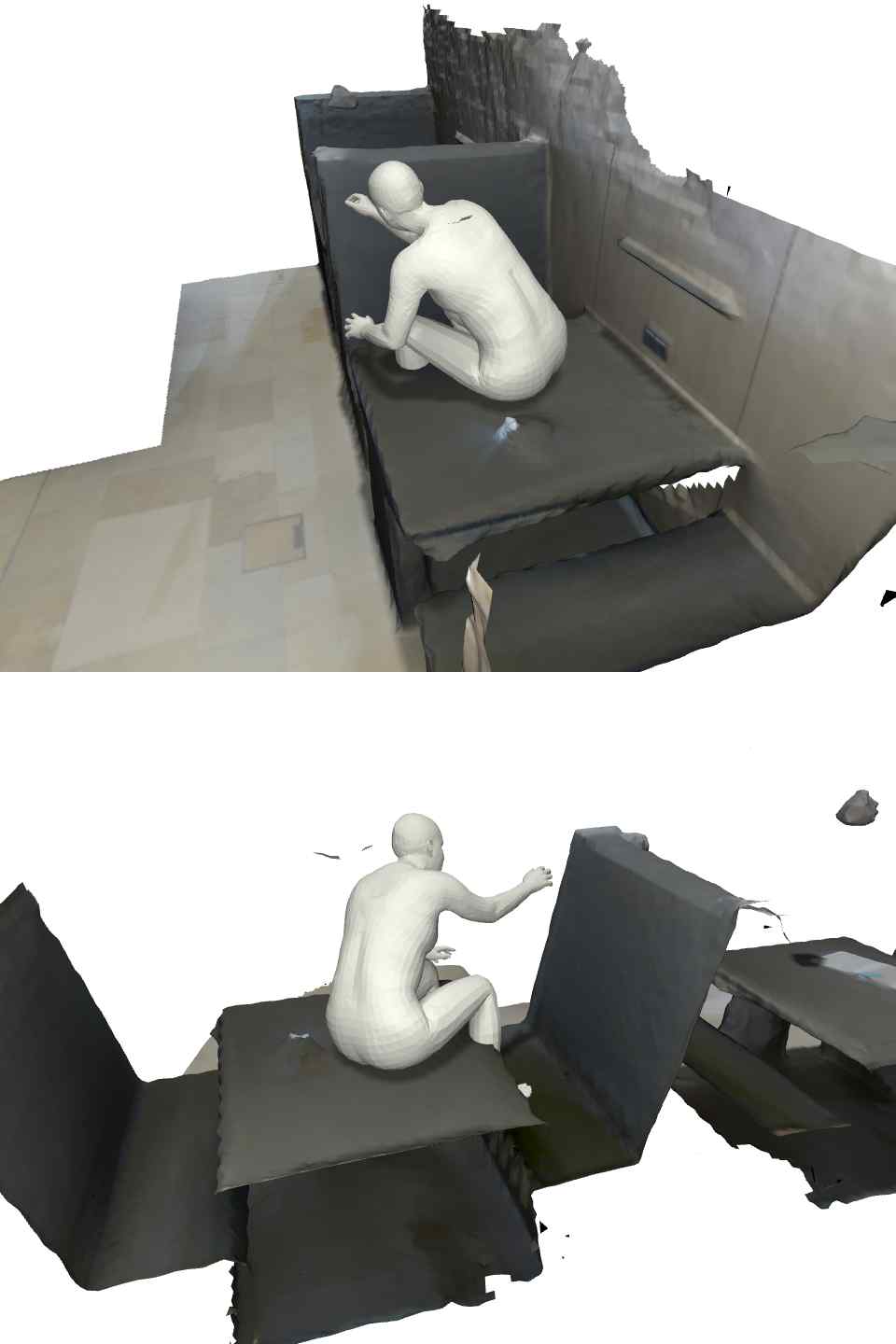}
    \end{subfigure}
    
    \vspace{1em}
    
    \begin{subfigure}[t]{\textwidth}
        \rotatebox{90}{\tiny sit on bed+touch wall}
        \includegraphics[width=0.15\textwidth]{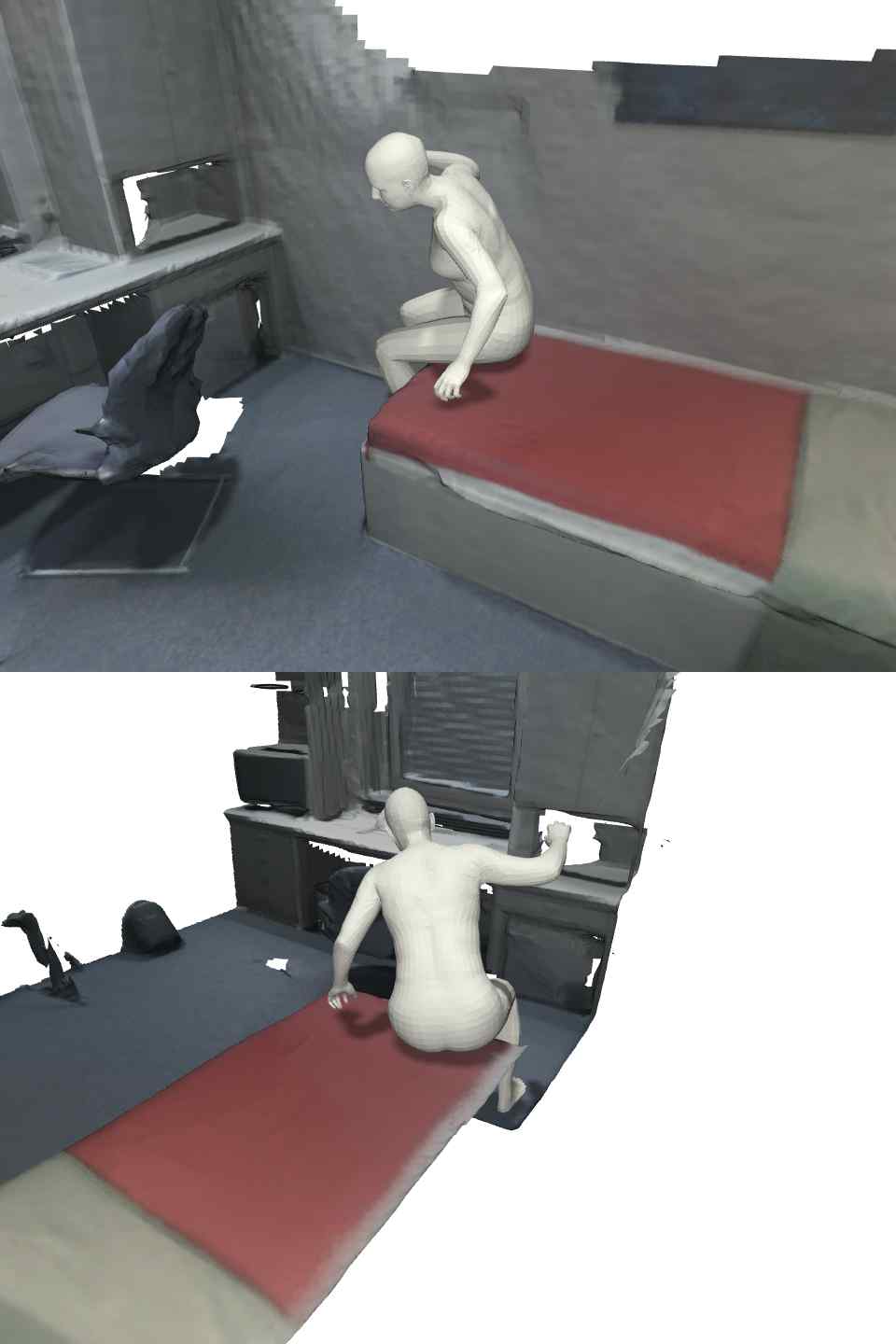}
        \includegraphics[width=0.15\textwidth]{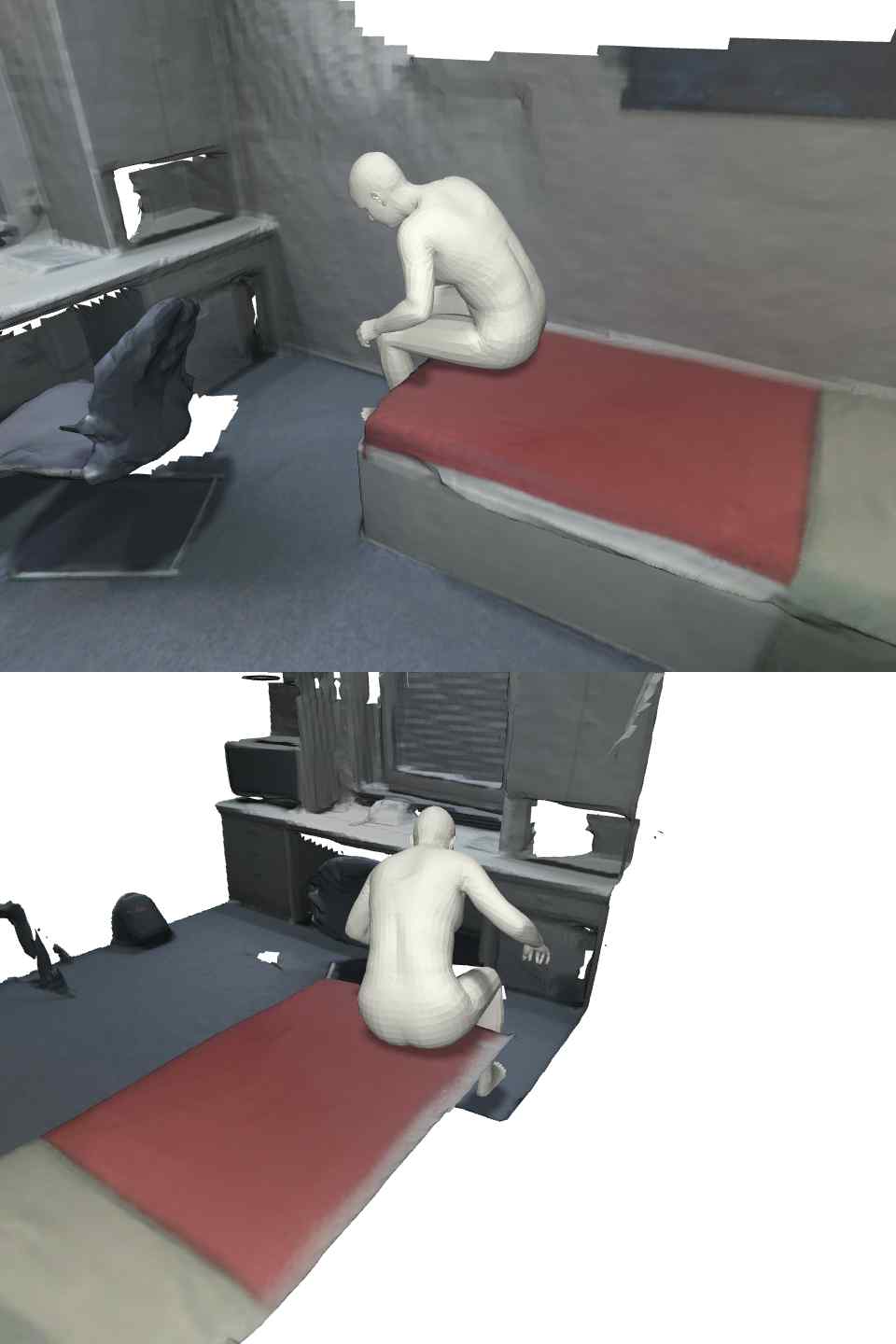}
        \includegraphics[width=0.15\textwidth]{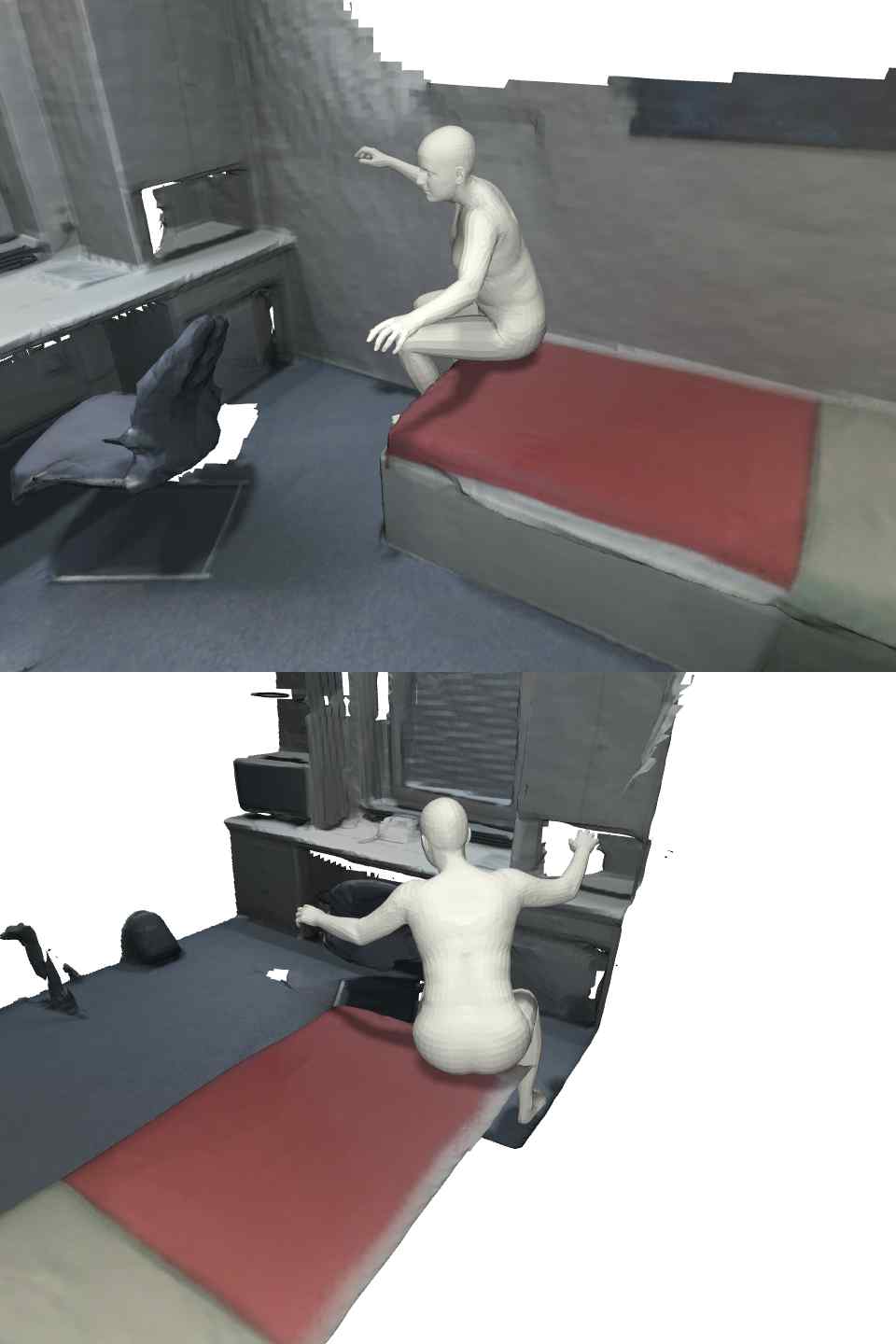}
        \includegraphics[width=0.15\textwidth]{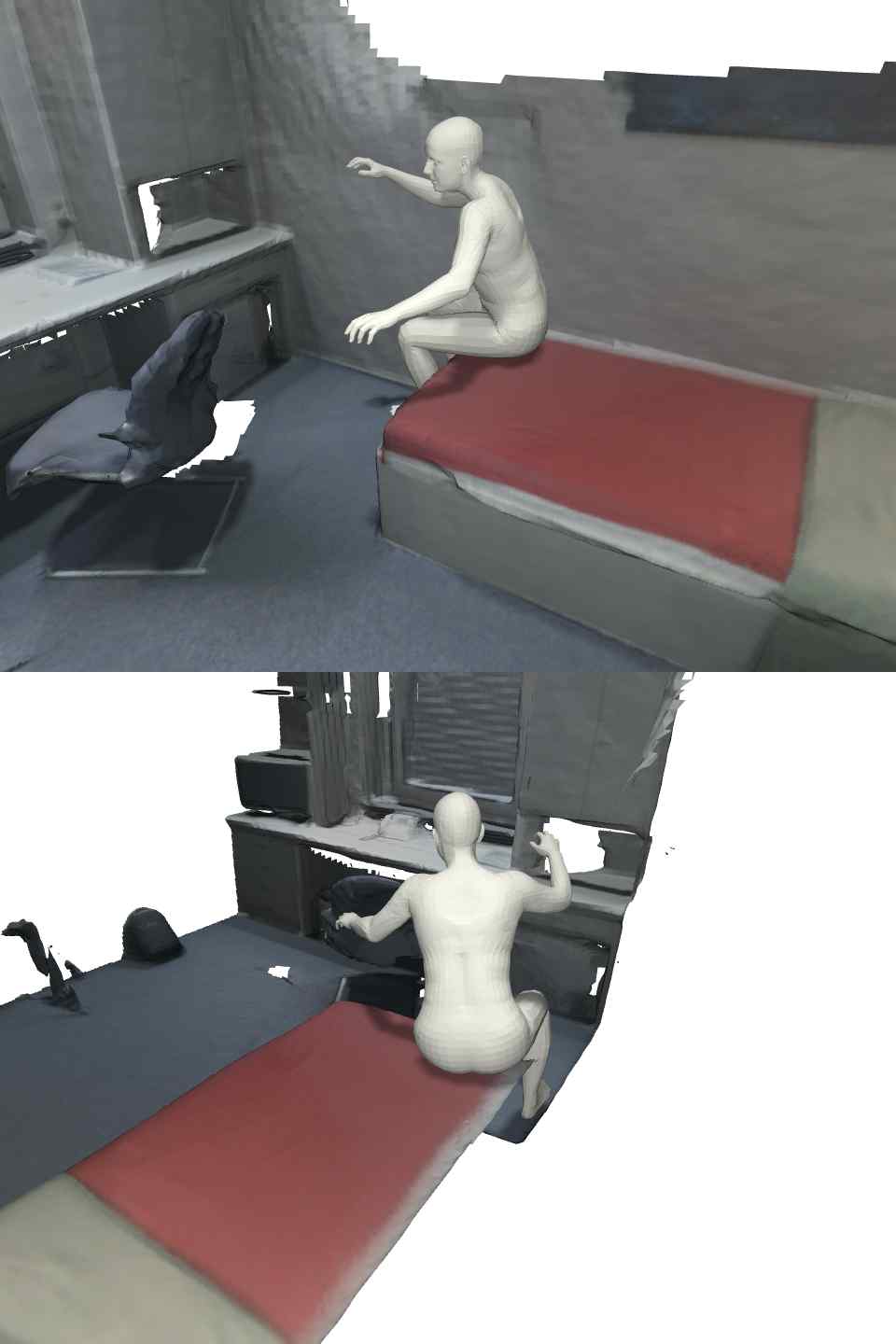}
        \includegraphics[width=0.15\textwidth]{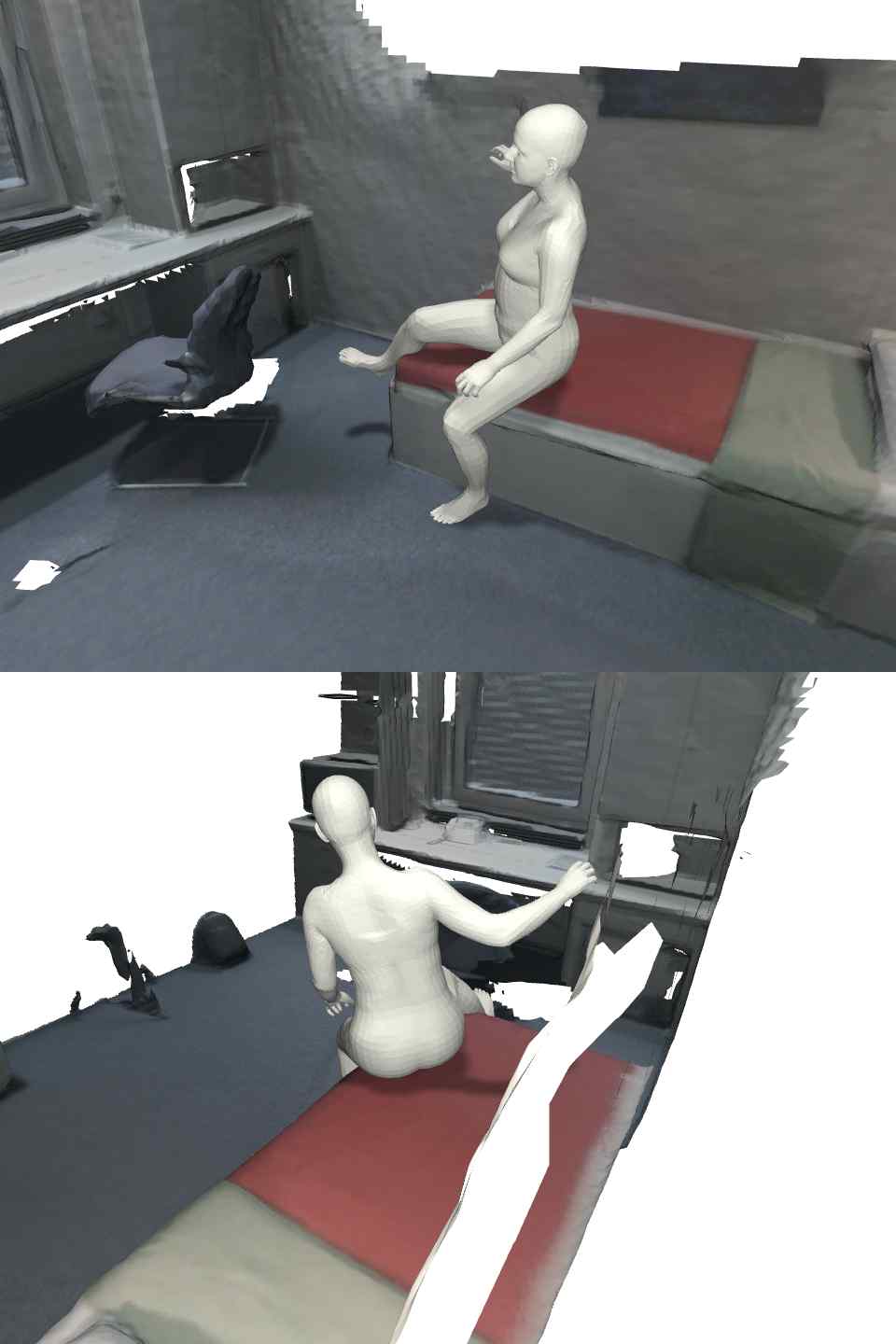}
        \includegraphics[width=0.15\textwidth]{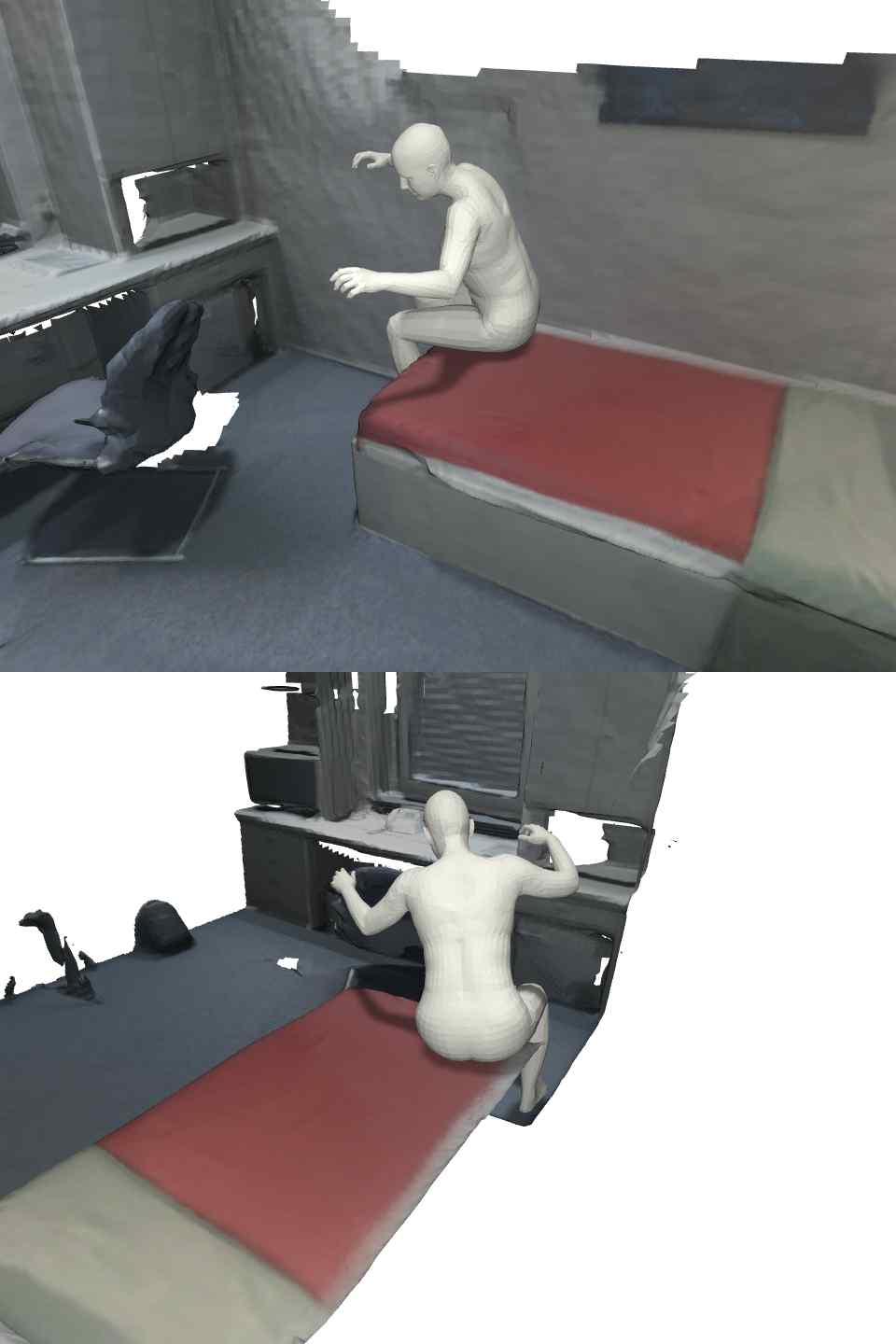}
    \end{subfigure}
    
    \vspace{1em}
    
    \begin{subfigure}[t]{\textwidth}
        \rotatebox{90}{\tiny stand on sofa+touch light}
        \includegraphics[width=0.15\textwidth]{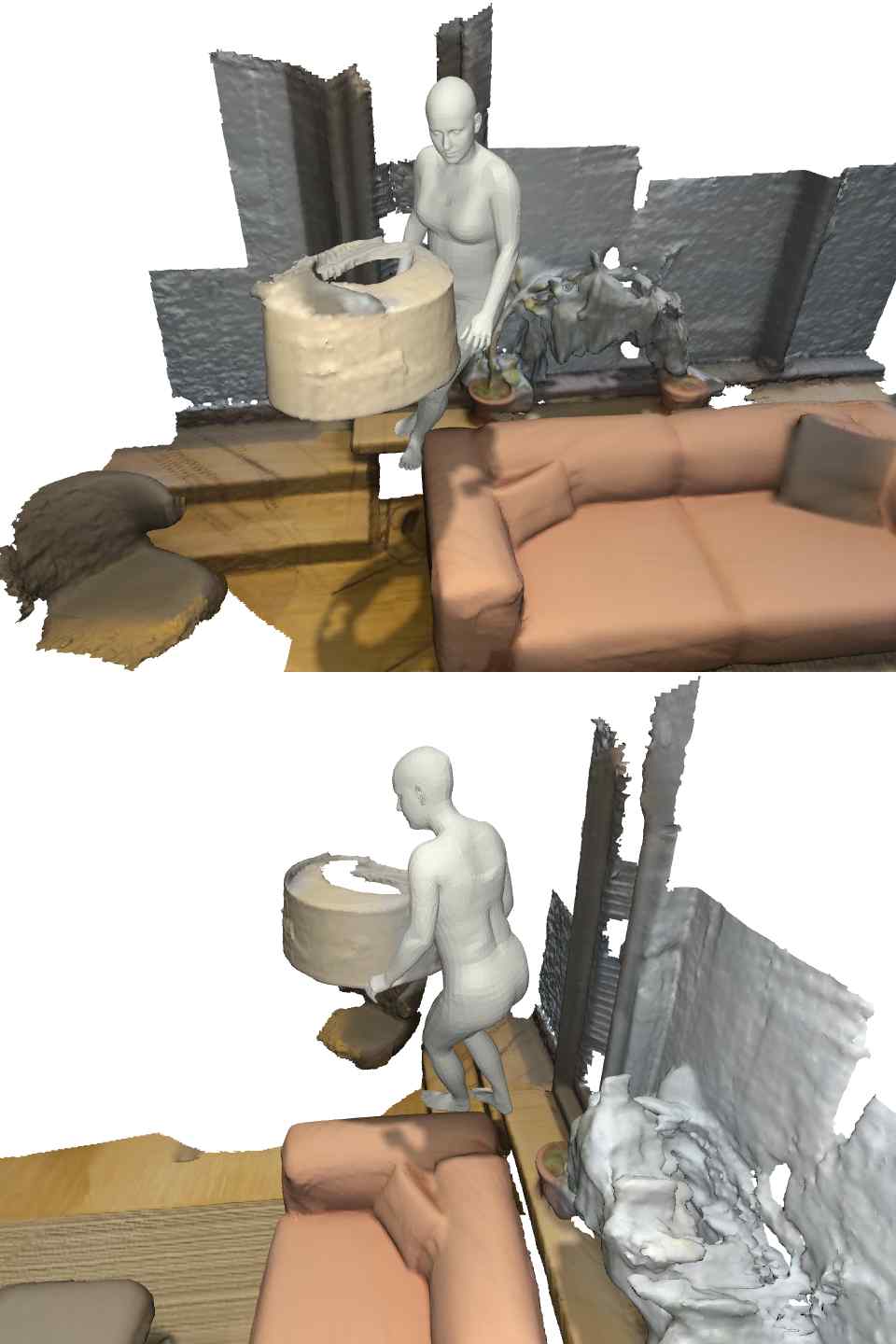}
        \includegraphics[width=0.15\textwidth]{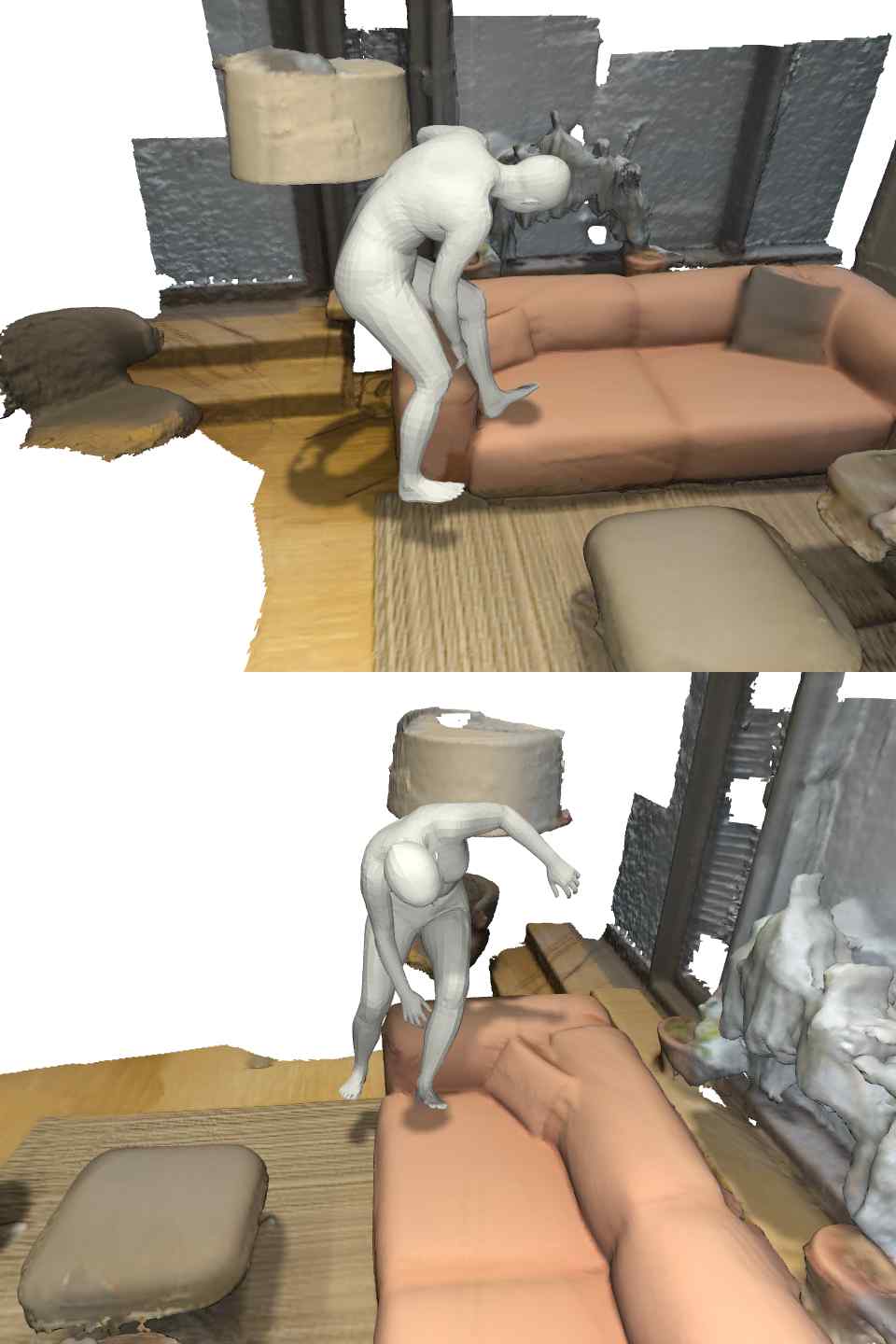}
        \includegraphics[width=0.15\textwidth]{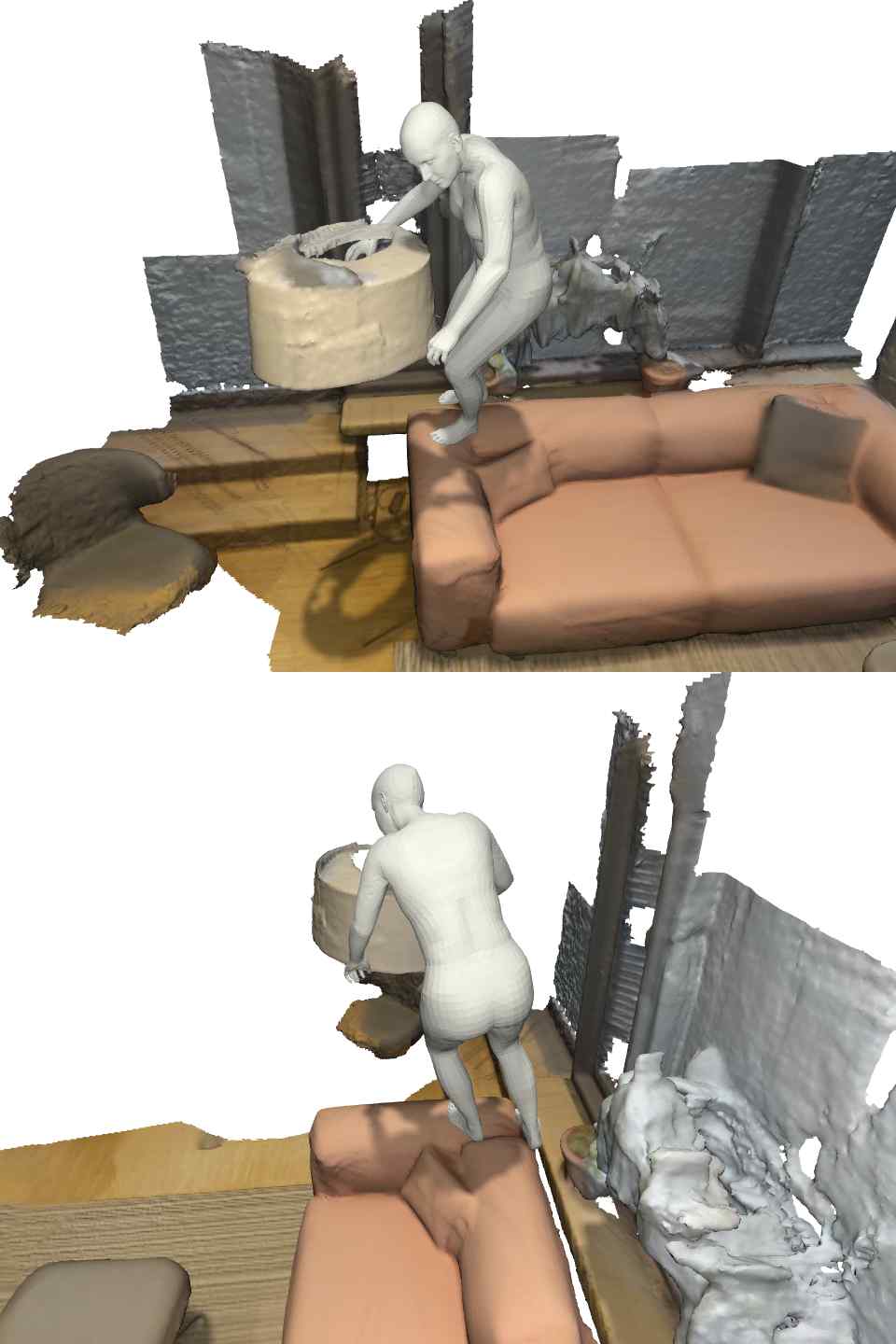}
        \includegraphics[width=0.15\textwidth]{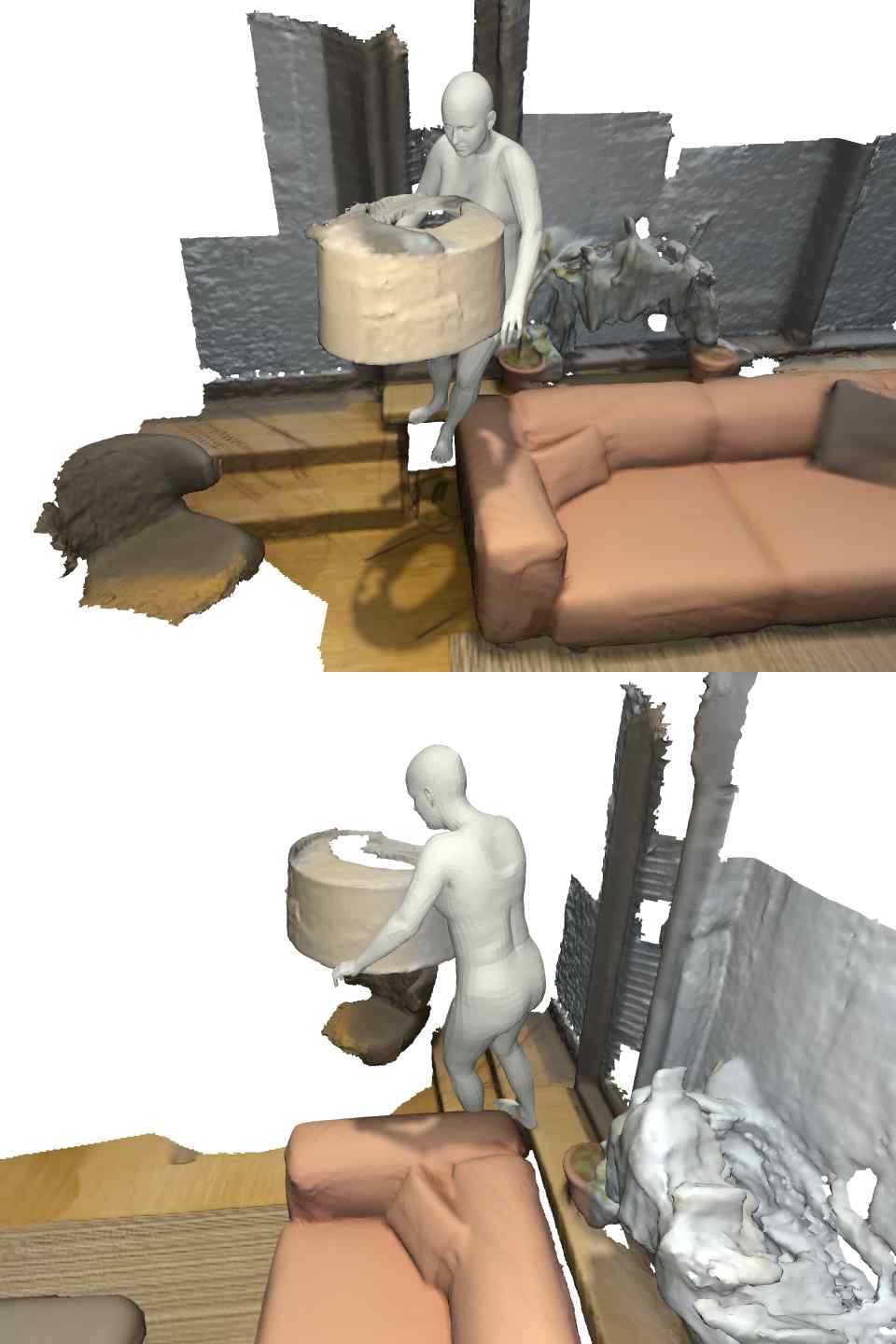}
        \includegraphics[width=0.15\textwidth]{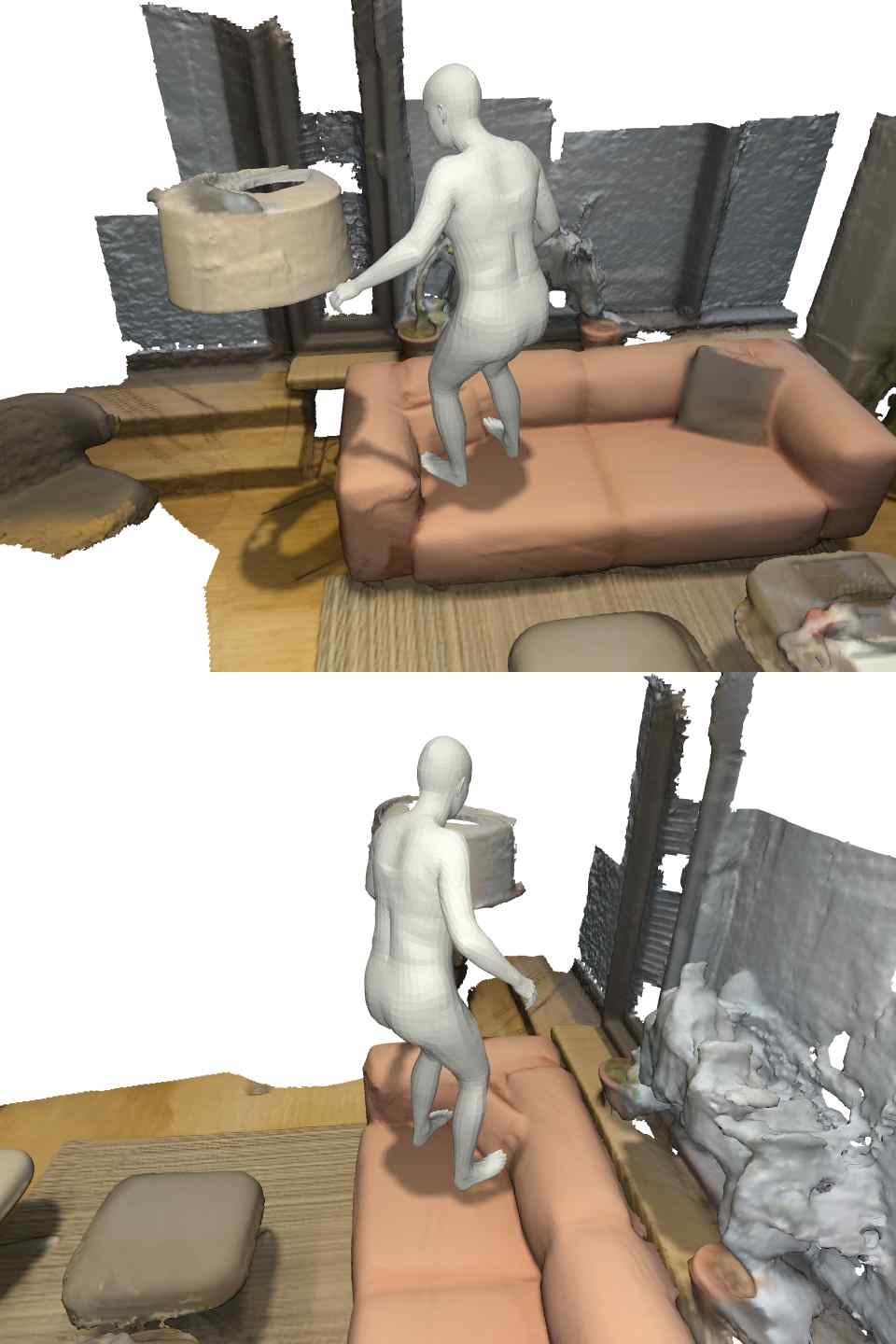}
        \includegraphics[width=0.15\textwidth]{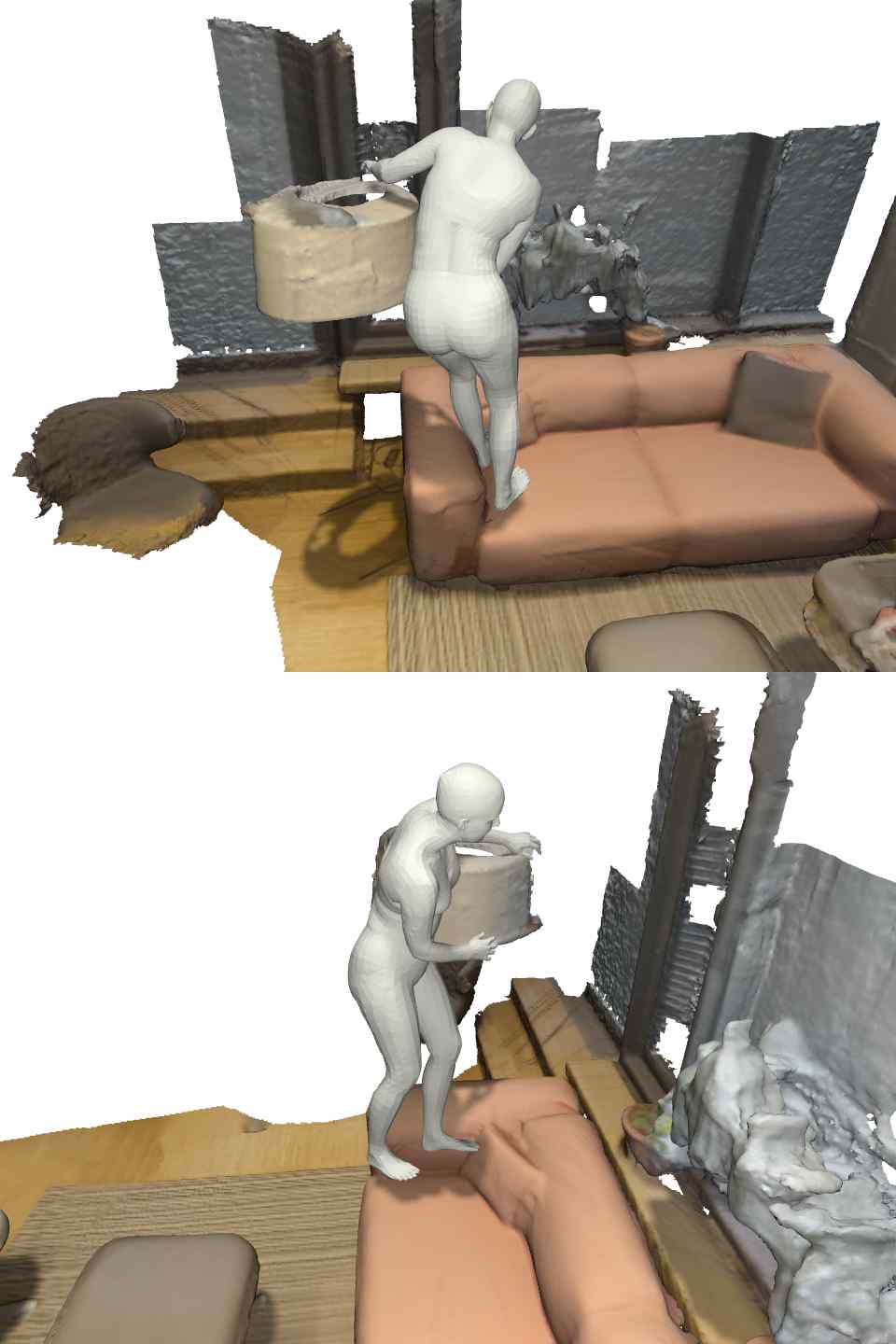}
    \end{subfigure}
    
    \caption{\textcolor{blue}{Randomly} sampled composite interaction from our method. Note that the model is not trained with the corresponding training data.}
    \label{fig:novel_composition}
\end{figure}

We show composite interactions randomly generated by composing atomic interactions in \cref{fig:novel_composition}. Our method is capable of generating composite interactions without corresponding training data.

\subsection{Synthesis in Scenes with Noisy Segmentation}
\begin{figure}[t]
 \captionsetup[subfigure]{labelformat=empty}
    \centering
    \begin{subfigure}[t]{0.24\textwidth}
        \adjincludegraphics[width=\textwidth, trim={0 {.5\height} {.5\width} 0}, clip]{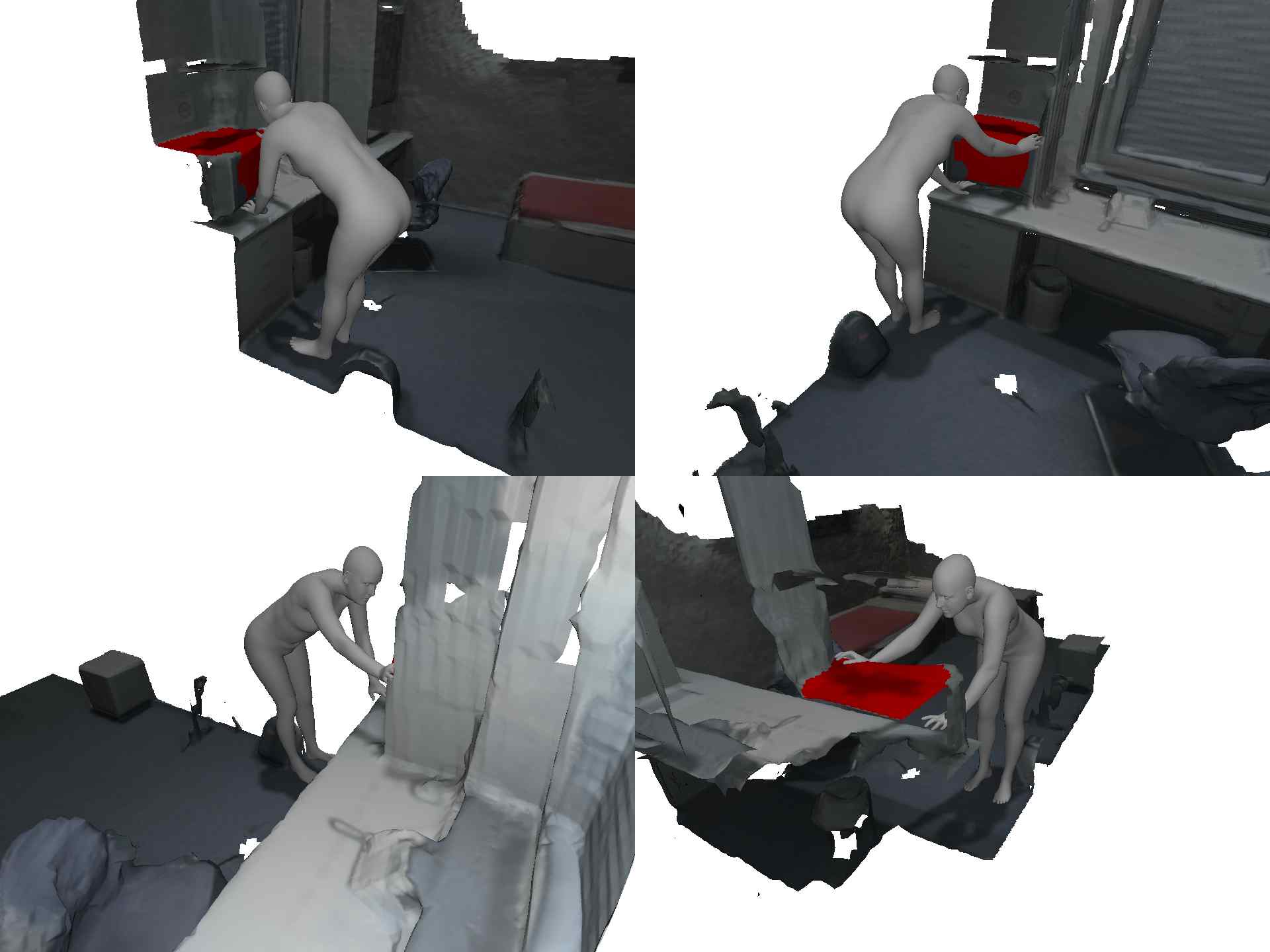}
       \caption{touch}
    \end{subfigure}
    \begin{subfigure}[t]{0.24\textwidth}
        \adjincludegraphics[width=\textwidth, trim={0 {.5\height} {.5\width} 0}, clip]{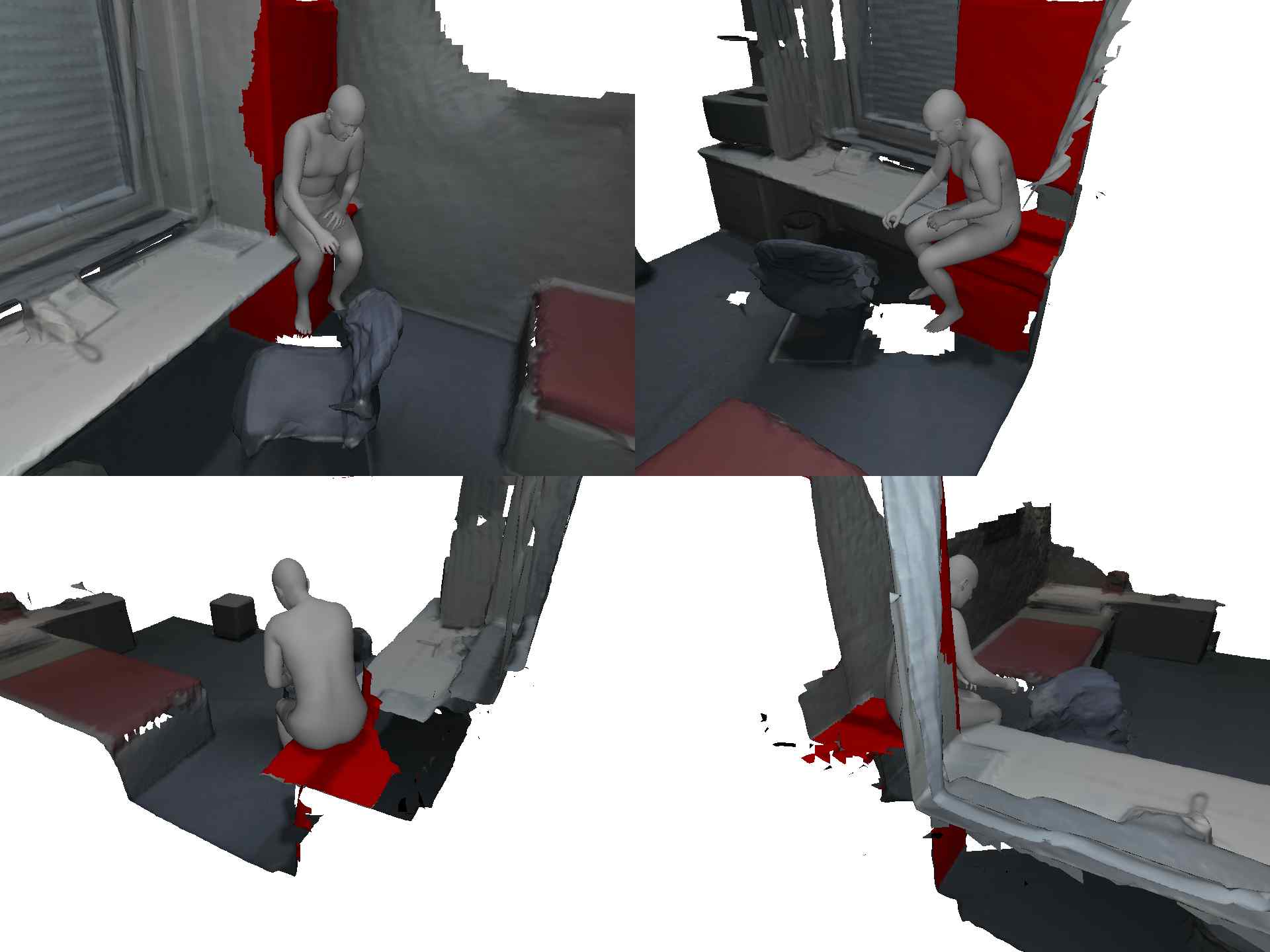}
       \caption{sit on}
    \end{subfigure}
    \begin{subfigure}[t]{0.24\textwidth}
        \adjincludegraphics[width=\textwidth, trim={0 {.5\height} {.5\width} 0}, clip]{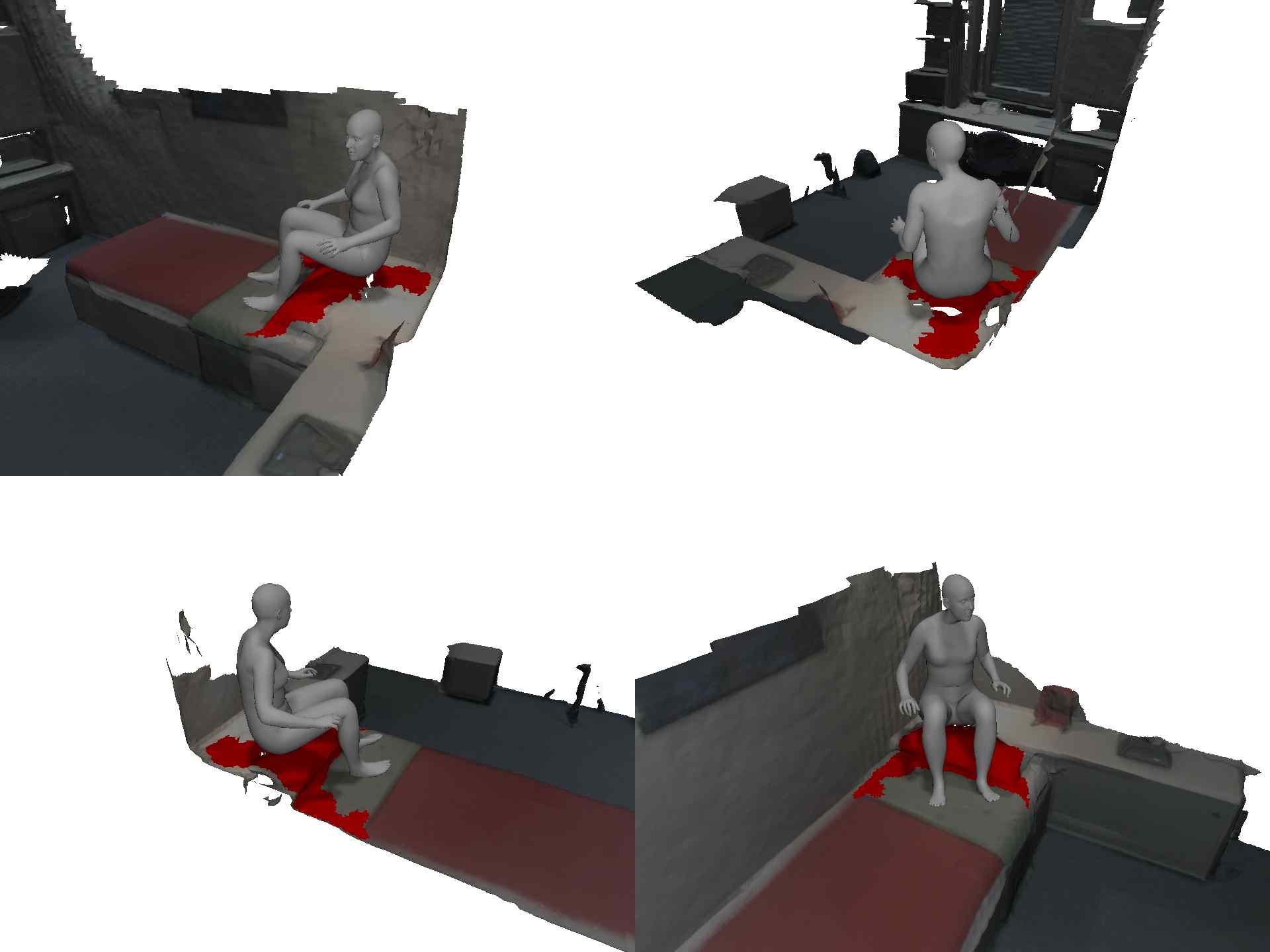}
       \caption{sit on}
    \end{subfigure}
    \begin{subfigure}[t]{0.24\textwidth}
        \adjincludegraphics[width=\textwidth, trim={0 {.5\height} {.5\width} 0}, clip]{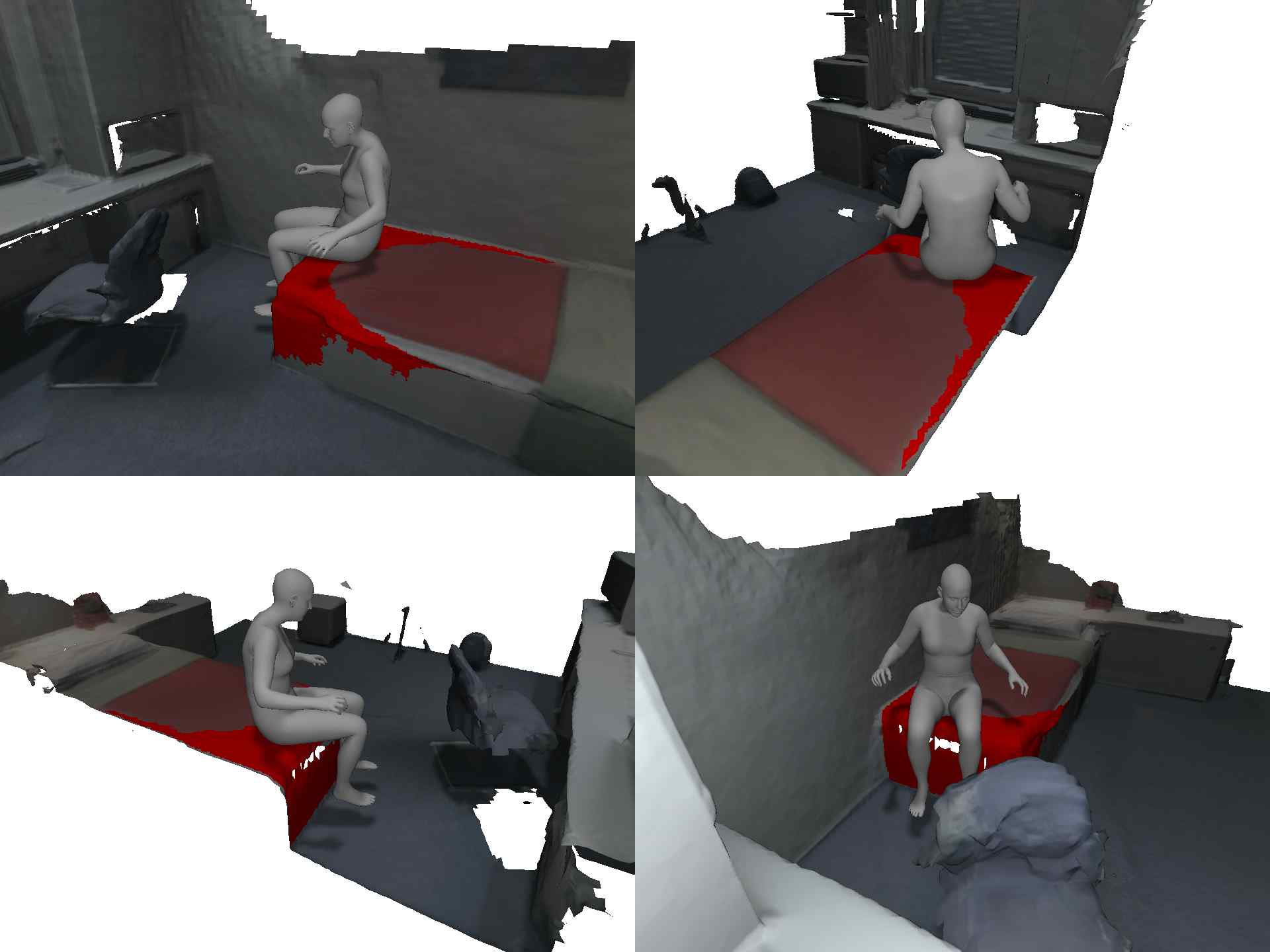}
       \caption{sit on}
    \end{subfigure}
    
    \caption{Synthesized interactions in PROX scenes with noisy object instance segmentation. The noisy interaction objects are highlighted as red.}
    \label{fig:prox_noisy}
\end{figure}

To show the possibility of generating interactions in scenes without ground truth segmentation using our method, we generate interactions in scenes with noisy segmentation obtained using off-the-shelf segmentation methods \cite{Nekrasov213DV, Chen_HAIS_2021_ICCV} where the object geometry can be incomplete and noisier than the scene segmentation we use. 
We show the generation results on noisily segmented PROX test scenes in \cref{fig:prox_noisy}. Our method can generate reasonable interactions given noisy objects as long as the object shape is not significantly different from training objects.

Regarding training with such noisy scene segmentation, we find it demands prohibitively  more effort in collecting interaction semantic annotation with noisy segmentation and conclude that a clean scene segmentation that is consistent with human perception is necessary. 

\subsection{Failure Cases and Limitation}
\begin{figure}[t]
    \centering
    \begin{subfigure}[t]{0.3\textwidth}
        \includegraphics[width=\textwidth]{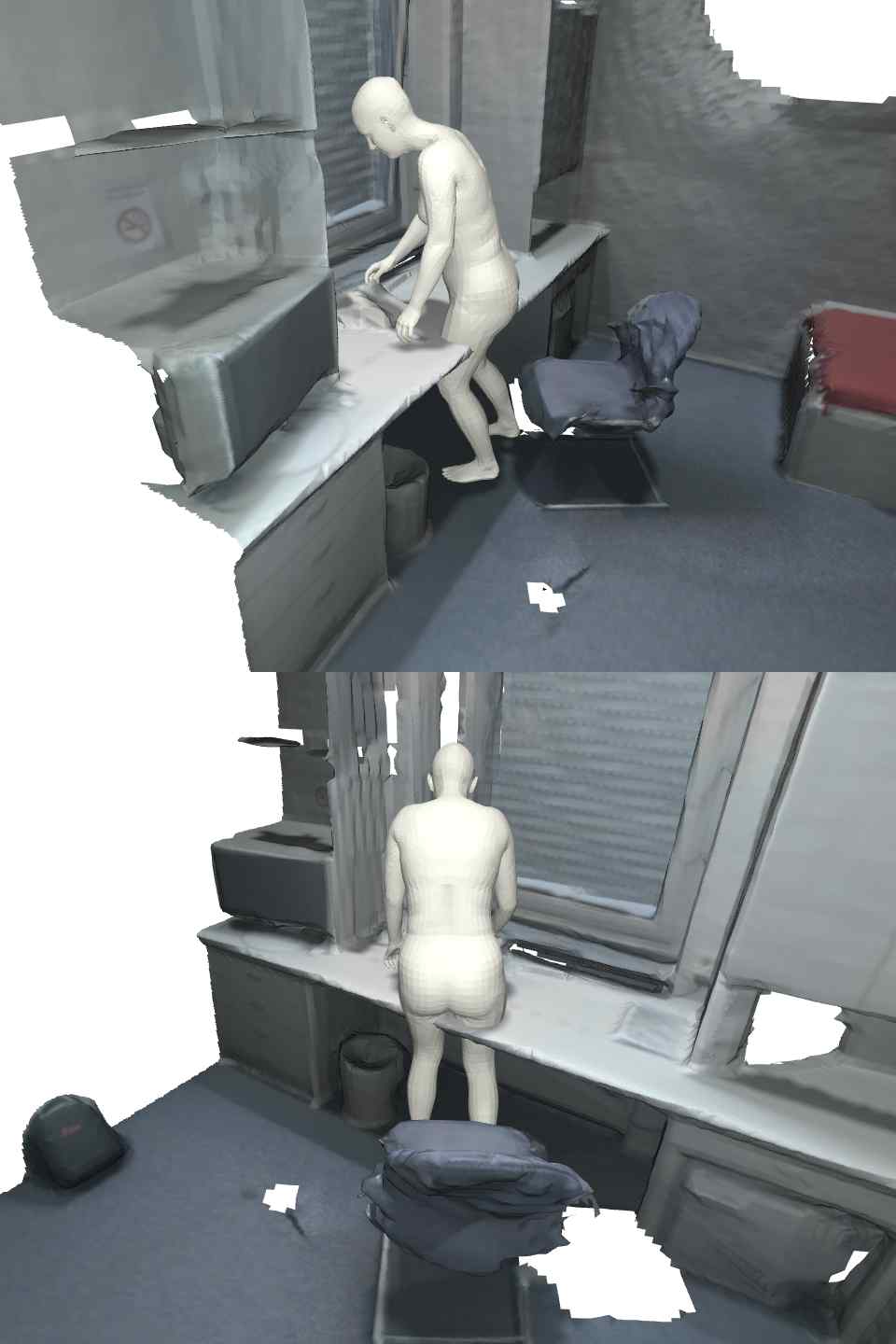}
        \caption{Penetration}
    \end{subfigure}
    \begin{subfigure}[t]{0.3\textwidth}
        \includegraphics[width=\textwidth]{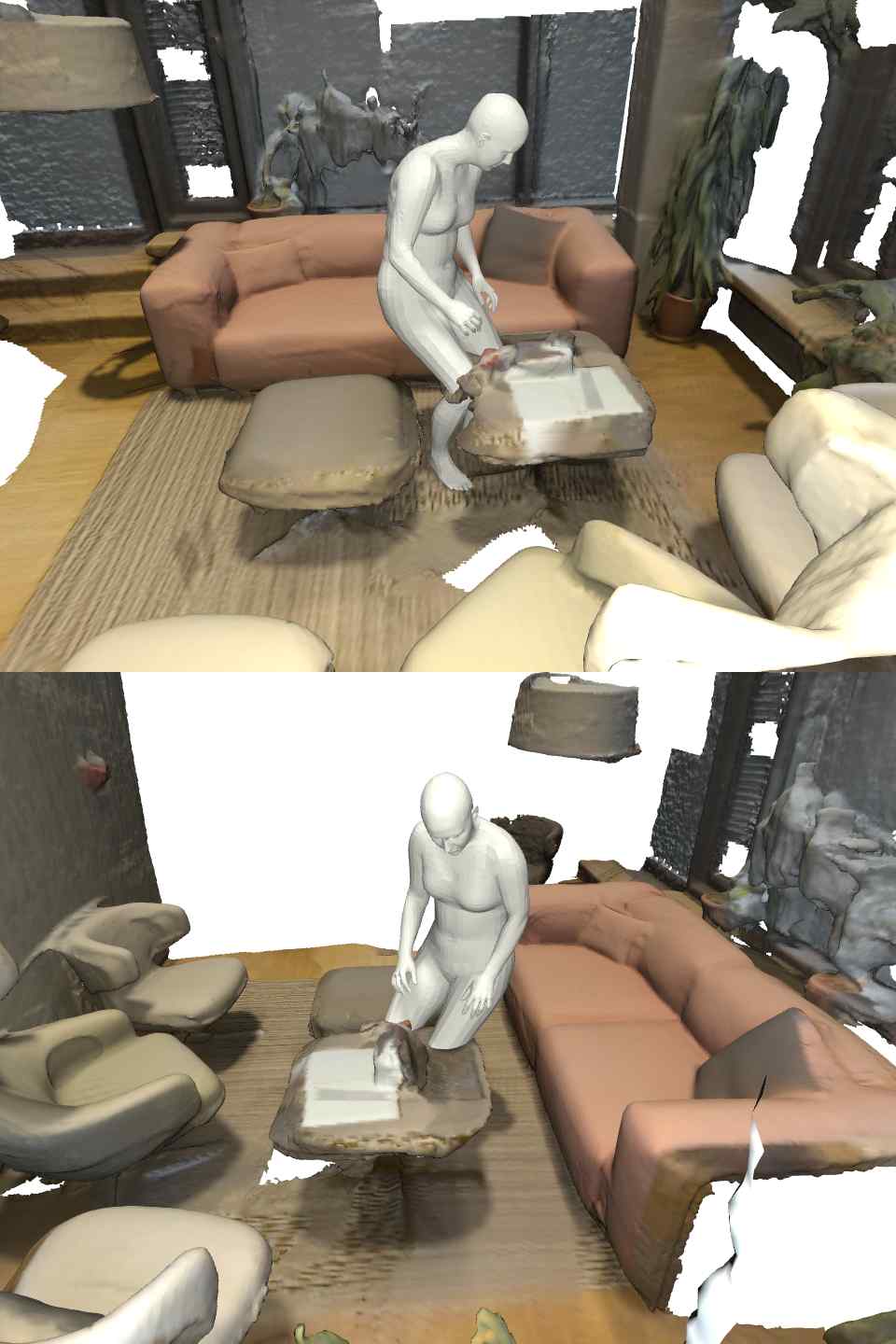}
        \caption{Geometry Variance}
    \end{subfigure}
    \begin{subfigure}[t]{0.3\textwidth}
        \includegraphics[width=\textwidth]{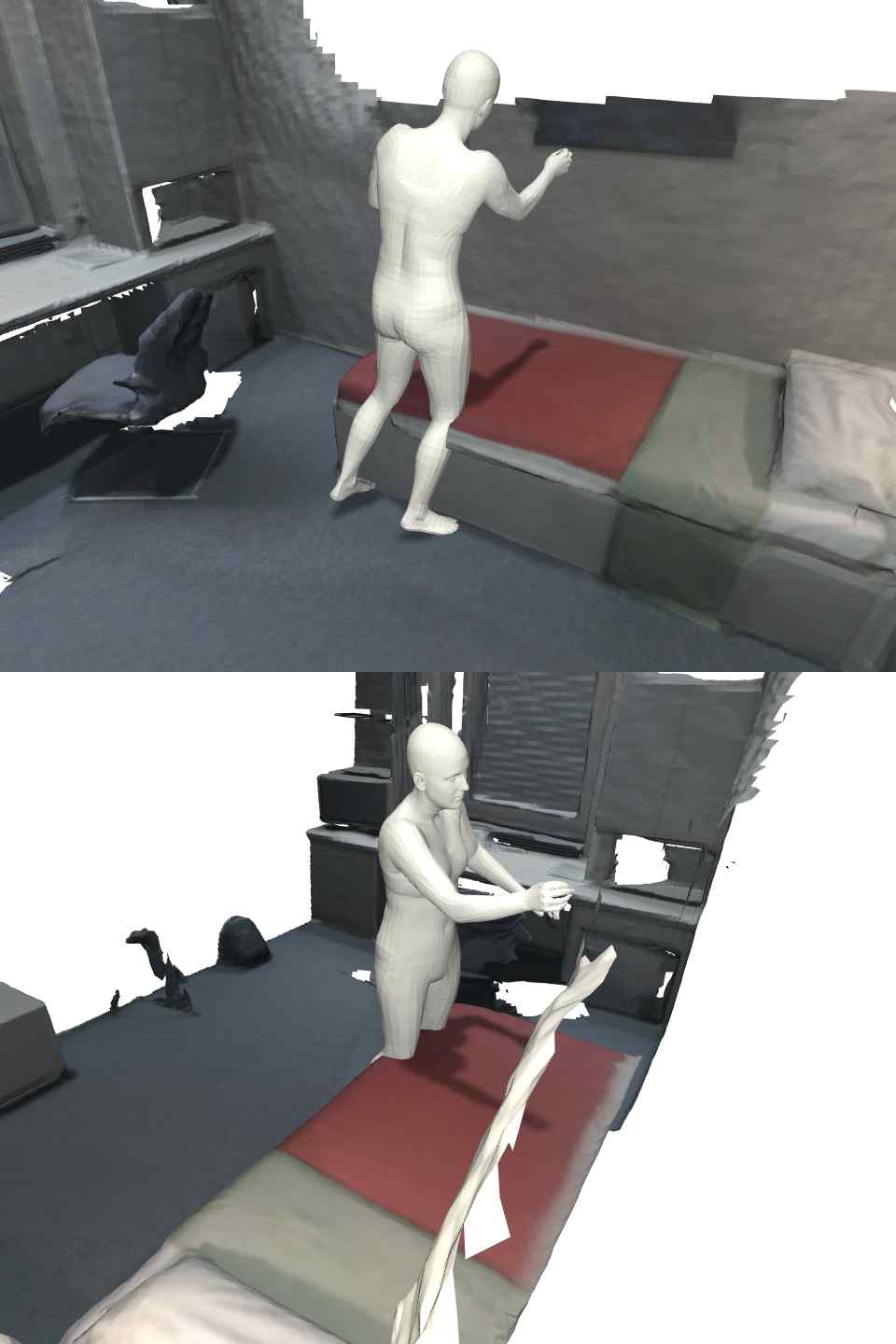}
        \caption{Local Minimum}
    \end{subfigure}

    \caption{Typical failure cases in our results. Example (a) shows  penetration with thin-structure objects where the scene SDF is not effective in resolving penetration. Example (b) shows the synthesis result of touching a table with significantly different geometry from tables in training data, i.e., much lower and smaller in size. Example (c) shows a failure case of being stuck in local minimum in optimization where the human is blocked by the bed from touching the wall.}
    \label{fig:failure}
\end{figure}

\begin{figure}[t]
    \centering
    \begin{subfigure}[t]{0.45\textwidth}
        \includegraphics[width=0.45\textwidth]{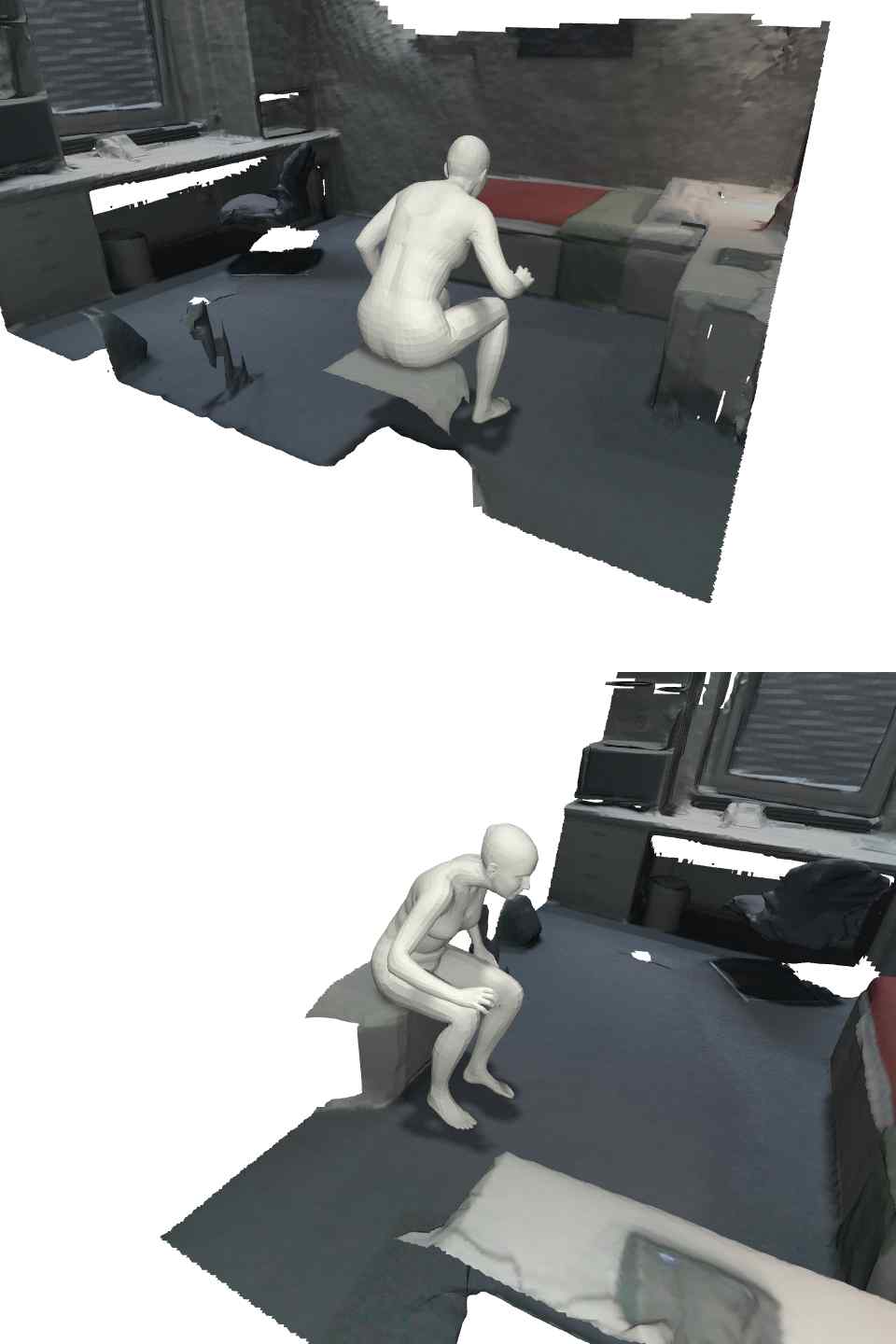}
        \includegraphics[width=0.45\textwidth]{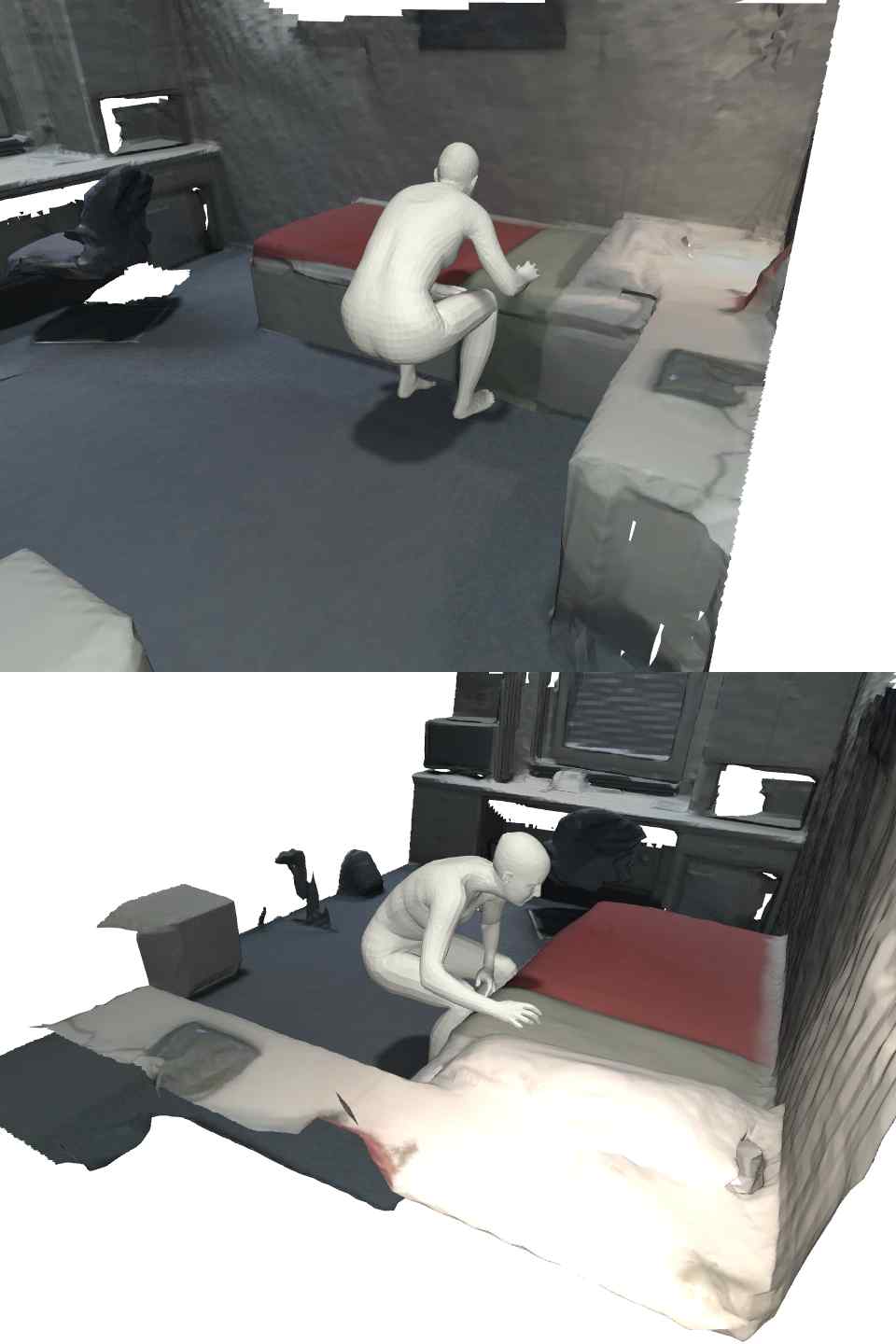}
        \caption{}
    \end{subfigure}
    \vrule
    \hspace{1pt}
    \begin{subfigure}[t]{0.45\textwidth}
        \includegraphics[width=0.45\textwidth]{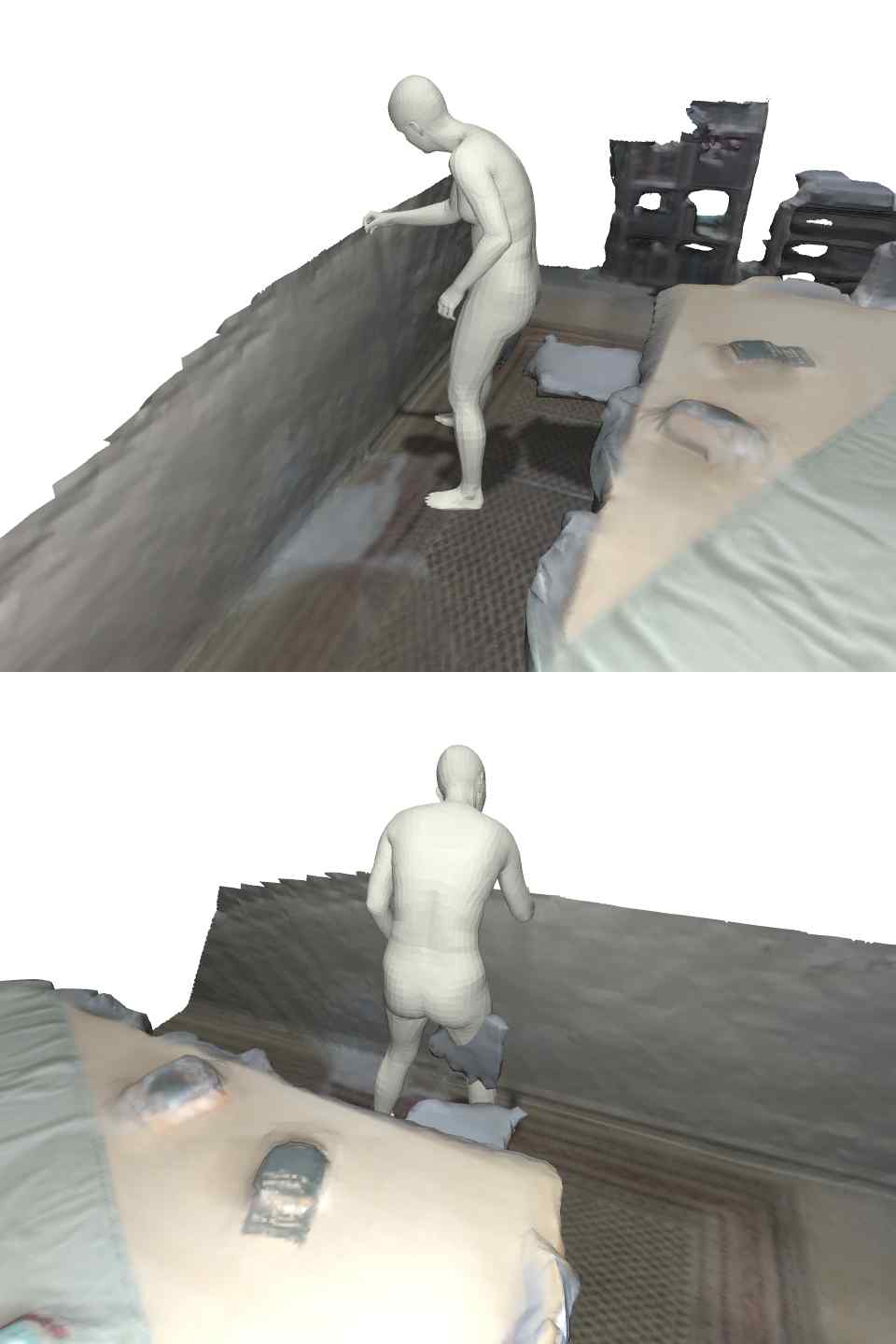}
        \includegraphics[width=0.45\textwidth]{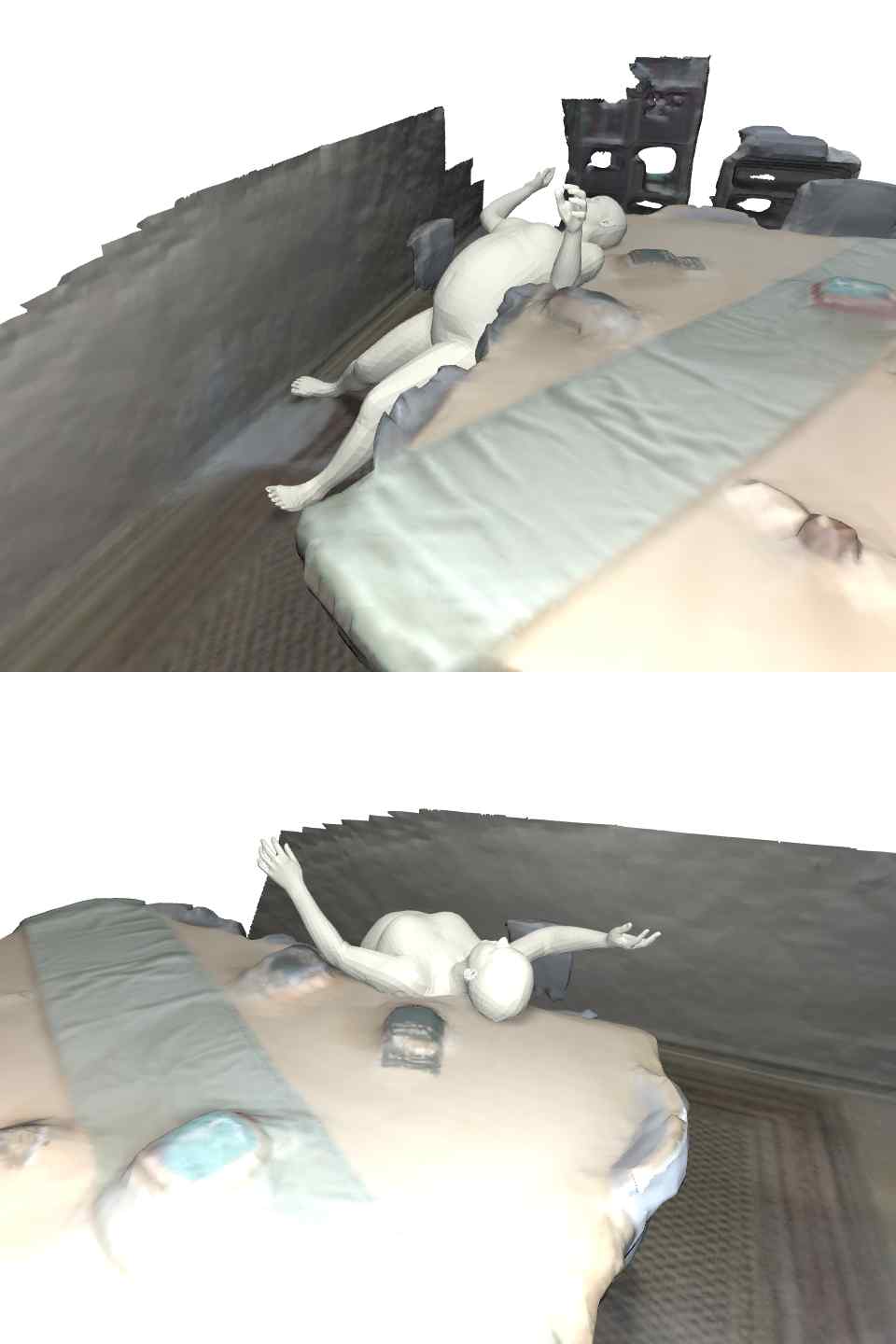}
        \caption{}
    \end{subfigure}

    \caption{Typical failure cases in generating novel interactions by semantic composition. Example (a) shows the failed composition of sitting on a cabinet and touching a bed far away due to being physically impossible. Example (b) shows the failed composition of lying on the floor and touching the wall. 
    }
    \label{fig:failure_composition}
\end{figure}

We show typical failure cases of generating interactions from semantic specifications in \cref{fig:failure} and failure cases of composing atomic interactions in \cref{fig:failure_composition}.

Our method has some limitations.
Firstly, our generative models currently ignore scene objects that are not explicitly specified in the input, which can lead to penetration with unspecified objects. 
We currently solve such penetration using post-processing based on pre-computed scene SDF grids. It is possible to get rid of the demand for scene SDF grid if we use recent human-occupancy methods \cite{Mihajlovic:CVPR:2022}, and learning obstacle-aware generative models is an interesting future direction.



Besides, we observe that hand-object contact in generated results of the touching action are not accurate enough, an issue caused by the low-quality hand estimation in used pseudo-ground truth data. Using hand-object interaction data with high-quality hand estimation can be future work to improve hand-object contact in synthesis results.

In addition, the action semantics considered in this paper is relatively coarse-grained due to the limited scale of interaction data in PROX. 
We do not distinguish left and right limbs in annotation due to limited data and can not generate fine-grained composite semantics such as touching the chair with the left hand while touching the table with the right hand.
Given larger scale interaction data, we expect to model more expressive interaction semantics and compose actions corresponding to finer-grained body parts segmentation, e.g., put the left palm on the table and lean on the right elbow on the table.

Moreover, we observe our mask-based composition method fails in two cases as shown in Figure \ref{fig:failure_composition}: 1) when physically impossible interaction combinations are specified as input, such as sitting on a cabinet while touching a bed 10 meters away. 2) when the data distributions of atomic actions have no intersection in the training data, such as the combination of lying and touching since none of the touching poses are simultaneously lying in our training data.

\end{document}